1928
K.N. Toosi University of Technology
Faculty of Electrical Engineering

Master's Thesis Submitted in Partial
Fulfillment of The Requirements
For The Degree of Master of Science in
Electrical Engineering

# A Probabilistic Framework for Dynamic Object Recognition in 3D Environment With A Novel Continuous Ground Estimation Method

Supervisor
**Prof. Hamid D. Taghirad**
by
**Pouria Mehrabi**

**Sep. 2018**

بسم الله الرحمن الرحيم

Is he who payeth adoration in the watches of the night, prostrate and standing, bewaring of the Hereafter and hoping for the mercy of his Lord to be accounted equal with a disbeliever? Are those who know equal with those who know not? But only men of understanding will pay heed (Al-zumar,9).

K.N. Toosi University of Technology
Department of Electrical Engineering

# Certificate of Acceptance

This is to certify that the dissertation entitled ***A Probabilistic Framework for Dynamic Object Recognition in 3D Environment*** by ***Pouria Mehrabi*** to K.N. Toosi University of Technology, in partial fulfillment of the requirement for the award of the degree of ***Master of Science in Electrical Engineering*** with specialization in ***Control Engineering*** is a bona-fide work carried out under my supervision. The dissertation fulfills the requirements as per the regulations of this University and in my opinion meets the necessary standards for submission. The contents of this dissertation have not been submitted and will not be submitted either in part or in full, for the award of any other degree or diploma and the same is certified.

**Supervisory Committee:**

Prof. Hamid D. Taghirad, Supervisor (Department of Electrical Engineering, K.N. Toosi University of Technology)

Prof. Babak N. Araabi, External Examiner (Department of Electrical Engineering, University of Tehran)

Dr. Mehdi Delrobaei, Internal Examiner (Department of Electrical Engineering, K.N. Toosi University of Technology)

K.N. Toosi University of Technology
Department of Electrical Engineering

# Declaration

I hereby declare that the dissertation ***A Probabilistic Framework for Dynamic Object Recognition in 3D Environment*** submitted by me to the School of ***Control Engineering*** of department of ***Electrical Engineering***, K.N. Toosi University of Technology, in partial fulfillment of the requirements for the award of ***Master of Science*** in ***Electrical Engineering*** with specialization in ***Control Engineering*** is a bona-fide record of the work carried out by me under the supervision of ***Prof. Hamid D. Taghirad***. I further declare that the work reported in this dissertation, has not been submitted and will not be submitted, either in part or in full, for the award of any other degree or diploma of this institute or of any other institute or University.

Sign:

---

Name:

---

Date:

---

K.N. Toosi University of Technology
Department of Electrical Engineering

# Copyright

Sign:

# A Probabilistic Framework for Dynamic Object Recognition in 3D Environment

Pouria Mehrabi

January 23, 2019

# List of Figures

# List of Algorithms

# Contents

# Chapter 1

# Introduction

## 1.1 What is *Vehicle Autonomy*?

Since the success of DARPA Grand and Urban Challenges [3] as well as Google's effort to promote autonomous vehicles and self-driving cars, expectations has been heightened and the belief is strengthen that autonomous vehicles will be able to operate in realistic complex environments. Autonomous driving tasks are highly complex containing many sub-systems performing different necessary tasks in order to make vehicles able to percept the environment wisely and execute control actions autonomously. The concept of the autonomous driving and driverless car is well-known and widely discussed. The media promises an introduction to the market in the near future, Thus automotive companies are competing to develop related new technologies. Recently software companies like Google and Apple are taking part in this heavy competition to confront to the giant vehicle manufacturers.

Vehicle's autonomy may refer to a vast variety of concepts according to the different usage and different environmental conditions. In fact, different degrees of autonomy may be defined according to different situations. For instance, the quality of autonomy needed for a car performing in a highway is different with the same car performing in urban environments. Often three primary actors are considered to be responsible for the driving procedure: the human driver, the driving automation system and other vehicle's systems and components. A given vehicle may be equipped with a driving automation system that is capable of delivering multiple driving automation features which are capable of performing in different levels of automation. Engaged features, will determine the exhibited driving automation level at each instance.

The Society of Automotive Engineers (SAE) provides a useful classification standard for automotive industries seeking common definitions to design their sys-tems [4]. This practice provides a taxonomy for motor vehicle driving automation systems that perform part or all of the Dynamic Driving Task (DDT). This performance may occur at the level of no driving automation (level 0) to a full driving automation (level 5) which is depicted in figure 1.1. According to the SAE's standard, different levels of driving automation are defined regarding to the specific role played by each of the three primary actors in performance of the DDT. Active safety systems such as certain types of driver assistance systems are excluded from the scope of driving automation parties because they do not perform driving task and in fact they provide momentary intervention during potentially hazardous situations. Intervention of active safety systems in driving performance does not eliminate the role of the driver in performing part or all of DDT, and thus are not considered to be driving automation [5].

Advanced Driving Assistant Systems (ADAS) [6] such as Adaptive Cruise Control systems (ACC) [7, 8, 9] and Parking Assists (PA) fall on level 3 of automation according to SAE's standard. In this level human intervention is still needed during the DDT. These ADASs are designed to best perform in low-dynamic traffic situations or well-structured environments such as highways. Other limited Advanced Assists are available for high-dynamic urban areas, like Pedestrian Collision Avoidance Systems (PCAS) [10] or Intersection Assistant Systems (IAS) [11]. Limited autonomy degree of the available Autonomous Driving Systems (ADSs), indicates that vehicle's autonomy specially in complex-structured urban areas, is an ongoing research. In this thesis our definition of Autonomous Land Vehicle (ALV) is according to level 5 autonomy, in which not only the ALV is responsible for monitoring the environment, it is responsible for further decision making and executions. At last, like all the researchers in this field, we aim at the one same shared goal: *To fully remove the human involvement in automotive driving tasks.*

## 1.2 Autonomous Vehicles System Design

Autonomous driving is a complex task, thus a complex system design must be introduced to tackle all the ambiguities of the autonomous driving task [12]. Thus functional requirements for an autonomous vehicle seems to be an important factor to shed light on system design complexities. These functional requirements must include: communication technologies, absolute and global localization, environmental and self-perception, mission execution and integration of human beings whether as the passenger or other traffic participants in public road traffic.

Each ALV is considered to be able to move freely, because it is not restricted to move on rails, bus bars or power supply lines and all the ALVs are considered to drive in public road traffic. The system design for an autonomous vehicle is assumed to be in the fashion that the operation procedure for the human in control of the vehicle, should be in a very intuitive level. Thus the vehicle shall be only instructed by an ***input of a mission*** which is often a transportation task. The mission input must be adaptable to the current needs of the operators especially if the vehicle is transporting humans, for example the vehicle must be ready for an emergency stop or adding a sudden stopover near next restaurant.

| Level | Name | Narrative definition | Execution of steering and acceleration/ deceleration | Monitoring of driving environment | Fallback performance of *dynamic driving task* | System capability (*driving modes*) |
|---|---|---|---|---|---|---|
| ***Human driver*** **monitors the driving environment** | | | | | | |
| **0** | **No Automation** | the full-time performance by the *human driver* of all aspects of the *dynamic driving task*, even when enhanced by warning or intervention systems | Human driver | Human driver | Human driver | n/a |
| **1** | **Driver Assistance** | the *driving mode*-specific execution by a driver assistance system of either steering or acceleration/deceleration using information about the driving environment and with the expectation that the *human driver* perform all remaining aspects of the *dynamic driving task* | Human driver and system | Human driver | Human driver | Some driving modes |
| **2** | **Partial Automation** | the *driving mode*-specific execution by one or more driver assistance systems of both steering and acceleration/deceleration using information about the driving environment and with the expectation that the *human driver* perform all remaining aspects of the *dynamic driving task* | **System** | Human driver | Human driver | Some driving modes |
| ***Automated driving system*** **("system") monitors the driving environment** | | | | | | |
| **3** | **Conditional Automation** | the *driving mode*-specific performance by an *automated driving system* of all aspects of the *dynamic driving task* with the expectation that the *human driver* will respond appropriately to a *request to intervene* | System | **System** | Human driver | Some driving modes |
| **4** | **High Automation** | the *driving mode*-specific performance by an *automated driving system* of all aspects of the *dynamic driving task*, even if a *human driver* does not respond appropriately to a *request to intervene* | System | System | **System** | Some driving modes |
| **5** | **Full Automation** | the full-time performance by an *automated driving system* of all aspects of the *dynamic driving task* under all roadway and environmental conditions that can be managed by a *human driver* | System | System | System | **All driving modes** |

**Figure 1.1:** SAE's standard for different levels of driving automation.

Since the ALV is performing in public road traffic and safe performance is an obligation in these areas, extra care must be taken about both environmental perception/awareness and driving decision/behaviors. Urban environments are highly dynamic, thus autonomous vehicles must robustly detect and classify the static and dynamic elements present in surrounding environment. Different traffic situations must be assumed to include both ***automated*** and ***manually driven*** vehicles. The locally defined road traffic regulations are of special interest as they define a minimal amount of environmental elements which have to be perceived and considered. Regulations in a connection with vehicle's situation, might specify the behavior in certain defined situations and environments. Therefore, the vehicle needs to be aware of its skills and abilities and has to act accordingly to its actual state. The estimation of the skills and abilities including the surveillance of hardware and software is another mandatory requirement (on-board diagnostics). Moreover, the vehicle has to be resistant against mis-use and manipulation. In summary autonomous vehicles must be capable of handling the following aspects [12]:

- **Operation:** The vehicle must be instructed by a human being.

- **Input mission:** The ALV must be capable of handling input missions by a human operator, and thus must accomplish the desired mission.

- **Communication and Map data:** communication and reading map data is necessary for route planning and navigational purposes.

- **Localization:** The vehicle must be aware of its global and local position and orientation (pose) to be able to navigate in the environment. Furthermore, the global localization is necessary for Vehicle to Vehicle (V2V) or Vehicle to Infrastructure (V2I) communications.

- **Environmental Awareness (Perception):** The vehicle must be able to perceive its surrounding environment, especially local and near static and dynamic objects to be able to react to intentions of other traffic participants. Furthermore, environmental perception enables the vehicle to be ensured that it does not initiate any danger to human and its environment.

- **Self-perception:** The vehicle must be aware of its current state, being functional capabilities, components and motion cues.

If the human driving behavior assumed to be the basis for modeling the driving task, certain studies used a three-level model to interpret human driving task [13, 14]. According to these models driving task can be considered to be decomposed into three different tasks: **navigating**, **guidance** and **stabilization**.

## 1.3 How the System Works?

First, the mission is processed on the highest level of system architecture where route-planning and navigation strategies are taking place. The input data for this task must be consisted of all the informations about relevant road networks. In the next level, vehicle's surrounding information and local scene must be taken into account to enable safe control of the vehicle. Furthermore, at the stabilization level, selected maneuvers of the guiding module are processed.

Two major approaches for system architecture were published in the field of autonomous driving systems. Most of common approaches related to the teams taking part in DARPA challenges are **localization-based** approaches as a detailed map was available in the DARPA challenge. Among these teams only [15] gave a **perception-based** method. **The perception-based** method is much reliable in urban environment driving scenarios as a detailed map of the environment **is not** available in ill-structured urban scenes. Thus, it is of great importance for every autonomous vehicle system design to suggest a structure for the processing of the environmental data. Methods like [16], [17], [18] and [19] have used a single perception block in their system design. Their system design lacked a unique hierarchical structure for environmental perception and have used sub-modules for processing all of the environmental data.

In urban driving scenes, an autonomous vehicle must consider various rules according to traffic regulations. Vehicles must also be capable of adjusting their perception to the lane-markings and traffic signs and also must be capable of co-operating with other vehicles. Thus, environmental modeling is not just as simple as tracking other vehicles but is a challenge of high complexity. Modularization and hierarchical structuring may be used to tackle this high complexity challenge. Many system architectures, adapted from DARPA Urban Challenge divide the autonomous driving task into multiple sub-tasks, but there is a lack of modularization and structuring in the entire system and how the system handles the processing of multiple incoming data e.g. map data, on-board perception system or even perception system of other traffic participants [2].

Another vital aspect of environmental perception is the up to datedness of the environmental data, which makes it necessary that a **real-time perception** system for the entire scene must be included in functional system design. Due to an interruption between data collection and map construction and usage of map data, the up-to-datedness of the environmental data is not always guaranteed. Therefore, a real-time perception system, either on-board or as part of a smart infrastructure is needed to ensure suitable real-time performance of the vehicle.

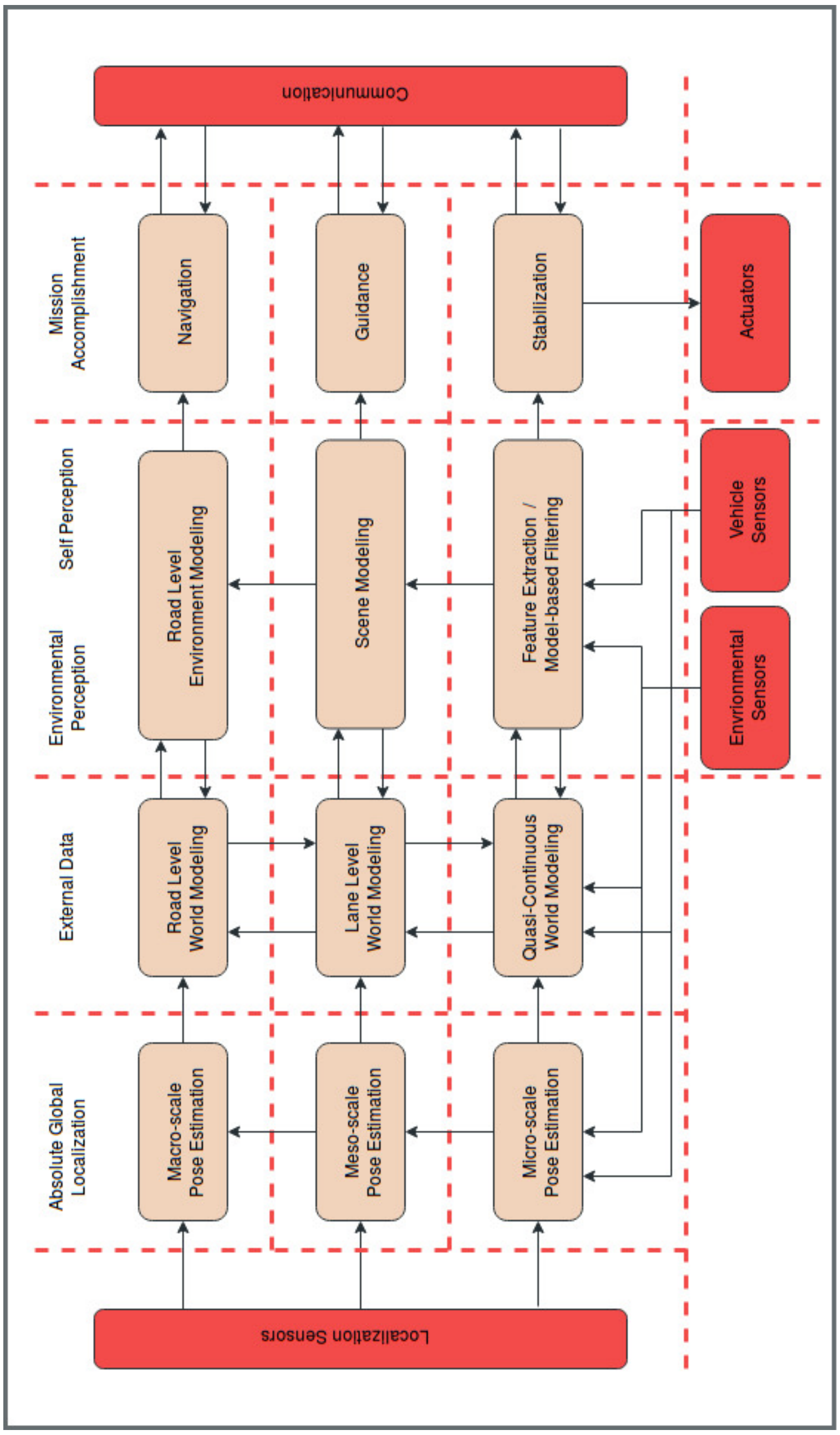

**Figure 1.2:** Functional system architecture for an autonomous on-road motor vehicle in the sense of a modular building block system.

A three-level design is assumed to be the basis for the review in this thesis which is originally developed by [12] and is depicted in Figure 1.2. These three horizontal levels differ mostly in their resolution, horizon and accuracy in time and space. Furthermore their tasks span different areas of actions. Levels of the system architecture are as follows:

- **Strategical level:** Planning, macro-scale resolution.
- **Tactical level:** Decision making, meso-scale resolution.
- **Operational level:** Reactive stabilization, micro-scale resolution.

It can be seen that, this architecture is divided into different vertical sectors as follows:

- **Absolute Global Localization**
- **External Data**
- **Perception**
  - **Environmental Perception**
  - **Self-perception**
- **Mission Accomplishment**

Furthermore, it can be seen that whole system design is separated into two different sector with regard to how modules interact with environment. Absolute Global Localization (AGL) and External Data falls into the sector relating to how the vehicle is in relation to the environment. Perception and Mission Accomplishment is related to how the environment is in relation to the vehicle.

### 1.3.1 Absolute Global Localization (AGL)

Global localization determines the pose of the vehicle in relation to the environment which is not necessary in many system design scenarios as well as ours, where the scenario is mainly **vehicle-referenced-view** based. Unlike the case of our interest in this thesis, the absolute global localization module may be of great importance in many fields of applications. For example in an environment where no local environmental features are available, like deserts or other ill-conditioned environments, AGL may be used for stabilization of the vehicle. Even in urban scenarios, integration of external data related to multiple traffic participants for several purposes like sharing common map data or V2V and V2I communication, the AGL would be of great importance. AGL may establish a central server-based platform to enable global automated map updates for the use of all traffic participants.

The accuracy of the localization solution determines the applicable level of the solution. **GNSS**[1] receivers have positioning errors up to 20 meters. As this accuracy will be only considered a macro-scale pose, these technologies are only applicable withing the strategical level. Other technologies like **DGPS**[2] may be used to improve positioning accuracy by fusing the motion estimation into position estimation. These methods will enable the use of conventional localization solutions in the tactical or operational level while being highly dependent on safety requirements.

### 1.3.2 External Data

All environmental data generated in or perceived from outside the vehicle are considered to be part of external data. Therefore, all data generated or perceived from other vehicles, provided from data storages outside the vehicle or received from radio communications, a world model defined in a global reference frame, data about the static environment e.g. map data, state of the traffic light, traffic signs, buildings, static objects in the road and etc, as well as the data about potential dynamic environment (so-called **movable**) like hazardous objects, road closures, traffic jams, object list, requests from and state information of other traffic participants may be assumed to be part of external data.

External data provides different levels of accuracy in different time spans regarding to abstraction level of the system design. For instance messages regarding traffic jams and temporal road closures which todays are being sent by the Internet or radio channels to vehicles, may be assumed as part of strategical and tactical level data. A list of hazardous objects near the vehicle may be assumed as an operational level data which the vehicle must be aware of and ready to act upon them. Different kind of input data may be considered for this architecture to be assumed as external data: An absolute global pose, local environmental data driven from environmental perception module, driver intentions received from V2X[3] module and map data from map suppliers.

[1] Global Navigation Satellite System
[2] Differential Global Positioning System
[3] Vehicle-to-X (other objects)

### 1.3.3 Perception

The perception module is the central module of an autonomous vehicle covering both environmental and self-perception as depicted in Figure 1.3. All of the information gathered in autonomous driving task about the surrounding vehicles and the vehicle itself are gathered in the perception unit. As the environment may be represented in different ways in external data unit, the kind of representation is highly dependent on abstraction level of the environmental perception. Vehicle sensors, environmental sensors and external available data are the main sources of input to the perception module as can be seen in Figure 1.2. Furthermore, modern autonomous cars may output their perception to other vehicle in the scene for cooperative goals. Mobile phone tracking and route finders are an example of collaborative perception sharing where the information is shared through a map platform and with a map-relative position. As the perception module is the most important part of an autonomous vehicle, at least for the sake of this thesis, different abstraction levels are discussed individually.

**Operational Level:** Operational level is considered as the lowest level of the abstraction. At this level the focus of the environmental perception unit lies on the precise and quasi-continuous feature extraction from incoming sensor data. Geometrical features, position, velocity and visual qualities like color of the surrounding objects may be inferred in this level, while self-perception unit processes data for inner vehicle state. Different methodologies that are prone to be availed to gain a fair perception of the environment along with different blends of the sensors available for the task makes the procedure for the environmental perception more anchored on designers taste and scene conditions.

**Tactical Level:** Tactical level may be considered as the main process for scene modeling. The main focus in tactical level perception module is to construct an associative context based on independently perceived environmental features. Static or stationary environment is constructed at the first, and movable environmental features are added to the scene afterward to build a complete scene. The semantic information is of greater importance in this abstraction level rather than geometric or topological informations.

**Strategical Level:** The environment is considered in the most abstract way on this level. A macro-scale level of observation is considered. Features are mostly road networks, macroscopic traffic flows, cross-roads and topological features. The connection of roads and topological structures of the environment are used in this level for macro-scale route planning while geometric and semantic informa-tions from other two lower levels are still necessary for optimized route planning and navigation.

### 1.3.4 Mission Accomplishment

The mission from passengers or other operators, construct the input to the mission accomplishment process. Three different abstraction levels of strategical, tactical and operational is again assumed to be present here. The results of mission accomplishment module have to be executed and communicated with other traffic participants through different kinds of channels. In the case of an automated vehicle the receivers are passengers of the vehicle, or in the driver assistance systems the driver itself. Communication may be performed acoustically like horn on tactical level, or optically like brake lights in operational level or also every other kind of communications in V2V.

At the strategical level, **planning** is accomplished. Macro-scale informations like road networks and traffic flows are the features being provided for this level. Furthermore, a passenger or driver can instruct an autonomous vehicle in the strategical level to operate as the operator may prefer. Navigational procedures are performed at the highest strategical level. The road network and current traffic flow is the needed input information for any navigational task. Based on this data and the provided destination and optimization criteria from passengers, an optimized route is planned which is usually accomplished one per mission input. The planned route can be adapted on-line due to changing traffic flow or further detection results from the road.

At the tactical level, **decision making** is accomplished. Meso-scale informations like abstracted local scenes consisting of static and dynamic movable objects are the features being provided for the use of this level. After the route planning within the strategical level is done, as the vehicle moves, the environment changes relatively to the vehicle. The output of strategical level to the vehicles guidance system is the next navigational point, which may be seen within today's driving assistance systems as the acoustic alarm to turn in specific direction. Thus, tactical or guidance level receives the mission indirectly. As the perception module provides the vehicle's pose within a scene, the situation assessment is able to analyze the scene with respect to the current mission. In this level, the system alarms other traffic participants if a hazardous situation is detected while in assistance systems warning is sent to the driver. The decision unit selects the driving maneuvers based on the current situation with respect to the traffic regulations [12].

At the operational level, **decision execution** is accomplished. Micro-scale information about exact geometric values are fed into this level to enable reactive collision avoidance and vehicle stabilization. The features used in this level is provided from perception unit. Environmental perception extracts features from model-based methods and feeds them into this level. The trajectory generation calculates a time and space-based nominal position of the vehicle and thus applies a closed-loop control of the vehicle considering the current environmental data. That is why the trajectory generation is classified as an overlaid closed-loop control [12].

## 1.4 Importance of Detection and Tracking of Multiple Moving Object

The task of Detection and Tracking of Multiple Moving Objects (DaTMO) falls into the perception module of an ALV and in the operational level. Operational level or the lowest level of perception module in abstraction, is directly fed with environmental sensor measurements and is in contact with real-world data and events. This level is responsible for feeding the upper-level perception modules with environmental states and information. Thus operational perception module is a vital component of any autonomous vehicle's system design because its output becomes the only reference for the two higher perception modules. Precision and robust algorithms and methodologies must be used tackling problems in this level. Therefore, DaTMO might be assumed to be one of the most vital tasks that an ALV must accomplish. Fundamentals and challenges of vehicle perception and DaTMO is the core concept of this thesis. Therefore the complete and detailed discussion of relating issues will be brought up in the second chapter.

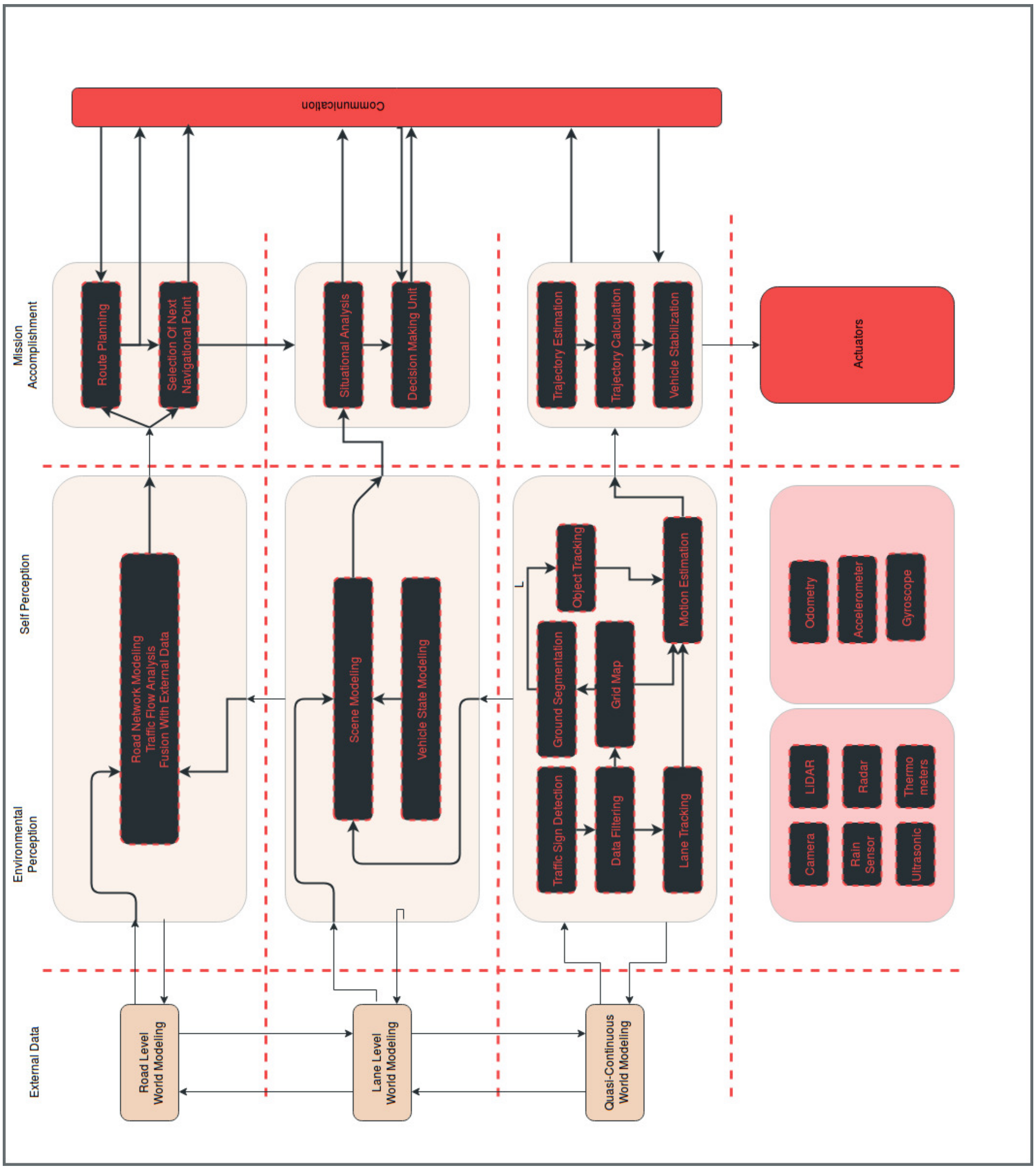

**Figure 1.3:** Detailed system architecture for autonomous land vehicles.

# Chapter 2

# Autonomous Vehicle Perception: Fundamentals and Challenges

## 2.1 Introduction

In the previous section the vital role of the perception module in an autonomous vehicle has been briefly discussed. In this chapter, main challenges toward successful employment of urban vehicle perception is discussed and state-of-the-art methods of urban autonomous driving perception is introduced. There are many important challenges which may be addressed in urban autonomous driving.

First, in urban driving scenarios, traffic participants must be classified and their future trajectory must be estimated regarding to safety issues and collision avoidance. These traffic participants may obey different motion models and are prone to change their behavior in a shortest time frame. This will arise a crucial need to overcome the uncertainty in traffic objects behavior because the vehicle is expected to truly realize future intentions of surrounding objects. Second, in urban scenarios there are more sensitive traffic objects in comparison to highways. Therefore, Vulnerable Road Users (VRUs) must be treated carefully and with especial methodologies. For instance the reference [20] depicts that although perception tasks like dynamic object recognition [1] may be done without an explicit representation of ground surface, the ground plane estimation becomes a key-task in urban environment modeling for reliable perception of the vehicle's surrounding objects. Third, and the most important issue in urban driving scenes is that significant variation of structures within the vehicle's surrounding environment makes the process of separating relevant objects more complex [16].

Tesla Motor, has been conducting many researches in autonomous driving field. Introduction of Tesla Motor's semi autonomous driving technology, called

[1]Recognition task includes both detection and tracking

Tesla Autopilot [2] changed the way that human can navigate and transport by incorporating state-of-the-art artificial intelligence [3] and hardware technology. Tesla Autopilot showed the possibility of utilizing environmental data to gain partial automation in highway driving scenarios [21]. However, Tesla's Autopilot is not trivially applicable in urban situations [2] because in urban environments, vehicle must tackle many complex perception issues to be able to act safely.

There are two possible approaches of vehicle perception. In the first approach *a priori* knowledge is utilized for vehicle perception while in the other, environmental perception must rely on the real-time data coming from on-board sensors. **perceptive-driven** method is more suitable for urban driving scenario due to the changing nature of the urban environments as the ***a priori*** information is not able to adjust itself with changing characteristics of the urban environment in real-time. Two different sensor modalities are used for autonomous vehicle perception: vision sensors and compact laser range finders are mostly used. While camera and RADAR sensors fail to do so, compact three dimensional rotating LIDAR scanners are especially suitable for the autonomous driving purpose as they are capable of collecting far-reaching high fidelity three dimensional surrounding spatial data.

While a human driver may easily assign a semantic classification to the visual perceptions coming from the outside world and driving the the car at the same time, with the process being nearly without error, performing this task is relatively impossible for an artificial intelligence unit with current state-of-the art technology. Without a priori knowledge the *machine* can give us precise location and geometric features of a traffic object, but it is not capable of giving us exact semantic clues about that object. Object classification or the so-called semantic classification of traffic participants is one of the primary tasks that an autonomous agent must be capable of doing. Geometrical and dynamic behavior clues might be used in order to perform semantic classification.

The inherent complexity of environmental perception in urban driving scenes, gives rise to the necessity of introducing uncertainty in modeling process. The lack of knowledge for semantic context deduction and incomplete measurements are major source of uncertainty in environmental perception. Sensor-related uncertainties are uncertainties which occur in the measured physical variables such as position, size and speed. Furthermore, uncertainties might be derived from false detections. Measurements regarding to a detected object might be erroneous and vehicle cannot be sure of existence of the detected object. Bayesian filtering and probabilistic robotics are used to address uncertainties in this field.

The first step for reliable environmental perception is the reliable detection procedure to locate potent location of traffic objects. After reliable detection has been performed, reliable semantic classification must be employed to enable tactical level decision making. Furthermore, reliable tracking of movable objects is of great importance as the autonomous vehicle must be capable of forecasting future intentions of other traffic participants. Semantic classification plays a very important part in this level because reliable semantic clues about particular object, enables us to predict its motion model more precisely. For instance, motion model regarding pedestrians in urban scene is different from car-like object.

[2] https://www.tesla.com/autopilot

[3] AI

## 2.2 Dynamic Object Recognition

Dynamic object recognition is the most important part of an environmental perception unit in any autonomous vehicle which is tending to perform safely in urban environments. In urban scenarios there are almost always multiple objects surrounding the vehicle, thus the problem of dynamic object recognition is actually a Multiple Object Recognition (MOR) problem. Effectiveness and accuracy of methods in this field are highly dependent on how the uncertainties are handled in the procedure, and in the operational level what combination of sensors are used to percept the environment. Most MOR methodologies in the literature are based on tacking-by-detection (TBD) procedure. Potential movable objects are detected using data provided by conventional sensors and position and velocity of dynamic objects are tracked afterwards [22, 23, 24, 25, 26, 27, 28, 29].

Continuous awareness of the kinematic states of the surrounding traffic participants are vital for modeling the perceived environment and furthermore for control actions and safety perseverance. This knowledge has to be real-time adjustable to be employable within reasonable time frames. For instance, movement of pedestrians are fast and somehow unpredictable. If a pedestrian is crossing by the vehicle, appropriate control action for collision avoidance must be adjustable on-line in order to react at the proper moment and real-time to guarantee the sufficient execution time.

Dynamic object recognition is consisted of different modules. The process starts with data gathering procedure from conventional sensors implemented on the vehicle for the sake of our thesis, and maybe other external sources in other different scenarios. Environmental raw data comes into the module, which is constructed of data regarding to different static and dynamic road participants, traffic signs, buildings and etc. in this data entry phase, occlusion detection can be performed if a segmentation method is already implemented in the raw data. Detection and tracking procedures are performed at the data after refinery steps data have been filtered through.

In the detection step, data is refined to not include outliers and the unnecessary data like ground points are excluded from main data frame. Furthermore, semantic information about the data is produced and the final result outputs semantic and kinematic information of different road participants in the scene. In the tracking suit, different estimations and sensor fusion methods are used to track the road participants. Tracker will hold all the information about all the road participants in a data storage to be sure nothing happens unexpected.

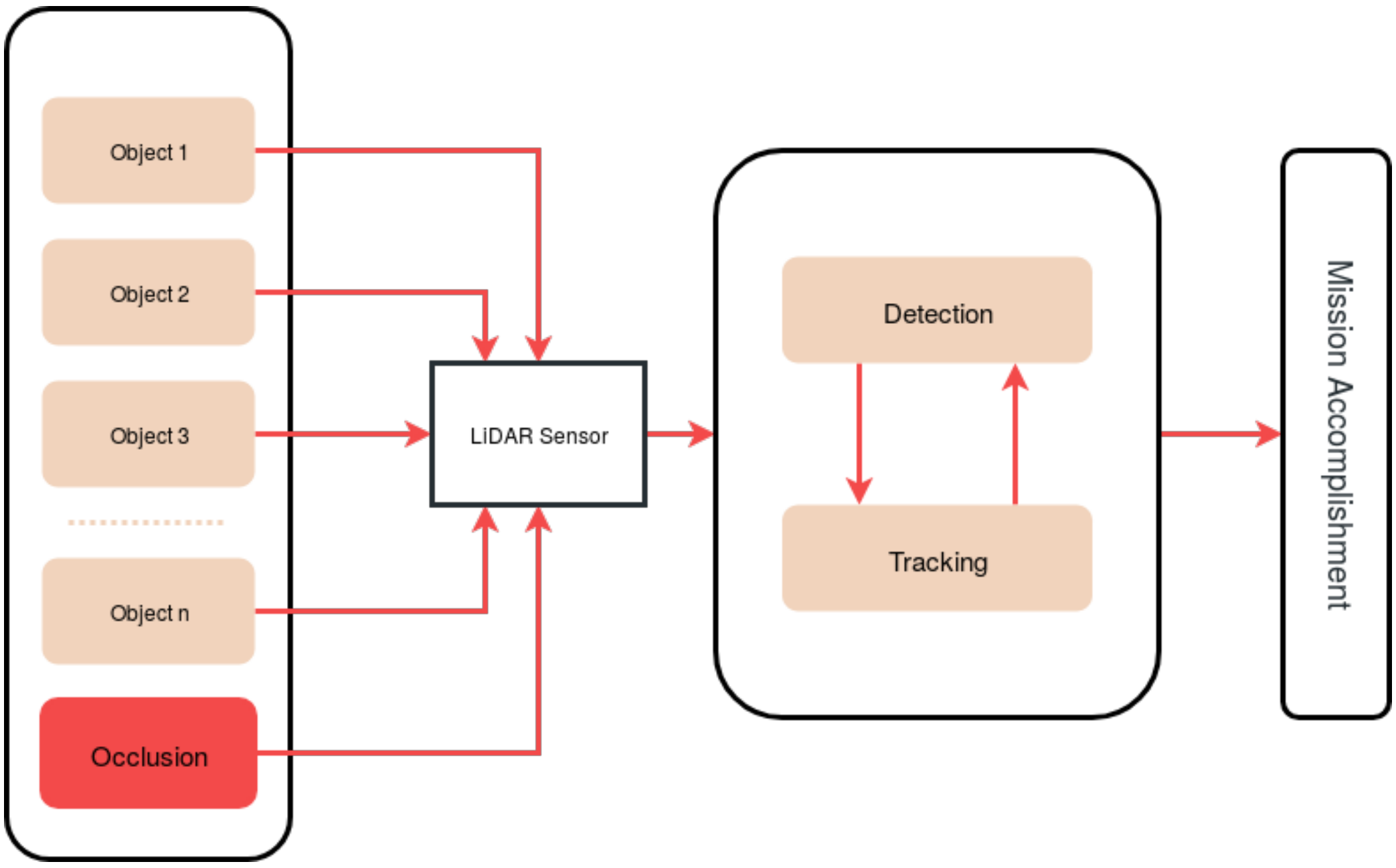

**Figure 2.1:** A schematic of how the multi-object detection and tracking works.

Early efforts toward Detection And Tracking of Multiple Maneuvering Targets[4] have been focused on the problem of tracking independent, disjoint point-wise targets. It was soon realized that there are main challenges to tackle this problem, the most important of them to be the procedure which lies in the process of obtaining the correct association between noisy measurements and object tracks [30]. Also, a challenge arise while in applications like autonomous driving, moving targets are usually deeply interacted with the background seen, thus their measurement is damaged with significant background clutter. Also, the appearance of the target objects is constantly changing due to the complex motion behavior of the ego-vehicle and objects related to each other.

Another major challenge regarding the problem of DAToMMT has got it's roots in the fact that all of the observations in this scenario are made relative to a *moving sensor*[5] which causes additional complexity for perception issues as static objects in the scene may seem dynamic due to occlusion and noise issues. Therefore, the problem of jointly detecting and tracking of all road participants seems not to be easy to implement in real world applications and must be treated with special care.

[4]DAToMMT
[5]Observing-While-Moving Scenario for Perception

## 2.3 MODaT fundamentals

The object perception problem is consisted of three main tasks: *segmentation* or *classification*, *detection* and *tracking* [31]. In the rest of this chapter, fundamentals of these different processes are discussed in details and different methods that are presented in different state-of-the-art publications are discussed. If the data is assumed to be preprocessed, the segmentation and classification is assumed to be done before the main **MODaT** procedure begins. Thus, multiple object recognition is performed on two different main stages: ***detection*** and ***tracking*** which is called Multi-Object Detection and Tracking for this reason. An overview of the procedure is depicted in The Figure 2.1.

In this thesis on the detection stage, raw three-dimensional data is segmented to distinguish static background from movable objects. Movable objects are the ones that are prone to move however they are currently static. Semantic classification enables labeling of these objects, for example a static car is different from a static tree. Furthermore, segmentation and successive semantic classification enables differentiation between current dynamic and static objects. After segmentation, object's poses are estimated due to their respective motion model.

Segmentation paves the way for object detection. For a complete object detection, meaningful attributes must be extracted form raw data regarding to the context and the aim of the method. In other words, detected objects must be classified semantically as the detected object may not *be static* or even may not *stay static*. Thus, dynamical behavior of detected objects are of great importance. Therefore, detected objects will be fed into a ***tracking*** module which enables the estimation of dynamical evolution of those objects in the surrounding environment. A motion model is then assigned to each particular dynamic object and regarding that motion model, kinematic attributes of the object is predicted in different time instances.

The detection process is an integrated part of the object tracking process. Thus, any fault in the detection process will adversely effect the tracking result. Occlusion related issues in raw measurement, uncertainties in measurement process, pedestrian blocking object, improper guess of orientation of objects and other issues may cause imperfect observation and thus imperfect geometric informations. In urban scenarios, sensor occlusions are not avoidable and often occurring multiple time, thus a method should be foreseen to take this occlusions into account.

The tracker is also a vital part of every MOR module, whereas a great amount of uncertainty is involved in the tracking process. The tracker algorithm tends to effectively estimate possible evolutions of each object which are suffering from modeling and measurement uncertainties. Optimal Bayesian filters such as Unscented Kalman Filter(UKF) or other probabilistic filter methodologies like Particle Filter (PF) may be utilized to handle the uncertainty.

As the vehicle continues its path toward different objects in the environment, detector algorithm will detect new objects in each incoming frame of data. Whether these new detected objects belong to previous detected ones, currently evolving in space or being new objects or just an occlusion error, the machine will not be able to tell the difference. In multi-target object tracking scenario, multiple objects with multiple trajectories might be found. The *Data Association* (DA) process is designed to handle this situation. DA assigns newly detected objects into previous tracks with an already established state filtering procedure, or starts a new tracking filter for them based on their motion cues. Data association tends to keep unique identity of the tracks while ensuring no dynamic object will be neglected during the procedure.

After all of these steps, spurious tracks might be found in the result. Existing tracks might be found that are not realistic or their estimation might not be realistic enough due to many major possible errors. Tracks might exist that their motion model is not assigned fairly due to the uncertainties in classification procedure. Thresholding in a *track management* module may help the machine to handle these issues. For example a threshold will be assigned as an existence measure. A certain track is assumed to exist if its track hypothesis is associated to a highly correlated measurement at least up to 90%.

## 2.4 Which One First, *Detection* or *Tracking*?

MODaT may be assumed to be performed due to two completely different paradigms in the literature: *tracking-by-detection* or *detection-by-tracking*.

**Tracking-By-Detection (TBD) Methods:** Many state-of-the-art multi-object detection and tracking algorithms are based on tracking-by-detection paradigm. Tracking-by-detection paradigm (TBD) is a two-step approach which is often assumed to be part of a larger traffic scene understanding framework. Furthermore, TBD often uses simple data association techniques [32]. Moving objects are detected first in each frame, then the trajectories are reconstructed for the plausible candidates.

Object detection procedure is based on discriminative feature extraction methods. These features are either extracted from pixels in vision-based methods or segments of point clouds generated from segmentation methods in LiDAR-based methods. Tracking targets are then classified in order to construct different clusters of traffic participants [33]. Vision-based multi-object detection and tracking methods are often based on tracking-by-detection framework.

A two step approach is used: In the first step, the algorithm detects objects and tries to link overlapping targets with similar measured features in successive frames into each other to form an individual track. At the second step, target tracks are managed and up to a predefined number of frames are tracked and maintained. Tracking-by-detection framework formulates tracking as a *target association* problem. Targets from each input frame are linked by tracker algorithms based on their similar appearance and motion cues to form long tracks [34].

Often detection results suffer from occlusion issues and miss-classifications. Furthermore, detection accuracy will effect whole-algorithm performance. TBD methods may be categorized into *on-line* and *off-line* tracking methods. Off-line MODaT methods utilize target detection results from all data frames in the past and future. In this methods, trackers are generated with regard to some successive frames and then iteratively associated to construct a long trajectories of objects in the entire time-sliding window. On the other hand, on-line methods tend to estimate object's trajectories based on current and past data which makes them more applicable for real-time applications e.g. robot navigation systems or autonomous vehicles [35].

In the on-line multi-object detection, the major challenge is how to associate noisy object detections in the current data frame (vision or LiDAR) with previously detected objects. This is the basis for any Data Association algorithm (DA): To find a similarity function between object detections and targets. Ambiguities in the association process is handled using different cues such as appearance, motion and location informations [36].

Object appearances are used as an important cues for data association which solves three vital assignment problems: *detections-to-detections*, *detections-to-trackers*, *trackers-to-trackers* [35]. Appearance cues seems to be not sufficient in multi-object detection and tracking scenarios especially in urban driving scenes because traffic participants are often consisted of many similar objects like pedestrians, car-like objects and bicycles. Thus, motion cues are often used for data association. Data association methods are discussed in details in proceeding sections.

**Detection-By-Tracking (DBT) Methods:** In [37], a method is proposed for detection and tracking of multiple vehicles using headlight and taillight blobs. The MODaT problem follows a detection-by-tracking approach in this article. The tracking problem is formulated as a *maximum a posteriori* problem [38]. Reference [39] Combines two detection and tracking different scheme in a detection-by-tracking framework to reduce false positive results of vision-based people detection and tracking in urban environments. The approximate articulation of each person is detected at every frame based on local features that model the appearance of individual body parts. Prior knowledge on possible articulations and temporal coherency within a walking cycle are modeled using a hierarchical Gaussian process latent variable model (hGPLVM) [39].

## 2.5 Detection Fundamentals

### 2.5.1 Overview

The detection problem is defined as the procedure of recognizing the presence of different objects in the environment. These detected objects are called *targets* from now on. Detection must result in information about unambiguous identity and state information of the targets because according to the tracking-by-detection framework, tracking results are highly dependent on the results from detection procedure [40]. Main challenges in object detection is the uncertainty of sensor observations and inevitable occlusion of scanned objects. Occlusion and self-occlusion issues may cause disappearance of detected target in some data frames or even may cause partial blockage of the object, making the algorithm to recognize that particular object as two different ones [22, 36, 41, 42, 32, 43, 44].

Different perception sensors are also available for these tasks. While many methods use single-sensor modalities for perception task, these methods have been showing their flaws since their advent. For instance, although RADAR provides useful information for detection and tracking of traffic participants, it does not provide good information for the classification task. Furthermore, RADAR may experience difficulties to detect non-rigid bodies in the environment like pedestrians [31].

Object detection using 3D LiDAR sensor is sequentially divided into two major steps: *pre-processing of raw data* and *estimation*. Pre-processing of raw LiDAR data must be done due to segmentation and classification. Furthermore in the estimation step, pose of target objects are obtained. According to many state-of-the-art methods in the literature, segmentation step is consisted of two different process: *ground removal* and *clustering*. The results of the ground removal algorithm are called *elevated point cloud* which is fed into clustering algorithm to be further used in pose estimation step [45, 46, 47, 16, 48, 49, 44, 50, 24, 51, 47, 52]. Schematic of the detection process based on common approaches in the literature is depicted in Figure 2.1.

This chapter will be structured as follows: First, different sensor modalities are discussed along with their advantages and limitations in MODaT applications.

Second, different segmentation and pose estimation methods available in the literature are discussed and most appropriate methods for multiple-object detection and tracking in urban environments are chosen to build the basis for our implementation.

### 2.5.2 Spatial Data Acquisition

Various types of sensors may be used for spatial data acquisition like conventional and stereo cameras, RADAR and recently LiDAR (**Li**ght **D**etection **A**nd **R**anging) sensors. In computer vision literature, object detection using vision sensors is a well-researched topic[53]. Cameras are affordable and more common compared to LiDARs while providing rich data with high resolution pictures. Different sensor modality is also available in vision-based MODaT methods as some methods tend to detect only frontal or rear object while some others utilize surround cameras. The main challenge in vision-based MODaT method for ALVs is the unstructured environment in which an autonomous vehicle is operating. Road environments may express many type of variations in illumination, different kind of complexities in background and main scene [52, 38, 54, 55, 56, 57].

Often for accurate real-time 3D tracking results, stereo-camera is used or surround structure is necessary. Computational processes are often performed onboard. The vehicle is embedded with a computer which has to be robust and cost-effective rather than being fast and precise. Surround-stereo-cameras imply processing of 4 combined images along with their disparity maps which effectively increases the computational cost of procedure. Detection and tracking of cluttered, far away, blurred objects demands high-resolution pictures which again increases the computational complexity of the procedure. While urban autonomous driving task needs an accurate, surrounding objects and environmental awareness, vision-based methods have to struggle to provide a method focusing on frontal, rear or side view object detection and tracking at the same time.

Recently, 3D LiDAR sensor have been introduced with the ability of acquiring massive three dimensional geometrical information. LiDAR sensors are fast and real-time (10Hz). They are also able to provide additional information about the environment like compactness and reflectance of the measured surface. In comparison to their vision-based counter parts, LiDAR sensors are more robust to environmental variations of illumination and has got wider range of view while suffering from low resolution measurements and limited angular resolution. 3D LiDAR also has limitations in adverse weather conditions and obtaining color information [58].

Classification of different objects must be performed based on dimension and geometric information available from LiDAR sensors and attributes like colors and texture which are available in vision-bases methods are not available with Li-DAR, had they done, object classification would be much more easier especially in complex scenes with e.g. occlusions. Often model-based object tracking scenarios prefer to use camera as their main sensor. Vision-based methods are also the best option for driving assist systems such as lane detectors and traffic sign detections.

In this thesis, accurate, long-range, surround spatial data gathered by a 3D LiDAR sensor is used in order to achieve a multiple-object tracking method in urban environments. Although camera is able to produce accurate and surround result, it is much more complex and computational costly due to setup requirements and the need for real-time processing of high-resolution stereoscopic data. LiDAR sensors often show comparable results with much lower computational effort needed for object tracking. On the other hand, LiDAR comes with some shortcomings.

Reconstruction of occlusion is more complex with LiDAR data because no attributes other than three-dimensional range, reflectance and shape information is available and no data can be derived from the target object in the case of occlusion. Disadvantages of using each particular sensor may be compensated by using different fusion-based structures like RADAR-LiDAR, Camera-LiDAR and etc. Fusion of camera an LiDAR adds color and other attributes to the LiDAR-based methods and results is more accurate tracks while, computational budget of the underlying structure must be considered in order to the system stays in real-time framework.

### 2.5.3 3D Laser Scanner: Velodyne HDL-64

The 3D Velodyne HDL-64 was originally designed for the use in DARPA Grand Challenge – a competition sponsored by the United States Department of Defense to support technological innovations in the field of unmanned ground vehicles. LiDAR is actually used for the measurement of the distance between a host vehicle and an observed object.

3D LiDAR has many advantages comparing to other vision sensors: First, it is capable of performing in any weather conditions and during the day and night; second its data is easier to segment; third it has a uniform measurement scale regardless of the distance. A Velodyne HDL-64E sensor is a High Definition LiDAR (HDL) sensor which is capable of capturing over 1.3 million points per second [59]. HDL-64E provides $360°$ horizontal field of view and $26.8°$ vertical field of view. LiDAR sensor produces laser pulses to take measurements of the environment and generates a three-dimensional map. 3D LiDAR measurement has a lower resolution than cameras. As LiDAR data is only consisted of a collection of points in 3D environments and reflectance information, plenty of CPU time remains untouched to be used by higher-level logics like classification and tracking algorithm.

**Sensor Technology** Automotive LiDAR sensors typically use the *time of flight distance measurement scheme*. After emission of a laser pulse, an internal sensor in the instrument measures the amount of time it takes for the pulse to bounce back to the sensor. the time elapsed between emission of the light beam and detection of the reflected beam is directly proportional to the distance of the vehicle to the object:

$$Distance = \frac{Speed\ of\ Light \times Time\ of\ Travel}{2} \tag{2.1}$$

Other attributes like reflectance is obtained from intensity of returning light. A high-definition LiDAR sensor with a rotating sensor head utilizes 64 semiconductor laser which enables it to generate 3D environments map at unprecedented levels of detail. LiDAR is similar to RADAR except that the sensor sends out and receives pulses of light in visible electromagnetic spectrum, ultraviolet or infrared instead of radio waves. This sensor uses a rotating head featuring 64 semiconductor laser,each firing up to twenty-thousand times per second. Each of the 64 lasers has its own dedicated detector.

Furthermore, each laser-detector pair is precisely aligned at predetermined vertical angle to produce the exact vertical field of view. HDL-64 is capable of providing 64 layers of LiDAR measurement which jointly constructs a 3D spatial representation of surrounding environment. Each layer of points are generated by the single rotation of a emitter-detector pair. These 2D layers then sums up to a 3D scan. This directly available three-dimensional spatial data along with the inherent immunity of the LiDAR against common environmental attenuations in urban scene make it more suitable for MODaT in urban areas.

### 2.5.4 Data Structure: 3D Point Cloud

Resulting points from each measurements of the 3D LiDAR sensor are accumulated in a frame with the help of Point Cloud Library representation (PCL) [60]. Point Cloud Library representation may be assumed to be a convenient container to store, process and visualize raw measurements of the 3D scanner [2]. PCL is a modern C++ library for 3D point cloud representation and processing. 3D point clouds are often consisted of *xyz-coordinates* of the measured points from each surface. Other attributes like RGB color or in the case of LiDAR reflectance Intensity can be added to the point cloud. Therefore, each point has information about its location regarding to the measurement origin along with other attributes:

$$p_i = \{x_i, z_i, x_z, R_i, G_i, B_i, I_i, ...\} \tag{2.2}$$

Then each point cloud frame is represented as a collection of its points:

$$P = p_1, p_2, p_3, ..., p_i, ..., p_n \tag{2.3}$$

Resulting point cloud data may be used and redistributed in any part of ALV's algorithm directly. Furthermore, many publicly available LiDAR point cloud datasets are available which enables doing research without physical presence of the sensor. On the other hand, publicly available data set enables cross-validation of different methods with identical data sets. One of these datasets is KITTI dataset[61] which provides the recording of various sensors in different urban scenarios. The KITTI data set is recorded from a moving platform while driving in different environments. Recordings include camera images, laser scans, high-precision GPS measurements and IMU rates. The recording captures real-world traffic situations ranging from rural areas to inner city and highways. Static and dynamic objects are labeled by hand which can be used as the ground truth for validation purposes. The LiDAR sensor used in the dataset is the Velodyne 64-E.

### 2.5.5 Point Cloud Segmentation

Each frame of point cloud data can be consisted of more than 1.3 million points. Thus, a large amount of point cloud data in a driving procedure which is consisted of many frames, demands a high computational effort to process. On the other hand, point cloud data is discontinuous and consisted of geometric information of points only. Semantic classification of these geometrical points into different meaningful groups of objects would be very beneficial for driving tasks. Before feeding this raw data into the detection process, a segmentation procedure must be performed on the raw data.

Segmentation process reduces the dimensionality of the raw data, filters outliers and enable the detection process to differentiate between non-important non-traceable objects and objects of interest for higher level logics like detection and tracking. Two major segmentation method may be found in the literature: Grid based method and Object-based method. The underlying tracking scheme declares the need of use of each method. The grid-based segmentation method is more suitable for detection-by-tracking schemes and object-based methods are used mostly for tracking-by-detection schemes. Exceptions may exist in the literature, for instance [1] uses grid-based segmentation approach in the preprocessing step.

**Grid-based Segmentation:** The grid based segmentation method is based on global occupancy grid maps. A grid cell network will be constructed and the probability of existence of an object in each grid cell is calculated. If the probability calculation shows the existence of an object in cell, the cell is called occupied. Often occupancy probabilities are updated with the use of Bayesian Occupancy Filter [62] while velocities may be obtained using Particle Filters [63, 64].

Grid-based segmentation process often results in simpler detection and data association procedures. Grid-based segmentation methods highly relies on occlusion handling procedures because if a moving object can not be detected due to occlusion, related cells will be mapped as static. Therefore, grid-based segmentation methods have limited ability for dynamic object representation and manipulations [65]. Since in urban environment situations, a detailed representation of dynamic objects is required for safe driving and environmental perception is highly dependent on a precise model of the complex surrounding area, grid-bases methods are not suitable for these tasks. Thus, in this thesis most of our attention is focused on object-based segmentation method.

**Object-based Segmentation:** Point-wise segmentation approaches are utilized in object-based segmentation scheme to describe different collections of the points. In this scheme, a separate pose estimation and tracking filter is needed to obtain dynamic characteristics of the segmented objects. Object-based segmentation methods are often consisted of two different steps: Ground Extraction and Object Clustering. Ground extraction is performed in order to exclude navigable or non-obstacle group of points from the point cloud. Ground extraction is followed by clustering which not only reduces the dimensionality of tracked objects but in come cases classifies them semantically.

**Ground Extraction** Ground extraction is an important preliminary step in object detection process. A point cloud coming from raw measurements of a 3D LiDAR data may be divided into two different kind of points: navigable or non-obstacle points and obstacles. In real-world application navigable areas are not restricted to be planar and elevations may be observed among them. Every frame of incoming data includes a huge number of points thus, ground extraction could be a time-consuming procedure without implementing special care.

Ground extraction methods may be divided into two different methods: grid-based and scan-based methods. In grid-based methods, ground extraction is performed by dividing the data into polar-coordinate-based cells. Then, height and radial distance of adjacent grid cells are analyzed with gradient-based methods to help the deduction of obstacles: if slope value between two adjacent cell exceeds a predefined threshold an obstacle is detected [19, 48]. In the scan-based methods often tends to extract a planar ground derived from a specific criteria. For instance some approaches take the lowest z value as the candidate for a ground point and apply Random Sample Consensus (RANSAC) fitting algorithm to determine the possible ground[66].

Grid-based approach enjoys the fact that ground contours are preserved in this method and they are able to present flat terrains better. Furthermore, scan-based methods are able to consider values of neighboring channels and thus prevent the method to produce inconsistent results.

**Clustering** After the implementation of ground extraction algorithm, *elevated* point cloud includes a large number of points. If the detection and tracking algorithm has to be implemented on this massive-scale data, the computational cost of the process excludes it from real-time criteria. This massive point cloud data is to be reduced into smaller clusters. Each cluster is consisted of multiple, spatially close-range and semantically coming from same object groups of points. Different kind of clustering schemes can be done either in 3D,2D or even 2.5D in which some z-axis information is retained in the procedure.

While 2D clustering methods are computationally efficient and have been shown to be sufficient for object tracking purposes, special care shall be taken as vertically stacked objects in this method could be considered as a single object and merged into single cluster which can cause problems in urban environments (e.g. pedestrian under a tree). 3D clustering methods will result in high-fidelity object cluster by incorporating the vertical axis features. Reference [67] proposed a 3D clustering method based on Radially Bounded Nearest Neighbor (RNN). Reference [68] also uses DBSCAN (Density-Based Spatial Clustering of Applications with Noise) to employ full 3D clustering. Considering real-time requirement and significant number of points involved, 2 or 2.5D clustering is more preferred [50] owing to the fact vehicle on-board computer likely to have limited computational power[2].

### 2.5.6 Pose Estimation

The first step of the detection process is to recognize the existence of object of interests in the scene which is done in segmentation process. After recognizing these objects, useful informations of them must be fed into tracking algorithm for further estimation issues. In order to extract usable information about recognized objects, pose estimation must be done to obtain object trajectory and orientation data. Object *pose* include the dimension, orientation, velocity and acceleration of the object. Pose estimation methods can be divided into model-based or feature-based methods. Model-based pose estimation tend to match a known geometrical model into the raw measurement data. On the other hand, feature-based methods tend to deduce object shapes from extracted features.

**Model-based** Optimization-based iteration is used to fit detected objects into known geometrical models as cuboid or rectangle in model-based pose estimation. In order to fit the clusters of points into geometrical shapes, edge-like features are extracted from the segments and the best fit is choose to assign a model. After parameterizing the geometrical shape, the most probable pose of the clustered object is extracted based on segment points. For instance, reference [69] formulates this problem as the minimization of the polar distance between the measured points and the visible sides to compute the best vehicle pose.

A major challenge in bounding box generation in model-based methods is the estimation of the orientation. Common approaches often calculate minimum area of clustered points to estimate vehicles orientation [70]. However in the presence of partial occlusion the results can be spurious in term of dimension and orientation accuracy. A method using I-, U- or L-like geometric classifiers are used to derive the most suited orientation in reference [20]. Another method to obtain the bounding box is to use convex hulls to estimate Object Oriented Bounding Box (OOBB) which gives the bounding box with the Minimum Area or OOMB[6]. The main idea of this method is to first find the convex hull of the objects points and then fit box to the shape which gives the minimum area bounding box.

Although model-based method gives an optimal pose estimation breakthrough it suffers from the pain point of being to computationally expensive for real-time applications. On the other hand, optimization methods may result a sub-optimal pose depending on the initialization conditions which is actually a local minimum but not the minimum of the function. To tackle the problem of computational cost and sub-optimal answers we use another method to fit a bounding box on our moving object candidates while estimating a fair orientation and pose for the object which is discussed in details in implementation section.

**Feature-based** Feature-based pose estimation methods use the edge features of objects point cloud to estimate the objects dimension and other attributes. For example, Himmelsbach et. al. [1] used RANSAC[7] algorithm to fit the dominant line of the points with the orientation of the object. Also Luo et. al. [53] uses a graph-based method to fit the point cloud scans into the predefined arbitrary shapes, however this method does not provide orientation estimation clues. Darms et. al. uses raw measurement to extract edge features to fit to a box model to finally represent the vehicle‘s pose[23]. Another approach is to first find all the points facing the sensor with the smallest radius among the others in the same observation azimuth. After extracting all of these points a L-shape polyline is fitted into these points using iteration endpoint algorithm[71]. A similar method is to find the edge points in segmented point cloud clusters to finally interfere the possible corner points[2].

[6]Object Oriented Minimum Box

[7]RANdom SAmple Consensus

In other scenario, positive training samples obtained from multiple viewpoints of the object to train the detector to find three-dimensional Harr-like features from clustered points are implemented using machine learning AdaBoost tool [72]. After the learning process is finished, trained detector is used to generalize a voxel-based box for each target. Also R-CNN[8] method is used in an sensor-fusion approach to estimate object orientation based on the joint proposal of stereo camera and LiDAR data.

Looking at examples given above, we realize that feature-based object detection methods offer both good computational time and pose estimation results, although these methods are highly dependent to the quality of the measurement. If the measurement process is unstable, the method is not going to address a well enough result for the object detection.

**Model-free** In the so-called model-free approach the detection process is based on motion cues. The main advantage of detecting dynamic objects based on motion cues is that no restriction is placed over the shape, size or color of the object. Furthermore, using model-free method needs no semantic information about the detected object. Actually movable objects are detected regardless of being a car, person, bicycle or etc. The main drawback of not using semantic informations is that potential movable objects may not be known ***a priori***.

A vast majority of examples using model-free methods were actually first deployed in the DARPA Urban Challenge. These methods usually tend to segment the laser range finder's raw measurement at the beginning, and then tend to extract geometrical features from the segmented data. Output of these processes are a number of moving-object candidates, and dynamic objects are extracted from them as objects with certain size, shape and kinematic features [16].

Toyota's tracking system [73, 74, 75] estimates a static representation of the environment alongside the detection and tracking of moving objects. These method actually combine SLAM[9] with dynamic object tracking methods. All of these methods take an occupancy grid representation of the environment then use the knowledge of occupancy probabilities from their map to propose locations of likely moving objects. In reference [76] an occupancy grid map is utilized to detect non-stationary objects by differencing the maps and making use of Expectation-Maximization (EM) algorithm. In reference [77] two occupancy grid maps are computed separately to take static and dynamic objects into account in parallel to each other, but no tracking method is presented for dynamic objects.

[8]Regional Convolutional Neural Network

[9]Simultaneously Localization And Mapping

### 2.5.7 Research Direction

The detection of dynamic objects in 3D environment is a challenging problem with a high complexity. These complexities arise due to measurement- and sensor-related problems, sparse data, occlusion issues and etc. State-of-the-art methods tackle these sort of complexities by using many simplifying assumptions. For example model-based methods assume dynamic participants of the environment to be a member of defined object classes and though they tend to capture their appearance by their predefined canonical models. Then everything depends on the probability of the perception holding its validity in that environment. If that high probability happen to be the case, the system's performance will be acceptable. There is of course a main drawback with these assumptions: in different environment, different assumptions must be made to handle the specific situation of that environment.

These simplifying assumptions are made for the motion cues in model-free approaches. One of these assumptions is that objects are often assumed to follow motions independent of each other. A major issue regarding these independent motion models is the need for capturing the motion interaction between different road participants, which requires semantic information about each object.

In this thesis, a model-free approach is used to perform the detection task. At the first step, statistical filtering is used to refine the raw input data. The ground segmentation process is applied next. A novel ground segmentation method is introduced which is capable of continuous estimation of the ground surface, thus besides removing planar counterparts without further hesitations, a continuous estimation of the ground surface is obtained which is capable of estimating ground height for every desirable point in the environment. Next, the clustering process is implemented based on nearest neighbor method. The output of the proposed detection module is a list of potent objects to the tracking module along with their size informations, placement and orientation information.

## 2.6 Tracking Fundamentals

Object tracking can be defined as the process of using on-board sensor measurements to determine location, future path and motion cues of arbitrary objects. In Autonomous Land Vehicle's (ALV) framework, tracking procedure is of great importance due to the safety and navigational issues. Trackers must take several attributes into account like velocity, location, pose and even semantic informations of target objects to be useful for navigational purposes. Tracking may be seen as the procedure that finally takes all the information provided by the sensors into actionable knowledge. The obtained knowledge will be used by the ALV to perform it's navigational tasks consciously and with safety. In order to establish a collision-free navigation procedure for outdoor robots, tracking procedure must be taken seriously. Tracking algorithms used for this purpose must be real-time applicable and capable of estimating trajectories of multiple objects with different motion models at the same time. Constant velocity model, constant acceleration model, and constant turning model are the motion models used for for trajectory estimation of multiple objects.

Real-time tracking algorithms often suffer from low accuracy and poor robustness when confronted with difficult, real-world situations [78]. Thus many robotic applications suffer from limited achievements due to cheap and unreliable track estimations. For example an autonomous vehicle performing in urban areas must be aware of pedestrians moving in it's vicinity to avoid hurting them or other moving objects to avoid collision.

Tracking using 3D data has been an important challenge for many years. While the LiDAR sensor is capturing 3D data from environment, the ego-vehicle is moving. Thus the relative position of objects and sensor is changing over the time. In tracking-by-detection scenario, the results of the detection which are often output of a clustering unit, suffer from occlusion and ambiguities. It occurs that clusters belonging to a same object may split into several parts and different segments belonging to different objects may merge into a single cluster. In addition, different contours of moving objects are available due to the fact that a laser scanner only is capable of seeing those parts of the object that are facing the laser been [71].

Traditional depth-based tracking methods often use classic Kalman Filters (KF) and tend to neglect almost all of the available 3D data by representing an object with it's centroid [79] or center of a bounding box [80, 16]. Although these methods are computationally efficient for real-time applications, they are not very accurate. Current state-of-the-art tracking methods gives noisy estimates of the position and velocity of these objects. Furthermore, occlusion and variating view points are major challenges for obtaining accurate tracking estimates. In addition, without robust estimates of the velocity of nearby vehicles, merging onto or off of highways or changing lanes become formidable tasks. Similar issues will be encountered by any robot that must act autonomously in crowded, dynamic and populated environments [78]. Compared to urban areas, where road network information is available for background separation and motion prediction, unstructured environments present a more challenging scenario due to low signal to noise ratio [28].

A reliable tracking algorithm must be capable of recognizing the same object and provide smooth and continuous trajectory of objects. It should also be capable of predicting the motion trends of the moving object. Performing highly intelligent actions, like avoiding collision spontaneously, overtaking other vehicles, and car following, all depends on understanding the motion of objects around the autonomous driving vehicle [71].

At least for the sake of vehicle perception, Multiple Object Tracking (MOT) is the most important active research area in the last decade. In tracking-by-detection scenarios like the one we have in this thesis, while object observations are known for the system, object tracking becomes a Data Association (DA) problem. Thus usually tracking methods in autonomous robotics literature are based on filtering techniques and state-space modeling while involving an association procedure like Probabilistic Data Association or Multiple Hypothesis Tracking (MHT) which is on of mostly used DA methods in the literature [81, 30, 82]. The joint probabilistic data association (JPDA) filter [10], and particle Filter are more efficient and widely used. They use a single frame data and previous tracking results, giving up reversing past ambiguities [71].

The multiple hypothesis tracking (MHT) is a multi-frame tracking method that is capable of handling ambiguities in data association by propagating hypotheses until they can be solved when enough observations are collected [80]. Furthermore, imperfections in detection may cause association errors which will lead to imperfect tracks [81]. Most encountered imperfections in detection stage are objects *merging* and *splitting* which are simply ambiguities that the tracker algorithm can not facilitate good estimation for them. The main drawback of the MHT method is that the number of hypotheses grow exponentially over the time, making it not computationally cost-efficient, because MHT propagates hypotheses until they can be solved with entrance of enough observations which brings great amount of computational burden [71].

The Joint Probabilistic Data Association (JPDA) filter is more efficient but prone to make erroneous decision since only single frame is considered and the association made in the past is not reversible. Other sequential approaches using particle filters share the same weakness that they cannot reverse time back when ambiguities exist [80]. In addition to these methods, recently batch MOT methods have attracted research attentions [83, 84, 80]. These methods are based on Markov Chain Monte Carlo (MCMC) technique which enables an efficient way for finding optimal solution in the entire solution space. For example, reference [83] uses this method to find temporal association between objects. Furthermore in reference [84] the MCMC method is used to solve the data association problem for a spatio-temporal solution space.

### 2.6.1 Tracking Unit Components

The necessary components of the tracking unit is depicted in Figure 2.2. Every new track is started by the entrance of the result list of the detection module. If the entered track exists, the measurement is checked if it is statistically likely to be correct measurement through gating process [2]. Therefore, every track that wants to enter the association process between measurements and tracks and further enter the estimation block, must first be capable of passing the gating block. Several classes of Bayesian modeling are used for different sub-processes of tracking module being estimation, data association or even dynamic modeling.

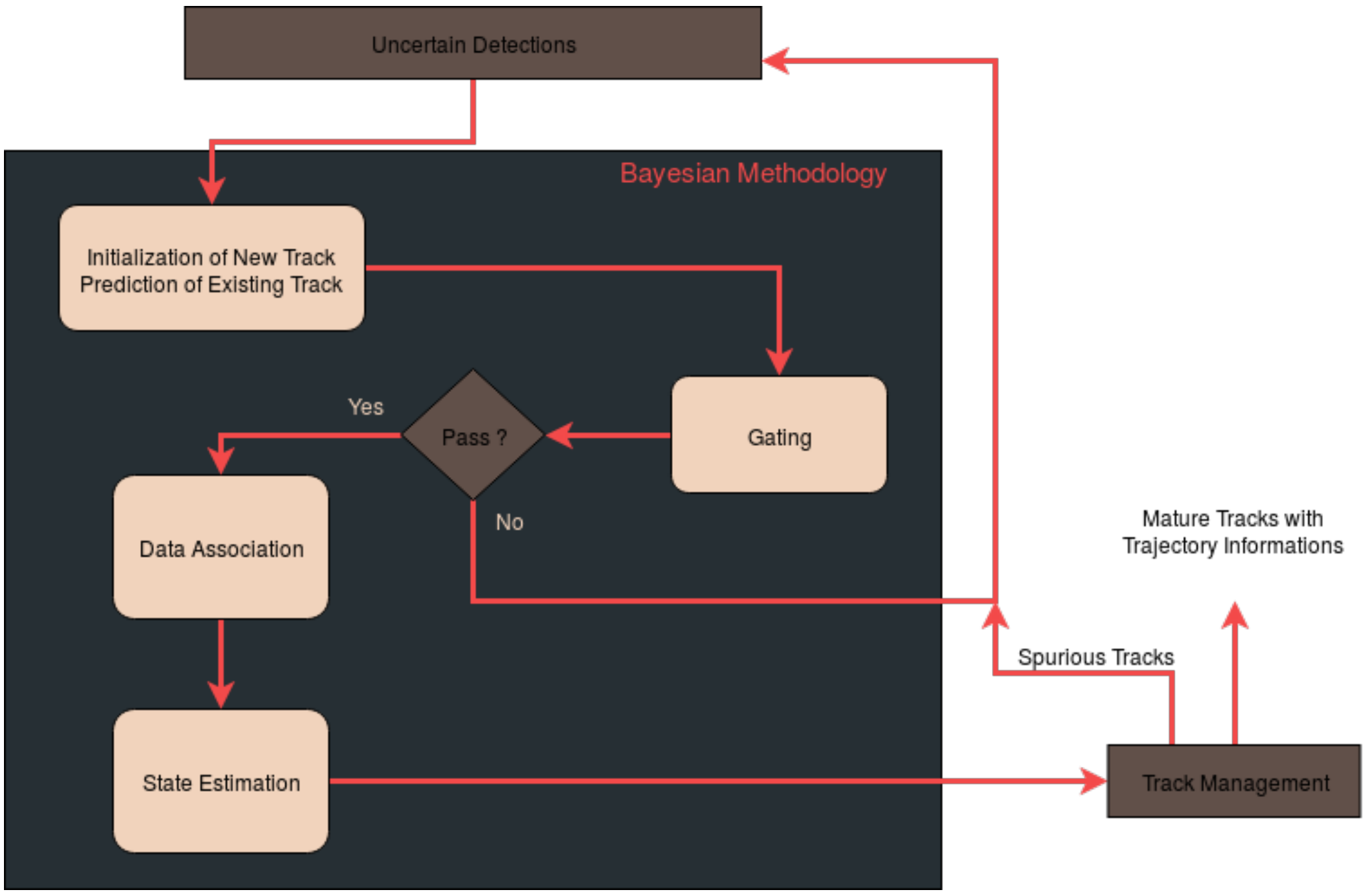

**Figure 2.2:** Tracking unit components. Bayesian framework is used to handle uncertainties in different steps.

### 2.6.2 Sensor and Object Model

The LiDAR sensor is placed on the top of ego-vehicle at a predefined height. The origin of any measurement and observation coming from the environment is the origin of the LiDAR. Like depicted in the figure, ego-vehicle produces rich data in each frame including all the road participants, back ground objects and infrastructure. The LiDAR sensor originally provides measurements in polar coordinates with range and bearing values. Velodyne digital processing unit like what we see in the KITTI data set, already provides these measurements in Cartesian coordinate system as depicted in Figure 2.4.

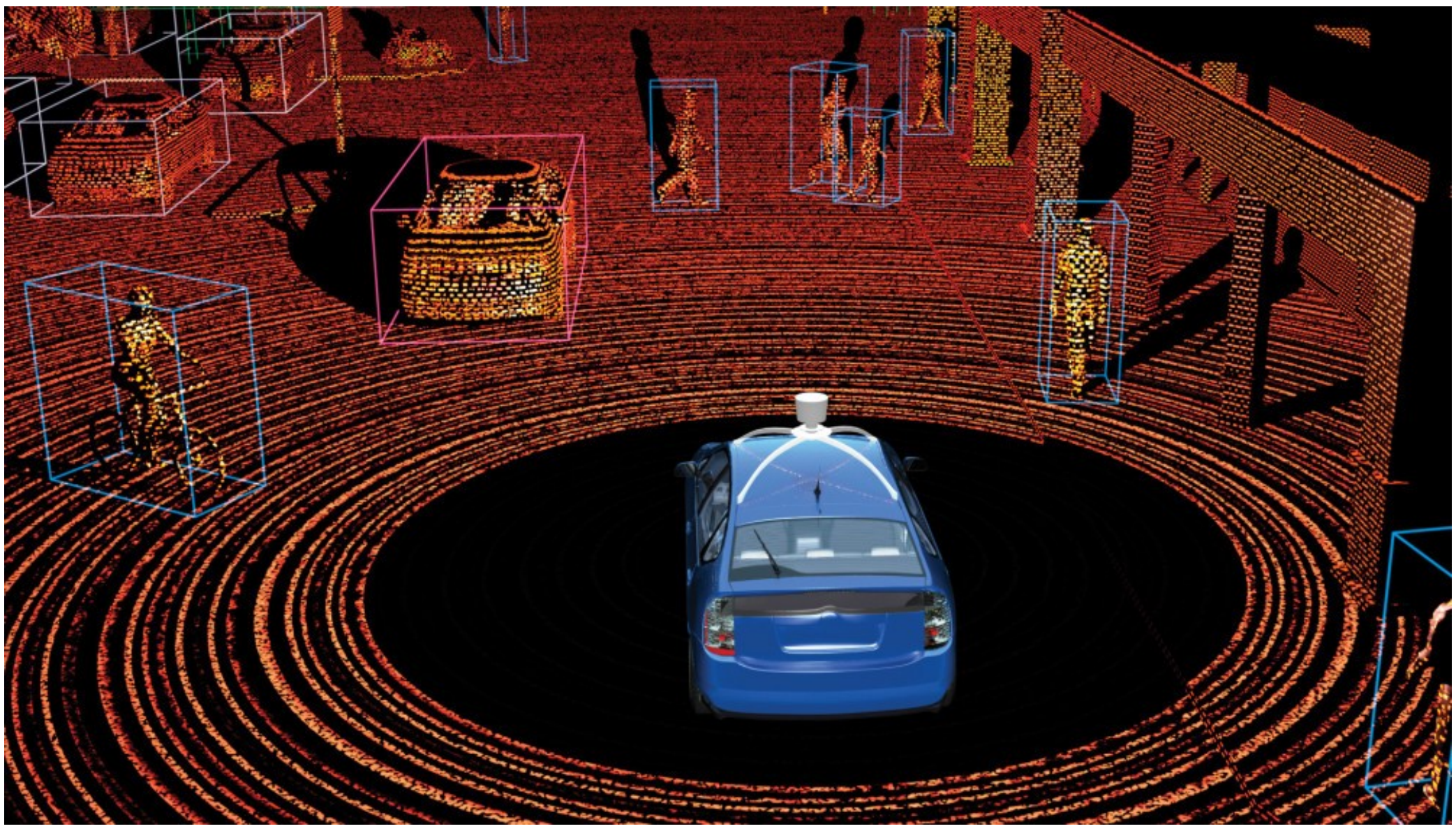

**Figure 2.3:** A time aggregated LiDAR point cloud frame from Uber engineering blog. Different road participants can be seen in the picture with different motion behaviors: bicyclists, pedestrians and cars.

In this thesis these measurements from KITTI data set are used by their Cartesian values [61] to enhance the measurement propagations. Up to this point, potent object clusters are detected in detection stage and already bounding boxes are fitted to each cluster. There are two major approach for object tracking: ***point tracking*** and ***extended object tracking*** which embeds states of the object with geometrical attributes. In this thesis, point tracking is chosen over the other method because LiDAR sensor promises to generate a very big and inconsistent measurement in each frame depending on the vehicle's view point, reflectance of the object and distance to the object which makes shape inference procedure a high-complexity computational effort. Therefore, incorporating objects geometrical at-tributes into estimation filter, makes estimation uncertainty bigger. Thus, although bounding box fitting is implemented to show the extent of each detected object, the geometrical attributes of the box are not going to be updated in the filtering process.

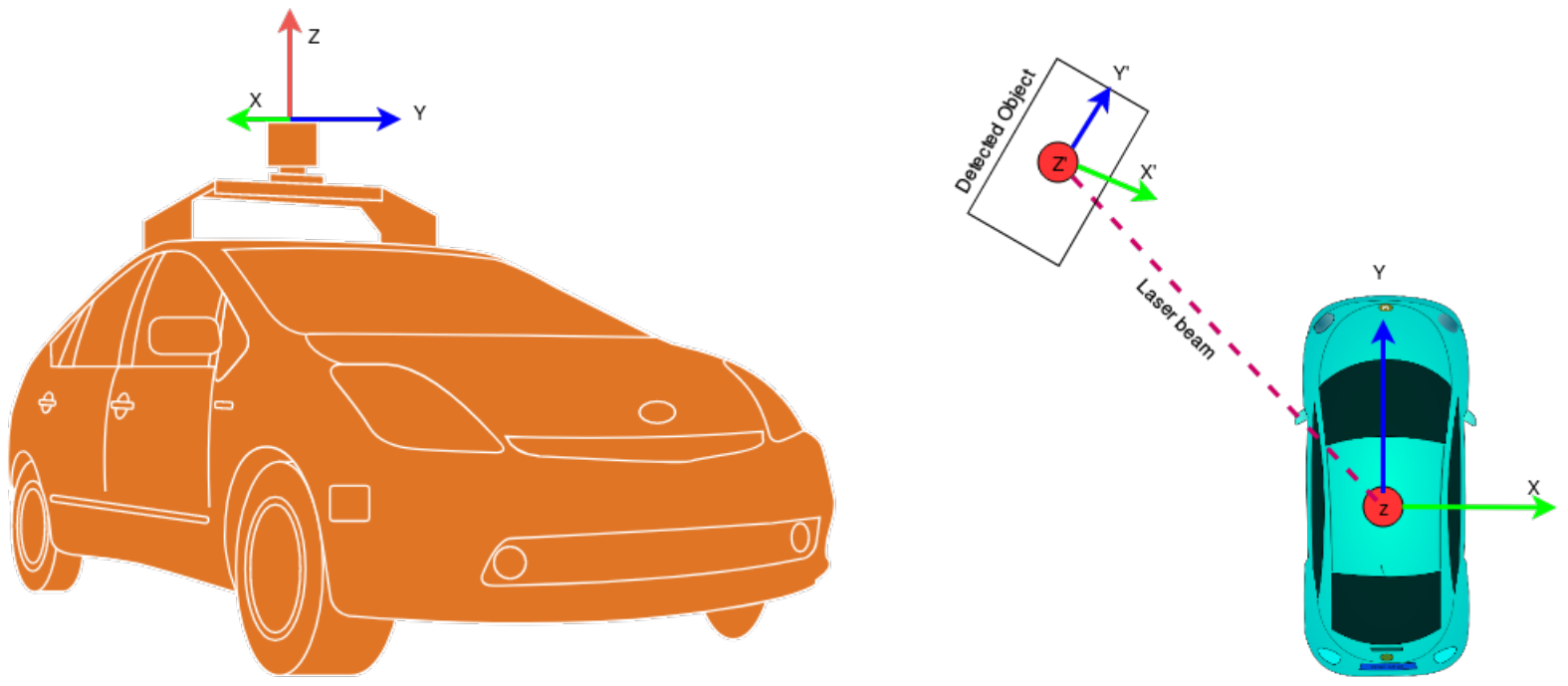

**Figure 2.4:** The sensor's coordinate frame which is placed the height of 1.73 meter in our utilized data set. Also, the measurement beam is shown in the right picture.

Furthermore, we simply assume that potent moving objects don't have any vertical displacement which is rational due to specification of objects moving in road or urban scene. Thus, any kinematic attribute along ***z-axis*** is assumed to be zero and our measurement of position vector **z** which comes in each discrete time instance with each frame, is defined to have values along ***x-*** and ***y-axis***:

$$z = \begin{bmatrix} pos_x \\ pos_y \end{bmatrix} \tag{2.4}$$

A stochastic, discrete-time state-space model is utilized to model the dynamic motion model of each target object:

$$x_{k+1} = f_k(x_k, u_k) + w_k \tag{2.5}$$
$$z_k = h(x_k, u_k) + v_k \tag{2.6}$$

In which $f_k$ is the system function at each time step $k$, $h$ is the measurement function which is assumed to be time-invariant, $x_k \in \mathbb{R}^n$ is the state vector and $z_k \in \mathbb{R}^p$ is the measurement vector. Process and measurement uncertainties are assumed to be mutually independent white Gaussian noises, with zero-mean and with covariance matrices $Q_k$ (for process noise) and $R_k$ (for measurement noise).

The state vector for our interest in dynamic object detection and tracking must be chosen in a fashion that be useful for further navigational and motion planning purposes for driving scenes. Thus, our often state vector includes position in ***x-axis***, position in ***y-axis***, heading angle or yaw, vehicle's speed (magnitude of velocity) and yaw rate. Thus, the state vector becomes:

$$x_k = \begin{bmatrix} pos_k^x \\ pos_k^y \\ \psi_k \\ v_k \\ \dot{\psi}_k \end{bmatrix} \tag{2.7}$$

Where $pos_k^x$ and $pos_k^y$ are positions of the object along the x- and y-axis, $\psi_k$ is the according yaw value, $v_k$ is the according speed value and $\dot{\psi}_k$ is the value of yaw rate at the time step $k$. Furthermore, we take measurement matrix $h$ to be linear, time-invariant:

$$z_k = \begin{bmatrix} 1 & 0 & 0 & 0 & 0 \\ 0 & 1 & 0 & 0 & 0 \end{bmatrix} x_k + v_k \tag{2.8}$$

### 2.6.3 Multiple Motion Models

When it comes to the road objects, they turn out to be more likely to not follow a same motion model to each other and even not a single motion model. For example, dynamical behaviors of a pedestrian is totally different in comparison to a bicyclist, and to a car. Furthermore, a single vehicle has got a different motion model when it is moving in straight direction for example between lanes (rectilinear motion), and when it is turning. Often in order to address this problem, multiple motion models are used to compensate for the uncertainties resulting from this fact. If one needs to neglect this effect, the process and measurement noise need to be extended to include this issue.

**Constant Velocity Motion Model** Constant velocity model is used in situations that the vehicle is moving in straight direction and it's speed is assumed to be near zero without any change in direction. Therefore the velocity can be assumed to be near zero. This model is only valid if vehicle's maneuver is free of any kind of turning. This kind of motion is modeled as:

$$f_k^{cv} = \begin{bmatrix} pos_k^x + v_k T \sin(\psi_k) \\ pos_k^y + v_k T \cos(\psi_k) \\ 0 \\ v_k \\ \dot{\psi}_k \end{bmatrix} \tag{2.9}$$

**Constant Turn Rate Motion Model** The constant turn rate model is often used for urban environments as we do not expect maneuvering vehicles in road scenes such as highways to perform any turnings, while turning maneuvers are very common in urban driving scenes. In this motion model the velocity again is assumed to be constant but not zero while turning rate is assumed to be constant too. This motion model is often used to describe turning maneuvers while it is useful for rectilinear motions with uniform acceleration, too. This kind of motion is modeled as:

$$f_k^{ctr} = \begin{bmatrix} pose_k^x + \frac{v_k}{\dot{\psi}_k}\big( -\sin(\psi_k) + \sin((T+1)\psi_k)\big) \\ pose_k^y + \frac{v_k}{\dot{\psi}_k}\big( \cos(\psi_k) - \cos((T+1)\psi_k)\big) \\ (T+1)\psi_k \\ v_k \\ \dot{\psi}_k \end{bmatrix} \tag{2.10}$$

**Random Motion Model** Many detected objects may not follow a particular motion model. Static objects like traffic signs, lights, trees and etc. are not moving at all so they do not obey any structured maneuver model. Furthermore some pedestrians tend to move so unpredictable and in a ill-structured manner that can be assumed to move randomly. These kind of motions are also seen in over-segmented clusters due to occlusion, because they too are going to disappear at an instance of time and may appear again at any other instance, making them also behave randomly. Therefore, using random motion model is unavoidable for driving task. In order to tackle this kind of uncertainty, we simply use stationary stochastic process which apparently has got a bigger process covariance function $Q_k$ due to bigger uncertainty of not having an exact model for the process. This

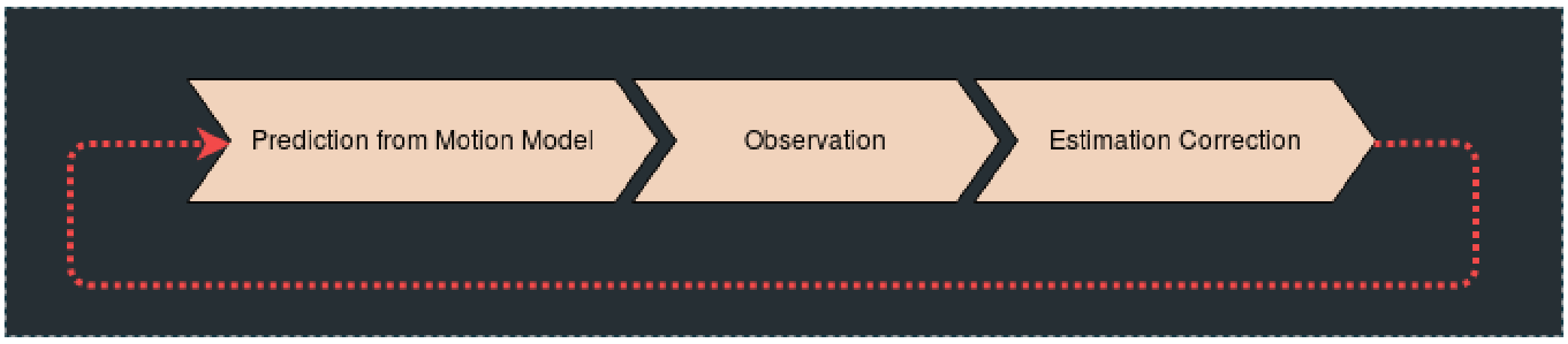

**Figure 2.5:** Bayesian estimation loop. First, based on information from motion model an estimation is made, then with advent of observations this estimation is sharpened.

kind of motion is modeled as:

$$f_k^{rm} = \begin{bmatrix} pos_k^x \\ pos_k^y \\ \psi_k \\ v_k \\ \dot{\psi}_k \end{bmatrix} \tag{2.11}$$

In future chapters, we see that these different motion models will be used in parallel in an Interactive Multiple Motion Model scenario to include all possible motion models for the detected objects.

### 2.6.4 Probabilistic Formulation of Tracking

Giving a probabilistic framework for tracking means to represent a methodology to ***estimate the posterior belief*** or ***state distribution*** based on sensor measurements. The posterior belief in our case is the introduced kinematic attributes of the target objects. A probabilistic framework for this goal enables us to use ***probability density functions*** as perfect means to represent information. The so-called Bayesian filtering is utilized in this thesis as the probabilistic framework which is shaped over two key ideas as depicted in Figure 2.5.

First, probability distributions can be used to represent our belief about the state of dynamical systems as depicted in Figure 2.6. Second, there can be a recursive cycle with three main elements that can lead to a fair prediction of the variables of our interest: ***predict from the motion model***, interacting the ***sensor measurements*** and finally ***correcting the prediction based on observations***.

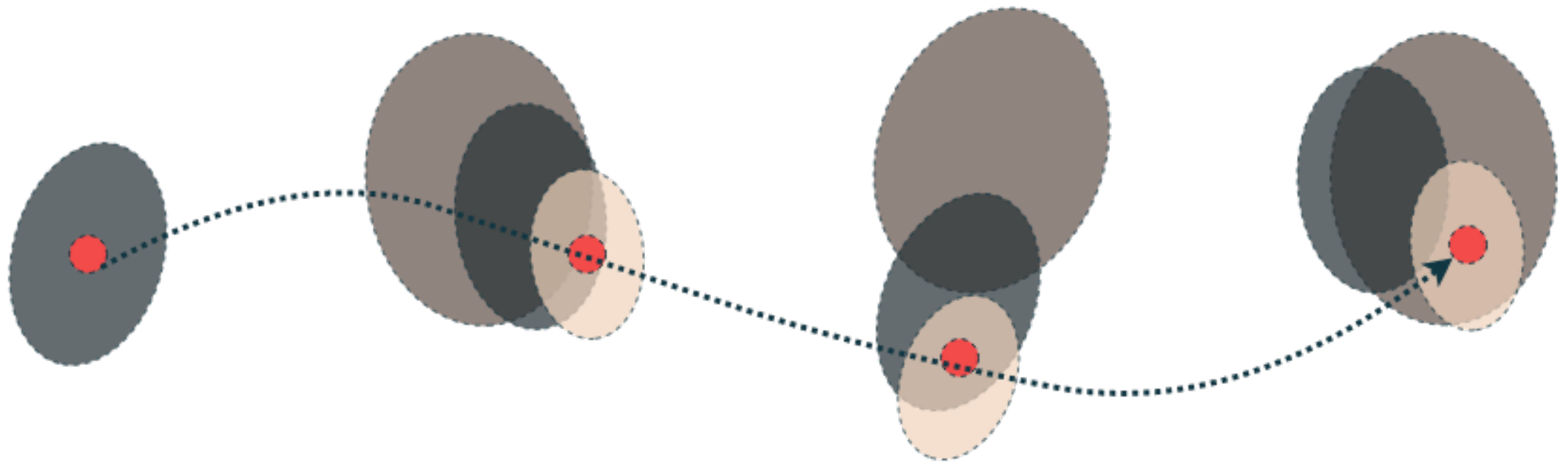

**Figure 2.6:** Time Step Evolution of target object's true position, estimated posterior belief and the measurement.

**Tracking:** Tracking can be defined as the process of inferring information based on a sequence of measurements over time. The problem with this process is that it is often inherently ambiguous and there is no optimum answer. In addition observations are always noisy, cluttered, misleading and incomplete while data association is not clear neither. The state vector at time step $t$ is denoted by $x_t$. $x_t$ might be position and orientation of a target object or even appearance and shape of it. The sequence of states from time step 1 to time step $t$ is denoted as:

$$x_{1:t} = (x_1, ..., x_t) \tag{2.12}$$

Furthermore, observation at time $t$ is denoted by $z_t$ which can include video streams, an entire image, range data or etc. The sequence of observation from time step 1 to time step $t$ is denoted as:

$$z_{1:t} = (z_1, ..., z_t) \tag{2.13}$$

**Posterior Distribution:** Regarding the problem of tracking, we are mainly concerned about the posterior distribution. The posterior distribution is the conditional probability distribution of states given the measurements:

$$p(x_{1:t}|z_{1:t}) \tag{2.14}$$

distribution is also called the ***full posterior*** because it gives all information about states at all time steps based on all of the measurements.

**Filtering Distribution:** The full posterior often is not needed to be calculated. Instead, the *filtering distribution* which is the probability of the the most recent state given observations up to that time is more useful:

$$p(x_t|z_{1:t}) \tag{2.15}$$

Given these distributions, two main tracking questions can be answered regarding motion tracking and state tracking. For state tracking we will have:

1. What is the most probable state? (most likely)

$$\max_x p(x_t|z_{1:t}) \tag{2.16}$$

2. what is the average state?

$$E[x_t|z_{1:t}] = \int x_t p(x_t|z_{1:t}) dx_t \tag{2.17}$$

Similarly for motion tracking we will have:

1. What is the most probable motion? (most likely)

$$\max_x p(x_{1:t}|z_{1:t}) \tag{2.18}$$

2. what is the average motion?

$$E[x_{1:t}|z_{1:t}] = \int x_t p(x_{1:t}|z_{1:t}) dx_t \tag{2.19}$$

Therefore, probabilistic tracking is the problem of efficiently computing the posterior or filtering distribution at each time step $t$. In this thesis, object tracking is formulated as a filtering problem in which dynamic states of the target objects are to be estimated from noisy, occluded and misleading measurements.

### 2.6.5 Defining The Posterior (Bayesian Theorem)

In this thesis, dynamic states of the target object, sensor measurements and every other aspect of the surrounding environment are modeled as random variables and all the procedures related to the motion of the ego-vehicle, measurement and perception are assumed to be a stochastic process. Therefore, probabilistic inference is needed to be utilized to enable us to decipher hidden laws governing on these procedures. Bayesian filtering is an inference methodology which is used for target object tracking in many sources.

The *Bayes' rule* may be applied to the posterior to express it in terms of *likelihood*, *prior* and *evidence*. The likelihood measures how well the observations match to the states. The prior tells us how likely the states are without any observation and at last, evidence is just a normalizing factor which does not depend on the states.

$$p(x_{1:t}|z_{1:t}) = \frac{\overbrace{p(z_{1:t}|x_{1:t})}^{\text{likelihood}}\overbrace{p(x_{1:t})}^{\text{prior}}}{\underbrace{p(z_{1:t})}_{\text{evidence}}} \tag{2.20}$$

The likelihood and prior must be constructed in order to be able to use the Bayes theorem for inference in tracking problems. In order to construct a formula for prior, we assume the motion model $p(x_t|x_{1:t-1})$ to be a ***first-order Markov model*** where the future is independent of the past, given the present:

$$p(x_t|x_{1:t-1}) = p(x_t|x_{t-1}) \tag{2.21}$$

Thus, the prior can be written as follows:

$$\begin{aligned} p(x_{1:t}) &= p(x_t|x_{1:t-1})p(x_{1:t-1}) \\ &= p(x_t|x_{t-1})p(x_{1:t-1}) \\ &= p(x_1)\prod_{i=2}^{t} p(x_i|x_{i-1}) \end{aligned} \tag{2.22}$$

In addition to this assumption, observations are assumed to be conditionally independent, given the states:

$$p(z_{1:t}|x_{1:t}) = \prod_{i=1}^{t} p(z_i|x_i) \tag{2.23}$$

These assumptions are made to simplify the inference process. The first-order Markov model assumption serves to the fact that the motion model can be obtained by a single-step-ahead motion model $p(x_t|x_{t-1})$, while likewise likelihood can be obtained by a single-step-ahead observation model.

### 2.6.6 Recursive Form for Posterior and Filtering Distribution

The two simplifying assumptions along with the use of some basic mathematics, enables us to derive a recursive form for both posterior and filtering distributions. Recursive forms are important because using them, all of the information which is kept in the past observations can be achieved using the previous distribution. Thus, with only access to the previous distribution and a new observation, the next distribution can be obtained. Recursive form for the posterior is of the following form:

$$\begin{aligned} p(x_{1:t}|z_{1:t}) &= \frac{p(z_{1:t}|x_{1:t})p(x_{1:t})}{p(z_{1:t})} \\ &\propto p(z_{1:t}|x_{1:t})p(x_{1:t}) \\ &= \prod_{i=1}^{t} p(z_i|x_i)p(x_1)\prod_{i=2}^{t} p(x_i|x_{i-1}) \\ &\propto p(z_t|x_t)p(x_t|x_{t-1})\boxed{p(x_{1:t-1}|z_{1:t-1})} \end{aligned} \tag{2.24}$$

Recursive form for the filtering distribution is of the form below, which leads to the definition of the ***prediction distribution*** $p(x_t|z_{1:t-1})$:

$$\begin{aligned} p(x_t|z_{1:t}) &= \int p(x_{1:t}|z_{1:t})dx_{1:t-1} \quad marginalization \\ &\propto \int p(z_t|x_t)p(x_t|x_{t-1})p(x_{1:t-1}|z_{1:t-1})dx_{1:t-1} \\ &= p(z_t|x_t)\int p(x_t|x_{1:t-1})p(x_{1:t-1}|z_{1:t-1})dx_{1:t-1} \\ &= p(z_t|x_t)\int p(x_{1:t}|z_{1:t-1})dx_{1:t-1} \quad factorization \\ &= p(z_t|x_t)p(x_t|z_{1:t-1}) \quad marginalization \end{aligned} \tag{2.25}$$

The prediction distribution is the distribution of the next state given the previous observations.

$$\begin{aligned} p(x_t|z_{1:t-1}) &= \int p(x_t, x_{t-1}|z_{1:t-1})dx_{t-1} \\ &= \int p(x_t|x_{t-1})p(x_{t-1}|z_{1:t-1})dx_{t-1} \end{aligned} \tag{2.26}$$

## 2.7 Bayesian Estimation for MOT problem

So far, a posterior and filtering distribution have been introduced in terms of a likelihood and a motion model. The task of tracking is to effectively compute this posterior. Solving the recursive Equation 2.25 is the main idea in object tracking filtering task. Different object tracking filters often differ in how they tend to choose realization for likelihood function and prior motion model. In the context of the object tracking in 3D environment, the prior density is derived by the object's dynamic equation and the likelihood is derived from sensor's measurement function.

In object tracking, quantities such as the state of the tracking targets, sensor measurements and other aspects of the surrounding environment are modeled as random variables. Random variables are often prone to sudden change and representing unsure values in each situation, thus probabilistic inference must be utilized to obtain the law which is governing evolution of those variables. Bayesian framework is an infamous probabilistic inference framework which is also well-known in MOT literature [85]. However, because measurements are often entering the system in a discretized-time manner, a recursive version of Bayesian inference must be utilized. For the MOT problem, Bayesian estimation is applied according to the following steps:

1. **State Prediction Step:** Current state of the tracking targets will be estimated using recent knowledge of states and measured values. A motion model is utilized to model the expected time-evolution of the system model, which in turn adds some uncertainty to the estimation. Since the states of the target objects and measurements are defined in a same coordinate system, the prediction step also includes a coordinate transformation between target object's coordinate and the sensor coordinate system. Thus, the measurement model is specified according to the sensor itself.

2. **State Update Step:** State prediction, results in a representation of the probabilistic distribution of track's state, based on recent measurements. In this step, the likelihood of the sensor's measurement is utilized with the current measurement to update the target track's states. The predicted state estimate is adjusted to current measurements to represent the best knowledge of the system at the current time step. A probabilistic distribution which is resulted from combination of information from latest sensor measurements and up-to-now accumulated time evolution of the system states, is the result of state update step.

Several relevant Bayesian filters are utilized to enable MOT. Basic fundamental of these filters are going to be discussed in the remaining of this chapter.

### 2.7.1 The Unscented Kalman Filter (UKF)

The Kalman filter (KF) is the analytical approach for the Bayesian inference method. Kalman filter results in an optimal estimation of the states, which is obtained in a recursive manner. KF assumes that object's dynamic motion model and the posterior densities are following Gaussian distribution and the system and measurement functions are linear, which is not true in real-world applications. Specially in object tracking problems, dynamic motions of the tracks can not be fully captured by a linear motion model. Other realizations of the Kalman filter such as Extended Kalman Filter (EKF) or the Unscented Kalman Filter (UKF) are utilized for nonlinear models. While a detailed explanation of different kinds of optimal Bayesian filters with applications in robotics and object tracking may be found in references [86, 87], here we delve into some details according to our interest in this thesis.

The Unscented Kalman Filter belongs to a bigger class of filters called Sigma-Point Kalman Filters. This class of filters use the statistical linearization technique. The statistical linearization technique is used to linearize a non-linear function of a random variable through a regression between n points drawn from the prior distribution of random variables. Actually the UKF is founded on the basic intuition that it is easier to approximate a probability distribution that it is to approximate an arbitrary nonlinear function. For example, the EKF linearizes the system and measurement function with a Taylor series approximation. The UKF instead, tends to avoid the computationally expensive task of linearization and uses an approximation based on sigma points drawn from a Gaussian distribution.

Instead of linearizing around the mean value and using the Jacobian as the system matrix, the UKF propagates the chosen sigma point through the original non-linear function. The new Gaussian distribution then can be recovered from these propagated sigma points. The results of these filtering method are not optimal results as the resulting distribution is just an approximation of a Gaussian. The benefit of using this method is that the UKF is shown to have more accurate results than EKF in the presence of severe non-linearity [88].

The UKF has a computational complexity comparable to the KF algorithm and a more accurate answer in comparison to the EKF algorithm, thus UKF has been often used for LiDAR-based object tracking [89, 90, 91, 26, 92]. The UKF algorithm may be summarized in three different steps which is discussed here: *sigma point sampling*, *prediction* and *update*.

**Sigma Point Sampling:** A set of sigma points are constructed as following:

$$
\begin{aligned}
\chi^0_{k-1|k-1} &= \chi^*_{k-1|k-1} \\
\chi^i_{k-1|k-1} &= \chi^*_{k-1|k-1} + \left(\sqrt{(L+\lambda)P^*_{k-1|k-1}}\right)_i,\ i = 1, ..., L \\
\chi^i_{k-1|k-1} &= \chi^*_{k-1|k-1} - \left(\sqrt{(L+\lambda)P^*_{k-1|k-1}}\right)_{i-L},\ i = L+1, ..., 2L
\end{aligned}
\tag{2.27}
$$

Where the $\left(\sqrt{(L+\lambda)P^{*}_{k-1|k-1}}\right)_i$ is the $i_{th}$ column of the matrix square root of $(L+\lambda)P^{*}_{k-1|k-1}$.

**Prediction:** In this step, the sampled sigma points are going to be propagated through transition function $f : \mathbb{R}^L \rightarrow \mathbb{R}^{|\chi|}$:

$$\chi^{*,i}_{k|k-1} = f\left(\chi^{i}_{k-1|k-1}\right), \; i = 0, ..., 2L \tag{2.28}$$

After this propagation, weighted sigma points are combined to reconstruct an estimation of the predicted state and covariance:

$$\hat{x}^{-}_{k|k-1} = \sum_{i=0}^{2L} W^i_s \chi^{*,i}_{k|k-1} \tag{2.29}$$

$$P^{-}_{k|k-1} = \sum_{i=0}^{2L} W^i_c \left[\chi^{*,i}_{k|k-1} - \hat{x}^{-}_{k|k-1}\right]\left[\chi^{*,i}_{k|k-1} - \hat{x}^{-}_{k|k-1}\right]^T + Q_{k-1|k-1} \tag{2.30}$$

The $W^i_s$ and $W^i_c$ values are the corresponding weight values which are computed according to the following equations:

$$\lambda = \alpha^2(L+k) - L \tag{2.31}$$

$$W^0_s = \frac{\lambda}{L+\lambda} \tag{2.32}$$

$$W^0_c = \frac{\lambda}{L+\lambda} + (1 - \alpha^2 + \beta) \tag{2.33}$$

$$W^i_s = W^i_c = \frac{1}{2(L+\lambda)} \tag{2.34}$$

$$\tag{2.35}$$

Where $\alpha$ and $k$ are related to the distribution of the sigma points and $\beta$ is related to the distribution of the states $x$.

**Update:** Another $2L+1$ other sigma points must be sampled here:

$$\begin{aligned}
\chi^{0}_{k|k-1} &= x^{-}_{k-1|k-1} \\
\chi^{i}_{k|k-1} &= x^{-}_{k-1|k-1} + \left(\sqrt{(L+\lambda)P^{-}_{k-1|k-1}}\right)_i, \; i = 1, ..., L \\
\chi^{i}_{k|k-1} &= x^{-}_{k-1|k-1} - \left(\sqrt{(L+\lambda)P^{-}_{k-1|k-1}}\right)_{i-L}, \; i = L+1, ..., 2L
\end{aligned} \tag{2.36}$$

This sigma points must be propagated through measurement function $h$:

$$Z_k^i = h\big(\chi_{k|k-1}^i\big),\ i = 0, ..., 2L \tag{2.37}$$

The predicted measurement can be computed now:

$$\hat{z}_k^- = \sum_{i=0}^{2L} W_s^i Z_k^i \tag{2.38}$$

The covariance of the predicted measurement:

$$P_z = \sum_{i=0}^{2L} W_c^i \big[Z_k^i - \hat{z}_k^-\big] \big[Z_k^i - \hat{z}_k^-\big]^T + R_{k|k-1} \tag{2.39}$$

The cross-covariance of the state-measurement is calculated as following:

$$P_{xz} = \sum_{i=0}^{2L} W_c^i \big[\chi_{k|k-1}^i - x_{k|k-1}^-\big] \big[Z_k^i - \hat{z}_k^-\big]^T \tag{2.40}$$

The UKF gain then is obtained as follows:

$$K_k^{UKF} = P_{xz} P_z^{-1} \tag{2.41}$$

Then the updated state and the covariance of the system can be calculated:

$$\hat{x}_k = \hat{x}_{k-1} + K_k^{UKF} \big(z_k - \hat{z}_{k|k-1}^-\big) \tag{2.42}$$

$$P_k = P_{k|k-1} - K_k^{UKF} x_k K_k^{UKF^T} \tag{2.43}$$

### 2.7.2 Interacting Multiple Models

The state estimation methodology relies on the motion model and measurement likelihood to predict and update the tracking object's state values. In urban environments moving objects are not moving with regarding to a well-defined pattern and are prone to follow different motion models at different time intervals while they will not even follow them precisely. This is actually what happens in a road junction.

In a common situation, different road users may change their motion model according to their navigational intention and position as depicted in Figure 2.7. A vehicle in a straight street may follow a constant velocity motion model while not changing it's lane. A bicycle but only follows a constant velocity model until it is not turning around, but as it tends to turn around it is following a constant turn rate model. A pedestrian often is modeled to follow a random motion model.

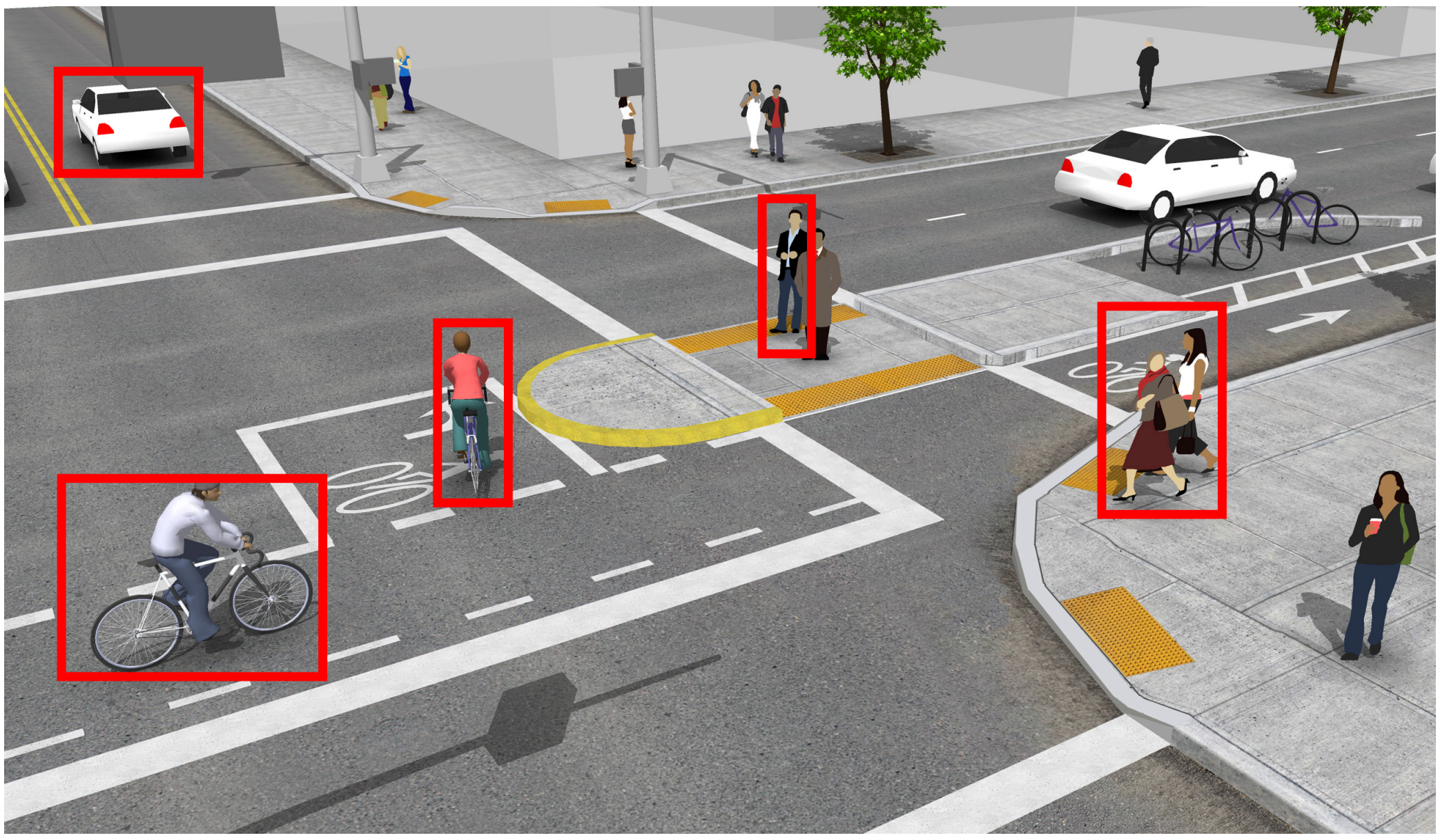

**Figure 2.7:** A figurative schematic of an urban intersection with different motion models.

In order to overcome this problem a filter must be applied on different maneuvering targets [92, 85]. This filter will also help the tracker to distinguish between static and dynamic objects as dynamic objects tend to move in a more predictable pattern than static objects which are actually noises in our model [93, 65]. Interacting Multiple Models filtering (IMM) is commonly used for the tracking of dynamic objects in urban areas [94, 95]. IMM filter enables better estimations for objects with ambiguous dynamical behavior. Different potential target maneuvers are represented using multiple motion models in the corresponding state estimation filter. Then, the most probable estimation results will be ranked in relation to each other. The output of the IMM filter either can be the filter with the highest probability or a weighted combination of the individual filters.

IMM filter is constructed from $j$ parallel filters with each of them using different motion model to represent time-evolution of states and the measurement. In the other words, for every single target object there are $j$ parallel running filters to maintain. Utilizing each individual filter estimation results, the predicted state vectors and covariance matrices are obtained with respect to predicted model probabilities. Next, each of the estimates is compared to current measurement to update the model-match probability. Computed probabilities of the dynamical target model, are added to the updated state vector and error covariance matrix information produced by the corresponding filter.

There is a combination step in which the mode probabilities are used. In combination step, state vectors and covariance matrices of each different motion models are merged into a single state vector and covariance matrix which in the process, new filtered state estimates, error covariance matrices and corresponding model probabilities are computed for each model using weighted state estimates.

**Formulation:** The system with an amount of uncertainty in it's motion model is assumed to evolve as a Markov Jump Linear System (MJLS). Then the stochastic process representing each of these $j$ different processes are as follows:

$$\begin{aligned} x_{k+1} &= f_j(x_k, u_k) + \omega_{j,k} \\ z_k &= h_j(x_k, u_k) + v_{j,k} \end{aligned} \tag{2.44}$$

Where all the definitions related to the noise parameters and state variables are identical as the definitions in Equation 2.5. Furthermore, $j = 1, ..., r$ which is part of the model set $\mathbb{M} = \{M_j\}_{j=1}^{r}$. IMM filter is a variant of MJLS which uses $r$ different motion models for running in parallel to estimate states. In the Bayesian framework, the posterior probability density function of the IMM may be inferred as:

$$p(x_k, M_k|z_k) \tag{2.45}$$

Which means that the estimate of the joint probability density function may be inferred by utilizing all the measurements up to time step $k$, state vector at time step $k$ and the $M_k$ for the motion model. using conditional probability lemma this probability distribution function can be further decomposed into:

$$\begin{aligned} p(x_k, M_k|z_k) &= \sum_{j=1}^{r} p(x_k, M_{j,k}) p(M_{j,k}|z_k) \\ p(x_k, M_k|z_k) &= \mu_{j,k} \sum_{j=1}^{r} p(x_k, M_{j,k}) \end{aligned} \tag{2.46}$$

With $\mu_{j,k} = p(M_{j,k}|z_k)$ is the *posterior mode probability* that the chosen object motion model actually matches the dynamics of motion model ran by the filter number $j$ at time step $k$. The probability density function can be inferred in a recursive form:

$$p(x_{k-1}, M_k|z_{k-1}) = \sum_{j=1}^{r} (x_{k-1}, M_{j,k}|z_{k-1}) \mu_{i|j,k-1} \tag{2.47}$$

Where

$$\mu_{i|j,k-1} = \frac{p_{ij} \mu_{i,k-1}}{\bar{\mu}_{j,k}} \tag{2.48}$$

where $\bar{\mu}_{j,k} = \sum_{i=1}^{r} p_{ij}\mu_{i,k-1}$ is the mode match probability of the filter $j$ at the time step $k$. $p_{ij}$ is a predefined transition probability from model index $i$ to index $j$ which is one the filter design parameters identified in the matrix $\Omega$.

$$\Omega = \begin{bmatrix} p_{11} & \dots & p_{1r} \\ \vdots & \ddots & \vdots \\ p_{r1} & \dots & p_{rr} \end{bmatrix} \tag{2.49}$$

**The IMM Full Cycle:** The IMM full cycle is consisted of 4 different stages: mixing, prediction, update and combination as is depicted in Figure 2.8.

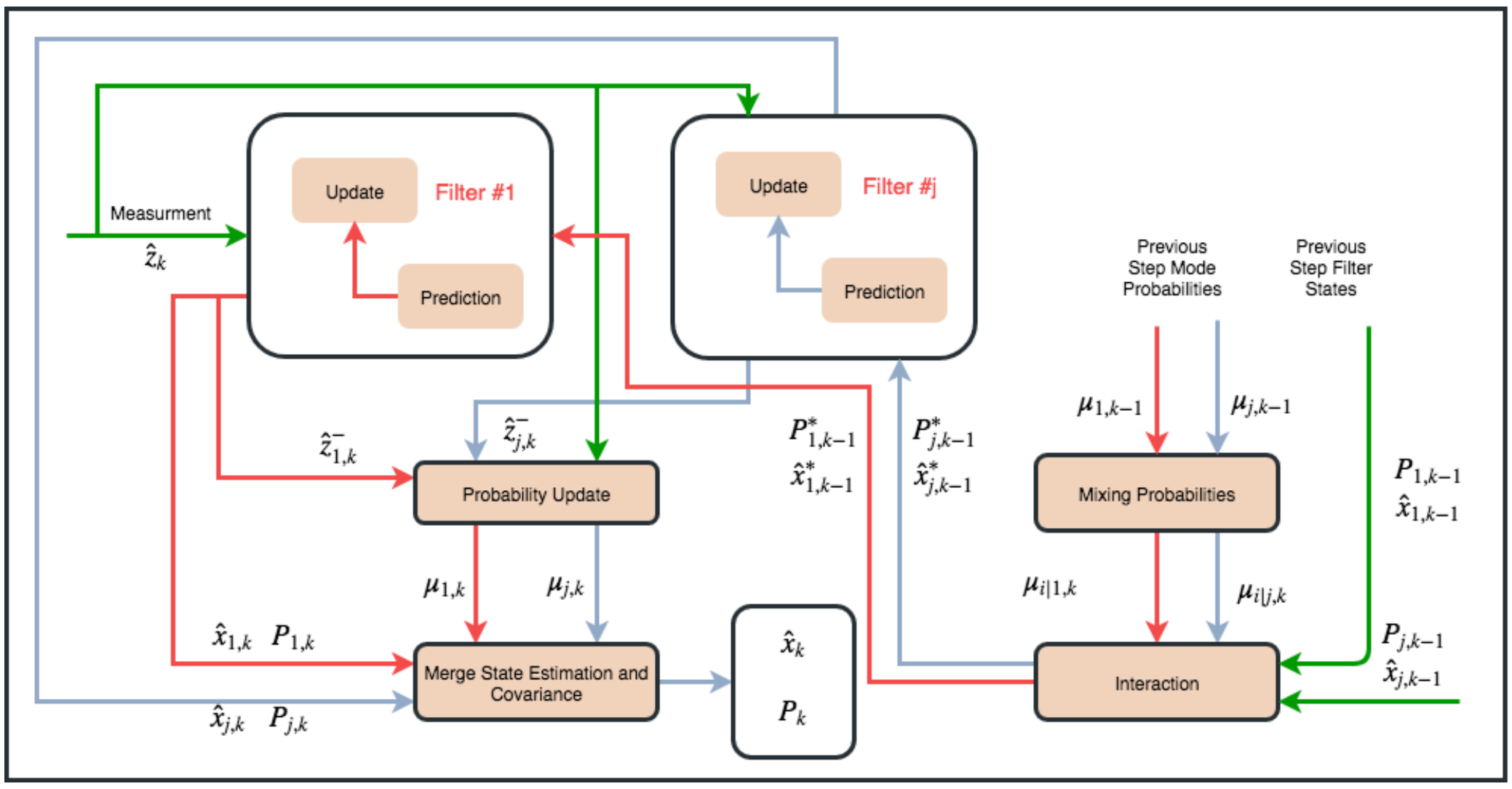

**Figure 2.8:** The IMM filter schematic. The depicted filter can be extended to $j_{th}$ filter with r-amount of different motion models.

**Mixing** The mixing stage is where IMM determines a single combined state estimate and its corresponding covariance matrix. The filter incorporates weighted average of each of $j_{th}$ model filter state $\hat{x}_{i,k-1}$ to determine the combined filter state $\hat{x}^*_{j,k-1}$ and the combined covariance $P^*_{j,k-1}$ by adding up each model's joint posterior density.

$$\hat{x}^*_{j,k-1} = \sum_{i=1}^{r} \mu_{i|j,k-1}\hat{x}_{i,k-1} \tag{2.50}$$

$$P^*_{j,k-1} = \sum_{i=1}^{r} \mu_{i|j,k-1}\hat{x}_{i,k-1}\big[P_{i,k-1} + (\hat{x}_{j,k-1} - \hat{x}^*_{j,k-1})(\hat{x}_{j,k-1} - \hat{x}^*_{j,k-1})^T\big]$$

**Prediction** Based on obtained mixed initial states $\hat{x}^*_{j,k-1}$ and mixed initial covariance $P^*_{j,k-1}$ and by utilizing individual $j_{th}$ Kalman filter employing it's own motion model, a prediction is made. An ordinary Kalman filter is used to carry out this prediction. The resulting model-specific predicted states are going to be denoted as $x^-_{j,k}$ and $P^-_{j,k}$. The model-specific predicted measurement will be denoted as $\hat{z}^-_{j,k}$ and the innovation matrix $S_{j,k}$ which we see in the figure 2.8 that is the input to the Probability update block.

**Update** A model-specific Kalman filter is used again for carrying out the update procedure. The resulting state estimation and innovation matrix are denoted in the Figure 2.8 by $\hat{x}_{j,k}$ and $P_{j,k}$. The parameter $\mu_{j,k}$ which is called the mode probability actually is a denotation of how much the current measurement fits into the active model $j$. The mode probability is updated after defining measurement's Gaussian likelihood:

$$\lambda_{j,k} = \frac{1}{\sqrt{(|2\pi S_{j,k}|)}} e^{\frac{1}{2}(z_k - \hat{z}^-_{j,k})^T S^{-1}_{j,k}(z_k - \hat{z}^-_{j,k})} \tag{2.51}$$

Then the mode probability us updated due:

$$\mu_{j,k} = \frac{\lambda_{j,k}\mu^-_{j,k}}{\sum_{i=1}^{r} \lambda_{j,k}\mu^-_{j,k}} \tag{2.52}$$

**Combination** The filter output of each j-filter is calculated by re-combination of the states:

$$\hat{x}_k = \sum_{i=1}^{r} \mu_{j,k}\hat{x}_{j,k} \tag{2.53}$$

$$P_k = \sum_{j=1}^{r} \big(P_{j,k} + (\hat{x}_{j,k} - \hat{x}_k)(\hat{x}_{j,k} - \hat{x}_k)^T\big) \tag{2.54}$$

## 2.8 Data Association Filter

Until now, proposed filters tend to obtain optimal estimates of target object's track. The problem with using these filters is that due to detection faults there is no guarantee that the tracked object is a real target object in the environment or even an existing object. In order to tackle this problem a subsequent classification and further validation of each estimated track is necessary. Data Association (DA) is simply a procedure of associating detection results into the tracking filter to prevent faulty tracking procedures.

Data association filters can be divided into two major classes: *deterministic* and *probabilistic* filters. The Nearest Neighborhood Filter (NNF) is an example of deterministic data association filters which updates each object with the closest measurement relative to the state values. Objects are associated with each known track based on the minimum Euclidean or the Mahalanobis[10] distance between the measurement and the track. The Probabilistic Data Association Filter (PDAF) is the infamous probabilistic method for data association which is well-known in robotics literature [96]. In the use of NNF algorithm, situations may rise where the algorithm encounters erroneous association decisions. These erroneous associations often happen due to a clutter problem in which multiple measurements are located near to each other thus a single measurement is used to update all of the surrounding objects. In order to avoid this erroneous association decisions PDAF performs a weighted update of all target object's state using all of the association hypotheses.

Based on their implementation, data association filters can further be divided into two classes: *single-* and *multi-target* tracking filters. The single-target data association filter must solve a $(n \times 1)$–*association* problem in which many detections must be associated to a single track. For deterministic single-track data association filter again nearest neighborhood method must be utilized which picks the closest detection for each track. For probabilistic single-track data association filter, PDAF is used in which a probabilistic weighting based on distance is utilized. The multi-target data association filter is a $(n \times m)$–*association* problem in which detection are associable to any track. For deterministic single-track data association the Global Nearest Neighborhood Filter (GNNF) is utilized in which all of the association hypotheses are taken into account and the one with smallest sum of distances are chosen. Furthermore, for probabilistic multi-tack data association filter the Joint Probabilistic Data Association Filter (JPDAF) is utilized in which marginalized joint probabilistic weighting is used based on distance. In the remaining of this section, different kinds of DA filters are to be discussed .

[10]Distance defined between a point $p$ and a distribution $D$

### 2.8.1 Probabilistic Data Association Filter (PDAF)

We start from the PDAF to later better understand how Joint Probability is incorporated to the PDAF to constructed the JPDAF. An ordinary Kalman filter with linear system and measurement functions are assumed. The PDAF algorithm consists of four different steps: *prediction*, *measurement validation or gating*, *data association* and *state estimation*. The measurement is assumed to follow a Gaussian probability distribution. PDAF uses the incoming measurement inside a validation gate to approximate the probability distribution function of the tracked objects after each update step. PDAF works with five basic assumptions:

1. Only one target of interest is present in the environment with the state vector $x \in \mathbb{R}^{n_x}$. Time evolution of this state vector can be modeled as:

   $$x_{k-1} = f_{k-1}(x_{k-1}) + \omega_{k-1} \tag{2.55}$$

   And the true measurement $z_k \in \mathbb{R}^{n_z}$ is given by:

   $$z_k = h_k x_k + v_k \tag{2.56}$$

   In this equations, $\omega_{k-1}$ and $v_k$ are zero-mean white Gaussian noises with the corresponding $Q_{k-1}$ and $R_k$ covariance functions.

2. The tracks are already initialized.

3. A sufficient statistic in the form of a Gaussian posterior is used to summarize all of the past informations about the tracks:

   $$p[x_{k-1}|z_{k-1}] = \mathcal{N}[x_{k-1}; \hat{x}_{k-1|k-1}, P_{k-1|k-1}] \tag{2.57}$$

4. If the target was detected and the corresponding measurement happens to be placed into the validation region, then according to equation 2.56, only maximum of one of the validated measurement can be used to originate the target.

5. All non-object originated measurements are assumed to be originated from a clutter that is uniformly distributed in space and Poisson distributed in time [2]. In the PDAF algorithm the prediction and measurement update are the same as a conventional Kalman filter, but two other steps being measurement validation and data association are specific to this algorithm according to the figure.

**Prediction:** one-step-ahead prediction of the state and the covariance matrices are calculated using a Kalman filter:

$$\begin{aligned} \hat{x}_{k|k-1} &= f_{k-1}(\hat{x}_{k-1|k-1}) \\ \hat{z}_{k|k-1} &= h_k(\hat{x}_{k|k-1}) \\ P_{k|k-1} &= f_{k-1}P_{k-1|k-1}f_{k-1}^T + Q_{k-1} \end{aligned} \tag{2.58}$$

Where the one-step-ahead covariance function is obtained from the Equation 2.57 while the innovation covariance matrix corresponding to the current measurement is given as:

$$S_k^{PDAF} = h_k P_{k|k-1} h_k^T + R_k \tag{2.59}$$

**Measurement Validation or Gating Process:** A validation gating or a windowing process of bad measurements must be implemented prior to the passing of measurements to the data association filter. In fact, the distances of all of the measurements to all of target tracks are calculated. Then, all the detections that exceed a predefined threshold for distance to closest tracks are excluded as depicted in Figure 2.9. These measurements are labeled as they are non-associable because it is so unlikely that these measurements are actually a representation for a already tracking target. Although, instead of throwing these measurements away, they are used to construct a new object hypothesis and those measurements are considered to be potential new tracks relating to new objects that are not yet tracked by the algorithm. Thus, non-associable detections are stored to further be used as the potential new tracks.

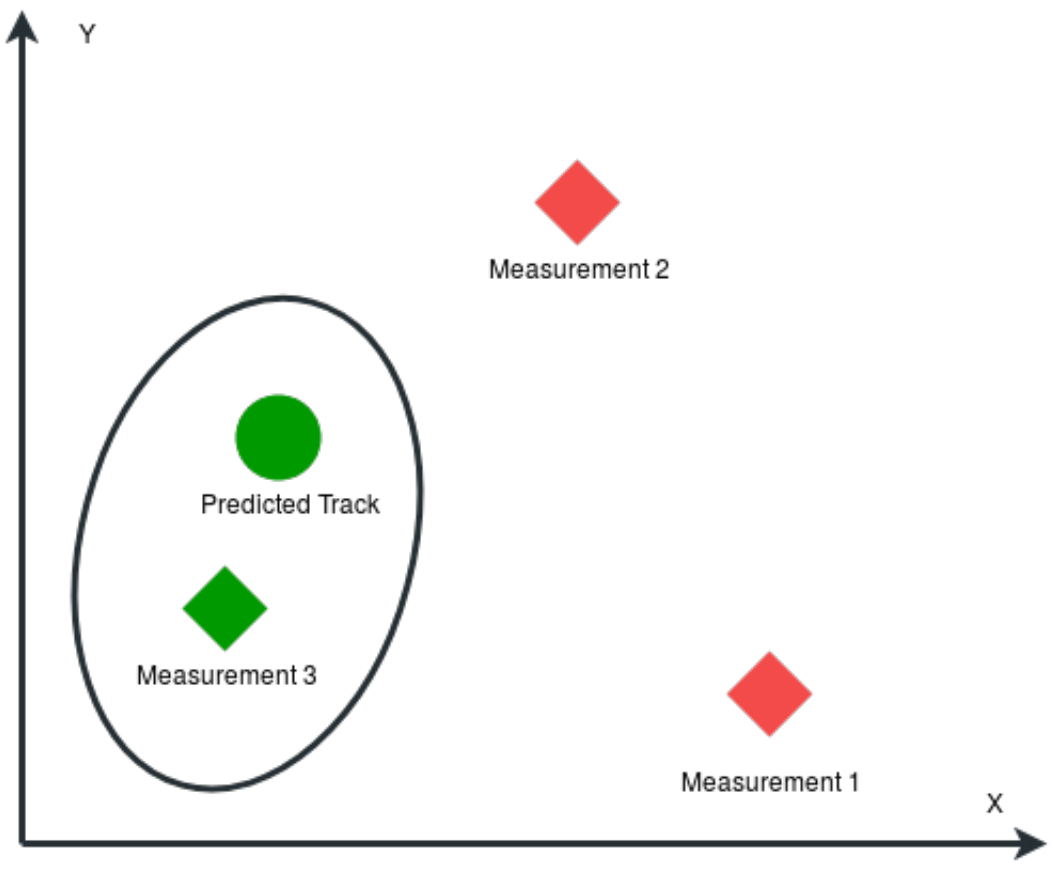

**Figure 2.9:** The gating or windowing process. The first two measurements are considered to be unlikely to be initialized from the tracking target, therefore they are excluded.

The gating problem, in this thesis is assumed to be a measurement selection problem [86]. The gating process tends to select a subset of measurements which contains object detections with high probability (gating probability) based on a priori knowledge of that the object is detected and exists in the environment. Measurements are assumed to be distributed according to a Gaussian distribution. The gating region is assumed to be a hyper-ellipsoid region. The Mahalanobis distance of the obtained measurement is computed and compared against that of predicted state vector according to method presented by [97]. The threshold for the gating process is called the **gating level** which is constructed using $\chi^2$ distribution. The gating area is given by elliptical region introduced in [96, 2]:

$$V(k,\eta) = \{z : [z - \hat{z}_{k|k-1}]TS_k^{-1}[z - \hat{z}_{k|k-1} \leq \eta\} \tag{2.60}$$

Where $\eta$ is the gate level value which is equal to $\left(\chi^2(P_G)\right)^{-1}$ where $P_G$ is the probability of the gate containing the true measurement. If an object is detected or the **validation gate probability** and $S_k$ is the innovation covariance matrix of the true measurement. The set of validated measurements are given as:

$$z_k^v = \{z_{i,k}\}_{i=1}^{m_k} \tag{2.61}$$

**Data Association:** A clutter density function must be chosen in order for data association process to be effective. We consider a nonparametric, uniform, diffuse prior clutter model that uses the number of returns in the track gate to estimate the clutter density and that is suitable for heterogeneous clutter environments as any number of false measurements is considered equi-probable [98]. Under these considerations, the association probability $\beta_{i,k}$ for the measurement $z_{i,k}$ (the measurement $i$ at time $k$) being the correct measurement given a target detection probability $P_D$ is calculated according to:

$$\beta_{i,k}(x) = \begin{cases} \dfrac{e^{\frac{1}{2}(z_{i,k}-\hat{z}_{i,k|k-1})^T S_k^{-1}(z_{i,k}-\hat{z}_{i,k|k-1})}}{\left(\frac{2\pi}{\eta}\right)^{\frac{q}{2}} \frac{m_k(1-P_D P_G)}{V_k P_D} + \sum_{j=1}^{m_k} e^{\frac{1}{2}(z_{j,k}-\hat{z}_{j,k|k-1})^T S_k^{-1}(z_{j,k}-\hat{z}_{j,k|k-1})}} & i = 1, ..., m_k \\ \dfrac{\left(\frac{2\pi}{\eta}\right)^{\frac{q}{2}} \frac{m_k(1-P_D P_G)}{V_k P_D}}{\left(\frac{2\pi}{\eta}\right)^{\frac{q}{2}} \frac{m_k(1-P_D P_G)}{V_k P_D} + \sum_{j=1}^{m_k} e^{\frac{1}{2}(z_{j,k}-\hat{z}_{j,k|k-1})^T S_k^{-1}(z_{j,k}-\hat{z}_{j,k|k-1})}} & i = 0 \end{cases} \tag{2.62}$$

Where $q$ is the size of measurement vector and $V_k$ is the volume of the validation region for the $q - dimensional$ measurement:

$$V_k = \frac{\pi^{\frac{q}{2}}}{\Gamma(\frac{q}{2}+1)} \sqrt{|\eta S_k|} \tag{2.63}$$

while $\beta_{0,k}$ is the probability that no measurements within the gate are correctly belonging to the track and model under consideration [30, 98]. Also, $P_D$ is the detection probability. $\Gamma$ is the infamous gamma function.

**State Estimation:** The state update equation is given by:

$$\hat{x}_{k|k} = \hat{x}_{k|k-1} + K_k v_k \tag{2.64}$$

The combined innovation $v_k$ is obtained as:

$$v_k = \sum_{j=1}^{m_k} \beta_{j,k}(z_{j,k} - \hat{z}_{k|k-1}) \tag{2.65}$$

The filter gain $K_k$ is given by:

$$K_k = P_{k|k-1} h_k^T S_k^{-1} \tag{2.66}$$

Where, associated covariance of the updated state is given by:

$$P_{k|k} = \beta_{0,k} P_{k|k-1} + (1 - \beta_{0,k})(P_{k|k-1} - K_k S_k^T K_k^T) + K_k \Big( \sum_{j=1}^{m_k} \beta_{j,k} v_{j,k} v_{j,k}^T - v_k v_k^T \Big) K_k^T \tag{2.67}$$

It can be seen that the probability weighting parameter of $\beta_{0,k}$ related to the correct measurement is taken into account in equation 2.8.1 and during the update step. Thus, PDAF algorithm is actually resembling the common KF updates in terms of state prediction and innovation matrix.

### 2.8.2 Tracking Multiple Targets in Clutter

The PDAF algorithm can not be used for the MOT problem because only distances to a single detected target is taken into account in this method. The MOT problem, on the other hand can be treated as a single object tracking problem with multiple trackers run in parallel. This is not a solution for the MOT problem in driving scenarios, because for running multiple trackers in parallel, target track's motions cues must be independent to each other. While, in urban areas, road participants tend to move in formation-like motion due to traffic conditions. Although, different road participants have different locations in the environment, their velocity and acceleration changes depending to each other and even may be identical to each other thus, their motions are highly correlated. This could be the main challenge for association of multiple measurements to multiple tracks.

The Joint Probabilistic Data Association Filter is an extension of the PDAF algorithm that operates under the assumption that the tracking is taking place under the clutter. In this clutter, the situation may happen that different tracks share the same measurement as depicted in Figure 2.10. Thus JPDAF may be assumed as the clutter-aware [11] version of the PDAF. The joint term comes from the fact that the algorithm uses inclusion global association hypotheses to calculate weighting parameters. The event that a measurement corresponds to a detected track is mutually is mutually exclusive among all of the tracks but this event is not necessarily mutual independent. Therefore, in a probabilistic view, the optimal filtering method that is used to tackle this problem, must consider all of the tracks. In order to tackle this problem some assumptions must be made:

- The number of already established target tracks is known as a priori.
- Any measurement of any target track can fall in the association region corresponding to another track and may cause persistent interference.
- An approximation consisting of some sufficient statistics of past situation of the system states like conditional means and covariance matrices of each target tracks are given to summarize the past situation of the whole system.
- Each target must have a dynamic motion model and measurement model.
- The states are assumed to be Gaussian distributed with means and covariances.

[11]The clutter-aware algorithm refers to a algorithm prepared to detect clutters and prevent the resulting faults.

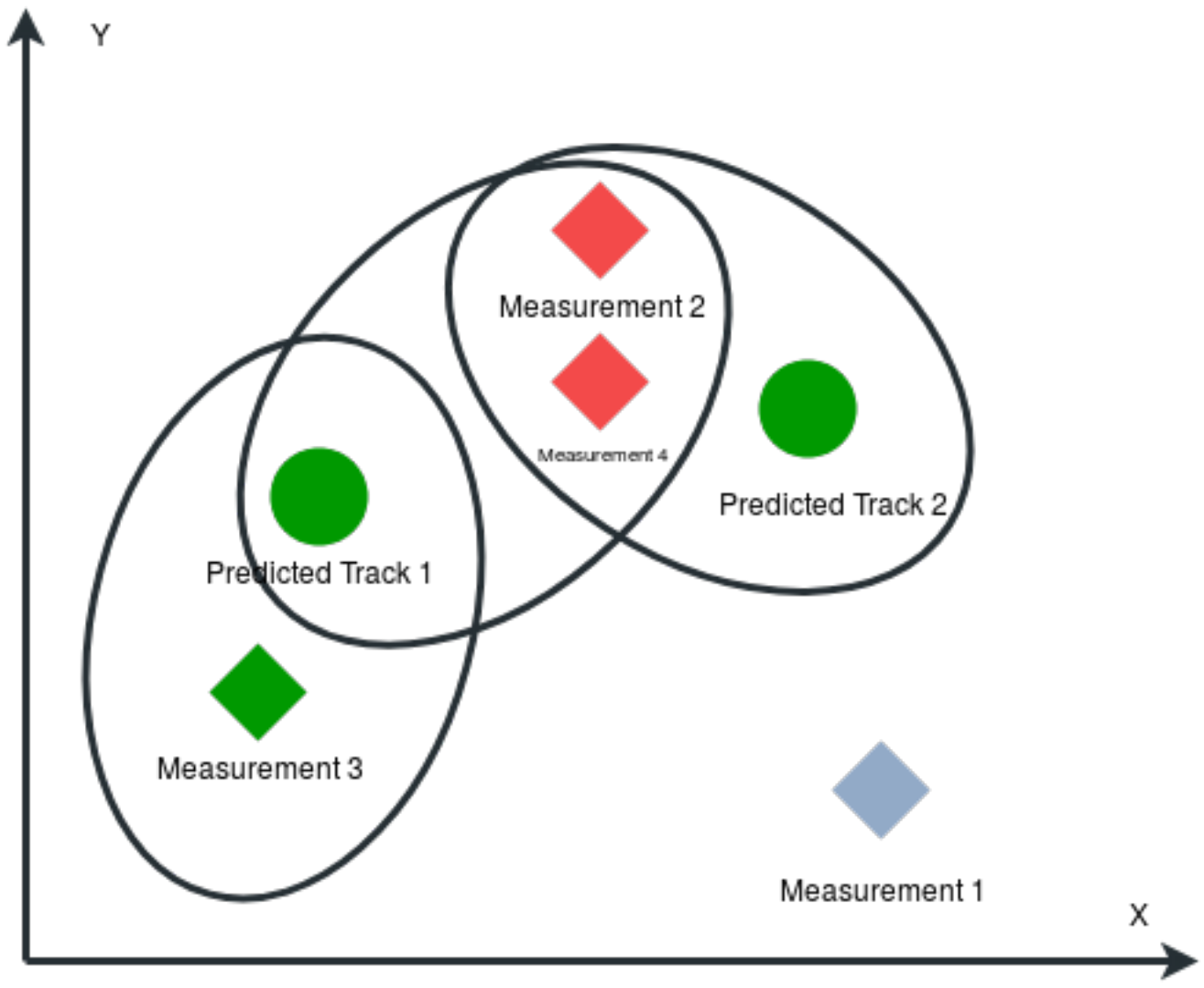

**Figure 2.10:** Gating procedure with clutter situation. Note that in this situation, measurement number 2 and 4 can be assumed to be associated with both tracks. wrong association in this situation may lead to wrong estimation of the states and even track lost.

Assume that there are some established tracking targets $t$ and some measurements $j$.The Gaussian likelihood of associating $j$ with $t$ in a joint event $\theta$ is given

by:

$$A_{j,t} = \frac{e^{\frac{-1}{2}(z_{t,j|k}-\hat{z}_{t,j|k-1})^T S_{t,j})^{-1}(z_{t,j|k}-\hat{z}_{t,j|k-1})}}{\sqrt{det(2\pi S_{t,j})}} \tag{2.68}$$

Therefore, the joint association probabilities $\gamma(\theta)$ would become:

$$\gamma(\theta) = \frac{\phi!}{V^{\phi}} \prod_j A_{t,j}^{\tau_j} \prod_t (P_d^t)^{\delta_t}(1-P_D)^{1-\delta_t} \tag{2.69}$$

where $V$ is the volume of the elliptical region in which the measurements are not considered to be in association with a target. These free measurements are assumed to be uniformly distributed. $\phi$ is the number of false measurements in event $A$, and $\tau$ is the binary measurement indicator. The parameter $\delta$ is binary target detection indicator. In order to provide an example about how these parameters are detailed in action, consider Figure 2.10. Feasible joint associations are: $\theta_{31}$ association of the measurement 3 to predicted track 1, $\theta_{22}$ as the association of the measurement 2 to track 2, $\theta_{42}$ as the association of the track 2 to the measurement 4. Then the binary target detection indicator for the target 1 becomes $\delta_1 = 1$ and for the target 2 $\delta_2 = 1$. The parameter $\phi$ or the number of false measurement is equal to $\phi = 1$. This procedure also applies for other feasible joint events where $j = (1, ..., m_k)$ and $j = (1, ..., N)$. States are assumed to be mutually independent while conditioned on the past observation. Thus, the JPDA association probability $\beta$ can be computed by simply marginalization of all feasible association probability $p(\theta_k|z_k)$ :

$$\beta_{j,t}(l) = \sum_{\theta_{j,t}\in\theta} p(\theta_k|z_k) \tag{2.70}$$

In which $p(\theta_k|z_k)$ is given by:

$$p(\theta_k|z_k) = \frac{\gamma(\theta)}{\sum_\theta \gamma(\theta)} \tag{2.71}$$

# 2.9 Outcomes

Until now, a literature review on different aspects of methods utilized in DaTMO has been conducted. Choosing corresponding method for each step of the whole DaTMO algorithm heavily depends on the application and desire of the designer. In what follows, applied methods for urban scene driving detection and tracking of multiple objects are detailed. This analysis is conducted in two different module: detection and tracking modules. In the detection module a novel method for efficient ground segmentation in rough urban environments is presented. The tracking module also consists of a novel modality of different methods which enables an efficient total solution for the DaTMO problem.

Necessary fundamentals of establishing a probabilistic framework for detection and tracking of multiple objects in urban environments have been discussed already and in details. In Chapter 3, a novel real-time method based on Gaussian process methodology is represented in which two jointly dependent Gaussian processes are used to estimate ground in continuous time. Also an input-dependent method is proposed. In Chapter 5 and Chapter 4, steps toward real-world implementation of these methods are discussed in details. System design and details of algorithms are discussed for both detection and tracking module while in Chapter 5 real-world implementation and results are presented. End of each chapter introduces some evaluation metrics for the algorithm package and our results are represented there to compare to other methods. Furthermore, Chapter 6 includes conclusion and further possible studies.

# Chapter 3

# Physically-Motivated, Real-Time, 3D Ground Segmentation: A Gaussian Process Approach with Local Characteristic Estimation

Autonomous Land Vehicles (ALV) shall efficiently recognize the ground in unknown environments. A Gaussian process based ground segmentation method is proposed in this thesis, which is fully developed in a probabilistic framework due to implementation of Bayesian inference. The data is segmented using a radial grid map. Two joint Gaussian processes are introduced to separately model the observation and local characteristics of the data. While, observation process is used to model the ground, the latent process is put on length-scale values to estimate point values of length-scales at each input location. Input locations for this latent process are chosen in a physically-motivated procedure to represent an intuition about the ground condition. Furthermore, an intuitive guess of length-scale values is represented by the assumption of the existence of hypothetical surfaces in the environment, that every bunch of data points may be assumed to be resulted from observations of these surface.

Bayesian inference is implemented using ***maximum a posteriori criterion***. The log-marginal likelihood function is assumed to be a multi- task objective function, to represent a whole-frame unbiased view of the ground at each frame. Simulation results shows the effectiveness of the proposed method even in an uneven, rough scene which outperforms similar Gaussian process based ground segmentation methods. While adjacent segments do not have similar ground structure in an uneven scene, the proposed method gives an efficient ground estimation based on a whole-frame viewpoint instead of just estimating segment–wise probable ground surfaces.

In the last decade, neural network methodology has been widely used for regression or classification problems. The main contributions of neural network methodology versus other routines in the statistical literature is their ability to tackle non-linear statistical problems while showing amazing flexibility facing different problems from different fields of research. Other decision-based or kernel-based methods failed to do the same in that regard. Flexibility of Neural Network methods comes at a price of more computational complexities as many parameters of the model should be determined from the data itself which can further induce over-fitting issues to the problem. While there has been methods introduced to overcome the over-fitting problem they are not fully reliable and the risk may be carried on the solution.

Bayesian methods on the other hand, tackle this problem by setting up themselves as a method to specify a hierarchical model. Bayesian framework uses the input data to construct an observation model and then utilizes this observation model as a reliable connection between two distributions: first they put a prior distribution on hyper-parameters which then they tend to specify the prior distribution of weights or function based on them. For example, in a regression problem assume that the observation is corrupted by an additive Gaussian noise. With arrival of an observed data, a posterior distribution model will be assigned to weights and hyper-parameters while in neural network methodology these posterior is not analytically tractable and computational method may be used instead, that are prone to produce results containing approximations. A very natural way to confront regression problems is to assign a prior distribution on all kind of the functions that we expect to observe and try to obtain a posterior after observation is happened, Therefore Gaussian Processes (GPs) are able to provide a natural way to induce non-parametric priors over regression functions.

In this thesis we see Gaussian Process Regression from the supervised learning point of view. In the machine learning literature the problem of supervised learning is divided into two different areas: *1- Regression* and *2- Classification*. The output for classification problems are discrete labels but in regression problems the output are predictions of continuous quantities. A detailed review of all concepts regarding Gaussian Process Regression ($\mathcal{GPR}$) that is utilized in this thesis is gathered in Appendix A. No further discussions are made about mathematical foundations of used materials in the following sections.

## 3.1 Introduction

The technology trends shows more interest in Autonomous Land Vehicles (ALV) with the growth of research interest into this subject. ALVs are able to provide many opportunities from empowering the ability of remote exploration and nav-igation in an unknown environment to establishing driver-less cars that are able to navigate autonomously in urban areas while being more safe-driven by compensating human driving faults. In order to establish a driver-less car capable of performing autonomously in urban areas, developed methods shall be reliable and real-time implementable.

Ground segmentation represents itself as a vital component of any algorithm pursuing further tasks in an unknown environment. A reliable ground segmentation procedure shall be applicable in environments with both flat and sloped terrains, while being realistic and real-time implementable, since often this task is only a prerequisite for other time consuming algorithms and it is an important basic part of ALV's perception of it's surroundings. Gaussian process regressions are useful tools for implementation of Bayesian inference which relies on correlation models of inputs and observation data [99]. They provide fast and fully probabilistic framework for non-linear regression problems. Although light detection and ranging (LIDAR) sensors are commonly used in ALVs, data resulted from these sensors, does not inherit smoothness, and therefore, stationary covariance functions may not be used to implement Gaussian regression tasks on these data.

Often ground segmentation in this scheme and with rich laser range data, involves prediction or construction of 2D model of the ground from rich 3D data which comes with a need for a high computational effort. Therefore it seems necessary to somehow reduce the complexity of the problem while saving computational resources or even improving the precision of the procedure. In this thesis, problem of ground segmentation is represented with the use of Gaussian Process Regression (GPR). The two-dimensional ground segmentation problem is reduced to many one-dimensional regression tasks in order to reduce the computational complexity while keeping the precision of ground model prediction. Polar Grid Map is used to partition the raw input data into many uneven bins. A candidate ground point is obtained in every bin and consequently through them in all of the space. Using these candidate ground points, a continuous estimation of the ground is represented using joint Gaussian process regression.

## 3.2 Review of Other Ground Segmentation Methods

Different methods for segmentation have been proposed in the literature [100, 101, 102, 103, 66, 1, 104]. In reference [100], ground surface is obtained in an iterative routine, using deterministically assigned seed points. In reference [101], the ground segmentation step is put aside to establish a faster segmentation based on Gaussian process regression. A 2D occupancy grid map is used to determine surrounding ground heights, and furthermore, a set of non-ground candidate points are generated. Reference [102] handles real-time segmentation problem by differentiating the minimum and maximum height map in both rectangular and a polar grid map. In reference [103] a geometric ground estimation is obtained by a piece-wise plane fitting method capable of estimating arbitrary ground surfaces. In reference [66] a Gaussian process based methodology is used to perform ground estimation by segmenting the data with a fast segmentation method firstly introduced by [1]. The non-stationary covariance function from [105] is used to model the ground observations while no specific physical motivated method is given for choosing length-scales. Reference [104] proposes a fast segmentation method based on local convexity criterion in non-flat urban environments.

These methods are either estimating the ground piece-wise and with local viewpoint or by labeling all the individual points with some predefined criterion. Except the method proposed in reference [66] none of the methods above, give a continuous model for predicted ground. Furthermore none of them give an exact, physically-motivated routine to extract local characteristics of non-smooth data, while efficient ground segmentation have to be done considering physical realities of the data including non-smoothness of the LiDAR data and ground condition in every data frame.

Due to the ability of Gaussian processes ($\mathcal{GP}$) to model correlations between data points, there is a growth of interest seen in the literature to use them with 3D point clouds. Furthermore, $\mathcal{GP}$s are capable of estimating functional relationships by considering correlations between observations and data points, even when no model is available and the function is prone to huge changes. The correlation is introduced to $\mathcal{GP}$s with covariance kernels. Covariance functions are key concepts in Gaussian process regressions as they define how data points relate to each other. Specification of covariance structures is critical specially in non-parametric regression tasks [106].

LiDAR data is consisted of three-dimensional range values which is collected by a rotating sensor, strapped down to a moving car. This moving sensor obviously causes non-smoothness in its measured data, which may not be taken into account using common stationary covariance functions. Although a Gaussian process based method for ground segmentation with non-stationary covariance functions is proposed in the literature to take input-dependent smoothness into account in reference [66], adjusting covariance kernels to accommodate with physical reality of the ground segmentation problem needs further investigation. Length-scales may be defined as the extent of the area that data points can effect on each other [107]. Length-scale values plays a significant role in the quality of the interpretation that covariance kernel gives about the data. A constant length-scale may not be used with LiDAR data due to non-smoothness of collected point cloud.

Different methods are proposed to adjust length scales locally for non-stationary covariance functions by assuming an exact functional relationship for length-scale values [108, 109, 110]. The ground segmentation method proposed by [66] assumes the length-scales to be a defined function of line features in different segments. This is not sufficient because no physical background is considered for the selection of functional relationship and this function might change and fail to describe the underlying data in different locations.

In this thesis, two Gaussian processes are considered to jointly perform the ground segmentation task, one to model height of the ground and the other to model length-scale values. A latent Gaussian process is set on the logarithm of length-scale values. Point estimates of length-scale values at each particular ground candidate location is calculated using a multi-task hyper-parameter learning scheme. Local estimation of length-scale values enables the method to consider both flat and sloped terrains. Furthermore, a whole-frame intuition about ground quality of each frame is injected into the optimization task by special treatment of selection process for pseudo-input set. Proposed method is tested on KITTI [61] data set and is shown to outperform similar Gaussian process based ground segmentation methods with much lower estimation error.

## 3.3 3D Point Cloud Segmentation With Polar Grid Map

In this thesis we use a method for 3D data segmentation which is computationally efficient and can be considered as a "quick segmentation step". This segmentation step must be quick because this is actually considered as a primary necessity for further segmentation task that is going to be implemented on the output of this level to classify or further segment other subjects or even the ground in our case, and these tasks are time and computational effort consuming themselves.

**Polar Grid Map Construction** The raw 3D LiDAR data is constructed of unorganized three-dimensional range data which we represent here by unordered set $P_t = P_1, P_2, ..., P_{N_t}$. This set consists of three-dimensional points represented by their euclidean coordinates $P_i = (x_i\ y_i\ z_i)^T$ reported with respect to the ego-coordinate system at the vehicles center of gravity with $y - axis$ pointing directly toward vehicle path and $z - axis$ pointing upward. In this thesis all the points that are collected at one scan are treated as they where collected by the sensor at the same exact time and no delay is considered in the data gathering process.

Often in other methods, the segmentation begins by binary labeling of all the points in each frame of measurements. A ground or non-ground label is used, which is indicating for each point whether it is a part of ground plane or not. In self-driving cars scenario local surface properties must be taken into account to consider both sloped and flat terrains. Therefore, some neighboring feature structures must be imposed on the data which is often not applicable in real-world situations because our scan of the environment carries so many points in it that needs a special partitioning treatment to be able to handle this kind of data.

In reference [1] an efficient method for partitioning 3D point cloud data is proposed. This method begins by assuming the $xy - plane$ to be a circle with infinite radius $r$. Then this radius is divided into several segments, although this method results in a new segmentation criteria and successfully divides the whole 3D space into several segments but 3D points are still unordered.

Thus, the unordered 3D point cloud set, will be organized due radial distances of the points from the origin of the coordinate system. This is done with the introduction of different bins in every segment. The parameter $\Delta\alpha$ is introduced as the only parameter that circular segmentation of the $xy - plane$ is dependent on. In fact $\Delta\alpha$ is the angle that every segment covers in the space. There would be a total number of $M = \frac{2\pi}{\text{ffl}\Delta\alpha}$ segments in this setup which are called $S_i : i = 1, ..., M$. Using this segmentation method enables the transformation of a two-dimensional ground segmentation problem into many one-dimensional $\mathcal{GP}$ regression problems in each segment and thus the rich LiDAR data can be processed in real-time. On the other hand making use of locally adaptive Gaussian processes with adapted length-scale in each segment of the circular polar grid map, enables the method to benefit from local necessary information of the data rather than rough approximations of global behavior.

This methodology also benefits from two facts: first the polar grid map despite of other methods, does not lose data when reduces the presentation from 3D to 2D and in this manner it retains all the valuable informations from the measurements. Thus the under-segmentation problem is avoided. Second, the polar grid map fits into the physical specifications of our sensing device. The polar grid map enables to consider both *sloped* and *flat* terrains and also the transition between them by calculating the sloped lines in each segment for candidate ground points. We tend to project all the points in our three-dimensional raw data in each frame to the x-y plane. We will further omit all the points beyond the radius *B* and assume that they are placed at infinity as depicted in figure 3.1.

### 3.3.1 Assigning M Segments

The circular $xy - plane$ has limited radius of $R_{def}$ and thus all the points with horizontal distance to the $xy - plane$ origin exceeds the value of $R_{def}$ are omitted from the data. The residual points are mapped into the $M$ different segments according to the angle that they make with the positive $y - axis$ of the vehicle coordinate system to have a polar grid map that is evenly split into M segments counter clock wise as depicted in figure3.2. The index to a segment that a certain point maps to is represented by $Segment(p_i) = S(p_i)$ and is calculated as:

$$Segment(p_i) = \left\lceil \frac{Angle(x_i, y_i)}{\Delta\alpha} \right\rceil \tag{3.1}$$

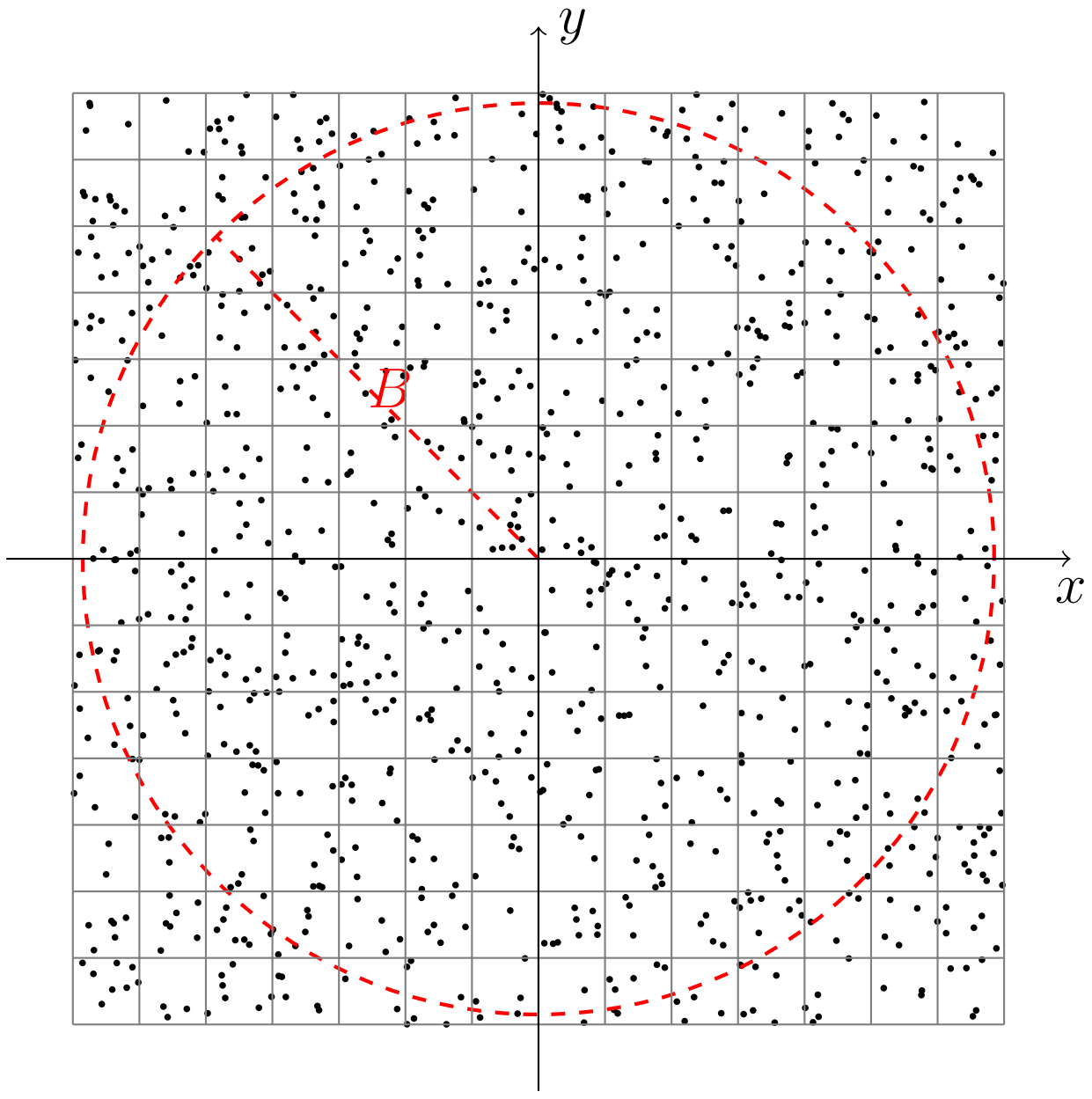

**Figure 3.1:** After projection, two-dimensional point cloud data is distributed on the $x - y\ plane$. All the point out the circle border with radius $B$ are neglected.

Where the term $Angle(x_i, y_i)$ represents the horizontal angle between positive $y - axis$ and the point $OP_i$ and is calculated with the use of $atan2$ routine function as $atan2(x_i, y_i)$ and it is sured to lie in $(0, 2\pi)$. $[.]$ is the rounding function. After that respective indexes for each point in each frame is obtained, a new set must be constructed that contains all the points belonging to the segment $m$. This set is denoted by $P_m$:

$$P_m = \{p_i \in P_t | S(p_i) = m\} \tag{3.2}$$

Even after these steps, all the points are still in three-dimensional representation bu t they are sorted due to their projected angular components in the space and into the different segments $S_m$.

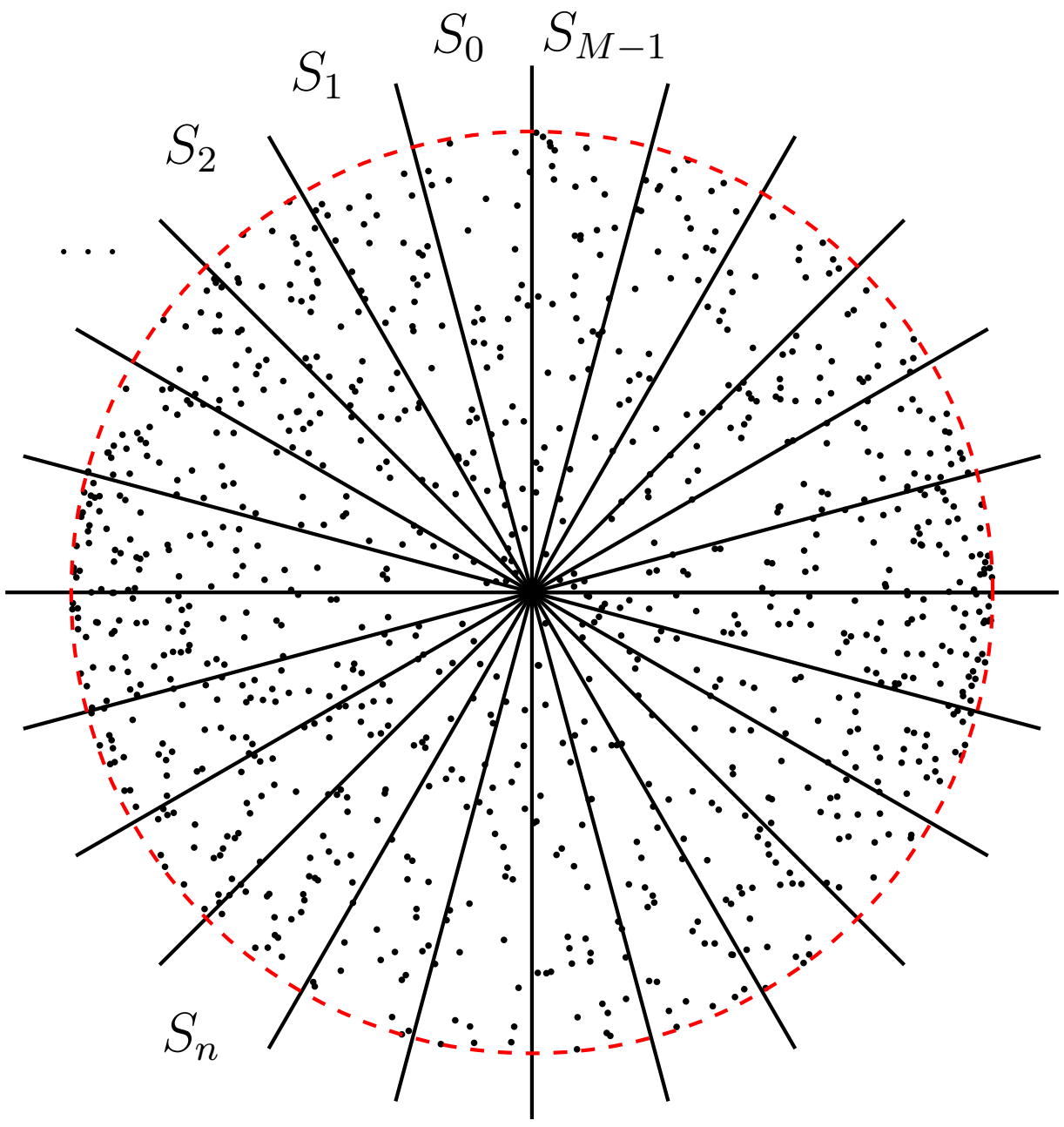

**Figure 3.2:** Different segments of each frame $S_m$. Each segment contains all the points withing specified angular range.

### 3.3.2 Assigning Bins

Until now, the three-dimensional space is divided into different angular segments. Each segment must further be divided into different bins. This bins must be arranged in radial order. This radial arrangement will enable us to implement two-dimensional line extraction algorithms on this segments. In order to proceed to the further segmentation of the bins in each segment, each particular segment will be unevenly divided into $N$ bins that results in discretization of range components in each bin. In each segment $n_{th}$ bin which is represented as a set by $b_n^m$, covers the range from $r_n^{min}$ to $r_n^{max}$. Thus, three-dimensional points $p_i = (x_i\ y_i\ z_i)^T$ will be mapped into different bins according to their radial coordinate as depicted in figure3.3. Shortly the point $p_i$ will be mapped into the bin $b_n^m$ if and only if its radial coordinate $r_i$ satisfies the following inequality:

$$r_n^{min} < r_i = \sqrt{x_i^2 + y_i^2} < r_n^{max} \ \wedge p_i \in P_m \tag{3.3}$$

The corresponding bins in different segments are the same in the sense of the range they cover but they differ in the sens of the corresponding point cluster that is mapped to them. The other important aspect of this kind of partitioning of the data is that it is compatible with the physical model of LiDAR data as the measurements collected with LIDAR sensor are sampled unevenly with respect to the ground and thus they become exceedingly sparse with increase of distance from origin of sensor. Furthermore, data may contain a lot of voids resulting from the occlusions. The set $P_{b_n^m}$ is constructed to represent all the 3D points that are mapped into the $n_{th}$ bin of the segment $S_m$. For every set $P_{b_n^m}$ there is a one-to-one set $P'_{b_n^m}$ of corresponding 2D points with $r_i$ being $\sqrt{x_i^2 + y_i^2}$ is the radial range of corresponding points:

$$P'_{b_n^m} = \{p'_i = (r_i, z_i)^T \mid p_i \in P_{b_n^m}\} \tag{3.4}$$

It is aforementioned in this thesis that complex three-dimensional data is transformed into the simpler two-dimensional representation which further enables us to reduce the two-dimensional ground segmentation problem into several one-dimensional regression tasks which are held in each segment of the new representation of the data. Thus we introduce the new set $PG_m$ to be the set containing all of the ground candidate points in each segment. In order to fully construct the set $PG_m$, in each bin of each segment, we take the point with the lowest $z_i$ from $P'_{b_n^m}$ to be the ground candidates in that bin, and thus all of the $p'_i \in P'_{b_n^m}$ in each segment, are chosen to be the elements of $PG_m$ as follows:

$$PG_m = \{p'_i \mid p'_i \in P'_{b_n^m}, z_i = min(\mathcal{Z}_n^m)\ for\ n = 1,..,N\} \tag{3.5}$$

Where $\mathcal{Z}_n^m$ denotes all of the $z - coordinates$ of all the points in the $n_{th}$ bin of the segment $m$.

Taking all of this actions until now, made it possible to ensure that all the points are ordered with respect to their ascending radial range value. Obviously the point to range reduction is not a one-to-one procedure while as we said before the three-dimensional bin to two-dimensional bin is a one-to-one procedure and even more $P'_{b_n^m}$ may be a empty set and we made sure to take this assumption in to consideration in our work. Instead of taking each bins lowest height point as the ground candidate, we can establish new measures but taking lowest height point shows a good result for ground segmentation purposes because these ground candidates best likely lie in a ground plane. partitioning data in this fashion brings us a very unique benefit which is that the complexity of the ground plane estimation is not dependent on the size of the original point cloud anymore and it is only a function of number of segments and number of bins in each segment.

## 3.4 Line Extraction

Line segments are known as the simplest geometric primitives. These creatures make it easy to describe even complex environments. Actually many algorithms have been proposed for line extraction task from two-dimensional range data. In this section, effort is taken to compare some mostly used line extraction algorithms and show why it is not efficient in this thesis's approach to use RANSAC which is more renowned in the literature[111]. As one may notice, there are three main problems concerning the task of line extraction in an unknown environment:

- Is it always possible to find a line and how many lines are there in the target set of points?
- What is the criterion to associate a point to a certain line model?
- How should one estimate the line model parameters?

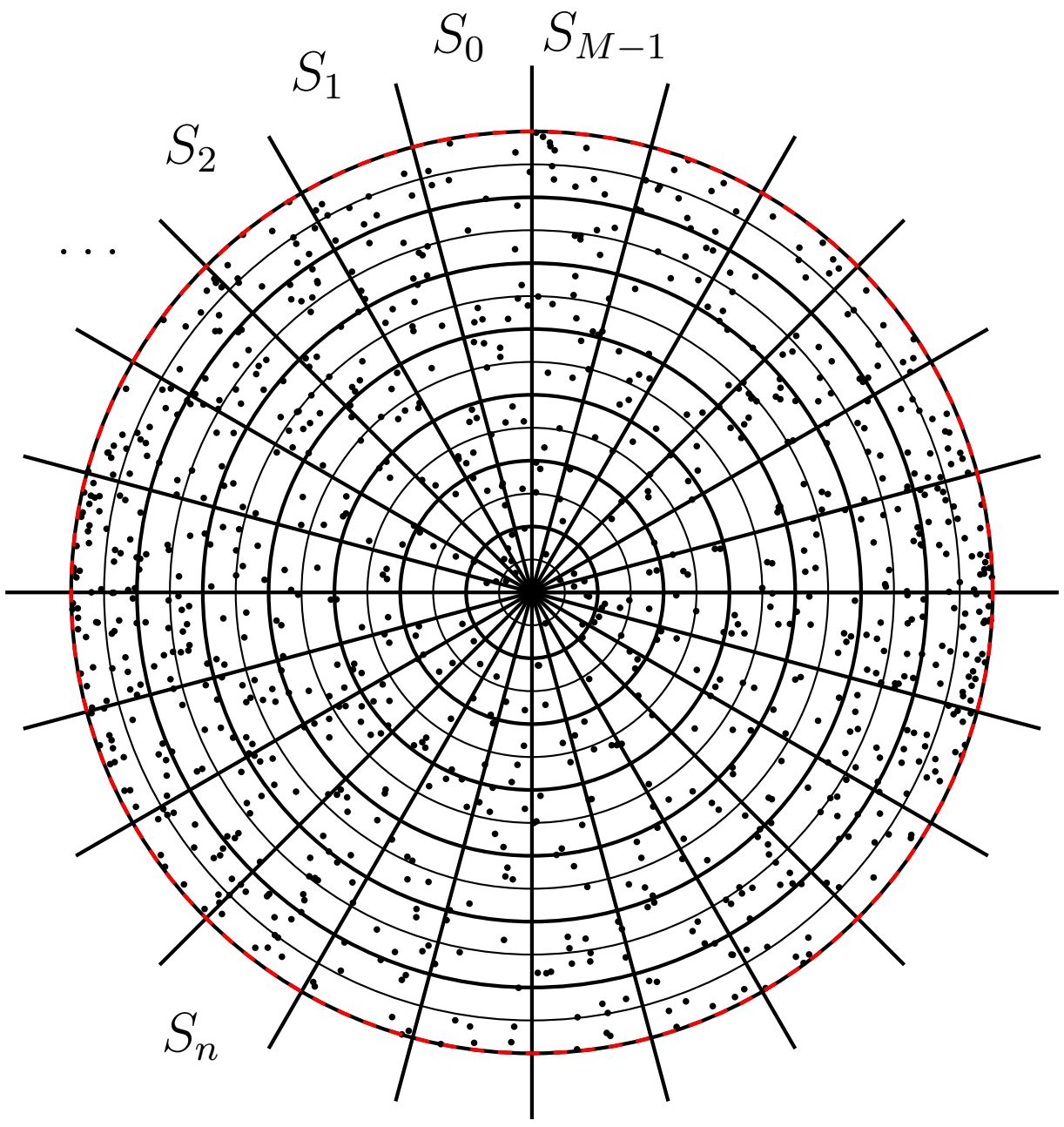

**Figure 3.3:** Each segment $S_m$ is divided into $N$ bins. Note that these bins have the same radial range, but actually they have bigger size as we move far from the origin. This feature is suitable for LiDAR data as it gets more sparse in long distances, thus less hesitation might be needed.

We should try to give an acceptable answer to these question during this section. Starting from the latter, the method we have used for calculating line parameters is called $total\ least\ squares$ method which is widely used in the literature and is represented in the proceeding parts.

A detailed comparison between common line extraction algorithms is adapted from [111] in order to modify the needed line extraction algorithm for this thesis.

These algorithms, and the reason why the simple *incremental algorithm* is chosen over the well-known, common *RANSAC algorithm* is discussed in Appendix D.

### 3.4.1 The Proposed Line Extraction Algorithm

In the previous sections the 3D LiDAR data is divided into meaningful and limited parts. Thus, our method is capable of controlling the run-time by choosing the appropriate values for parameters like $\Delta\alpha$ and $N$. After the implementation of the polar grid map, segments of two-dimensional organized data representing ground candidates in each angular part of the environment are available. Having this ground candidate points along each segment, enables us to further implement line extraction algorithms to estimate one-dimensional candidates for the ground plane in each bin of each segment. First we will describe what we will consider as a ground plane. Consider the line $y = mx + b$. what should the parameters of this model represent to be considered as a part of our ground plane?

1. Although the ground structure is assumed to consider both the sloped and flat terrains, ground candidates are not expected to show vertical structures. In order to implement this fact into the line extraction algorithm a threshold is put on each line's slope $m$. Slope values must not exceed a certain value $\zeta_m$.

2. Ground planes are not expected to show a high level zero slope area or to show a very flat area in high places. In urban scene driving areas the data is not to suddenly show a plateau. A threshold for smaller slope values must be imposed to the algorithm. If a certain lines' slope values fall beneath the value of $T_{m_{small}}$, a sure criterion that lines' absolute $y - axis$ intercept $b$ must not exceed a certain threshold called $\zeta$.

3. The root mean square error of the fit must not exceed a certain threshold $T_{RMSE}$.

**Least Square Fit Modeling for Line Extraction** A straight line in a two-dimensional plane is represented by $ax + by + c = 0$. Then in every segment $m$, The process of line detection is started with the first two points and a line is

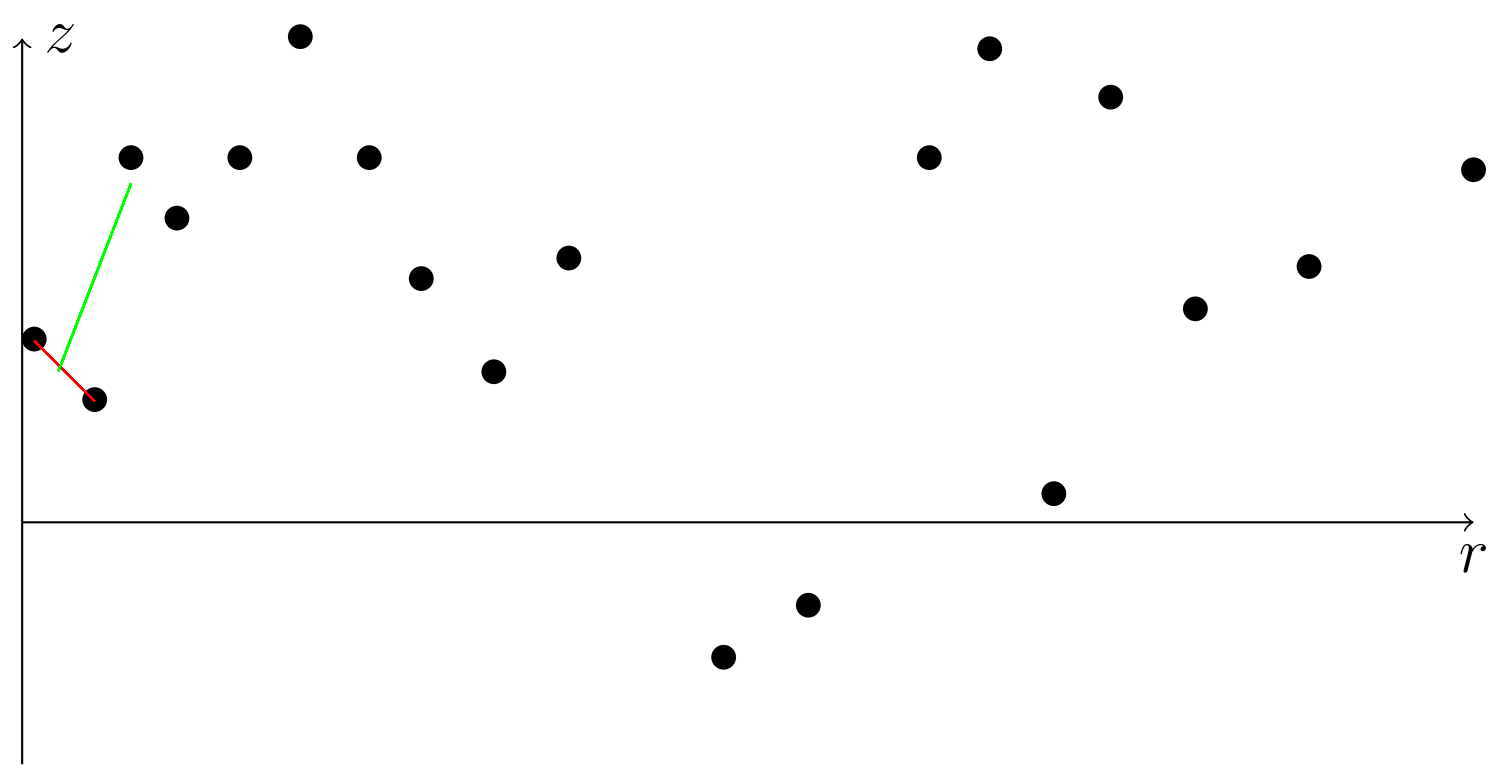

**Figure 3.4:** Ground candidate points of a certain segment are mapped into two-dimensional set of $(z, r)$ with $z$ representing the height and $r$ the radial distance of the origin. In this figure, for instance line parameters for the first two points are calculated using equation (3.6) and are shown with the red line. After adding the third point, line parameters are calculated again which is shown with the green line. As those line parameters does not fit, the algorithm decides to establish a new line.

constructed by the following equations which are already a least square fit:

$$\begin{aligned} a &= \left(\sum_i^n x_i \sum_i^n y_i^2\right) - \left(\sum_i^n y_i \sum_i^n x_i y_i\right) \\ b &= \left(\sum_i^n y_i \sum_i^n x_i^2\right) - \left(\sum_i^n x_i \sum_i^n x_i y_i\right) \\ c &= \left(\sum_i^n x_i y_i\right)^2 - \left(\sum_i^n x_i^2 \sum_i^n y_i^2\right) \end{aligned} \tag{3.6}$$

Above equations may be used to compute line parameters for any number of point that is required to. After obtaining line parameters for the first two point, the third point is added to the first two and the parameters are recalculated. These calculations are done for all of the three points at this time. Then, if the new line models satisfies the current line condition we move forward to the next point in the raw, but if the lines condition were not satisfied, we omit the last point and recompute the line parameters and return the line value. This procedure will be continued each time from the first point in the raw that was not considered in the last line model as depicted in algorithm 1. This fast and efficient method is actually implementable for our method due to the existence of an ordered ground candidate set for each segment.

Algorithm 1 represents a fast method for line extraction which does not suffer from under-segmentation too. The main contribution of this novel method

**Algorithm 1** Line Extraction in segment m

**INPUT:** $M$, $N$, $\zeta_m$, $\zeta_b$, $\zeta_{RMSE}$
**OUTPUT:** $PG_m$, $L_m$
$k = 0$
**for** $j = 1\ to\ M$ **do**
    **for** $i = 1\ to\ Size(PG_m(j))$ **do**
        **if** $P'_{b^m_n} \neq \emptyset$ **then**
            $PG_m(j)$ = Set of all ground candidate points in segment(j)
            $d_L = PG_m(j)$
            **if** $|d_L| \geq 2$ **then**
                $(a_k, b_k, c_k) = Fitline(d_L)$
            **end if**
            **if** $|-\frac{a_k}{b_k}| \leq \zeta_m$ **then**
                $L_m(j) = L_m(j) \cup \{a_k, b_k, c_k\}$
                $k = k + 1$
                $d_L = \emptyset$
            **end if**
        **end if**
        $D_j = GradientFilter(L_m)$
    **end for**
**end for**

shall be the the two-dimensional presentation of environment in the domain of un-organized 3D points. Extracting the ground plane lines $L_m = \{(a_i, b_i\})$ in each segment is based on incremental method which is represented in algorithm [15]. The line fitting part has utilized the total-least-square method pf equation3.6. Having our lands on the lines $L_m$ which are describing the ground plane withing one segment of our polar grid map, we may now have a criterion to label the points in the segments and thus divide ground and non-ground points. The line extraction algorithm is actually going to be implemented on all the *ground candidate* points in each segment. *ground candidate* points are the points with the minimum height in every bin of each segment. The comparison thresholds for these lines are given to the algorithm as the parameters $\zeta_m$ or the maximum value for the lines slope, because we do not expect our ground points to show a vertical specification and $\zeta_b$ as the minimum threshold as we do not expect our ground points to be very flat or show a plateau area.

The process of line extraction is shown as schematic in figure (3.4). First, algorithm calculates the line parameter for the first two points, then adds the third point and once again calculates the parameters for the set of these three points and finally compares these two line parameters and if it fails to find a significant di-version, it assumes these points to share the same line parameter that is calculated recently. On the other hand if it finds a significant diversion in line parameter values, it returns the previous line parameters as the line parameters for the points except the last one, and again starts the calculation from new starting point to find the next line.

## 3.5 Ground Segmentation: One Dimensional Regression Method

Although Gaussian process preliminaries are mentioned in the Appendix A, some matters are going to be re-stated here in order to fulfill problem presentation needs. Nonlinear Gaussian process regression problem is to recover a functional dependency of the form $y_i = f(x_i)+\epsilon_i$ from $n$ observed data points of the training set $\mathcal{D}$. The set $PG_m$ contains all of the two-dimensional ground candidate points in the segment $m$ that we suspect them to be a part of the ground. However, this set might be contaminated by the obstacle points until we make sure it is not by somehow filtering the set and putting obstacle points away. After the pre-processing steps, the three-dimensional data is reduced to the current two-dimensional pseudo input data. Now in every segment of this polar grid map a one-dimensional ground model will be constructed using Gaussian process regression that yields the predictive distribution of the ground heights at arbitrary locations of the segment. The set $PG_m\{(r_i, z_i)\}$ contains locations of suspected ground points. The outliers in these sets are going to be omitted and a residual sample set $\mathcal{R}_m\{(r_i, z_i)\}_{i=1}^{n}$ is going to be constructed for each segment that contains only certain ground points. Using this residue points the predictive ground model $P_m(z_*|r_*, \mathcal{R}_m)$ is built for segment $m$ which will give us in return the predictive distribution of the height $(z_*)$ at the arbitrary test input location $r_*$.

While these outlines are not gonna change in different problem setups, three different inference method is introduced in this thesis for this task: Z-only method, $Z\mathcal{L}$ method and $Z\mathcal{L}\mathcal{I}$ method. These three different methods differ in how they involve Gaussian process regression task into the solution. In the Z-only method, no local characteristic of the data is taken into account. Ground segmentation problem is assumed to be a non-linear Gaussian regression problem. In the $Z\mathcal{L}-method$ To joint Gaussian processes are assumed to model the ground to include input-dependent smoothness of the data. In the $Z\mathcal{L}\mathcal{I}-method$, three joint Gaussian processes are assumed to include both input-dependent noise and smoothness. These three methods are discussed in details in the proceeding sections, while their gradient evaluation calculations are detailed in Appendix C.

## 3.6 Ground Segmentation With Z-Only Process

In the Appendix A, the $\mathcal{GP}$ regression procedure is completely described. The main purpose of the regression task is to learn a model for the ***predictive distribution*** $P(z_*|r_*, \mathcal{R}_m)$ of new target values $z_*$ at the location of $r_*$ given the training set $\mathcal{R}_m$. It was shown that according to the definition of Gaussian process the joint distribution of the training samples can be expressed as Gaussian process as well. In fact in our problem setup, the set $\mathcal{R}_m\{(r_i, z_i)\}_{i=1}^n$ is our training set of $n$ samples and the joint distribution of these training samples can be written as:

$$P(Z|\mathcal{R}) \sim \mathcal{N}(\mu, K) \tag{3.7}$$

Where $Z = [z_1, ..., z_n]^T$ is the corresponding training heights vector and $\mathcal{R} = [r_1, ..., r_n]^T$ is the corresponding training radial coordinate vector as well. The mean vector is represented by $\mu$ which is often taken to be zero and the covariance matrix is represented by $K$ which represents the relationship between the random variables and was described for the isotropic and stationary case in Appendix A. Furthermore, often for isotropic and stationary stochastic processes, the squared exponential covariance matrix is used which is discussed in details.

### 3.6.1 Covariance Function

Matrix $K$ is often specified in terms of a covariance kernel $k$ and the noise variance $\sigma_n^2$ is assumed to be independent, normally distributed noise term with constant noise variance parameter as follows:

$$K(r_i, r_j) = k(r_i, r_j) + \sigma_n^2 \delta_{ij} \tag{3.8}$$

Where $\delta_{ij}$ is the infamous Kronecker delta function. The kernel function for the basic $Z - only\ method$, is the stationary, isotropic covariance function which is called squared-exponential covariance function:

$$k(r_i, r_j) = \sigma_f^2 \exp\left( - \frac{(r_i - r_j)^2}{2\sigma_l^2} \right) \tag{3.9}$$

Where $\sigma_f^2$ is the signal covariance and $\sigma_l$ is the length-scale. As it is aforementioned, the free parameters $\sigma_f^2$ and $\sigma_l$ are considered to be the hyper-parameters of the Gaussian process. The main drawback of using this covariance function, is the assumption of constant length-scale over all space, which is indeed incorrect in our problem setup, also another problem of using $squared\ exponential$ covariance function in ground segmentation problem is that this covariance function is not able to take the uneven distribution of length-scales in the environment into account, in fact this covariance function assumes that the length-scale is constant in the whole input space while in ground segmentation problem this assumption does not hold because of different terrains that vehicles might found themselves in. However in the first method this covariance function is used for $\mathcal{GP}$ regression task. The value for the length-scale $\sigma_l$ is chosen to be the value of extracted line multiplied by some constant values to give our covariance function a touch of meaning from physical reality, but not so precise at this step.

Therefore, to overcome these problems, the *non-stationary, isotropic* covariance function proposed by [112] is used to increase the accuracy of segmentation method:

$$K(r_i, r_j) = \sigma_f^2 (\mathcal{L}_i^2)^{\frac{1}{4}} (\mathcal{L}_j^2)^{\frac{1}{4}} \left( \frac{\mathcal{L}_i^2 + \mathcal{L}_j^2}{2} \right)^{-\frac{1}{2}} \exp\left( - \frac{2(r_i - r_j)^2}{\mathcal{L}_i^2 + \mathcal{L}_j^2} \right) \tag{3.10}$$

In the *Z-only method*, length-scale values are adapted assuming a constant value multiplied by a measure of *local characteristic*, which is obviously a lesser precise method comparing to *Z-$\mathcal{L}$* method. In [66] it is assumed that in order to calculate the characteristic length-scale for the input $r_i$ in the $PG_m$ set, it is efficient to find the lines in the set containing all the extracted lines of $PG_m$ (the set $L_m$ which is computed by the method of algorithm 1). The closest line in the set to the input point is obtained. The gradient of the corresponding line is assumed to as a characteristic value. Then the length-scale regarding to that point using this information is calculated as following:

$$\mathcal{L}_i = \begin{cases} a \times \log\left( \frac{1}{|g(r_i)|} \right) & If \ |g(r_i) > g_{def} \\ a \times \log\left( \frac{1}{|g_{def}|} \right) & otherwise \end{cases} \tag{3.11}$$

Where $a$ is just a scale parameter that have to be learned in hyper-parameter optimaztion part. According to the definition of Gaussian process, we can find the joint probability distribution of training output $z$ and the test output $z_*$ at the given test input $r_*$ by the following equation:

$$\begin{bmatrix} z \\ z_* \end{bmatrix} \sim \mathcal{N}\left( 0, \begin{bmatrix} K(R,R) & K(R,r_*) \\ K(r_*,R) & K(r_*,r_*) \end{bmatrix} \right) \tag{3.12}$$

The main idea of this method is that to model every finite set of samples $y_i$ from the process jointly Gaussian distributed such that the predictive distribution $P(z_*|r_*, \mathcal{R}_m)$ at arbitrary query point $r_*$ would be Gaussian too. Thus the predictive distribution for Gaussian process takes the following form:

$$\bar{z}_* = K(r_*, R) K(R,R)^{-1} z \tag{3.13}$$

$$v[z_*] = K(r_*, r_*) - K(r_*, R) K(R,R)^{-1} K(R, r_*) \tag{3.14}$$

Where $\bar{z}_*$ is the mean value of the test output and $V[z_*]$ is the covariance of test output and $K(R,R) = (K(r_i, r_j))\ with\ 1 \leq i,j \leq n$. Also the $n_{nh}$ element of $K(r_*, R) \in \mathcal{R}^{1\times n}$ is equal to $K(r_*, r_n)$. In the $\mathcal{GP}$ regression literature, covariance matrix plays a vital role as it defines the effective relationship between input and target values, thus one should be very hesitating while choosing a covariance function for certain regression task and must not forget to consider physical realities of the real-world problem, as in this simple model for $Z - only\ method$ we did this simply by assigning length-scale values with line parameters.

## 3.7 Ground Segmentation With Input-Dependent Smoothness (Z-$\mathcal{L}$ Joint Processes)

Different kind of terrains might be wended by an ALV, in different kinds of environments and applications. In this section a method based on Gaussian process regression is developed to consider both flat and sloped terrains in rough road driving scenes. This so called Z-$\mathcal{L}$ method, estimates ground for separated segments with an objective function specially treated to take whole-frame condition of ground into account. In Gaussian process regression methods, covariance function actually defines how the parameters are related to each other and how they are being affected by each other due to the mathematical model at different extents.

Although simple Gaussian process regressions are a very powerful tools for Bayesian inference, they fail to consider *local-smoothness*, since in their simplest version they assume to have stationary kernel functions. LiDAR data inherits *input-dependent smoothness* meaning that its data does not bear smooth variation at every part and direction of the environment, thus the stationarity assumption fails to describe this data precisely. Therefore, covariance functions with constant *length-scales* are not suitable for LiDAR point cloud since flat grounds must have a larger length-scale than a rough ground. The non-stationary covariance function originally represented by [105] is used for ground segmentation. Local characteristics of this covariance function may be calculated using local line features by assuming length-scales to be of the form $\mathcal{L}_i = ad_i$, where $a$ is assumed to be one of the hyper-parameters [66] or as called the Z-only method in previous section. Although this method is fast, the locally changing characteristics of the ground are not taken into account.

In this section, point estimates [107] of the length-scales are calculated locally and in real–time. Estimating length-scale values locally, takes into account the changing correlation features of points with respect to each other, and enables more realistic prediction of ground surfaces. Furthermore these point estimates are obtained with regard to a physical motivated procedure for pseudo-input col-lection which enables us to intuitively consider *ground quality* of each frame to produce more realistic results. While local estimation of length-scales ensures precision, the *whole-frame* treatment of objective function and pseudo–input selection ensures more realistic results with regard to real condition of ground in each frame. The proposed method outperforms ordinary LIDAR-based ground segmentation methods, while being more robust and real-time implementable.

### 3.7.1 Non-smoothness of the LiDAR Data

The important thing about the LiDAR input data, which all of the regarding inferences of this thesis is based on, is that it does not bear the *locally smoothness* quality with itself for sure. In last section we didn't hesitate to take this fact in account, had we taken we would be able to enjoy more desirable results of our algorithm. Gaussian process regressions are in fact a powerful tool for Bayesian regression tasks with input dependent smoothness. The basic assumption in Gaussian process manifest, is the stationarity of kernel process. This means that the covariance between two function values only depends on the distance between their independent variable and not on their direct values. The main drawback of this assumption is the lack of adapting ability for the $\mathcal{GP}$s task to the variable smoothness in the target function which is necessary encountering problems with ***input-dependent smoothness***.

There are methods to deal with this problem by putting a latent process on local smoothness characteristics, then jointly estimate or learn the latent and observed process together. Estimated means of local smoothness are used to apply integration based *Monte Carlo* methods to do the inference. Using Monte Carlo based methods, is not computationally cost efficient for large datasets and thus not applicable in real-world applications. Instead, *efficient point estimates* of *local smoothness* are carried out in this thesis, which will enable The use of *gradient-based optimization* methods for inference tasks to further enable point-wise estimation of local-dependent smoothness parameters in the covariance kernel.

The main advantages of using point estimates of local smoothness with latent $\mathcal{GP}$s is that it remains fully in the $\mathcal{GP}$ framework so $\mathcal{GP}$ inference methods can be held. In addition, analysis of the estimated covariance kernel is enabled using this method which causes the ability to reach great knowledge about the problem specifications. All of the gained knowledge, are obtained without taking the integral over all of the latent values but just with the use of the predicted mean values of length-scales. *Length-scales* are in fact the extent of the area that observations strongly influence each other, thus they may be chosen in the way that does not show constant values over the domain of our observations.

No functional form for the length-scale values as $\mathcal{L}(r)$ at the location $r$ are specified. Instead a *Gaussian latent* is put on the regarding values. In fact, an in-dependent $\mathcal{GP}$ is used to model the logarithms of this function namely $\log(\mathcal{L}(r))$. This new process is denoted as $\mathcal{GP}_l$. The logarithm is chosen to avoid the negative values in the process, as length scales may not bear negative values. This new process itself, is governed by a different covariance function which is specified by the new hyper-parameters $\theta_l = \{\bar{\sigma}_f, \bar{\sigma}_l, \bar{\sigma}_n\}$. Furthermore, a set of $m$ support values $\bar{\mathcal{L}}$ are assumed to form a part of new model parameters. This is because because the new covariance kernel must be defined with some control parameters. The new covariance is denoted as ***$\mathcal{L}$-kernel***. This new covariance kernel, must be precisely defined for the calculations to hold their reliability. Furthermore, a new input set for the $\mathcal{L}$-kernel must be defined from the current data available data. A part of the point cloud data, which is now transformed in to the set $\mathcal{R}_m$, is taken as the credential input subset to this new process, to be further used for the calculations related to the $\mathcal{L}$-kernel.

Consequently two Gaussian process must be confronted in the new problem setup: one is namely called $\mathcal{GP}_z$ which is the observed Gaussian process and the other one, the $\mathcal{GP}_l$ process which is the latent Gaussian process assigned to the input-dependent smoothness characteristic values.

### 3.7.2 Covariance Kernel for $\mathcal{L}$-Process

In order to increase the accuracy of segmentation, the non-stationary, isotropic covariance function which is originally proposed by [112] is chosen to model the regression problem. This covariance kernel further addresses the varying correlation problem for each segment of the data:

$$k(r_i, r_j) = \sigma_f^2 |\Sigma_i|^{\frac{1}{4}} |\Sigma_j|^{\frac{1}{4}} \left| \frac{\Sigma_i + \Sigma_j}{2} \right|^{-\frac{1}{2}} \exp\left[ -d_{ij}^T \left( \frac{\Sigma_i + \Sigma_j}{2} \right) d_{ij} \right] \tag{3.15}$$

With $d_{ij} = r_i - r_j$. In this equation for covariance kernel, for a local point $r'$ a local Gaussian kernel matrix $\Sigma$ is appointed and the covariances between target values $z_i$ and $z_j$ are calculated by averaging between this kernels at the input locations $r_i$ and $r_j$. This setting helps the influence of local characteristics of both locations to influence the modeled covariance of the corresponding target values. In the isotropic case, eigenvalues of the assigned kernels are aligned to the coordinate axes, thus their eigenvalues would be equal. This means that matrices $\Sigma_i$ simplifies to $\mathcal{L}_i^2$ which is the notion for the length-scale values for unidimensional input values. The expression $\mathcal{L}_i^T \mathcal{L}_i$ is used for the multidimensional input case. In the unidimensional case equation 3.15 will transform to the following form:

$$K(r_i, rj) = \sigma_f^2 (\mathcal{L}_i^2)^{\frac{1}{4}} (\mathcal{L}_j^2)^{\frac{1}{4}} \left( \frac{\mathcal{L}_i^2 + \mathcal{L}_j^2}{2} \right)^{-\frac{1}{2}} \exp\left( -\frac{2(r_i - r_j)^2}{\mathcal{L}_i^2 + \mathcal{L}_j^2} \right) \tag{3.16}$$

Which in multidimensional input case will be presented by the following form:

$$
\begin{aligned}
K(r_i, rj) &= \sigma_f^2 \cdot \left[\mathcal{L}^T \mathcal{L} I_n^T\right]^{\frac{1}{4}} \left[I_n^T \mathcal{L}^T \mathcal{L}\right]^{\frac{1}{4}} (\frac{1}{2})^{-\frac{1}{2}} \\
&\times \left[\mathcal{L}^T \mathcal{L} I_n^T + I_n^T \mathcal{L}^T \mathcal{L}\right]^{\frac{-1}{2}} \exp\left(\frac{-s(R)}{[\mathcal{L}^T \mathcal{L} I_n^T + I_n^T \mathcal{L}^T \mathcal{L}]}\right) \quad (3.17)
\end{aligned}
$$

Note that by choosing this covariance kernel, the covariance between two outputs $z_i$ and $z_j$ is calculated by averaging the two length-scales at the corresponding input locations $r_i$ and $r_j$. Furthermore, each input location has its own length-scale which we will get into the procedure of calculating those values, later in this chapter.

Assigning individual length-scales for each input location, enjoys the fact that by using local specifications the local characteristics can be taken into account. However in this methodology the length-scale value of each point must be obtained by assigning a new $\mathcal{GP}$ process regression to the latent length-scale itself. Length-scales are calculated and updated with the use of unidimensional gradient information in each segment of corresponding polar grid map. The covariance function that is introduced here seems fairly desirable for the application purposes, however it demands further hesitation in order to be able to fit to our problem which we are going to discussed here. It is of great importance to note that considering the ***Z-$\mathcal{L}$ process***, brings us two different functional relationship that we want to infer them based on our data and the assumption we have made about their behavior. The first one is the so called *target function* which is inferred based on the *measurement process* as depicted in Figure 3.5.

The second Gaussian process which is assumed to be jointly related to the *target function* process, is the ***latent length-scale*** **or $\mathcal{L}$ process** as depicted in Figure 3.6. This two functional relationship are jointly coupled to be fed into the inference procedure. The first functional relationship or *Z-process* is considered to obey the covariance kernel which is represented in 3.15. For the second process, the simple covariance kernel of Equation 3.8 is assumed to model the process. Parameters of these covariance kernels construct *hyper-parameters* of the new regression task. Furthermore, $\bar{\mathcal{L}}$ should be taken as a hyper-parameter because no variation can happen in the log marginal likelihood function without a variation in the $\bar{\mathcal{L}}$. This actually makes the gradient evaluation task a more difficult task to do, because $\bar{\mathcal{L}}$ is actually a vector, this in derivation procedure we will encounter tensors.

Furthermore, the subset from refined $PG_m$ two dimensional data, which is going to be fed into second process must be well-defined. Thus the set $\bar{r}$ is introduced. In order to cut some of the data in each segment to feed our $\mathcal{L}$-process covariance functions, we consider a slight devision from the mean value of an-gle between vectors $z$ and $r = \sqrt{x^2 + y^2}$ to represent the criteria for this subset. Every data point that falls into this section, will be selected as the regarding pseudo-input set for that segment as depicted in Figure 3.7. This measure is chosen because it could be a good representative of height deviations in each segment.

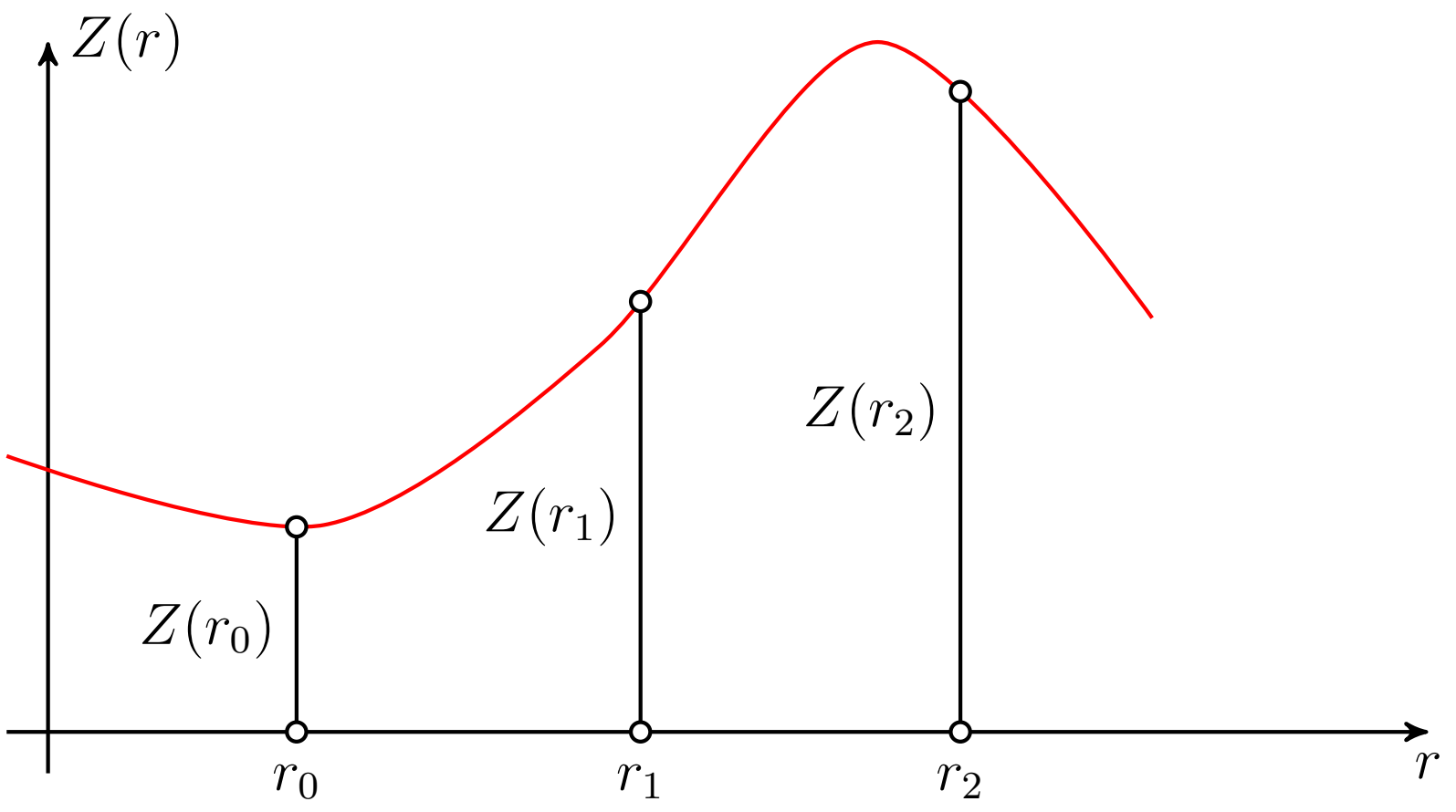

**Figure 3.5:** The so called Z-process (measurement process). Our observations of the environmental data can help us to infer the functional relationship for the ground. This functional relationship is hypothetically depicted in the figure. The input for the $\mathcal{GP}$ process at this level, are the ground candidate locations.

### 3.7.3 Pseudo-Input Selection

A common assumption using LIDAR sensors is that measurements coming from a certain surface must be somehow more related to each other. This assumption yields that if certain area of point cloud data resembles a surface, covariance related characteristics must be similar in that area and differ from other neighboring data points which does not show dependency to the same surface. Therefore if some measure is introduced for all the points from certain hypothetical surface in point cloud data, this measure can represent a fair measure of length-scale value for that certain area of the data, loyal to the shared surface.

In the reduced two dimensional data set of each segment represented by Equation 3.5, surfaces are reduced to lines. Therefore every set of points loyal to a certain line in each segment are assumed to have been measured from a common surface. The length of each line is assumed to be a fair observation of the length-scale for its loyal points. Thus these values are assumed to be a fair intuitive guess of $\bar{y}$ in equation 3.39.

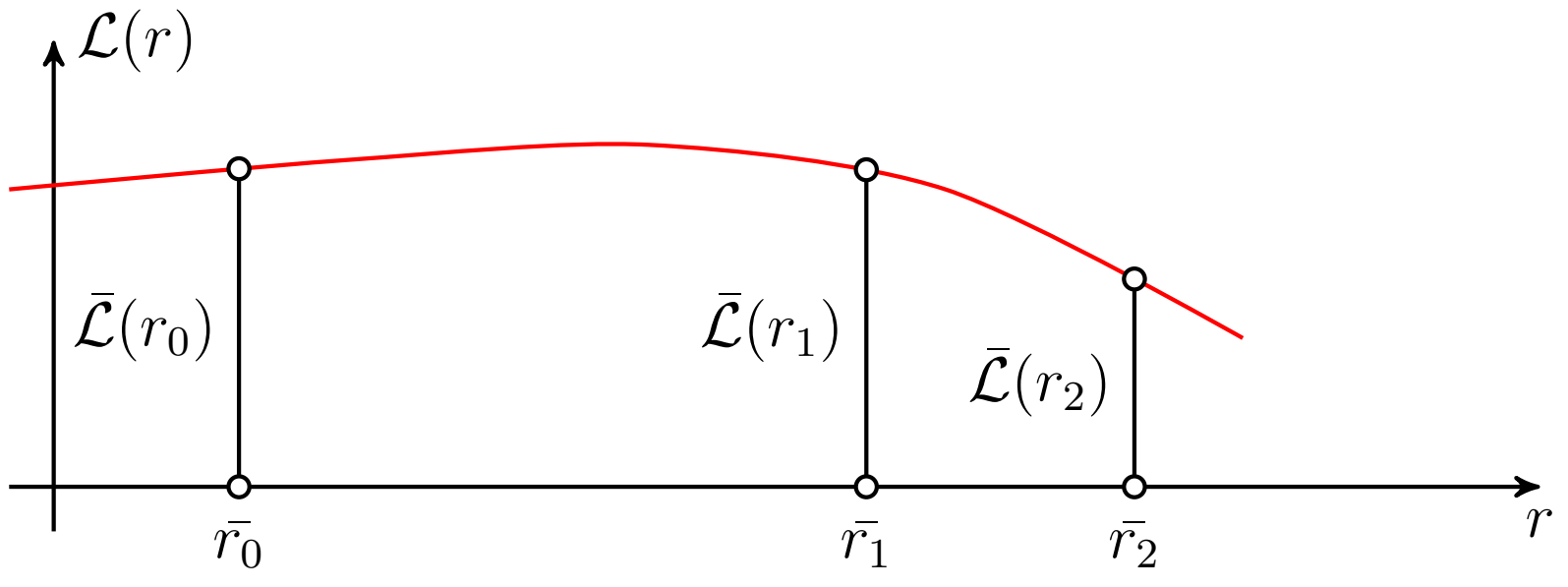

**Figure 3.6:** The so called $\mathcal{L}$-process or the process which is put on length-scale values. This process take the pseudo-input values as the input. This pseudo-input values are chosen with certain criteria to represent local conditions of the ground.

In every frame of the LiDAR data, data points are scattered in all possible areas of the *polar grid map*. The pseudo-input subset for the segment $S_n$ is to be found. Thus the mean value of angle between $r_{PGM} = \sqrt{x^2 + y^2}$ and $r_{Spherical} = \sqrt{x^2 + y^2 + z^2}$ is assumed to be as a measure of smoothness in height measurements in each segment. After finding this *mean value* as depicted with the red line in Figure 3.7, the pseudo-input subset $\bar{r}$ is defined to be the points within the angular distance $\zeta_h$ of this mean value.

### 3.7.4 Making Predictions

In the extended model which utilizes two joint Gaussian process regressions $\mathcal{GP}_z$ and $\mathcal{GP}_l$, the model prediction procedure is intractable if we want to integrate over all possible latent length-scales. Instead the most probable length-scale estimates are managed to be inferred. In order to obtain the predictive distribution for the target values, the distribution is integrated over all possible latent length-scales:

$$P(z_*|R_*, \mathcal{R}_m, \theta) = \int\int P(z_*|R_*, \mathcal{R}_m, \exp(\mathcal{L}_*), \exp(\mathcal{L}), \theta_z) \\ P(\mathcal{L}_*, \mathcal{L}|R_*, R, \bar{\mathcal{L}}, \bar{R}, \theta_l) dl dl^* \qquad (3.18)$$

Where $z_*$ represents the regressed parameters at the arbitrary location, $R_*$ represents the matrix of vector query locations, $\mathcal{R}_m$ represents the training dataset,$\mathcal{L}^*$ is the mean prediction of length-scale at $R_*$ and $\mathcal{L}$ is the mean prediction of length-scale at the locations in $\mathcal{R}_m$ and $\theta$ represents the hyper-parameters.

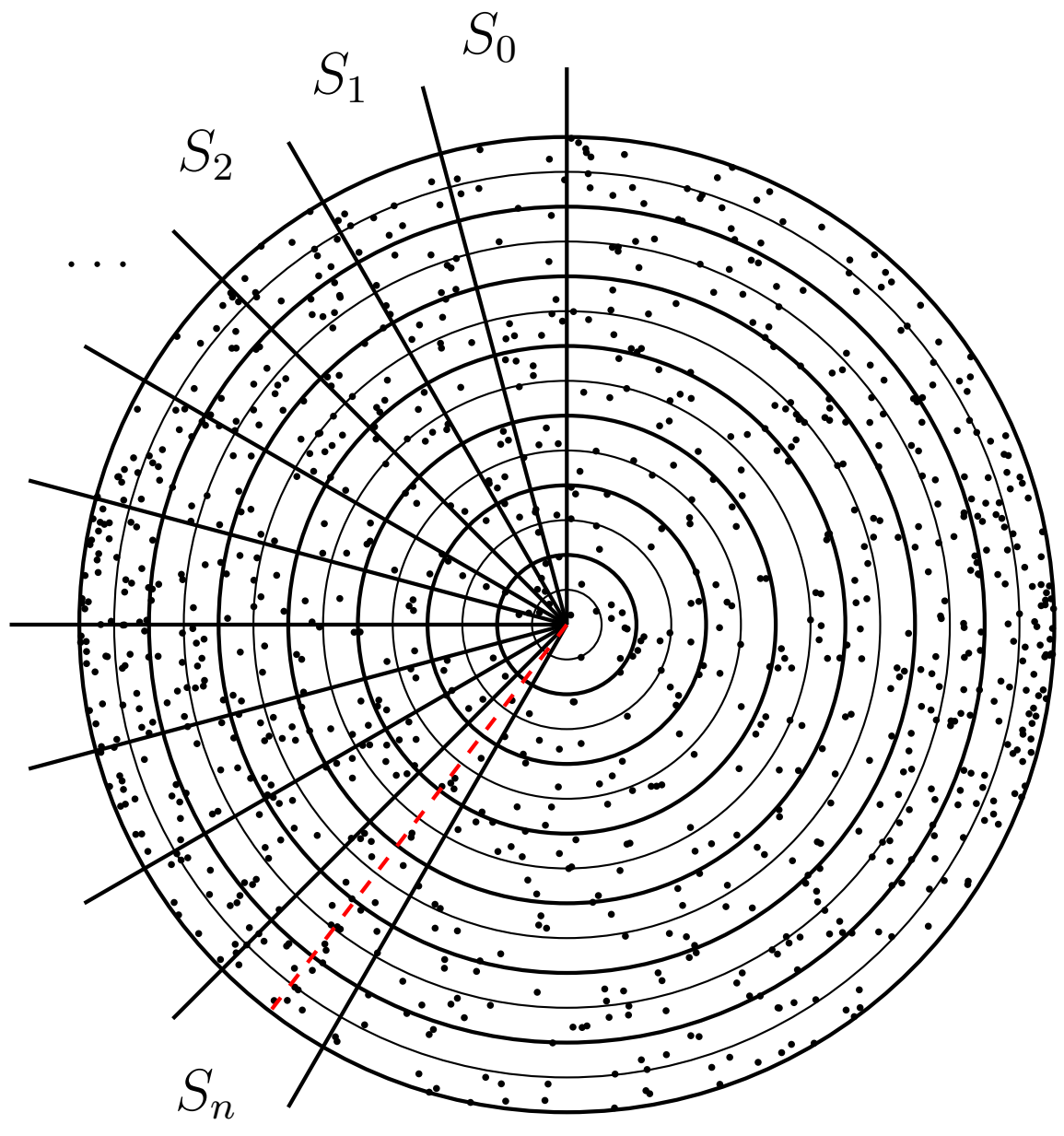

**Figure 3.7:** The criteria for pseudo-input selection. In each segment, a slight deviation of mean vertical angle value is chosen to be the sector containing all of the pseudo-input values. This gives us a rough measure of physical situation of the ground.

Unfortunately this marginalization is intractable.Some sources try to find the solution by Monte Carlo methods. In this thesis, instead of trying to solve this marginalization, the solution is obtained using the most probable length-scale estimates:

$$P(z_*|R_*, \mathcal{R}_m, \theta) \approx P(z_*|R_*, \exp(\mathcal{L}^*), \exp(\mathcal{L}), \mathcal{R}_m, \theta_z) \tag{3.19}$$

The two mentioned, joint Gaussian processes are independent. Thus, predictions can be made independently from common Gaussian regression equations. First, prediction on $\mathcal{GP}_l$ is is made to find $\mathcal{L}^*$ and $\mathcal{L}$ values. Then the prediction for another Gaussian process $\mathcal{GP}_z$ is to be made by treating $\mathcal{L}^*$ and $\mathcal{L}$ as fixed parameters. The test points will be all the points in the original $\mathcal{R}_m$ set or $\bar{r}^* \in \bar{R}^* \equiv r \in \mathcal{R}$, length-scales $\mathcal{L}^*$ are to be estimated at the points that the main Gaussian process $\mathcal{GP}_z$ must be obtained. The $\mathcal{GP}_l$ process is actually going to introduce functional relationship between support value $\bar{\mathcal{L}}$ and $\bar{R}$ and tries to infer the parameters related to this functional relationship from the unknown random field. In These equations, the parameter $\bar{\mathcal{L}}$ is called the *point-estimate* of length-scales at the locations defined in the subset $\bar{R}$.

$$\mathcal{GP}_l :\Rightarrow P(\mathcal{L}^*|\bar{R}^*, \mathcal{R}_m) \sim \mathcal{N}(\mathcal{L}^*, \sigma_*^2)$$

$$\mathcal{L}^* = \mu_{\mathcal{GP}_l} = \log \mathcal{L} = (\bar{K}(r,\bar{r}))^T[\bar{K}(\bar{r},\bar{r}) + \bar{\sigma}_n^2 I]^{-1}\bar{\mathcal{L}}$$

$$\Rightarrow \mathcal{L} = \exp\left[(\bar{K}(r,\bar{r}))^T[\bar{K}(\bar{r},\bar{r}) + \bar{\sigma}_n^2 I]^{-1}\bar{\mathcal{L}}\right] \tag{3.20}$$

With covariance kernel being as follows:

$$\bar{K}(\bar{r},\bar{r}) = \bar{\sigma}_f^2 \exp\left(-\frac{1}{2}\frac{s(\bar{R})}{\bar{\sigma}_l^2}\right)$$

$$\Rightarrow \bar{k}(\bar{r}_i,\bar{r}_j) = \bar{\sigma}_f^2 \exp\left(-\frac{1}{2}\frac{(\bar{r}_i - \bar{r}_j)^2}{\bar{\sigma}_l^2}\right) \tag{3.21}$$

Also:

$$\bar{K}(r,\bar{r}) = \bar{\sigma}_f^2 \exp\left(-\frac{1}{2}\frac{s(R,\bar{R})}{\bar{\sigma}_l^2}\right)$$

$$\bar{k}(r_i,\bar{r}_j) = \bar{\sigma}_f^2 \exp\left(-\frac{1}{2}\frac{(r_i - \bar{r}_j)^2}{\bar{\sigma}_l^2}\right) \tag{3.22}$$

Now for the second Gaussian process $\mathcal{GP}_z$ the predicting distribution would be of the form below:

$$\mu_{\mathcal{GP}_z} = \bar{z} = K(r^*,r)^T\left[K(r,r) + \sigma_n^2 I\right]^{-1} z \tag{3.23}$$

$$K(r,r) = \sigma_f^2 . \left[\mathcal{L}^T\mathcal{L}I_n^T\right]^{\frac{1}{4}}\left[I_n^T\mathcal{L}^T\mathcal{L}\right]^{\frac{1}{4}}(\frac{1}{2})^{-\frac{1}{2}}\left[\mathcal{L}^T\mathcal{L}I_n^T + I_n^T\mathcal{L}^T\mathcal{L}\right]^{\frac{-1}{2}} \times \exp\left(\frac{-s(R)}{[\mathcal{L}^T\mathcal{L}I_n^T + I_n^T\mathcal{L}^T\mathcal{L}]}\right) \tag{3.24}$$

In this representation, the input of the $\mathcal{GP}$ regression algorithm is assumed to be multidimensional so the parameters dimensions would be as $\mathcal{L} \in \mathbb{R}^{d\times 1}$ for every data point. In the matrix form it would be as $\mathcal{L} \in \mathbb{R}^{d\times n}$ where $n$ is the number of data points.

Length-scales of Z-process covariance functions are assumed to be in a exponential relationship with actual point estimates of $\mathcal{L}$-process, resulting the relationship of the form $\mathcal{L}^* = \log \mathcal{L}$. Thus, the **mean prediction of $\mathcal{L}$-process** actually gives us the ***logarithm of length-scales*** $\mathcal{L}$ not the length-scale itself. Inputs are assumed to be unidimensional values of $r$ to the $\mathcal{GP}$ regression algorithm. Thus in this problem setup, the definition of the term $\mathcal{L}^T\mathcal{L}1_n^T$ in unidimensional input case is of the following form:

$$\Omega_1 = (\mathcal{L}_{1\times n})^T \mathcal{L}_{1\times n} \mathtt{1}_n^T = \begin{bmatrix} l_1^T l_1 \\ l_2^T l_2 \\ \vdots \\ l_n^T l_n \end{bmatrix} \begin{bmatrix} 1 & 1 & \dots & 1 \end{bmatrix}$$

$$= \begin{bmatrix} l_1^T l_1 & l_1^T l_1 & \dots & l_1^T l_1 \\ l_2^T l_2 & l_2^T l_2 & \dots & l_2^T l_2 \\ \vdots & \vdots & \ddots & \vdots \\ l_n^T l_n & l_n^T l_n & \dots & l_n^T l_n \end{bmatrix} \quad (3.25)$$

Also the term $\mathtt{1}_n \mathcal{L}^T \mathcal{L}$ in one-dimensional input case is of the following form:

$$\Omega_2 = \mathtt{1}_n (\mathcal{L}_{1\times n})^T \mathcal{L}_{1\times n} = \begin{bmatrix} 1 \\ 1 \\ \vdots \\ 1 \end{bmatrix} \begin{bmatrix} l_1^T l_1 & l_2^T l_2 & \dots & l_n^T l_n \end{bmatrix}$$

$$= \begin{bmatrix} l_1^T l_1 & l_2^T l_2 & \dots & l_n^T l_n \\ l_1^T l_1 & l_2^T l_2 & \dots & l_n^T l_n \\ \vdots & \vdots & \ddots & \vdots \\ l_1^T l_1 & l_2^T l_2 & \dots & l_n^T l_n \end{bmatrix} \quad (3.26)$$

The third term which is in fact the sum of $\Omega_1$ and $\Omega_2$ is going to be of the form:

$$\begin{aligned} \Omega_3 &= (\mathcal{L}_{1\times n})^T \mathcal{L}_{1\times n} \mathtt{1}_n^T + \mathtt{1}_n (\mathcal{L}_{1\times n})^T \mathcal{L}_{1\times n} \\ &= \begin{bmatrix} l_1^T l_1 + l_1^T l_1 & l_1^T l_1 + l_2^T l_2 & \dots & l_1^T l_1 + l_n^T l_n \\ l_2^T l_2 + l_1^T l_1 & l_2^T l_2 + l_2^T l_2 & \dots & l_2^T l_2 + l_n^T l_n \\ \vdots & \vdots & \ddots & \vdots \\ l_n^T l_n + l_1^T l_1 & l_n^T l_n + l_2^T l_2 & \dots & l_n^T l_n + l_n^T l_n \end{bmatrix} \end{aligned} \quad (3.27)$$

The last element in equation 3.24, is the result of element-wise division of $-\frac{S(r_i,r_j)}{\Omega_3}$. In addition, power operators are all element-wise for all values except for the power $-1$ which is indicator of the inverse operation. These calculation are necessary as in the next section we have to calculate gradient of log-marginal likelihood with respect to hyper-parameters.

### 3.7.5 Learning Hyper-Parameters

Until now, all the decisive parameters of the model are assumed to be known before the solution procedure is started, or as it said a ***a-priori***. However in real-world applications one may not be always able to be so sure about the value of the joint hyper-parameters $\theta$ of the whole process. Instead these values are to be estimated using observations. The training data is consisted of $n$ observations $z$ at locations $R$. The *Maximum A Posteriori* argument is used to find the hyper-parameters. This means that the hyper-parameters that maximize the probability of the likelihood of observing $z$ given $R$ are chosen to be the right ones.

$$Maximize\ P(z|R,\theta) = \int P(z|R,\mathcal{L},\theta_z)P(\mathcal{L}|R,\bar{\mathcal{L}},\theta_l)d\mathcal{L} \tag{3.28}$$

This integral is also intractable. The solution must be found by maximizing *a-posteriori* probability of the latent length-scales:

$$\log P(\mathcal{L}|z,R,\theta) = \log P(z|R,\exp(\mathcal{L}),\theta_z) + \log P(\mathcal{L}|R,\bar{\mathcal{L}},\bar{R},\theta_l) + cte \tag{3.29}$$

Where $\mathcal{L}$'s are the mean predictions of $\mathcal{GP}_l$ regression task. In order to find the corresponding solutions, the gradient of this object functions with respect to hyper-parameters $\theta$ are utilized in the so called gradient-based optimization method scheme. In the experiment the parameters $\bar{\sigma_f}$, $\bar{\sigma_n}$ and $\bar{\sigma}_l$ of the latent kernel width are assumed to be independent of the parameters related to the Z-process which are assumed to be $\theta_l = \{\sigma_f, \sigma_n\}$ and $\bar{\mathcal{L}} = \{l_1, ..., l_2\}$. Thus in the optimization process, it is assumed that $\frac{\partial\mathcal{L}(\theta)}{\partial\star} = 0$, because their *independence* must be taken into account, where $\star$ denotes one of the hyper-parameters. Furthermore, the locations $\bar{R}$ of the latent kernel width variables were calculated in previous section from original data input set $R$ as the pseudo-input set.

The only critical thing to be remained, is the choice of the objective function which is of great importance for the optimization process. The objective function must be chosen to include all the real-world conditions as possible for the method.

### 3.7.6 The Objective Function

If optimization process is to be effective enough, it shall take all the segments into account to form an *objective function*. This would be an example of *multi-task* regression problem. The objective function in multi-task problems is equal to the sum of all objective functions regarding to different tasks of the problem that share the same hyper–parameters. Therefore, we assume that all the segments share the same *hyper–parameters* and establish a global view to our frame data in order to have the results to be whole–frame inclusive.

**Single-Segment Objective Function** After assignment of the objective function, the only remaining task would be to maximize this function with respect to the joint hyper-parameters as well as the latent support values $\bar{\mathcal{L}}$ in order to be able to learn the hyper-parameters of the problem correctly. The first term in the log-marginal likelihood represented in equation 3.29, could be realized as the standard Gaussian process objective function:

$$\begin{aligned} \log P(z|R, \exp(\mathcal{L}), \theta_z) &= -\frac{1}{2} z^T (K(r,r) + \sigma_n^2 I)^{-1} z \\ &\quad -\frac{1}{2} \log |K(r,r) + \sigma_n^2 I| \\ &\quad -\frac{n}{2} \log(2\pi) \end{aligned} \tag{3.30}$$

Where as it is detailed before, $K(r,r)$ is the noise-free process covariance matrix for training locations $R$ which is assumed to be of the form of equation 3.24. The second term of the log-marginal likelihood objective function of equation 3.29 can be written as:

$$\log P(\mathcal{L}|R, \bar{\mathcal{L}}, \bar{R}, \theta_l) = -\frac{1}{2} \log |\bar{K}(\bar{r}, \bar{r}) + \bar{\sigma}_n^2 I| - \frac{n}{2} \log(2\pi) \tag{3.31}$$

The second $\mathcal{GP}$ regression or the point-wise estimation of length-scales approach, considers the most likely latent length-scale $\mathcal{L}$ as the mean prediction of $\mathcal{GP}_l$ at locations $R$. Now we can combine the two extended term to get the objective function of equation 3.29 as follows:

$$\begin{aligned}
\log P(\mathcal{L}|z,R,\theta) &= \log P(z|R,\exp(\mathcal{L}),\theta_z) + \log P(\mathcal{L}|R,\bar{\mathcal{L}},\bar{R},\theta_l) + cte \\
&= -\frac{1}{2}z^T(K(r,r)+\sigma_n^2 I)^{-1}z - \frac{1}{2}\log|K(r,r)+\sigma_n^2 I| \\
&\quad -\frac{n}{2}\log(2\pi) - \frac{1}{2}\log|\bar{K}(\bar{r},\bar{r})+\bar{\sigma}_n^2 I| - \frac{n}{2}\log(2\pi) \\
&= -\frac{1}{2}z^T(K(r,r)+\sigma_n^2 I)^{-1}z - \frac{1}{2}\log|K(r,r)+\sigma_n^2 I| \\
&\quad -\frac{1}{2}\log|\bar{K}(\bar{r},\bar{r})+\bar{\sigma}_n^2 I| - n\log(2\pi) \\
&= -\frac{1}{2}\Bigg(z^T(K(r,r)+\sigma_n^2 I)^{-1}z + \log|K(r,r)+\sigma_n^2 I| \\
&\quad +\log|\bar{K}(\bar{r},\bar{r})+\bar{\sigma}_n^2 I|\Bigg) - n\log(2\pi) \\
&= \alpha\Bigg(z^T A^{-1}z + \log|A| + \log|B|\Bigg) + \beta
\end{aligned} \tag{3.32}$$

Where $\alpha$ and $\beta$ are real valued constants and $A := K(r,r) + \sigma_n^2 I$ and $B := \bar{K}(\bar{r},\bar{r}) + \bar{\sigma}_n^2 I$ are the corresponding covariance functions of Gaussian processes. The noise-free covariance function $K(r,r)$ is calculated with the covariance kernel of the equation 3.24 which is isotropic.

The isotropic case is chosen only in the matter of simplicity. In previous sections we presented this covariance kernel function both for unidimensional inputs and multi-dimensional inputs. Until here, the procedure that is going to be taken for joint prediction of the combined Gaussian processes $\mathcal{GP}_z$ and $\mathcal{GP}_l$ to get desiring results is fully discussed. However a part of the discussions remains incomplete if we fail to mention the procedure for mean prediction of length-scales. In the above equation the mean prediction for the length-scale values will get the form:

$$\mathcal{L} = \exp\left[(\bar{K}(r,\bar{r}))^T[\bar{K}(\bar{r},\bar{r}) + \bar{\sigma}_n^2 I]^{-1}\bar{\mathcal{L}}\right] \tag{3.33}$$

Which represents the length-scale values. In order to get precise results we should be very careful in choosing mathematical notations and also mathematical analytical results, therefore as we mentioned before the main covariance function in which we will represent our noise-free $z-process$ or $K(r,r)$ will be of the form:

$$\begin{aligned} K(r,r) &= \sigma_f^2.\Big[\mathcal{L}^T\mathcal{L}I_n^T\Big]^{\frac{1}{4}}\Big[I_n^T\mathcal{L}^T\mathcal{L}\Big]^{\frac{1}{4}} \\ &\times(\frac{1}{2})^{-\frac{1}{2}}\Big[\mathcal{L}^T\mathcal{L}I_n^T + I_n^T\mathcal{L}^T\mathcal{L}\Big]^{\frac{-1}{2}} \\ &\times \exp\left(\frac{-s(R)}{[\mathcal{L}^T\mathcal{L}I_n^T + I_n^T\mathcal{L}^T\mathcal{L}]}\right) \end{aligned} \quad (3.34)$$

The covariance functions for the latent length-scale process $\mathcal{GP}_l$ becomes of the following form which is decided to be in the squared exponential form.

$$\bar{K}(\bar{r},\bar{r}) = \bar{\sigma}_f^2 \exp\Big( -\frac{1}{2}s(\bar{\sigma}_l^{-2}\bar{R}^2)\Big) \quad (3.35)$$

And for the predictions the covariance function would be of the following form:

$$\bar{K}(r,\bar{r}) = \bar{\sigma}_f^2 \exp\Big( -\frac{1}{2}s(\bar{\sigma}_l^{-2}R, \bar{\sigma}_l^{-2}\bar{R})\Big) \quad (3.36)$$

In our setting the function $s(\bullet)$ is designed in the following fashion:

$$s(\bar{\sigma}_l^{-2}\bar{R}^2) = \frac{(\bar{r}_i - \bar{r}_j)^2}{\bar{\sigma}_l^2} \quad (3.37)$$

We actually did use the line parameters we have obtained in previous sections, as our initial value for our optimization algorithm where we set corresponding initial parameters for input locations $\bar{r}$ as their corresponding *line* values which were extracted in ***line exaction*** algorithm.

**Whole-Frame Objective Function** Although the represented single-segment objective function, results in an estimation for the hyper-parameters for the whole–frame problem, it is actually the objective function designed with respect to each certain segment. In the ground segmentation task many segments would be confronted. Therefore all of them must be taken into account if the optimization process is to be effective enough in global point of view. This would be an example of ***multi-task*** objective function which is equal to the sum of all objective functions in the problem setting that share the same hyper-parameters. Therefore, all the segments are assumed to share the same *hyper-parameters*. In order to deliver inclusive results the interpretation of ***length-scale*** parameters have to be re-defined to have same shared parameters for each data frame.

In order to add a whole-frame physical intuition of ground quality to our ground segmentation algorithm, a multi-task objective function is proposed to take all the segments once at the same time into account for ground estimation. Different segments of each frame must share all hyper-parameters and at the same time $\bar{\mathcal{L}}_i$ set must represent ground candidate set of each segment individually. The $\bar{\mathcal{L}}_{wf}$ vector is defined to contain all the individual parameters of segments and at the same time. This vector is uniquely defined and shared in whole space.

A set of points at each segment's ground candidate set are chosen to explain the ground quality in that segment. This definition for pseudo-input set $\bar{\mathcal{R}}$, enables the latent variable vector $\bar{\mathcal{L}}$ to be a vector defined on the whole point cloud space, rather than just on one segment:

$$\bar{\mathcal{L}}_{wf} = \left[\bar{\mathcal{L}}_1, \ldots, \bar{\mathcal{L}_M}\right] \tag{3.38}$$

Where $\bar{\mathcal{L}}_i$ represents corresponding latent variable vector for $i_{th}$ segment with length $l_m$. This latent variable vector is defined over pseudo-input set in each ground candidate set which is not yet defined.

In order to introduce pseudo-input set $\bar{r}_i$ set for each segment, a slight deviation from the mean value of angle between vectors $z$ and $r = \sqrt{x^2 + y^2}$ are considered as a measure of height deviations at each segment. In every segment all the ground candidate points with defined radial distance to the mean angle value of that segment, will be the corresponding pseudo-inputs for that segment and their latent Gaussian process values of length-scales are assumed to be part of whole–frame vector $\bar{\mathcal{L}_{wf}}$. Length-scales are assumed to be in a functional relationship which is put on pseudo-input radial values and the latent Gaussian process tends to find this relationship:

$$\bar{y} = \bar{\mathcal{L}}_i(\bar{r}) + \epsilon_r \tag{3.39}$$

Where $\bar{y}$ is the corresponding pseudo-observation and $\epsilon_r$ is the observation noise.

Then the whole–frame objective function is defined as follows:

$$
\begin{aligned}
L(\theta) &= \sum_{m=1}^{M}\Big(\log P(\mathcal{L}^m|z^m, R^m, \theta)\Big) \\
&= \sum_{m=1}^{M}\Big[\alpha_m\Big((z^m)^T A_m^{-1} z^m + \log|A_m| + \log|B_m|\Big) + \beta_m\Big] \\
&= \alpha_m \sum_{m=1}^{M}\Big((z^m)^T A_m^{-1} z^m + \log|A_m| + \log|B_m|\Big) + \sum_{m=1}^{M}\beta_m \\
&= \alpha_m \sum_{m=1}^{M}\Big((z^m)^T A_m^{-1} z^m\Big) + \alpha_m \sum_{m=1}^{M}\Big(\log|A_m|\Big) \\
&\quad +\alpha_m \sum_{m=1}^{M}\Big(\log|B_m|\Big) + \sum_{m=1}^{M}\beta_m \\
&= \alpha_m \sum_{m=1}^{M}\Big((z^m)^T A_m^{-1} z^m\Big) + \alpha_m \log\Big(\prod_{m=1}^{M}|A_m|\Big) \\
&\quad +\alpha_m \log\Big(\prod_{m=1}^{M}|B_m|\Big) + \sum_{m=1}^{M}\beta_m \\
&= -\frac{1}{2}\sum_{m=1}^{M}\Big((z^m)^T A_m^{-1} z^m\Big) - \frac{1}{2}\log\Big(\prod_{m=1}^{M}|A_m|\Big) \\
&\quad -\frac{1}{2}\log\Big(\prod_{m=1}^{M}|B_m|\Big) + \sum_{m=1}^{M}\Big(-\frac{n_m}{2}\log(2\pi) - \frac{\bar{n}_m}{2}\log(2\pi)\Big) \\
&= -\frac{1}{2}\sum_{m=1}^{M}\Big((z^m)^T A_m^{-1} z^m\Big) - \frac{1}{2}\log\Big(\prod_{m=1}^{M}|A_m|\Big) \\
&\quad -\frac{1}{2}\log\Big(\prod_{m=1}^{M}|B_m|\Big) - \frac{\log(2\pi)}{2}\Big(\sum_{m=1}^{M}(n_m + \bar{n}_m)\Big) \qquad (3.40)
\end{aligned}
$$

### 3.7.7 Ground Segmentation Based on $\mathcal{GP}$ Regression Algorithm

In every segment *m*, a candidate two-dimensional ground point set $PG_m$ is constructed for every LiDAR frame which can be contaminated by obstacle points as outliers. The typical Gaussian process regression task assumes that all of the data in $PG_m$ is ground points with a few outliers. Ground segmentation problem is formulated as obtainment of one regression model with the ability of out-lier rejection for each segment in our polar grid map. Furthermore, an iterative learn-ing method is adapted to build the local ground model in every segment which benefits at the same time from both desirable approximation ability and out-lier rejection. The algorithm for implementing this method is depicted in Algorithm 2.

This algorithm starts with receiving a 3D scan of environment as a set of point clouds $P = \{P_1, ..., P_k\}$ which is consisted of $P_t$ frames at time $t$ and outputs the label of each point in each point cloud frame as **ground** or **obstacle**. This algorithm can be divided into six different steps:

1. Polar Grid Map Representation
2. Line Extraction
3. Seed Estimation
4. Gaussian Process Regression
5. New Seed Evaluation
6. Point-Wise Segmentation

The algorithm 2, takes the input data and at the first step applies the Polar Grid Map segmentation on it to produce desirable representation of the data to be more reliable for the presented ground segmentation method. In the second step, line extraction algorithm will be used to extract critical lines in every segment. In the third step, the seed estimation, the goal is to find the initial ground points as the training data. Thus the points with absolute height values less than $T_s$ within a fixed radius B of the sensor are chosen as the initial seeds. In the fourth step, the ground model is obtained based on the Gaussian process regression methodology. Given equations for $\mathcal{GP}$ are used to calculate the covariance matrix between the points of the seeds $s_p$. In the fifth step, the function ***eval*** uses the Gaussian process expressions for covariance matrices which will be used to calculate $K(r^*, r^*)$ and $K(r^*, s_p)$ for every query test point $(Z^*, r^*)$ and then Gaussian Process equations for the prediction distributions can be used to calculate the mean and variance of the output $z^*$ at the test input location $r^*$. In order to recognize the candidate ground points or new seeds in our proposed algorithm from outliers in the set $PG_m$, we will use the following attribute:

$$\begin{aligned} V[z] &\leq \zeta_m \\ \frac{|z^* - \bar{z}|}{\sqrt{\sigma_n^2 + V[z]}} &\leq \zeta_d \end{aligned} \tag{3.41}$$

Where $\zeta_m$ represents the threshold of the variance $V[z]$ of the output $z^*$ and $\zeta_m$ represents the normalized distance between the real output $z^*$ and the expected

**Algorithm 2** Ground Segmentation

**INPUT**: $P_t = \{p_1, p_2, ..., p_l\}, M, N, B, \zeta_s, \zeta_r, \bar{\sigma}_f, \bar{\sigma}_n, \bar{\sigma}_l, \bar{\mathcal{L}}, \sigma_f, \sigma_n, \zeta_m, \zeta_d$ **OUTPUT**: Lable of each point $p_i$ $i = 1, 2, ..., l$

$(PG_m, P'_{b_n^m}) = PolarGridMap(P_t, M, N)$
**for** i=0:M-1 **do**
  $L_i = LineExtraction(PG_i)$
  $\mathcal{S}_{new} = \mathcal{S}_p = \emptyset$
  $\mathcal{S}_{new} = \mathcal{S}eed(PG_i, B, \zeta_s)$
  **while** $Size(\mathcal{S}_{new}) > 0$ **do**
    $\mathcal{S}_p = \mathcal{S}_p \cup \mathcal{S}_{new}$
    Model = $\mathcal{GPR}(\mathcal{S}_p, L_i, \sigma_f, \sigma_n)$
    Test = $GP_i - S_p$
    $\mathcal{S}_{new} =$ **Eval**$(Model, Test, \zeta_d, \zeta_m)$
  **end while**
  **for** $j = 0 : N - 1$ **do**
    **Segment**$(Model, P'_{b_j^i}, \zeta_r)$
  **end for**
**end for**

mean value $\bar{z}$ of the output at the test input $r^*$. The test point $(z^*, r^*)$ is able to be classified into the new seed set if and only satisfies both of the inequalities at the same time. The seed evaluation algorithm starts from the initial seed and all the seeds are accumulated one by one untill no more seeds are found. The sixth step or point-wise segmentation is related to the function ***Segment*** in the algorithms, for the segment m the final seeds that are found using iterations above, are used to model the local ground. For the $j_{th}$ bin in segment $i$ or the bin $b_j^i$ the radial coordinate of the bin is $r_j^i = \frac{r_j^{min} + r_j^{max}}{2}$ and thus this value can be used with the Gaussian process equation for mean prediction to predict the mean height of the ground $\bar{z}_j^i$ in the bin $b_j^i$, and this value further is assumed to be the reference ground height of the bin $b_j^i$ or $H_{ij}$.
Now in order to lable all the point in our initiative point cloud, for every point $p'_k$ in the set $P'_{b_j^i} = \{p'_k = (z_k, r_k) | p_k \in P_{b_j^i}\}$ we evaluate the height $|z_k - Hij|$ and if this value is less than $\zeta_r$, the point as assumed to be a **ground point**, otherwise it is classified as an **obstacle** point.Remember that the relation between the two-dimensional representation of our point cloud and three-dimensional raw data there is an one-to-one relation ship so the 3D points in the $P_{b_j^i}$ have the same label as the 2D points in the $P'_{b_j^i}$ and thus all the labels of the points in 3D scan can be found.

In reference [107] it is discussed that the input-dependent noise must be taken into account in Gaussian process regressions in order to make the estimation reliable. While, the LiDAR data is easily considered to bear the input-dependent noise within its data, the literature lacks the discussion for this matter. Furthermore, the problem of input-dependent noise is not discussed neither for the ground segmentation problem nor for the Gaussian process regression task. Thus, a method using three different underlying processes is proposed in this thesis for this problem. This method is represented in Appendix E in details while the necessary gradients are presented in second section of Appendix C.

## 3.8 Accuracy Evaluation Of Simulations

These three different methodologies that had been presented for ground segmentation are tested on one point cloud data, In order to give a measure for algorithms accuracy in mean predictions, we will make use of ***standardized mean squared error***:

$$SMSE := \frac{1}{n}\sum_{i=1}^{n}\left(var(z)^{-1}(z_i - \mu_i)^2\right) \tag{3.42}$$

Also to give a measure for comparison between and data fit of whole predictive distributions we will make use of ***negative log predictive density***:

$$NLPD := \frac{1}{n}\sum_{i=1}^{n}\left(\log P_{model}(z_i|r_i)\right) \tag{3.43}$$

The corresponding results are depicted in Chapter 5, where the implementation of the detection module is discussed.

# Chapter 4

# System Design and Architecture Proposition: Tracker Module

The ultimate output of the detector module, is a list of bounding boxes along with their respective dimensions, yaw angle (orientation) and the position of the center of each bounding box. The main contribution of the tracker module would be to maintain a unique and reliable identity of these bounding boxes across all of the data frames. Therefore, the tracker module is consisted of two separate sub-components: position tracker and the bounding box tracker. In this thesis, the tracking module is developed to track points instead of extended tracking. Thus, the position tracker tends to track the position of each object's center while the bounding box tracker tries to maintain dimensions of each object.

The probabilistic Bayesian filtering is used in position tracker to deal with motion model uncertainties and center point association. On the other hand, the bounding box tracker uses a sort of heuristic logic-based filtering method to deal with bounding box tracking task.

## 4.1 The Position Tracker

Object tracking in urban environments with LiDAR-only measurements is an intricate task to accomplish. The measurement output of a LiDAR scanner, mounted on the navigating vehicle, is expected to include incomplete spatial data. The reason for this incompleteness would be occlusions from other static or dynamic objects or even ill-posed prospective of the ego-vehicle. On the other hand, in crowded urban areas objects are not expected to behave like they do in the less-crowded ones. For example in a crowded scene, pedestrians, bicyclists and vehicles are not expected to follow the linear constant velocity motion model of the less-crowded scene and they tend to follow different, multiple motion models.

Another challenge arises with the necessity of MOT task where multiple objects are to be tracked simultaneously in a cluttered environment.

Two different trackers are utilized to tackle these uncertainties: the moving-object and the clutter-aware filter. The clutter-aware filter is the filter incorporating the probabilistic approach for data association problem. On the other hand, the moving-object filter is the filter which incorporates the filtering process based on different motion models. This trackers are implemented in different parts of the tracker module [65, 125, 92, 126].

### 4.1.1 IMM-UK-JPDAF

The least requirement for the multi-object tracking problem in the urban environment is the ability to track multiple-moving objects with different, uncertain motion models with the use of data coming from uncertain sensor measurements and of course with the presence of occlusions. These different kinds of uncertainties will be taken care of by the use of different filtering methodologies. In the first step, the Interacting Multiple Model (IMM) method is used to handle different motion models. Furthermore, since the motion models are not restricted to be linear, the Unscented Kalman Filter (UKF) is used to handle the nonlinear state estimations. Then, the Joint Probabilistic Data Association Filter (JPDAF) is used in data association step to handle object tracking in cluttered environment with noise detection in the incoming data. These different filters, work along together to form a so called IMM-UKF-JPDAF filter. In many references, different derivatives of this coupled filtering solution may be found with other names, such as work done by [95, 91, 65, 92] and many other works in the literature.

The IMM-UKF-JPDAF filter consists of four main filtering steps: interaction, prediction and measurement validation, data association, mode probability update and combination. The mathematical and theoretical background of further section are discussed in Section 2.6, therefore minimum explanation about mathematical aspects of the filtering process is represented here. According to the Equation 2.5, $j$ different filters, each for different motion models are used, each given by a state space model with $x_k$ being the states and $u_k$ being the corresponding measurements at time step $k$. The process noise and measurement noise are assumed to be mutually independent zero-mean, white, Gaussian noise with corresponding covariance matrices.

According to reference [30], switching between different modes may be assumed to be a Markov process. The transitions between different IMM motion models (corresponding indexes $i$ and $j$) are denoted with $p_{ij}$. Therefore the transition probability matrix is given by Equation 2.49. Different implemented motion models are discussed in in Section 2.6.3. Different steps of the couples IMM-UKF-JPDAF process are to be discussed in proceeding sections while the full

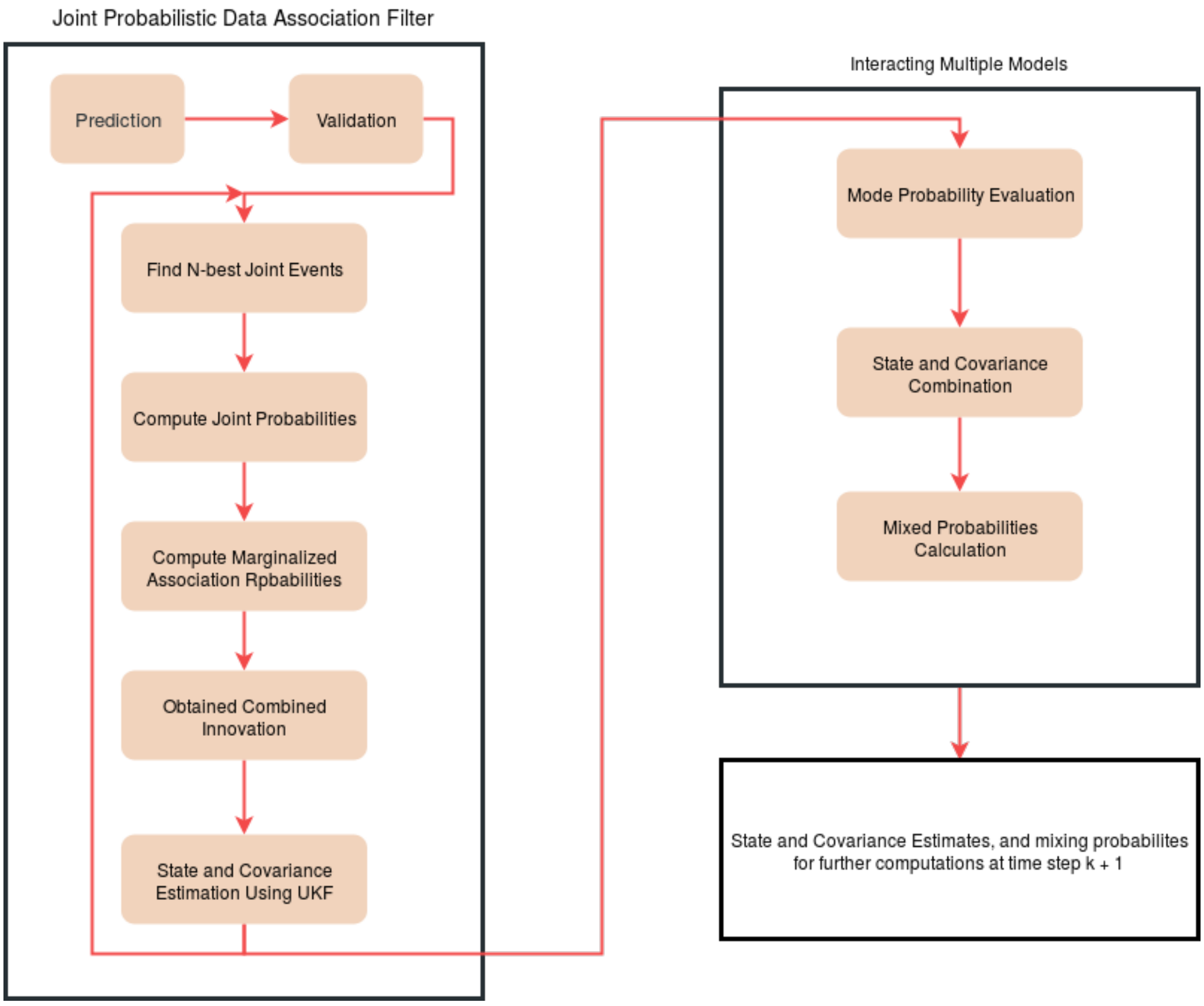

**Figure 4.1:** The corresponding flowchart for the IMM-UKF-JPDAF filtering process. The IMM actually performs the filtering task, while the JPDAF performs the correlation. The input-output schematic of the flowchart is identical to the Figure 2.8.

process cycle is depicted in the Figure 4.1.

**Interaction Step** In this step, the IMM probability mixing process is implemented. The probability that the $i_{th}$ mode has been started in the previous cycle, given that the $i_{th}$ mode is active in the current frame is calculated using equation 2.48. The prior mode probabilities $\mu_{j,k}^{-}$, at the current time step is given by the prediction of the mode probabilities of the previous time step. This actually denotes the probability that a mode transition occurs from mode $i$ to mode $j$. Then with the help of these j individual filters, the combined initial states and the covariance matrices are mixed from each particular filter to form a single initial state and covariance matrix which is given in Equation 2.51. This process actually gives the initial state for each separate filter, which is computed in the previous time step.

**Prediction and Measurement Validation Step** The Unscented Kalman Filter (UKF) is utilized for the prediction task. Evaluated sigma points are to be propagated through the system function $f$. Then the weighted sigma points are to be recombined to produce the predicted state and the corresponding covariance. New sigma points are subsequently chosen by the use of $P^-_{k|k-1}$ value, which are going to be propagated through the measurement function $h$. Furthermore, the wighted recombination of the sigma points is used to produce the predicted measurement and covariance matrix. All the formulation stuffs are detailed in section 2.7.1.

**Data Association and Model-specific Filtering Step** The data association and state filtering step, are done exactly the same with equations provided in section 2.8.1 or the Probabilistic Data Association Filter (PDAF). Actually the conventional PDA filter is used and each model is incorporated in the process. The updated state is simply the predicted state plus the innovation weighted by the Kalman gain. The associated innovation term is multiplied with association probabilities $\beta$.

**Mode Probability Update and Combination Step** The likelihood of the measurement to fit to a model, is used to update the mode probabilities. The details of this step is discussed in section 2.8.1 under the title: Data Association.

## 4.2 Track Management

The track management is a necessary part of the system due to the fact that a large amount of target tracks must be maintained during the process, with their corresponding uncertainties. At the first sight, track management seeks to dynamically limit the number of spurious target list and thus preventing false data associations. Furthermore, if some detections are lost in some data frames, the track management unit must maintain those tracks. Another important responsibility of the track management unit, is to semantically classify the target objects based on their attributes. Since in our case we do not tend to semantically identify object types, this part will be used to classify objects due to their static or dynamic behavior and track maturity.

### 4.2.1 Track Initialization

A new track is started when a statistically nearby measurement is found for it. For this to happen, the multiple trackers must be placed nearby the measurements from the beginning. Therefore, a unique method must be proposed for track distribution from the beginning of the task. At the first time step, all of the track are

**Algorithm 3** Track Recession Compensation

```
INPUT: listTrack
historyGate ← Value for history of gate level
thresholdDistance get Value for distance threshold
for i = 1:sizeof(listTrack) - 1 do
   for j = 1: sizeof(listTrack) + 1 do
      historyDifference ← obtainDifference(listTrack(i), listTrack(j))
      euclideanDistance ← obtainEuclideanDistance(listTrack(i), listTrack(j))
      if historyDifference < historyGate or euclideanDistance < historyGate
      then
         if LifeTime(listTrack(i)) < LifeTime(listTrack(j)) then
            Status(listTrack(i)) ← Invalid
         else
            Status(listTrack(j)) ← Invalid
         end if
      end if
   end for
end for
```

assumed to be distributed evenly in the tracking region. This tracking region is the same adjustable detectable region defined by assumption of a radius in detector module. At the next time step, different tracks are relocated with respect to the last known measurement points which are not yes associated with the existing track. If all measurements have been associated with existing track, the rest of uninitialized tracks are to be relocated randomly in the tracking region[127].

### 4.2.2 Track Recession

If the neighboring tracks share the same measurement, the JPDA filter is prone to shuffle [128], which may cause a duplicate track to be associated with the same object. Thus, a track recession procedure is applied to the tracker to prevent this. The track recession process is implemented based on track history and Euclidean distance of the neighboring tracks. For each individual track, the last $n$ Kalman filter states are stored in the history. Then the difference between corresponding history values of the states toward all other tracks, is calculated. a *history gating level* value is introduced. If the cumulative sum of standard derivation, is less than the history gating level, then the track is considered to be duplicate. Obviously, for duplicate tracks, the track with shorter life is to be deleted to preserve track continuity. On the other hand, the Euclidean distance may be used to further overcome this problem. The Euclidean distance of each tracks are computed toward each other. Then the newer track will be deleted the computed distance is less than physically possible threshold of distance between two moving traffic objects in urban environment.

**Table 4.1:** Track Status Number Description

| Status Number | Name | Description |
|---|---|---|
| 0 | Invalid | Track with invalid or out of the range measurement or even uninitialized track |
| 1-4 | Initializing | Track with newly associated measurement |
| 5 | Already Tracking | Track with associated measurement that passes gating for more than 3 frames |
| 6-10 | Drifting Track | Track with lost measurement. If the corresponding measurement returns, it may return to Tracking status |

### 4.2.3 Track Status and Maturity

Each individual track is assigned a finite status number, to distinguish their track status. The status number, represents the validity and maturity status of each target track. Different values for the status number are represented in Table 4.1. For instance, if a track has its first successful association at time frame k, its status number will change to $1$ as *Initializing* track. If at the time frame $k + 1$ another successful association happens for this track, the status number will be incremented by 1, until it reaches the value $3$. And this trend goes on until the track reaches the *tracking* status. Its obvious that a track in the tracking status will always remain in this status if a valid measurement is valid for it. The same procedure is applied to a track which is in *tracking* status. For instance, if a track is in tracking status at the time step $k$ and it loses its measurement at the time step $k + 1$, its status number will be incremented by 1 and it enters the *drifting* status. If no measurement is associated to the track, until it reaches the status number 10, it will be assumed as an *invalid* track. The important thing is that for every, the state is being filtered as long as the status number is not invalid. This means that a lost track can be recovered using this method, because the track is moving with the lost object until it appears again at the predicted location [2].

### 4.2.4 The Classifier

The classifier tends to classify objects based on whether they are dynamic or static. In urban environment, and in driving scenario a static object is simply an object with zero relative velocity with respect to the ego-vehicle. Although, this physical fact may be useful in other non-real-world applications, in our scenario with all uncertainties and occlusions this can not form the basis for our static object classifier algorithm. Instead, we may make use of the IMM filters. The IMM filter lets

**Algorithm 4** Classifier

```
INPUT: listTrack, historyBox
velocityThreshold ← Value
k ← The current time step
n ← The number of steps to look back
for every Track ∈ listTrack do
  if StatusOf(Track) = Mature & LifeTimeOf(Track) > lifeTimeThreshold &
  ConvergenceOf(Track) = true then
    relativeVelocity ← obtainAverageHistory (Track.realtiveVelocity, history-
    Box, n)
    if relativeVelocity < velocityThreshold & (Track.probablity.M3 <
    Track.probablity.M2 || Track.probablity.M3 < Track.probablity.M1) then
      Track.isDynamic ← false
    else
      Track.isDynamic ← true
    end if
  end if
end for
```

us to distinguish between static and dynamic objects to some degrees. The trick lies in the fact that the static and noisy object is expected to have higher probability to evolve under the random motion model. The regarding random motion model is a stationary stochastic process model with a large process covariance. The algorithm for classifying task is detailed in Algorithm 4.

### 4.2.5 Track Initialization

Informations from the IMM probabilities are combined to velocity informations to form the criteria to classify dynamic and static objects, as in reference [65]. The process is simple: an object with zero or negligible relative velocity is assumed to be static. In addition, another checking is done based on the IMM probabilities of each three motion models. If, the IMM predicts that the probability of the object moving under the random motion model is higher than two other motion models, then it is statistically likely that the object would be static. Another problem is that the estimated velocity is not always smooth. On the other hands, the predicted IMM probabilities may converge after some amount of time. To tackle these problems, the average past velocity is used as a criteria for static/dynamic classification and the IMM filter is checked to determine whether it has converged to a steady-state value before using it as a classifier criteria.

## 4.3 The Bounding Box Tracker

As we have mentioned in previous sections, the result of the IMM-UKF-JPDAF filter is the estimated pose for the point-wise objects. The dimensional and heading information of the corresponding bounding boxes must be further incorporated to the filtering process. A logic-based algorithm is implemented in order to tackle this problem. Three different major tasks must be accomplished in this algorithm/procedure:

1. The detected boxes must be associated to an existing track. This is done due to retain the detector's best known output.

2. Update the retained bounding box parameter if a better detection output is available.

3. Geometrical corrections to include occlusions and ego-vehicle perspective change.

The range and point-of-view-dependent occlusion characteristics of the LiDAR sensors are used to derive the logical rules needed for the bounding box tracker. The track maturity information from the IMM-UKF-JPDAF tracker is also used to distinguish noise from the real object.

### 4.3.1 Bounding Box Association

We have said that in the regarding filtering process of the JPDA tracker, multiple measurements are associated with the tracker. This is done due the need for augmenting update step of the filtering. Unfortunately, this may not be extended to the bounding box attributes, since the bounding box dimensions are not among the filtered states. Thus, the bounding box attributes must be mapped to each track, exclusively. At last, the object tracker can not give a hypotheses pr probabilistic estimation of the situation of bounding boxes to the higher-level perception module. A single best estimate of each bounding box must be passed to other sections of the perception module. An algorithm for association of the box tracking procedure is provided in this thesis.

The bounding box association is only happened when the target track is a mature track according to the regarding thresholds. In addition, the target track must have passed a certain lifetime threshold to be associated to a bounding box. If a target track is associated only with a single measurement, no further check is required because the on-to-one condition is fulfilled in this situation. Furthermore, the Mahalanobis distance is computed in order to obtain the right measurements. Therefore, it can be assured that the generated bounding box will place within

**Algorithm 5** Bounding Box Association

**INPUT:** listTrack, listBox, distanceThreshold, lifeTimeThreshold
**for each** Track $\in$ listTrack **do**
  **if** ***StatusOf***(Track) = Mature & ***LifeTimeOf***(Track) $>$ lifeTimeThreshold **then**
    nearestMeasurement $\leftarrow$ ***getNearestEuclideanCenterPoint***
    **if** ***EuclideanDistance***(nearestMeasurement, ***CenterPointOF***(Track)) $<$ distanceThreshold **then**
      measurementIndex $\leftarrow$ getIndex(nearestMeasurement)
      Track.AssociatedBox $\leftarrow$ listBox(measurementIndex)
    **else**
      Track.AssociatedBox $\leftarrow$ empty
    **end if**
    **if** ***SizeOf***(Track.MeasuredCenterPoints) $>$ 1 **then**
      ***AmbiguityOf***(Track) $\leftarrow$ true
    **else**
      ***AmbiguityOf***(Track) $\leftarrow$ false
    **end if**
  **end if**
**end for**

the gating area and may be assumed to be statistically likely. Therefore by using this trick, this methods prevents a clutter-related box to be associated with a valid track.

On the other hand, if a track is associated with multiple measurements, the box with the closest Euclidean distance is preferred over the ones that are further. Gating process based on adjustable distance threshold is also done to prevent impossible association. The association criterion is purposely made to be simplistic, since the measurement has actually passed more rigorous gating by the IMM-UKF-JPDA position tracker. Therefore, although a false association is possible, the likelihood such association coming from a mature track that has been cruising over certain time threshold is expected to be low [2].

### 4.3.2 Dimension and Heading

In the preceding sections, it is mentioned that the detection module passes an initial list of bounding boxes along with their attributes to the tracking module. Furthermore, it is mentioned that regarding attributes for each bounding box contains the height, width, length and the yaw heading value. These box parameters where obtained based on minimum volume box fitting and later a yaw correction step, based on rotating calipers method is implemented to give the correct values for the orientation of boxes. This methodology is guaranteed to correctly work for a complete measurement of the target object. But, how we can be sure of the results of this calculations if our measurement is not complete (all parts of the object are not observable due to different issues)?

Complete measurements of the LiDAR sensor can only be obtained if the object falls within a certain range of the sensor. Furthermore, no occlusion must be happened if a measurement is to be assumed complete.In other words, as the object goes closer to ego vehicle, more part of a detected object is expected to be seen by LIDAR sensor, and the non-observable part would be reduced [2]. A major challenge, regarding LiDAR measurement is that when an occlusion happens no distinguishable edge line will be found in the cluster. The orientation correction method is no longer valid for this kind of clusters and the result will be erroneous. On the other hand, although in the motion model incorporated in UKF position tracker, the yaw and yaw rate value are being updated they are not applicable to the bounding boxes. The reason is that, the UKF is updating the yaw values based on the assumption that the target track is an infinitely small object or a point. Therefore, these point-wise estimated values for the yaw, are not same as the bounding box heading. Furthermore, road participants are not to necessarily travel toward their yaw heading. An example for this effect is the side-slip angle, happening for a vehicle moving at relatively high-speed [11].

Another example would be the significant discord between the yaw heading and the bounding box heading which is caused by the moving sensor effect. If the reference of the sensor is assumed to be moving constantly, when the ego-vehicle rotates, the surrounding targets will seem as they are maneuvering by the sensor. Thus, the yaw estimate will differ with the original yaw, as the yaw estimation is changing due to this effect while the target object itself does not have any change in its orientation. Another example would be of the parked objects. a car parked diagonally should not have its bounding box heading set to its yaw estimate. To summarize; yaw estimate heading is distinct with bounding box heading and the bounding box should reflect detection results.

Different scenarios for bounding box behavior is depicted in figure 4.2. As the time step increases, the detected objects 1 and 2 are coming closer to the ego-vehicle. As these objects are moving closer into the proximity of the ego-vehicle, the bounding box grows larger as more parts of objects would be observable. This trend would be the same if the detected objects go away from the ego-vehicle. If the detected object moves away from the effective range of the LiDAR sensor, the bounding box gets smaller. The wrong yaw estimate can be observed on bounding box fitting of ”Detected Object 2” at time step k, as there is no L-shape to fit the box, the resulting bounding box has an incorrect heading. Finally, another scenario in which the bounding box would go smaller is when occlusion and/or over-segmentation occurs. ”Detected Object 3” is split into 2 boxes due to tree-like ”Occluding Object”.

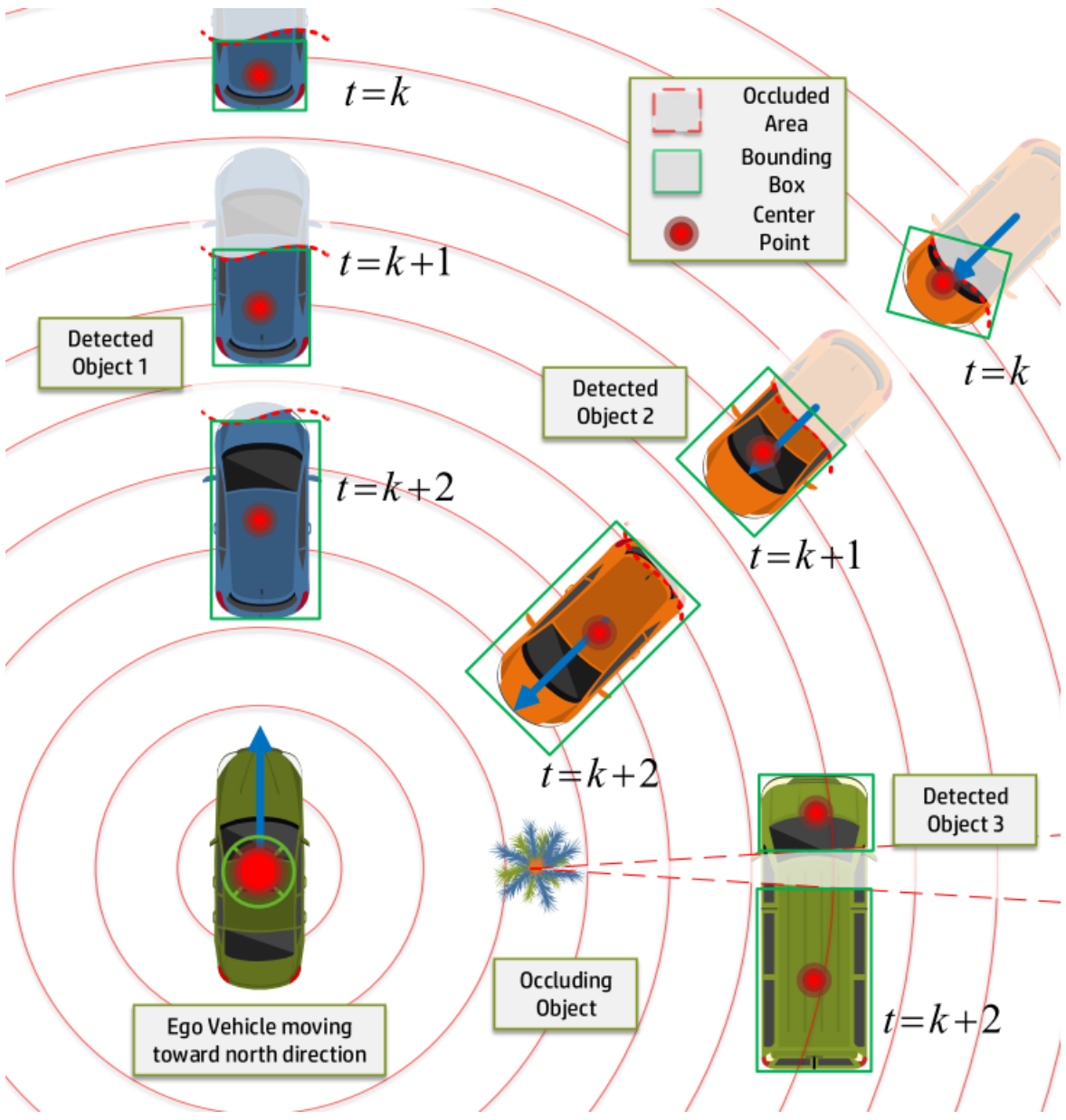

**Figure 4.2:** Different scenarios for bounding box behavior. Figure is adapted from [2].

In comparison to the extended object tracking methods, which try to track objects along with their dimension and geometrical shapes as state values, the rule-base bounding box tracking method enjoys the benefit of being more computationally efficient. On the other hand, the kind of uncertainties involved in bounding box detections are not so complicated to demand more complex tracking methods. Figure 4.2 shows that dimension and yaw heading of the detected bounding boxes are constantly changing due to different causes. In the discussion, we have tried to give general heuristic rule for these changing behavior. This rules are meant to give us an exact routine to update regarding attributes for every bounding box in each time frame. Three general heuristic rule are elaborated during the implementation phase of this thesis:

1. Bounding boxes are not expected to shrink in general.
2. According to the behavior depicted in the figure, further we can conclude that the length value for the bounding box is going to change when self-occlusion happens.
3. Bounding boxes are not expected to rotate so fast. it is physically improbable for car-like (without loss of generality) object to change yaw heading

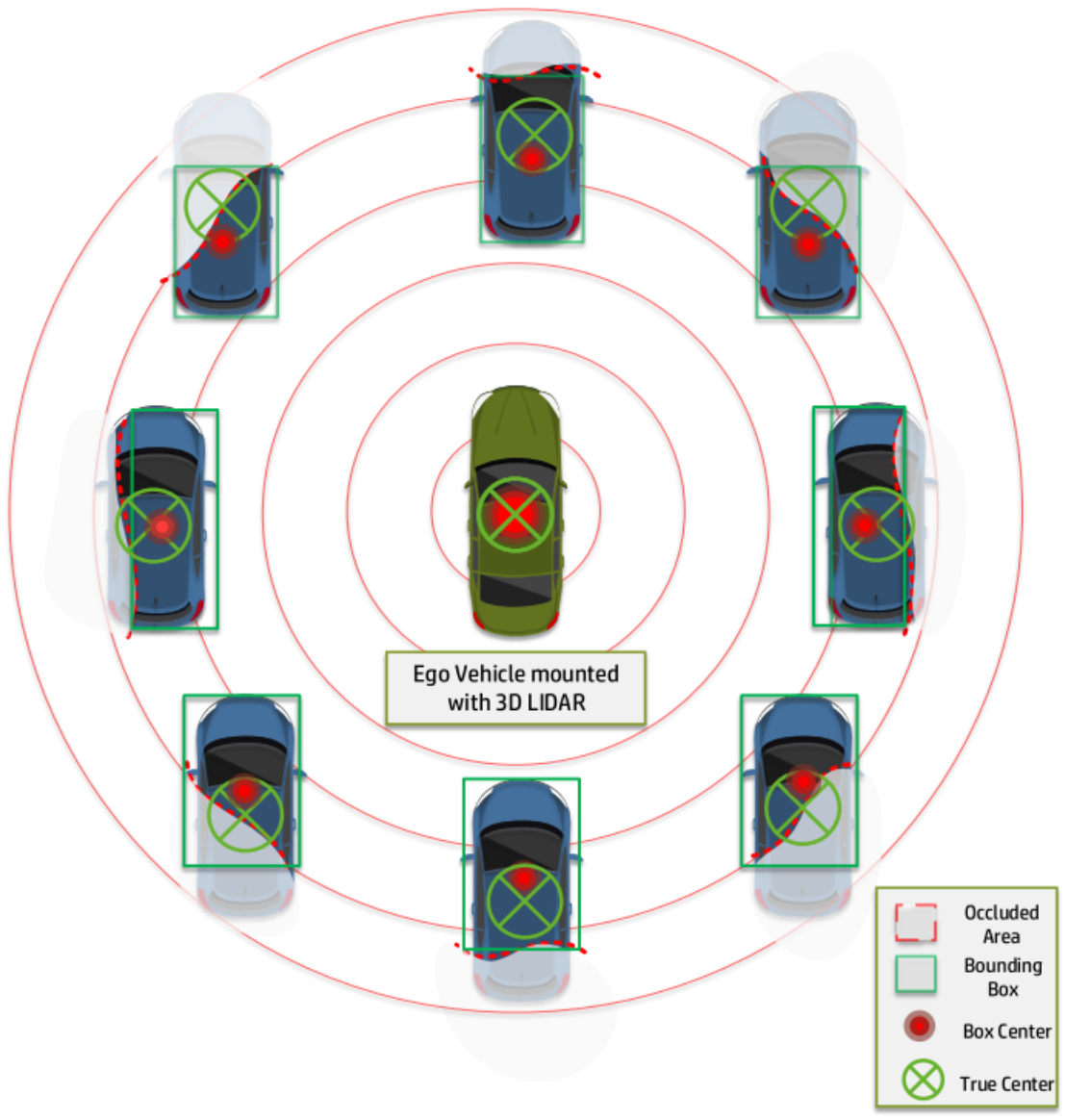

**Figure 4.3:** Self-occlusion of the detected objects due to changing view-point of the ego-vehicle. It is obvious that, self-occlusion may lead to center point displacement.

significantly between sensor sampling time (100 ms) on a normal road.

4. Bounding boxes are not expected to show a sudden movement in the opposite direction. If the a bounding box is moving in a certain heading direction, in the next time step, it is not expected to change that suddenly.

5. If an occlusion of shrinkage happens, a sudden shift will be observed in the placement of the center point of the bounding box. This may lead to an unnatural backward movement of the detected object.

These five rules are going to simply describe the uncertain behavior of the output coming from the detector, while occlusions are sporadic. Those bounding boxes which violate these simple rules are assumed to be just a noise. Some real-world-based insights may be made into the algorithm. For example, The dimension of a certain bounding box never gets larger than the real object. An exception may happen on the case of under-segmentation of the possible case of failing the over-segmentation handling. Another unlikely situation to happen with target objects is the sudden movement of the object in the opposite direction. Traffic objects are not to simply jump back within ordinary LiDAR time fram ($100ms$). There is no clear boundary to distinguish the real yaw change due to object's maneuver and the yaw change due to the detection noise. This causes the rule-based filtering method to be only capable of catching the large erroneous yaw changes. The algorithm, uses a priori history for each data frame. The information provided from IMM-UKF-JPDA tracker is used to augment the box tracker. The bounding box tracker, stores the last good bounding box at the time step $k-1$ to further compare it to the newly generated bounding box at the time step $k$. The bounding box regarding the history data, which is said to be the last-known good bounding box is actually obtained during the bounding box association process. Then, these two different bounding boxes are compared to each other, and if the newly generated one, which is generated at time step $k$, passes the rule-based filter of this section, then necessary update steps will be taken place. The dimension, the yaw or both of them may be updated for the regarding bounding box. If rules, reject the new bounding box, the last-best-know bounding box will be retained.

If the occlusions are not sporadic, or stay constantly, this rule-based algorithm is not capable of getting good updates for the yaw changes. Persistent occlusions happen for example when an object is moving further away from the ego-vehicle and therefore shows a similar relative velocity in all frames. This challenge may be tackled by taking informations of the bounding box from few steps back. An adjustable look-back-steps parameters $\zeta$ will be assumed to control the number of steps to look back to detect if bounding box informations are not changing. Therefore, box informations from $k-\zeta$ backward steps are analyzed to see if the box has not been updated for $\zeta$ steps back. This indicates that the regarding bounding box attributes, would need to be updated. For the static objects which have zero relative velocity, the velocity information from the position tracker is used to rule out change of yaw and dimension reductions.

**Algorithm 6** Bounding Box's Dimension and Heading Update

**INPUT:** listTargetTrack, listDetectedBoxes
$k \leftarrow$ The current time step
$\zeta \leftarrow$ The amount of look-back step
boundingBoxSmallToleration $\leftarrow$ A small threshold for change in the bounding box
boundingBoxYawChangeThreshold $\leftarrow$ Adjust the parameter
boundingBoxAreaChangeThreshold $\leftarrow$ Adjust the parameter
boundingBoxVelocityThreshold $\leftarrow$ Adjust the parameter

### 4.3.3 Self-occlusion Handling

Self-occlusion of target tracks is a common source for detection noise which is highly dependent on the perspective of the ego-vehicle with respect to the target

**Algorithm 7** Self-occlusion Handling

**INPUT:** currentTargetTracks, boxHistory
boundingBoxVelocityThreshold $\leftarrow$ Adjust the parameter
**if** currentTargetTrack.AssociateBoundingBox.pose.x $>$ 0 **and** currentTargetTrack.relativeVelocity $<$ -boundingBoxVelocityThreshold **then**
  shiftBoundingBoxForward(currentTargetTrack.AssociateBoundingBox)
**else if** currentTargetTrack.AssociateBoundingBox.pose.X $<$ 0 and currentTargetTrack.relativeVelocity $>$ boundingBoxVelocityThreshold **then**
  shiftBoundingBoxBackward(currentTargetTrack.AssociateBoundingBox)
**end if**

object. The box center points are effected directly by the self-occlusion. The position tracker sub-module is only aware of the position of the bounding box's center point, therefore any inconsistency in their position will be very harmful to the tracking module. If self-occlusion happens, a center point correction procedure must be applied to the box tracker. The last-best-known box information form the previous subsection is used to correct the center point position. The center point's location $(x, y)$ is reported with respect to the ego-vehicle's navigation frame. On the other hand, the shift in the occluded bounding box which is reported in local coordinate frame $(x', y')$, must be projected to the global frame and then to the ego-vehicle's frame.

The center point shifting is illustrated in Figure 4.3. The mathematical relationship between the center point of the newly shifted bounding box $(x_s, y_s)$ and the original center point of the occluded bounding box is of the following form:

$$\begin{bmatrix} x_s \\ y_s \end{bmatrix} = \begin{bmatrix} x_{original} \\ y_{original} \end{bmatrix} + \begin{bmatrix} \cos(\psi) & -\sin(\psi) \\ \sin(\psi) & \cos(\psi) \end{bmatrix} \begin{bmatrix} \Delta'_x \\ \Delta'_y \end{bmatrix} \tag{4.1}$$

The geometrical informations of the differences between length $l$, width $w$ and the position of the center point $(x, y)$ is used to determine the center point shift values $\Delta'_x$ and $\Delta'_y$.

$$\begin{aligned} \Delta'_x &= (x_{TB} \pm \frac{1}{2} l_{TB}) - (x_{OB} \pm \frac{1}{2} l_{OB}) \mp \frac{1}{2}(w_{TB} - w_{OB}) \\ \Delta'_y &= (y_{TB} \pm \frac{1}{2} w_{TB}) - (y_{OB} \pm \frac{1}{2} w_{OB}) \mp \frac{1}{2}(w_{TB} - w_{OB}) \end{aligned} \tag{4.2}$$

The shifting direction of the bounding box will determine the usage of $\pm$ operator. In the case of back occlusion, the $+$ operator is used to shift the box to toward its top edge. Also, in the case of frontal occlusion the $-$ operator is used to shift the box toward its bottom edge. The shifting direction itself is determined by the object location in Cartesian coordinate relative to ego-vehicle. The main challenge for this algorithm is to distinguish the real shifted boxes with the imperfect boxes which are not caused by self-occlusion. In order to tackle this problem, the relative velocity in the longitudinal direction is checked. If the object is moving in the same direction with the ego-vehicle, the shift toward the vehicle is incorrect and vice versa. On the other hand, even static objects would have some velocity in longitudinal direction due to the noise, therefore a threshold is set to manage the corrections. These procedures are summarized in Algorithm 7.

# Chapter 5

# System Design and Implementation: Detector Module

## 5.1 Overview

Until now, theoretical foundation and mathematical tools for multiple object detection and tracking is represented. In this chapter, design process and implementation steps are discussed in details. The architecture of the system is presented to give a big picture about how the algorithm is going to work. After that, each of the system's building blocks are represented in details. In each section of this chapter, it is detailed that how theoretical aspects of the design is implemented in practice.

### 5.1.1 System Architecture

The DaTMO framework consists of two major building blocks as depicted in Figure 5.1: Detector and Tracker. The detector module tends to produce valid and unambiguous segmented objects from the input raw LiDAR data. The responsibility of the tracker module is to obtain and maintain dynamic attributes of these segmented objects across all of incoming data frames. The hierarchical structure and data flow in the designed system is depicted in the figure. The input to the system is the raw LiDAR data gathered from surrounding environment. The regarding spatial data is represented by three-dimensional point cloud data and in Cartesian coordinates. The output of the system can be the list of objects with their embedded dynamic attributes or trajectories in the environment.

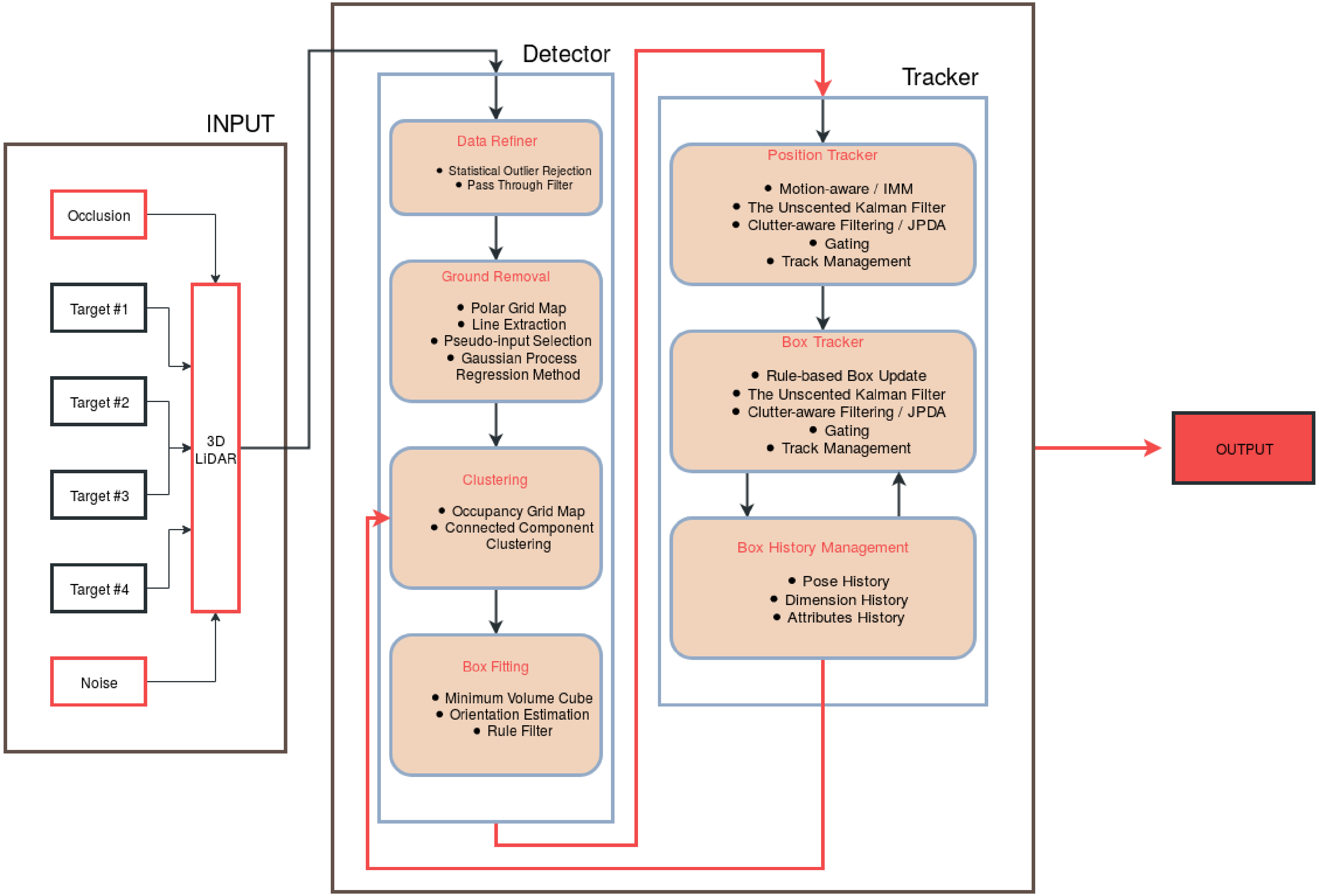

**Figure 5.1:** The DaTMO System Design.

The detection module consists of four different sub-modules: *data refiner*, *ground removal*, *clustering*, *box fitting*. The data refiner is the first procedure implemented on the raw input data. It rejects the out-lier points that may be found in the data, also neglects all of the data placed further than a defined radius. Ground removal is necessary to reduce the dimensionality of the raw LiDAR data. This procedure helps to delete all of the points which are not interesting to their irrelevance to any object. In addition, ground estimation is necessary for the further navigational goals of the ALV, because it provides the ALV with the deeper understanding of the ground it's maneuvering on.

Clustering module segments the data into different coherently grouped clusters of points. The box fitting module, assigns a simple representation frame for each clustered group of point clouds. This representation is designed to be the Minimum Volume Bounding Box (MVBB) fitted to the point cloud. In addition to the dimensions of the fitted box, the heading estimation is performed to compensate the errors in heading value computation.

The tracker module receives the list of bounding boxes for each frame. Then, it is responsible for keeping and maintaining the time-evolution of the dimensional and spatial attributes of each bounding box. The position tracker and bounding box tracker sub-modules are responsible for this task. The box's dimension is expected to stay constant by heading change, while other event like changing view point or occlusion may change the dimensions. Furthermore, spatial evolution of the boxes are expected to change according to the motion model. As both values are perturbed by noise and uncertainties, a probabilistic framework such as Bayesian filtering must be utilized in position tracker algorithm. The box history is responsible for storing of each track's information. The box tracker sub-module uses this stored information to update the bounding box list using noise rejection.

This system design is highly inter-related. Each iteration of the system is highly sequential. Each sub-modules of the MODaT system relies heavily on the proceeding sub-module to perform its task. The architecture is based on Tracking By Detection (TBD) scheme which is introduced in Section 2.4. Furthermore, The Bottom-Up Top-Down approach introduced in reference [26] is used. The top component is the detector which is in a feed-back relationship with the bottom component or the tracker. The feed-back is constructed to reduce false detections.

### 5.1.2 Development Framework

The whole package is developed in C++ programming language. The Qt Creator [1] and Qt Toolkit[2] is used as the cross-platform. Furthermore, the Robot Operating

[1] https://doc.qt.io/qtcreator/

[2] https://www.qt.io/

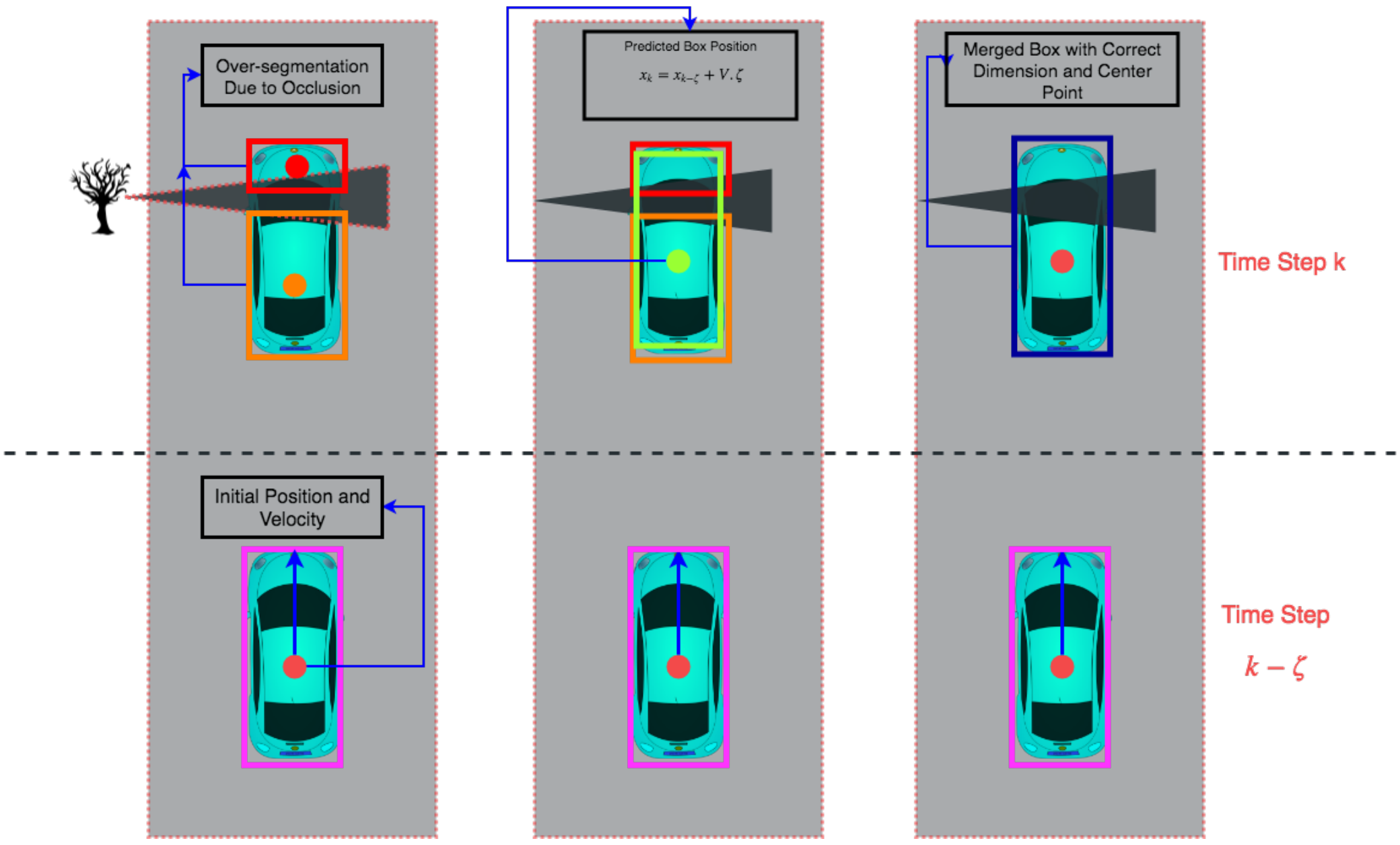

**Figure 5.2:** Over-segmentation handling procedure.

System (ROS)[3] is used to add different components to each other. The benchmark device for results is a ordinary-grade developer laptop (Intel® Core™ i7 6700HQ Processor with 12 GB of RAM, running UBUNTU GNU/Linux 16.04.4 LTS). The Qt Toolkit and the UBUNTU operating system is used in order to make the software package usable for other industry related experiments. The C++ implementation, makes the package to run as a pre-compiled binary program which prevents the overhead in runtime. Some third party libraries are used in the package. The Point Cloud Library(PCL) [60] is used for point cloud manipulations. Eigen3 [4] is used for matrix manipulations. ROS package is used for systematic formulations regarding to robot-based implementation.

The software package is intended to be further implemented on a car embedded computer. Therefore, the system must be real-time implementable with ordinary computing power. In this thesis, the designed tracking system is ensured to be faster than the LiDAR sensor sampling rate ($10Hz$).

[3] https://www.ros.org/
[4] https://eigen.tuxfamily.org/

## 5.2 Detector

The detector is responsible for preprocessing of the raw LiDAR data. The main task of the detector is to finally pass the list of bounding boxes along with their dimensional and heading attributes to the tracker. Each sub-module of the detector component is discussed ind details.

### 5.2.1 Ground Removal

The raw, refined LiDAR data consists of about $5 \times 10^5$ data points with the majority of them being ground points carrying no information about the objects in the scene. Although, this majority of the data points does not carry meaningful information, truly detecting them is of great importance to the MODaT task. False detections in this sub-module may lead to neglecting meaningful information by deleting point data regarding to an object. On the other hand, failure to detect a vertical structure belonging to a rough ground scene, may lead to unwanted accidents. Another issue arises in urban environments, while driving scenes are not limited to show flat driving paths. In other words, a successful ground removal algorithm must be capable of estimating both flat and sloped terrains, while in the literature often the sloped terrains are neglected. In thesis a novel method for ground segmentation is represented which is discussed in details in Chapter3. A physically-motivated method based on Gaussian process regression is presented which is capable of real-time estimation of the ground even in rough scenes. In this section, implementation issues and results of the algorithm is discussed.

The ground removal sub-module starts with the polar grid map, which is of great importance for the speed and precision of the ground segmentation procedure. The theoretical background of polar grid mapping is discussed in Section 3.3. After the implementation of the polar grid map the line extraction algorithm is implemented. These extracted lines will be further used for pseudo-input selection. The pseudo-input selection process produces the physically-motivated input sets for the second Gaussian process used for length-scale updates. A full library is written in C++ for general Gaussian process regression task in gradient-based optimization. Using this library, the inference is done. After removing the ground points from the data, the remaining points are called *elevated points* in the literature. A schematic of the elevated points, regarding to line extraction in each segment is depicted in Figure 5.3.

Some methods, tend to compute slope between adjacent cells, to get an insight about elevated points [1]. This approach is good for smooth terrains, but in rough terrains or even urban scenes, a slight road bump or even a smooth step in the road will be classified as elevated point. Our method, on the other hand tends to estimate a continuous model for the ground. By assuming all of the lines in each segment, and introducing the length-scale values and the pseudo-input set, a general-view-based estimation is obtained. The algorithm for ground segmentation is represented in Algorithm 2.

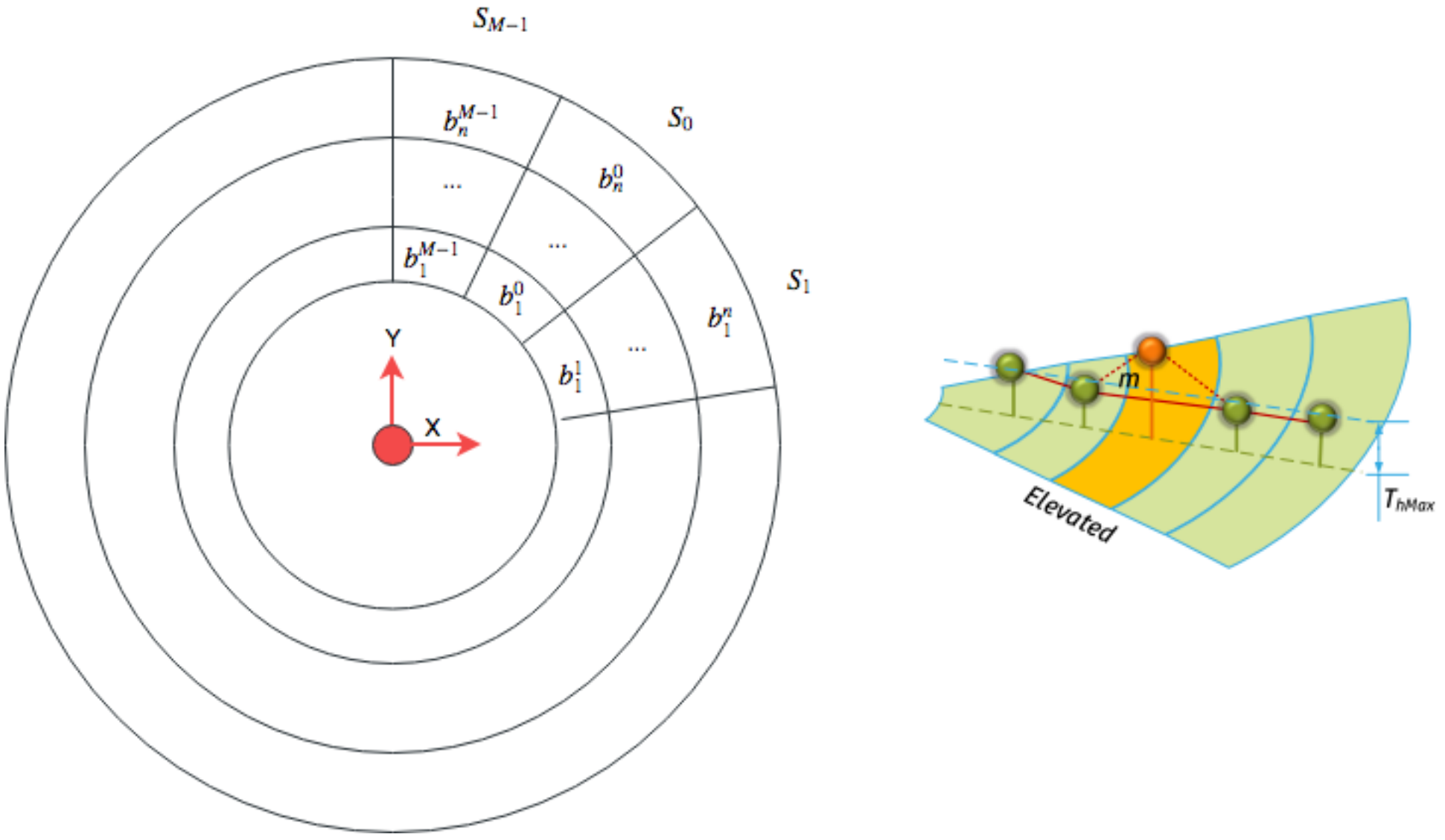

**Figure 5.3:** The polar grid map along with the elevated points. The polar grid map (left) and a cut from a segment, containing adjacent cells (right). The right figure is adapted from [1].

**Implementation Results** For implementation purposes the input space is divided into $M = 180$ different segments covering a 2–degree portion of the environment and each segment is divided into $N = 120$ bins. For line extraction purpose $\zeta_m$ is set to $0.02$ radians and $\zeta_b$ is set to $15^{cm}$, therefore, any line structure with height difference less than $0.15^m$ and less than $0.02$ radians deviation from mean angle are assumed to be loyal to a unique line. Regression prediction error is validated using standardized mean squared error by

$$\text{SMSE} \equiv \frac{1}{n}\sum_{i=1}^{n}\left[\text{var}(y)^{-1}(y_i - \mu_i)^2\right] \tag{5.1}$$

All the results are obtained on a laptop with Intel Core i7 6700HQ processor and simulations are implemented using point cloud library (PCL) and with C++, while figures are prepared by MATLAB.

Figure 5.4 shows two different adjacent segments and their ground segmentation result, in which, red circles represent the estimated ground for data points, while black squares are the raw data. As it can be seen in the first row, Figure 5.4 $a$ and $b$, predicted ground for two adjacent segments have a flat structure, although segment $a$ shows a sloped structure in that area and segment $b$ shows an uneven depth. While the ground candidate set showed in the figure by connected blue line proposes other structure, ground segmentation algorithm predicts a flat ground for this area as a result of considering all the segments together and having a whole-frame. This results shows that general point of view does not obey a segment–wise logic. On the other hand in segments shown in part $c$ and $d$ of this figure, although data in segment $d$ covers a wider region, algorithm tends to estimate detailed structure of ground in both segments regardless of what local ground candidate set may impose based on segment–wise logic. This feature enables the algorithm to truly recognize ground in radial distances ranged from $6$ to $8$, although segment $d$ offers a more smooth ground shape from its data in that range.

**Table 5.1:** Ground Prediction Results

| $n_{PG_m}/n_{\mathcal{L}}$ | $\sigma_f^2$ | $\sigma_n^2$ | $\bar{\sigma}_f^2$ | $\bar{\sigma}_l^2$ | $\bar{\sigma}_n^2$ | SMSE$_{Z-\mathcal{L}}$ | SMSE$_Z$ | $t_{Z-\mathcal{L}}(s)$ | $t_Z(s)$ |
|---|---|---|---|---|---|---|---|---|---|
| 36/23 | 0.3663 | 0.8607 | 1.415e+6 | 2.250e+6 | 1.187e+6 | 2.4258e-27 | 1.4978e-23 | 0.5432 | 0.2335 |
| 52/30 | 0.3117 | 0.7207 | 4.729e+6 | 1.347e+6 | 2.058e+6 | 1.2466e-28 | 1.1231e-23 | 1.0045 | 0.4356 |
| 68/16 | 0.4896 | 1.1544 | 1.381e+6 | 3.881e+6 | 4.501e+6 | 2.7561e-26 | 1.4102e-21 | 0.3283 | 0.1963 |
| 81/14 | 0.1103 | 1.6487 | 4.588e+6 | 3.521e+6 | 4.402e+6 | 4.2903e-19 | 8.6733e-22 | 0.2732 | 0.3412 |

Furthermore, ground segmentation method of [66] is implemented on the same data. Detailed results of the implementation is given in Table 5.1. The first two rows are corresponding estimated values for figure 5.4 $a$ and $b$ with $n_{PG_m}/n_{\mathcal{L}}$ being the number of ground candidate points versus number of selected pseudo-input set. It is important to notice that reducing $n_{\mathcal{L}}$ by weakening pseudo-input selection criteria will increase speed of the given algorithm with the expense of reducing the precision. This enables a trade-off between precision and speed of the algorithm. Given results shows that in these two segments, choosing a large pseudo-input set gives a more precise ground estimation result than the other method while making it slower. However it is still real-time applicable and implementable for urban driving scenes. The last two rows are corresponding results for figure 5.4 $c$ and $d$ while it can be seen that by choosing a small pseudo-input set relating to ground candidate set $14/81$ makes our prediction faster while reduces the precision. This comes from the fact that as the size of pseudo-input set gets larger, the gradient-based optimization step becomes more time consuming. The hyper-parameters and time values reported in table 5.1 are recorded for ground segmentation in the corresponding frame while in figure 5.4 just selected segments are depicted.

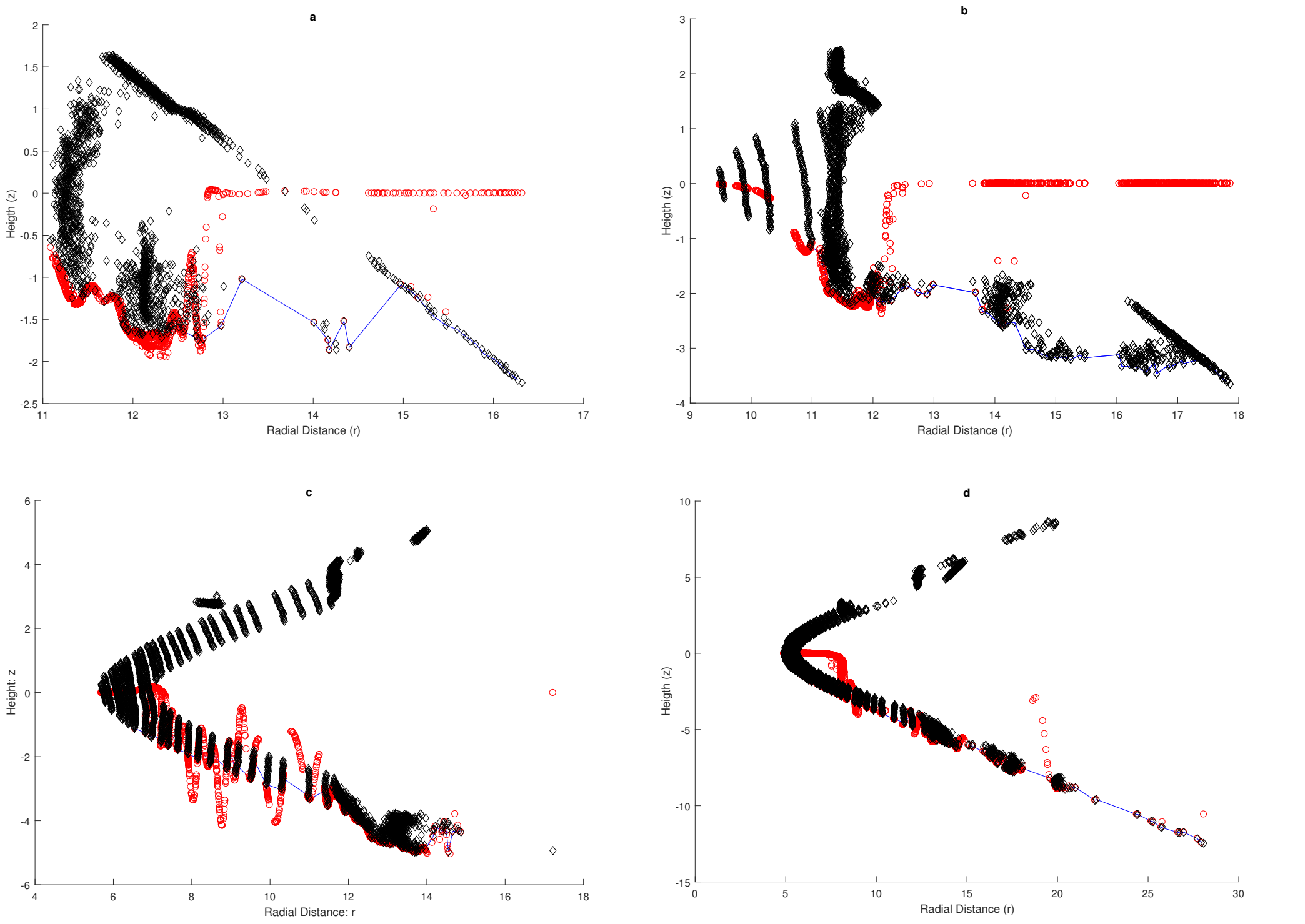

**Figure 5.4:** Ground segmentation results for two different adjacent segments: Red circles represent estimated ground for data points and black squares are the raw data. Blue line is the ground candidate set for each segment.

As it is seen in these comparison results the proposed method denoted by $Z - \mathcal{L}$ outperforms the conventional method in precision, while being fast enough for urban driving scenarios and robust to the locally changing characteristics of input point cloud. Calculated SMSE error shows that the proposed method is more precise than that of the method given in [66], and its speed is related to the ratio of number of points in pseudo–input set to the number of points in ground candidate set. The SMSE error for each segment is reported as the mean value of 100 iterations of ground prediction. Furthermore, values of hyper-parameters are reported. In order to consider different scales of covariance kernels, while scaled gradient-based optimization method is used to find optimal hyper–parameters. This real-time applicable ground segmentation procedure may be applied efficiently in applications, where fast and precise clustering and ground segmentation is needed to enable further real-time processes like path planning or dynamic object recognition and tracking especially in rough scenes. Furthermore, taking

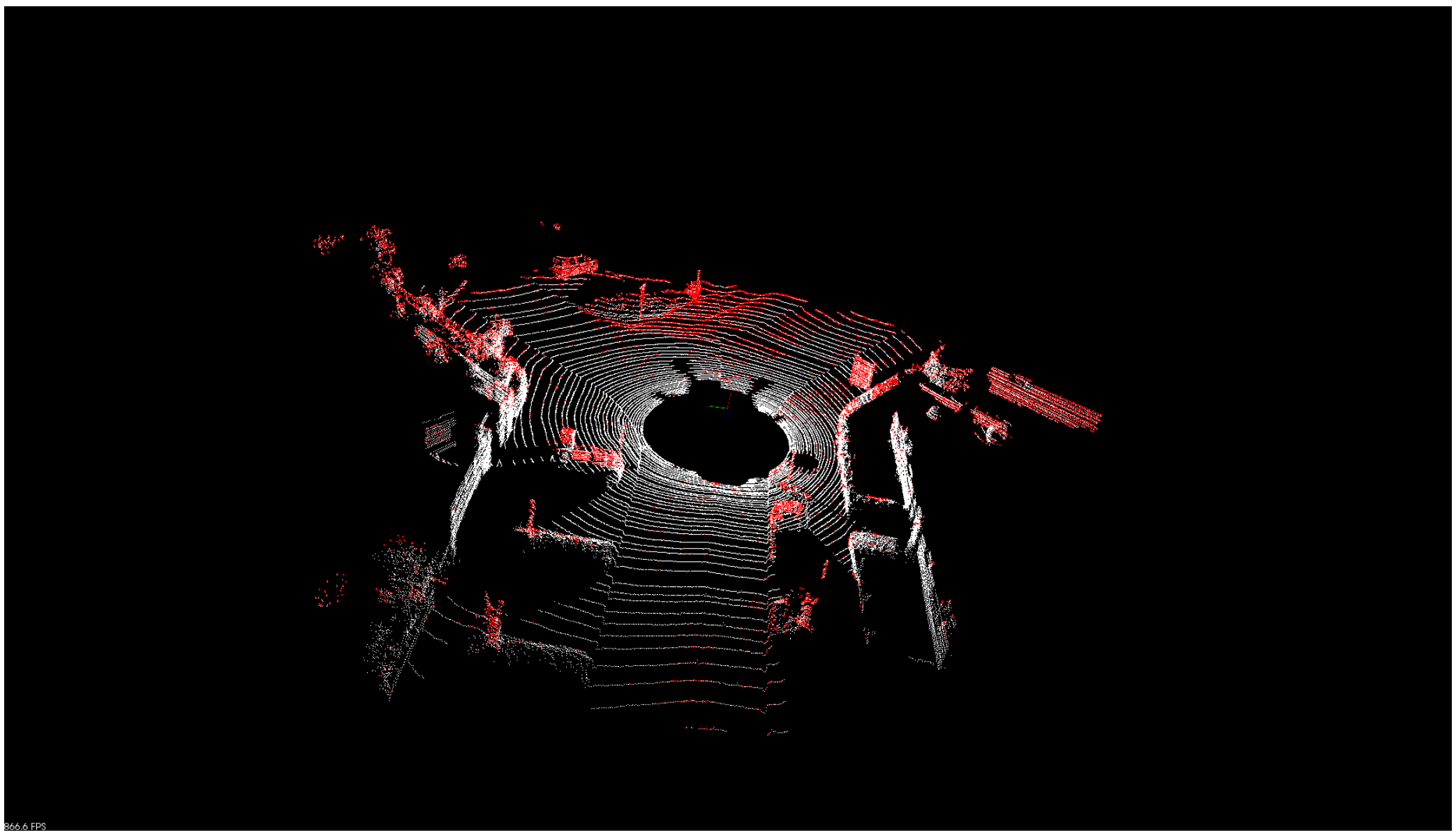

**Figure 5.5:** Ground estimation result implemented for a single data frame. White–labeled points are indicators for the ground points and red–labeled points are identified as elevated points or non–ground points. A video of the implementation process is available at https://aras.kntu.ac.ir /mehrabi/thesis-materials

.

location-dependent characteristics of non-linear regression into accounts, enables the method to show better performance in rough scenes. In addition, the initial guess of length-scale parameters which is based on fast line extraction algorithm, increases the time efficiency of the proposed method by providing physical motivated initial guess for the $\bar{\mathcal{L}}$ vector. Furthermore, the physical motivated procedure of choosing each segment's length-scale vector, gives the method a sense of intuition which is related to the ground quality. This intuition which is behind choosing the most important parameters of the regression method, ensures the fair calculation of ground at each individual frame with respect to its specifications.

**Conclusions** A physically-motivated ground segmentation method is proposed based on Gaussian process regression methodology. Non-smoothness of LIDAR data is introduced into the regression task by choosing a non-stationary covariance function represented in equation 3.24 for the main process. Furthermore, local characteristics of the data is introduced into the method by considering non-constant length-scales for these covariance function. A latent Gaussian process is put on the logarithm of the length-scales to result a physically-motivated estimation of local characteristics of the main process. A pseudo-input set is introduced for the latent process that is selected with a whole-frame view of the data and based on the ground quality of the data in each segment by assuming the related correlation of data points that are gathered from same surface.

It is verified in this section that the proposed method outperforms conventional methods while being realistic, precise and real-time applicable. Furthermore, presented results shows that proposed method is capable of effective estimation of the ground in rough scenes. While the ground structure in the given example in figure 5.4 may be assumed to be rough as it contains bumpy structures and sloped obstacles, the proposed method is capable to detect ground model efficiently and precisely.

---

**Algorithm 8** Clustering

---

$m \leftarrow$ Number of grid rows
$n \leftarrow$ Number of grid columns
***Function: RCC***(initialCartesianGridMap,clusteredGridMap)
initialCartesianGridMap $\leftarrow$ minusOne(initialCartesianGridMap)
clusterID $\leftarrow$ 0
FindRegardingComponent(clusteredGridMap, clusterID)
***End RCC***
***Function: FindRegardingComponent***(clusteredGridMap, clusterID)
**for** cellRow = 1:m **do**
    **for** cellCol = 1:n **do**
        clusterID += 1
        PerformSearch(clusterGridMap, clusterID, cellRow, cellCol)
    **end for**
**end for**
***End FindRegardingComponent***
***Function: PerformSearch***(clusterGridMap, clusterID, cellRow, cellCol)
clusteredGridMap(cellRow, cellCol) $\leftarrow$ label
Neighbors $\leftarrow$ gatherNeighbors(cellRow, cellCol)
**for** (x,y) $\in$ Neighbors **do**
    **if** clusteredGridMap(x,y) = -1 **then**
        ***Function: PerformSearch***(clusterGridMap, clusterID, cellRow, cellCol)
    **end if**
**end for**
***End PerformSearch***

---

**Algorithm 9** Over-segmentation Handling

```
INPUT: listCluster, listTargetTracks, boxHistory
The current time-step is obtained → k
Adjust the parameter → sharedPercThreshold
Adjust the parameter → totalAreaThreshold
Create an empty list → mergeCandidates
for Each cluster ∈ listCluster and each box ∈ boxHistory(k-1) do
   commonArea ← computeCommonArea(cluster object, box object)
   if commonArea > sharedPercTrheshold × cluster.MinimumArea then
      mergeCandidates.AddCluster (cluster)
   end if
end for
for Each cluster ∈ listCluster and each predictedBox ∈ mergeCandidates do
   mergedCluster ← mergeCluster(mergeCandidates.clusters)
   if mergedCluster.area > totalAreaThreshold × predictedBox.Area then
      listCluster.push(mergedCluster)
      listCluster.erase(mergeCandidates.clusters)
   end if
end for
```

### 5.2.2 Object Clustering

The output of the ground removal sub–module is the set of all elevated points of each regarding data frame. Next, each different potent object in these elevated points must be distinguished. The clustering procedure has the responsibility to perform this task. The Connected Component Clustering method [113] is a method represented for finding connected components of the 2D binary images. In driving scenario, potent objects are not expected to represent vertical displacement, thus LiDAR point cloud data regarding these objects may be treated as 2D binary images for clustering purposes. The other issue that enables the utilization of this method for 3D LiDAR data in driving scenarios is that in driving scenario objects are not expected to pile vertically. Therefore, a top-view of the LiDAR data may be treated as a binary image.

All the 3D data is projected into the $x - y$ plane. Then the connected component clustering algorithm is implemented on the 2D data. The value of height may be retrieved after clusters are detected, therefore, no information is lost in this process. The height information is of great importance because it is used in the next step, while bounding box fitting is taking place. Using this method enables the MODaT software to remain real-time while performs precisely and acceptable for driving expectations. The two-pass method [114, 115] is used to perform the task. One pass is assumed to the temporary label of *connectedness* and the second pass is assumed to be the *unique cluster ID* as depicted in figure 5.6.

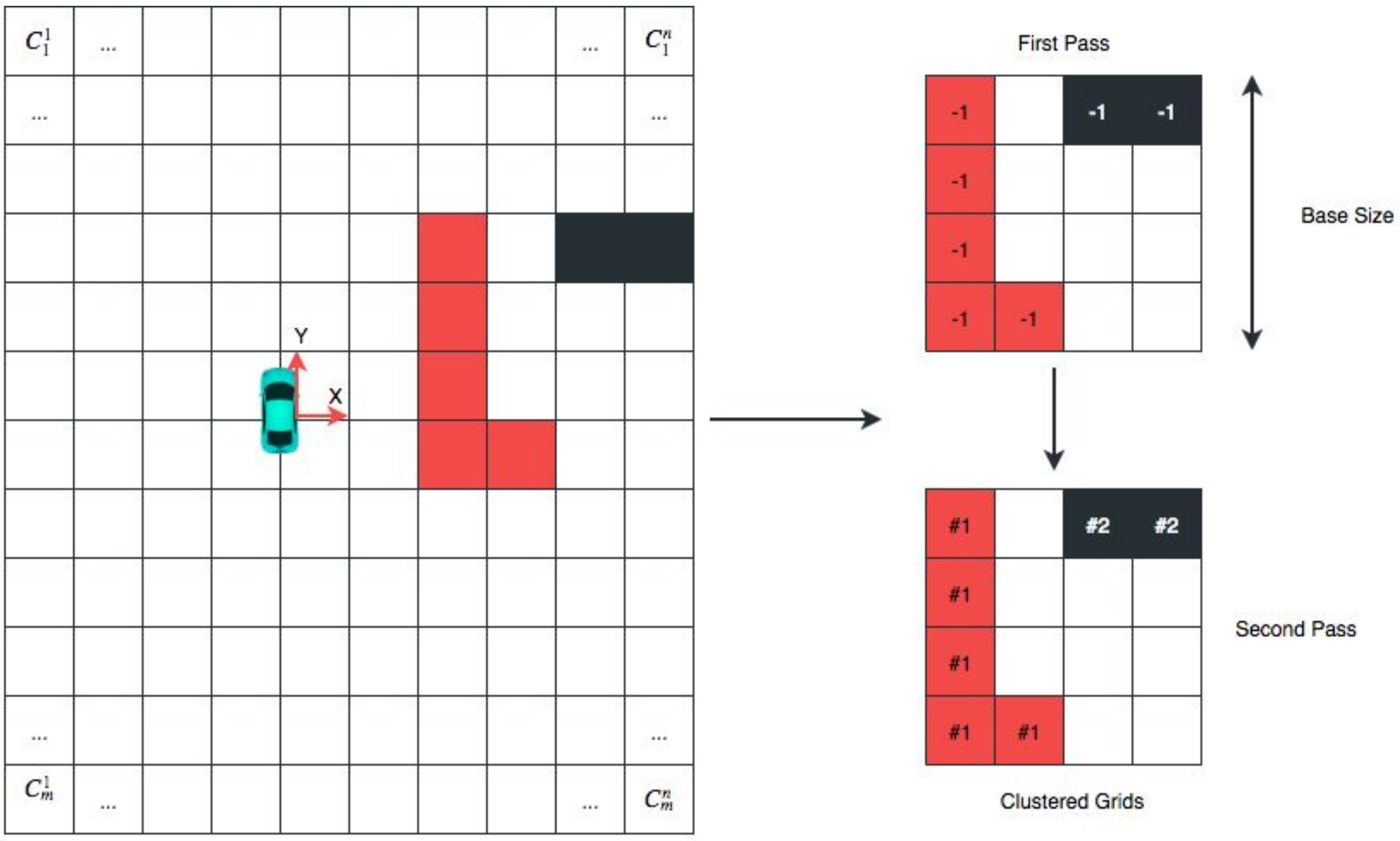

**Figure 5.6:** The constructed occupancy grid and clustering process based on two–pas method. The left is a schematic of the gridded data. In the right, two different clusters are detected which are further labeled as cluster # 1 and cluster # 2.

According to figure 5.6, the $x - y$ plane of mapped elevated points is divided into $m \times n$ different cells. Two binary values are assigned to each cell regarding their true situation: the value $0$ is assigned to the empty cells and the value $-1$ is assigned to the occupied ones. A single cell is is picked at the center of the grid map, and the cluster ID variable initiates for that cell. Then, all of the adjacent cells to that central cell, are tested for their occupancy status and if occupied, the same cluster ID will be assigned to them. If an empty cell happens to be adjacent to a cluster, the value $0$ will be assigned and the next cluster ID starts by an increment to the last one. The algorithm for clustering procedure, is presented in Algorithm 8 which is consisting of three different sub-functions. The results of the implementation of the method on the LiDAR data is further depicted in figure 5.7.

As it is seen in the Figure 5.7, different parts of the main point cloud are shown in different colors. Every bunch of the points with the same color, are the points which belong to a certain cluster. Heuristic rules are used to tune the classifier. This clustering method is suitable for the purpose of the method in this thesis because except for the occluded objects, no other potent movable object is expected to be disconnected in the scene, Therefore all the connected components are classified and are fed into next sub-modules for further processing steps.

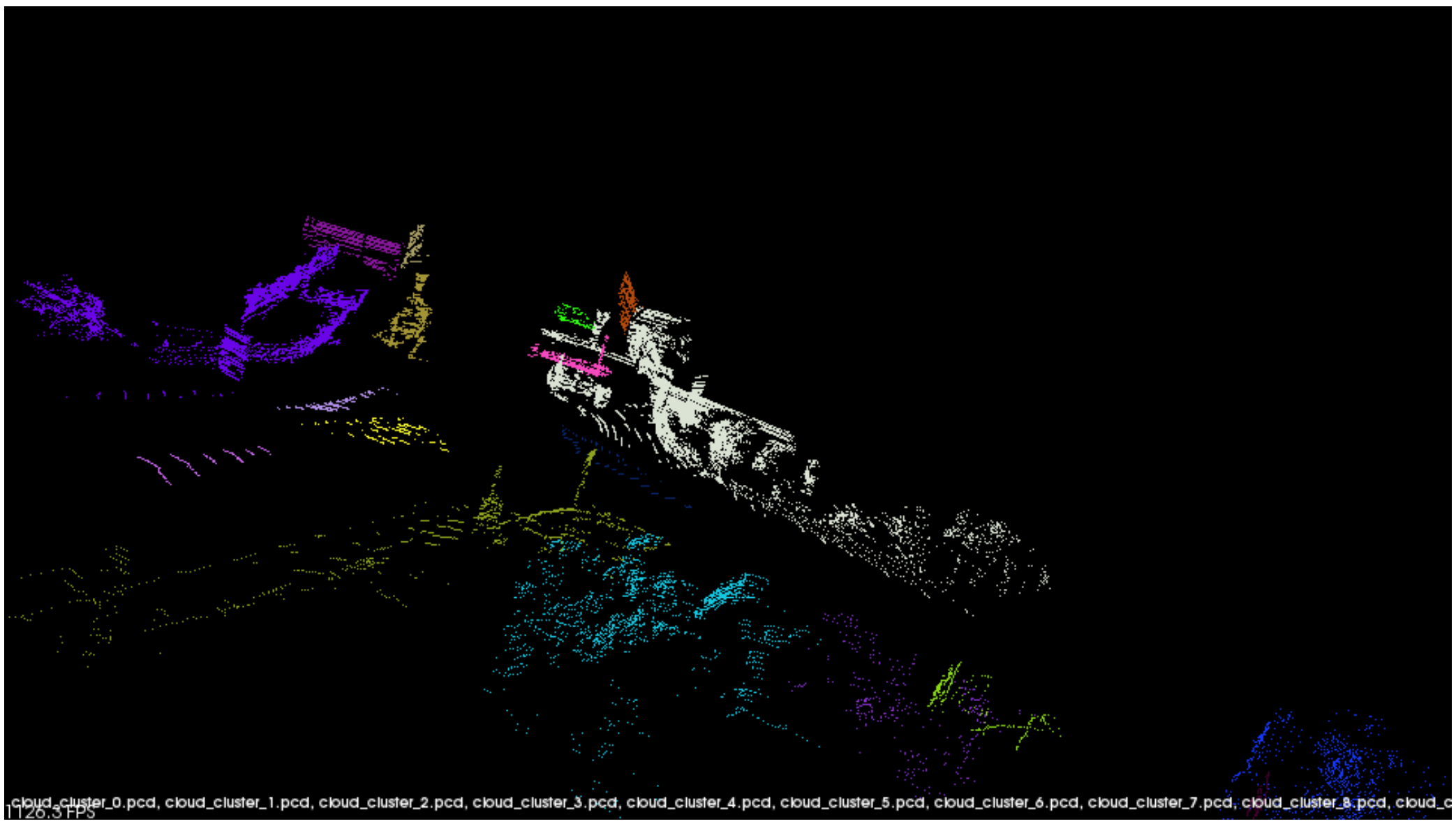

**Figure 5.7:** The result of clustering for the same frame in figure 5.5 after ground removal process is depicted. Different shad of colors represent different clusters extracted from the data.

### 5.2.3 Bounding Box Fitting

The output of the clustering sub-module is a bunch of different clusters of point clouds. Until now, there are no useful information about this different clusters other than knowing what data points construct a certain object. The bounding box fitting sub-module fits a box to each cluster to further produce in-hand information about each object. These informations in our case are dimensions of the box and the orientation of the detected object. Often 2D boxes are fitted to the two-dimensional mapped point cloud data. In this thesis, 3D boxes are fitted to each different cluster with three different values for height, length and width. Then the orientation is estimated using rotating calipers method.

Computing of the Minimal Volume Oriented Bounding Box (MVOBB) for a given three-dimensional point cloud is a hard problem in computer science. There are some exact algorithm to solve this problem with cubic order of the number of data points. In reference [116], a method is represented for efficiently approximation of the MVOBB for three-dimensional data. Which is much faster than the conventional algorithms. On the other hand, computing the minimum volume bounding box ensures the fact that issues regarding accident and obstacle avoidances are discussed more precisely, simply because more precise estimation of the dimension of the objects are in reach. Furthermore, using a direct method for fitting bounding box to the three-dimensional data, ensures our method to not to lose any height information. The volume of each computed bounding box is also calculated to further be used in rule-based filtering procedure to deleted unrelated objects due to their size.

For orientation estimation, the method proposed by [117] is used to update the orientation for each cluster. The method is called L-shaped fitting. The L-shaped method is efficient in driving scenarios for vehicles because the contours of vehicles in LiDAR data is often represented by a L-shaped object. This feature may be used to estimate the orientation of the vehicle. In [118] a method is suggested to solbe the L-shape extraction problem. The closest point to the sensor is chosen a the corner point for the L-shape. Then, some lines are used to connect this point to the borders of the object. Reference [119] truly mentions that in this method, side differentiation for the L-shape is highly sensitive to the measurement uncertainties. Furthermore, situation may happen that the vehicle has round corners in which the wrong corner point is selected by this method. Other methods may utilize optimization-based iterations to find the definitive L-shape [120]. Furthermore, in reference [121] error minimization is implemented for L-shaped fitting. A combination of error minimization and iteration-based re-estimation is used in [119]. In this thesis, the so-called *Search-based Rectangle Fitting (SBRF)* is used to extract L-shapes.

**Search-based Rectangle Fitting** In reference [120] the L-shaped fitting problem is formulated as an optimization problem. The SBRF algorithm starts by fitting a rectangle to four separate lines which encompass all the points in a 2D cluster of points as depicted in figure 5.8. At the next step, the rectangle is rotated. The number of rotations are determined by $\delta$. Then, for each rotation the fitting process is repeated to include all of the points. Three main criteria are assumed for a solution to be an optimal one: minimum area criteria, point-to-edge closeness maximization and the squared error of the point-to-edge minimization. Based on these three criteria, a weighted score is calculated for each rotation. The angular rotation with the best score is chosen to be the optimal solution for the L-shaped fitting problem. It is mentioned in [120] the closeness and squared error minimization give best performance for the vehicle pose estimation in driving scenario [117].

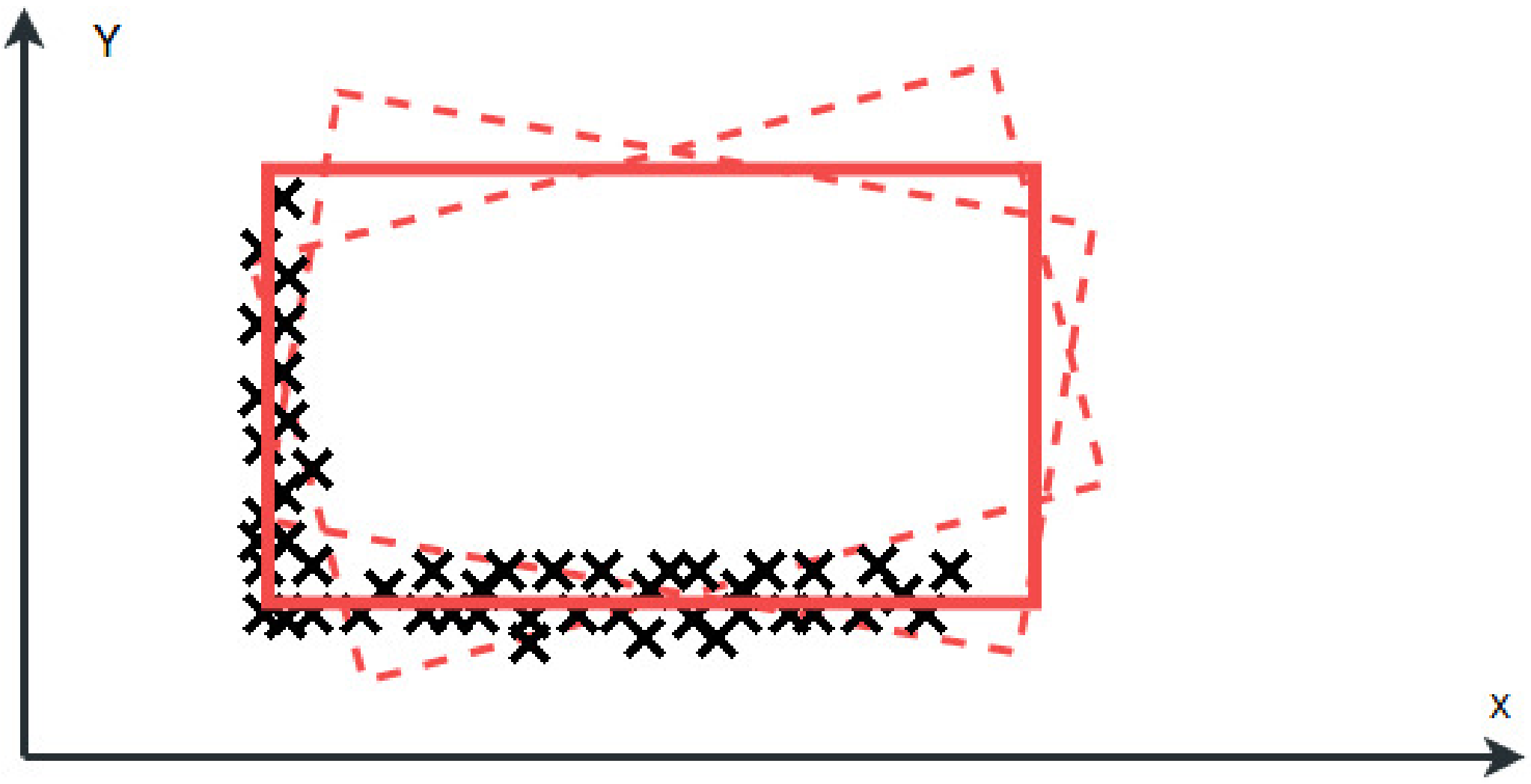

**Figure 5.8:** The fitted rectangle forms a bounding box for all the points in a cluster and the rotation of said rectangle around the points, from $\theta = 0$ with an increment $\delta$. The rectangle is re-calculated in each pose $\theta$ to always include all the points.

**Convex Hull** A convex hull for a set of points is the smallest convex region, containing all the points. An example of a convex hull is depicted in the Figure 5.9. The convexity is due the fact that for every two point in the set, the line segment between them will be part of the convex hull. An incremental algorithm based on the methods presented in [122, 123, 124] is implemented to find the convex hull in this thesis. The algorithm first, sorts the regarding set of point in ascending order based on their x coordinate. Then the algorithm iterates through these points while checking if the points should be added to the hull. First, two separate hulls are constructed which later will be merged to create the full convex hull: The upper and the lower hull. In the upper hull, points are added if the last point and the next point create a clockwise turn, and in the lower hull points are added if the last point and the next point create a counter-clockwise turn [117].

Therefore, the bounding box fitting starts with fitting simple boxes to all of the clustered objects in the scene based on the difference between their minimum and maximum point, as an example of the results can be seen in the Figure 5.10. As it is seen in the figure, Different segments from the clustering sub-module are imported to the rule-based filter sub-module. Then, the heuristic rules are applied to each of this segmented clusters. Inordinate clusters, like very big ones or very long ones are deleted from the list of clusters due to not suitable for being assigned as a potent object. Furthermore, a volume check is performed on the clusters to delete inordinate ones. This is actually implementable as the average

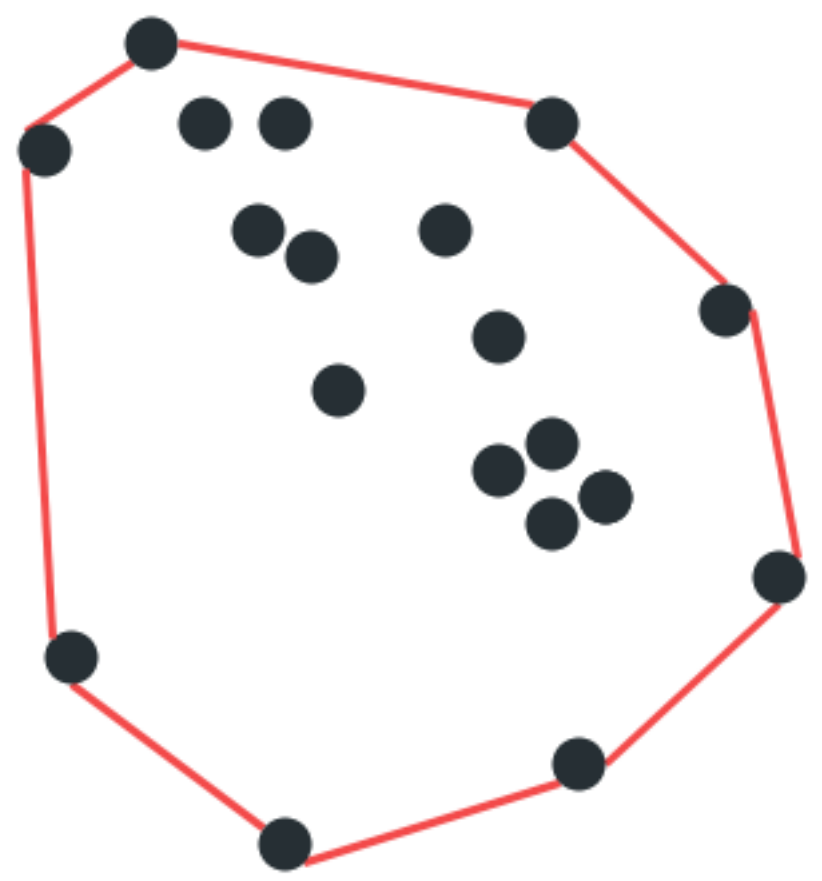

**Figure 5.9:** An example of a convex hull for a set of points.

size of vehicles are known as standard values.

Then, based on their volume some of the out-lier boxes will be deleted. The simple box fitting algorithm is used, to be more time-efficient. The problem with simple box fitting algorithm is that the fitted box is not oriented with the actual heading angle of the object. Therefore, the minimum volume box fitting is implemented on the remaining objects. Then, all of the points in each remaining clusters are mapped to the xy-plane. Then, regarding convex hull for each cluster of points are calculated and based on this convex hull, a bounding box is fitted to the cluster. Then, the L-shape fitting process is used to correct the orientation of the objects as depicted in the Figure 5.11.

It is shown in the figure, how the ordinary bounding box fitting method is successful to provide a size-wise rational bounding box for the object, while it fails to include the right orientation. The blues rectangle represents the bounding box induced to the cluster by the ordinary method. The white rectangle is the rectangle which is introduced to the cluster by L-shaped-based orientation estimation method. The white box is shown to be precisely aligned with the car.

### 5.2.4 Rule-based Filter

Until now, there are multiple boxes fitted to multiple target candidates. These candidates are necessarily contaminated by some non-interest objects like walls, bushes, trees and etc; as depicted in figure 5.12. This majority of non-interest objects are deleted from the list by using a dimensional thresholding method. Five different measures are used to eliminate these objects: width, length, height, the ratio between the length and width and the volume of the assigned bounding box.

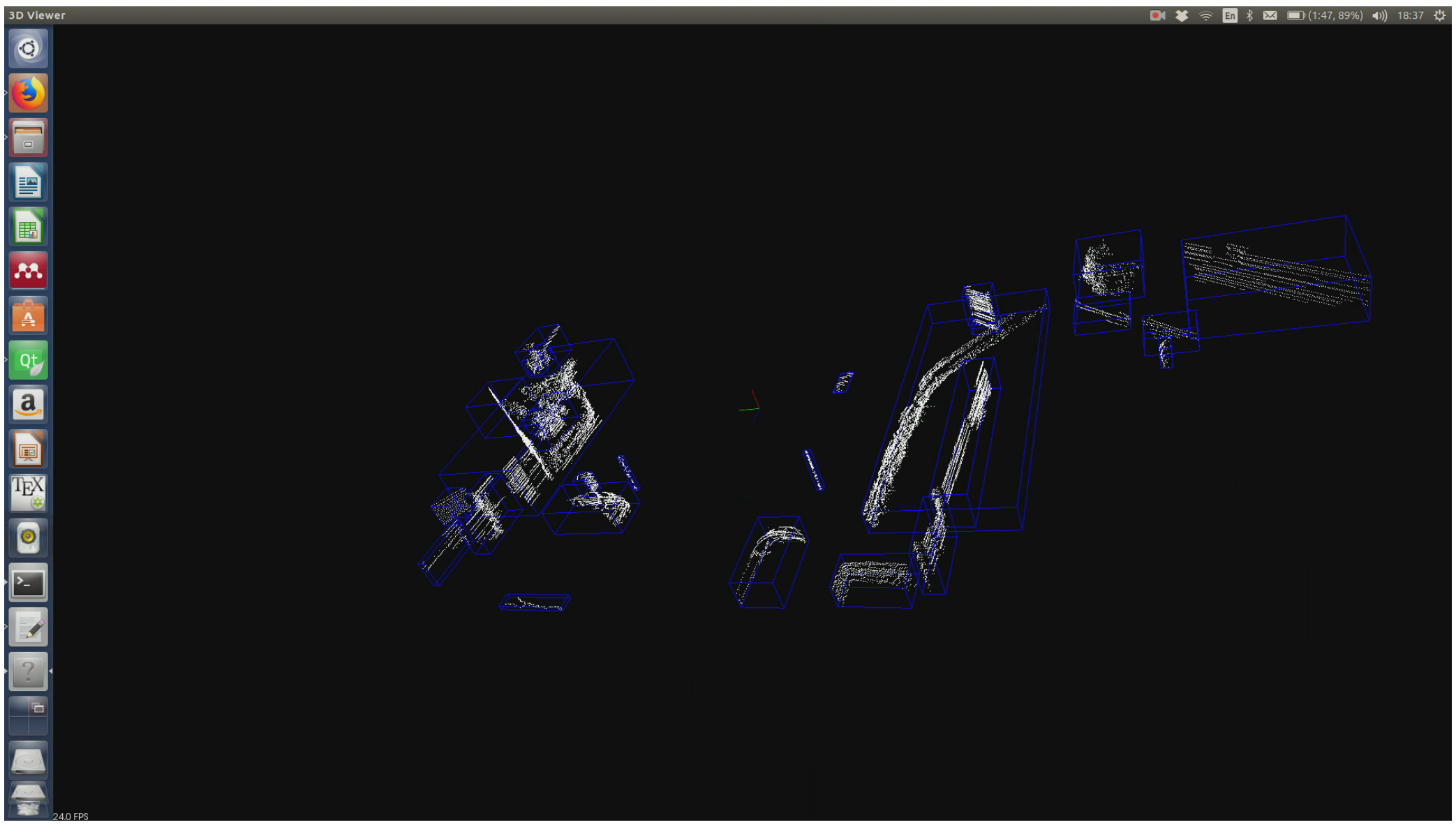

**Figure 5.10:** Simple box fitting results of the same data frame. The boxes are fitted due to the value of the minimum and maximum point of each cluster. The volume of each box is computed and based on this volume, non-relevant objects are deleted from the list of the objects of that frame.

Furthermore, as some references suggests, the number of LiDAR points and the point density (i.e. point per $m^3$) are consider as a thresholding measure. In order to prevent false eliminations, conservative values are chosen for these parameters. Therefore, we do not expect our algorithm to filter objects with similar dimensional profiles such as a bush and a car, or a pole and a pedestrian. The ultimate aim of the rule-based filter is not to prevent all false positives to happen, but to reduce the number of non-interest objects that are being passed to the tracker, in each frame. In addition, occluded objects are not to have a full bounding box. Therefore, the ration check will fail for over-segmented bounding boxes. This, will make the ratio check only applicable for larger objects such a very long, thin wall to distinguish them from fragments of real, over-segmented objects. The result of applying the rule-based filter can be seen in the Figure 5.12 while the list of all thresholding parameters are listed in the Table 5.2.

### 5.2.5 Over-segmentation Handling

In the detection module, it is detailed that the connected component clustering algorithm, is highly dependent on the quality of data. Over-segmentation may happen if there would be no well-defined space between cluster points. Even

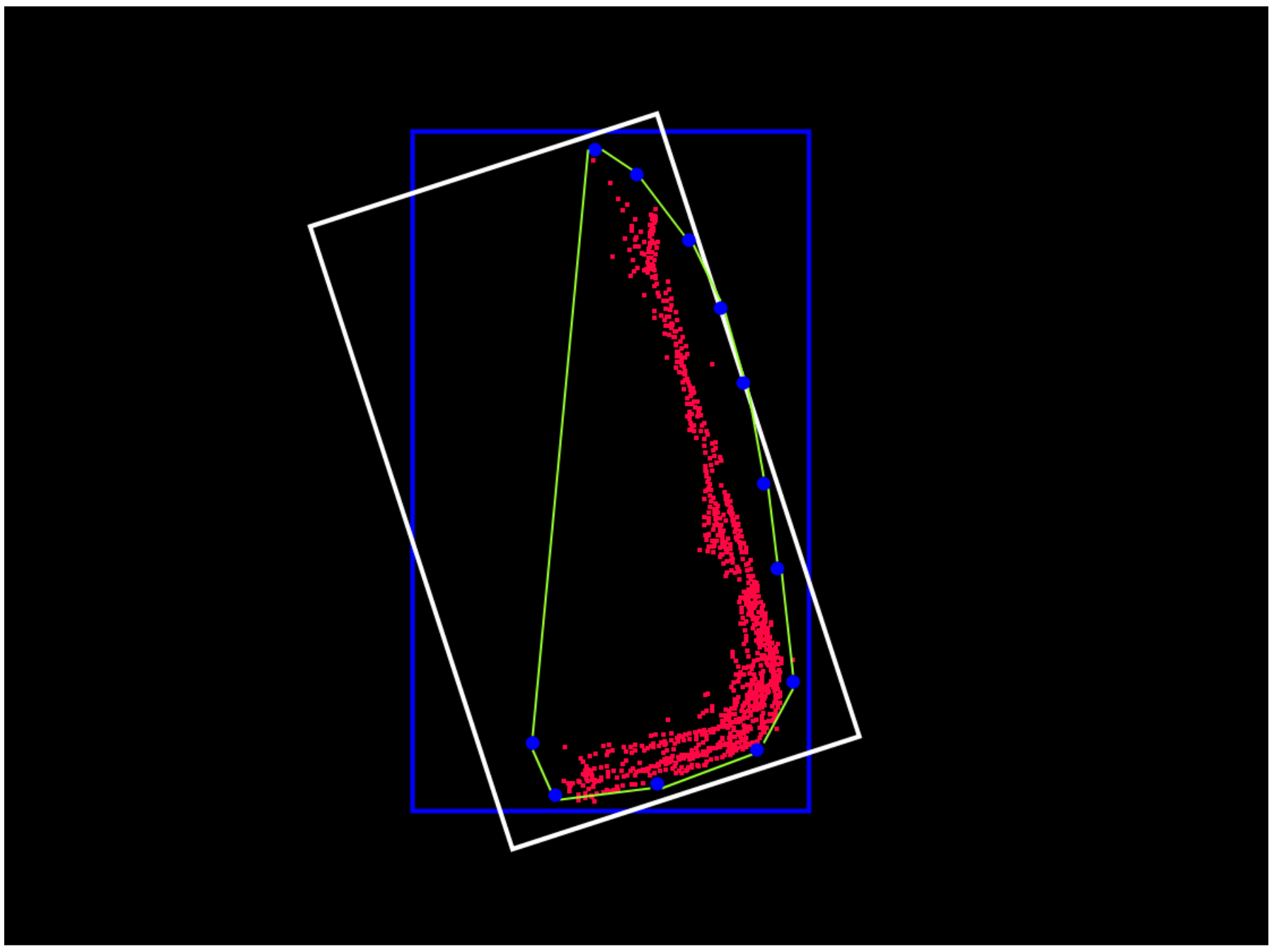

**Figure 5.11:** Fitted bounding boxes. The blue box is the simple box fitted to the cloud with the use of minimum volume method which is unable to align itself with the real orientation of the car. The white box is the bounding box fitted with the help of convex hull, which is aligned with the true orientation of the car.

for a well-structured cluster data, an abnormal space between data points may be created due to the occlusion caused by other objects. In this case, the occlusion means that an external object is blocking the line-sight of the LiDAR sensor like the tree-like object which is blocking LiDAR rays in the Figure 4.2. It is obvious that, in this case the occluded area becomes a blank space which causes the over-segmentation. A good way to overcome this issue, is to get help from the tracker to assist the detector. The over-segmentation may be predicted before it happens by using object heading direction and the last-best-known bounding box dimensions which is discussed in details in Section 4.3. Another similar approach, is proposed by [26] which uses the area percentage threshold to avoid repeating the clustering steps. However, we tend to dynamically change the connected component clustering parameter in order to tackle this problem. The over-segmentation handling procedure can be summarized by the following:

According to the Section 4.3, the full dimensioned bounding box in the time step $k - \zeta$ is already stored. Furthermore, the tracker module provide the relative velocity information for all tracked objects in any time step. For instance, in time step $k$ the detector uses the predicted position prior which is coming from the tracker, which has been happened the bounding box fitting. Thus, if the predicted box encloses or overlaps significantly with newly generated clusters, then such clusters are to be merged. The main drawback of such a procedure is that it may further induce the under-segmentation. For example, two correctly clustered objects placed in each others proximity, may be merged into a bigger cluster. To address this, the dimension of potential merged cluster is made sure to not exceed that of predicted box before the merging is finalized. The over-segmentation handling procedure is depicted in Figure 5.2.

**Table 5.2:** List of Thresholds for Rule-based Filter

| Number | Description | Variable |
|---|---|---|
| 1 | $T_l^{max,min}$ | Min and Max value of the object's length |
| 2 | $T_h^{max,min}$ | Min and Max value of the object's height |
| 3 | $T_w^{max,min}$ | Min and Max value of the object's width |
| 4 | $T_v^{max,min}$ | Min and Max value of the object's volume |
| 5 | $T_a^{max,min}$ | Min and Max value of the object's top-view are |
| 6 | $T_r^{max,min}$ | Min and Max value of the object's ratio of length to width |
| 7 | $T_d^{min}$ | Min density value for each assigned bounding box |
| 8 | $T_{r,check}^{min}$ | Min length value of the object for ratio check of parameter No. 6 |

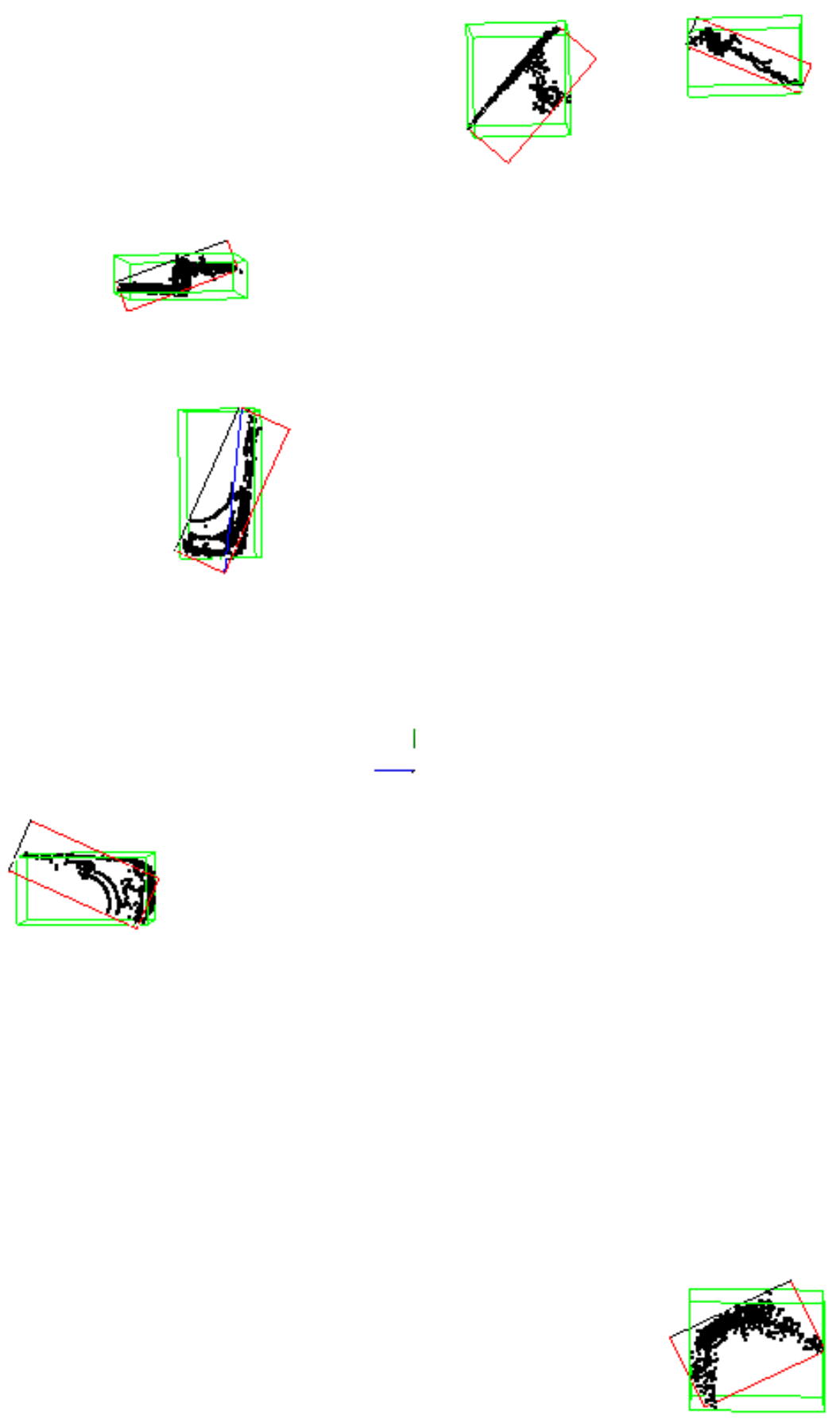

**Figure 5.12:** Rule-based filter result is depicted in the figure. It is shown that along with the two target objects (cars) a bunch of other dimensional-wise similar objects are passed to to the filter. In the bottom picture, the mapped points to the x-y plane can be seen. On the other hand, in the above pictures, corrected orientation estimation may be seen.

# Chapter 6

# Conclusions and Future Work

This chapter elaborates the contributions and integration-based solutions resulting from this work. First, contributions and the solution is summarized. Then the initial research objectives of this thesis will be discussed on how much the work was successful to fulfill them. The potential improvements of the research in the future will be elaborated last.

## 6.1 Summary of Results

Autonomous driving is an emerging technology at the advent of the 2020. Until now, many steps toward establishing a fully autonomous driving vehicle has been taking by researchers all around the world. These steps was well-persuasive for the investors, high-tech corporations and giant corporations to invest huge amount money on this technology. All around the world, legal and insurance institutes have taken huge steps toward establishing infrastructures for the advent of this technology to be more easily performed. Full autonomous driving demands a very specific and precise object tracking solution to be awarded to perform freely in the human-encountered areas.

Furthermore, object tracking in urban environments is a more complex task, because larger uncertainties are included in the scene. On the other hand, object tracking in urban environments demands surround-awareness which is not real-time applicable in camera-based solutions with the state-of-the-art solutions. However, LiDAR-based methods prone to be successful for completing surround-awareness task duo to their method of data gathering, but only if to be successful to handle data association problem and present a solution for clutter- and occlusion-related problems.

On the other hand, the more important the tracking might seems for the driving task, the more detection gains importance. The success of tracking is highly dependent on the detection task. If the detector module fails to prepare a reliable list of objects for the tracker, the tracker will be further failed to perform its task. Furthermore, the reliable list of objects, from each data frame is important for the data association-related tasks. All of these, gets more substantial when the DaTMO task is to be performed in urban environments. Either, no specifically-tailored detection module is proposed in literature for the urban environment driving automation.

The method, proposes a system architecture for the DaTMO task based on current works of different researchers in the field. The system architecture is set to be regarding to tracking-by-detection framework and due to the top-down approach. The system is designed to be a part of further design of a full autonomous car, therefore the module-based-view is maintained in all of the sections of the thesis. Furthermore, for the detection module a novel ground estimation method is proposed which is specifically designed for the urban environments and the regarding issues. Furthermore, different clutter- and occlusion-related problems are fully addressed and recoded to gain further insights about the system.

## 6.2 Research Objectives

1. ***Identifying the requirements of a robust DaTMO procedure against uncertainties and limitations in urban environments.*** These requirements are discussed in details in the introduction, fundamentals and implementation sections. In summary these requirements are as follows:

   (a) The detector must be capable of removing unnecessary data points from the point cloud being outliers or the ground points.

   (b) A continuous ground estimation method must be employed to be be capable of estimating both flat and sloped terrains due to unwanted rough areas that may be found in urban scenarios.

   (c) The detector module must be capable of detecting all of the potent movable objects in the presence of occlusions due to range or the view-point missing data.

   (d) The tracker must be capable of tracking different movable objects with different maneuvering behaviors, at the same time.

   (e) The tracker must be capable of perform the tracking in the presence of the clutter.

   (f) The proposed architecture must enable the whole process of detection and tracking to be performed all together and in real-time. The sensor sampling rate (10 Hz) is defined as the real-time deadline.

2. ***A probabilistic framework containing system design and algorithms is proposed for both modules: Detection and Tracking.*** Methods must be introduced in both frameworks to handle the clutter and occlusion for the method to be considered as a surround-awareness solution.

   The detector module, utilizes a novel ground segmentation method which is very important in two aspects: First, the method is designed to take local characteristics of the ground into account while giving a whole-frame continuous-location estimation of the ground surface. Therefore, there is no planar approximation for the ground surface that might neglect sloped or rough areas. Second, the whole-frame, continuous-location method for ground estimation enables the method to even remove the occluded ground parts of the data, efficiently. Furthermore, the orientation estimation and clustering method are designed to not neglect the miss-orientation of occluded objects. The accuracy of detection module is further enhanced by using logic- and rule-based filters. The detector module is fully implemented using C++ and ROS. Used algorithms are presented in the corresponding section and the corresponding results are depicted.

   While the presence of uncertainties are often neglected in different casts of methods toward detection process for ALV's, in this thesis this presence is taken into account by using machine-learning-based methods such as Gaussian regressions. Furthermore, the presence of uncertainties in the tracking problem is addressed using optimal Bayesian filters. States of the target tracks are assumed to be random variables of evolving stochastic processes. The data association problem is tackled using the joint probabilistic data association filter. The unknown dynamics of potent vehicles in the surrounding environment are treated as they are different kind of motions with uncertainties.

3. ***Real-time implementation and calibration of the detector module*** The detector module is implemented on a public available data set (KITTI) data set. Results are shown to outperform the same ground estimation methods by the means of introduced error criteria. While the method remains in real-time criteria. The result enjoys the benefit of being adjustable in the matter of speed and precision, by choosing introduced adjustable variables.

## 6.3 Future Works

The whole so-called perception unit in this thesis, is a research in motion. Several improvements may be introduced to the work. On the other hand, by the improvement in the computational power of available hardwares, new methods may be proposed to tackle limitations in the sort of works. Here, we tend to introduce two classes of such improvements: first, improvements and remaining works in order to establish a full software implementation of DaTMO and second, some more general research questions.

Beside implementations represented in this thesis, further implementations and improvements to the current ones may be accomplished as follows:

**Ground Segmentation with Input-dependent Noise:** Often, for the detection module no uncertainty rejection method is introduced in the literature. This is due to the fact that driving tasks until now, were more concentrated on tracking problems. By the advancements of the method, now several improvements may be made into detection modules by considering uncertainty in different detection sub-modules. In this thesis, the presence of uncertainty is considered in ground segmentation process by using a machine-learning based method to include non-smoothness of the data.

Another improvements may be introduced into this part of work by assuming an input-dependent noise for the ground estimation process. The notion of introducing of the input-dependent noise to the Gaussian regression process has been remained untouched in the literature. The theoretical aspects of this work is done and is represented in Appendix E with gradient evaluations available in Appendix C. Implementation of the method remains for the future works.

**Ground Segmentation Using Non-reduced 3D LiDAR Data:** All theoretical steps of the ground segmentation procedure, are developed based on multi-dimensional input/output for the Gaussian process regression. Theses is done due to the potential better results of carrying out the estimation based on Cartesian $x - y$ values as input to the estimation process and the height $z$ values as the output, while in current version data is projected into radial coordinate and the input is set to be the value of radial coordinate $r$. This will enable the method to escape the loss of data and further a one-to-one relation between input and output will be held.

**Three-dimensional physically-motivated Gaussian processes based ground segmentation with the use of sparse Gaussian process for pseudo-input set selection** Optimization task is the most time-consuming part of our proposed ground segmentation algorithm. In the Gaussian process regression literature, methods are proposed for optimal regression size reduction for large scale data. sparse Gaussian process methodology might be used in order to increase computational efficiency of our proposed method, especially in order to choose an optimal pseudo-input set.

**More Primitive Shape Detection:** Currently the detector sub-module only considers the L-shaped feature as carrying useful information for bounding box detection. Other shape detection methods are proposed in the literature [47] including U-, V-, and I-shaped features which will enable better detection results.

**False Detection Improvement:** A Joint Integrated Probabilistic Data Association Filter is proposed in [129] which enables the method to better handle false detections.

For more general aspects of the future works and aspects regarding designed framework, following improvements may be necessary:

**State and Parameter Estimation , at The Same Time** Included Optimal Bayesian Filters are tuned by hand in current work. The parameter estimation may be added to construct a coupled estimation scheme toward more precise results in tracking module design.

**Using More Broad Motion Models:** In the current tracking module design, motion models are not complete. Other useful motion models such as Constant Steering Angle and Velocity (CSAV) motion model,Constant Speed Changing Rate and Constant Turn Rate (CSCRCTR) and Constant Curvature and Acceleration (CCA) motion models are not included in the algorithm [130]. A particle filter based method for integrating different motion models is represented in reference [94] which is notable to review for this subject.

**A Novel, 3D Tracking Scheme** Initial goal for this thesis research plan was to establish a probabilistic framework for 3D dynamic object recognition based on a LiDAR-only scheme. Motion models utilized for this purpose have to be capable of handling three-dimensional motion issues since, three-dimensional pose estimation problem requires special mathematical treatments. Common pose estimation approaches in the literature are based on two common assumptions: smoothness of the measurement process and membership of the state-space to an commutative geometry space, neither of them are true for a three-dimensional LiDAR-based MODaT problem because first, LiDAR measurements are often non-smooth since they are coming from a moving sensor and second, 3D motion variables are smooth manifolds which may be parameterized by rotation or transformation matrices which are member of an Euclidean group.

Furthermore, different kind of three-dimensional motion models suffer from singularities. Since three-dimensional poses do not belong to an commutative space, additive uncertainty does not seem to be able to give an accurate answer.

Thus, uncertainties had to be handled differently since they act differently in three-dimensional environment. Among other problems concerning 3D object recognition in 3D environment, handling these singularities and uncertainties seems most untouched in the literature.

Grisetti [131] notes that, in the estimation of a stationary nonlinear system in optimization problems, using rotation matrices as parameters might result non-valid solutions. Furthermore, using other minimal representations like Euler angles suffers form singularity issues. Thus, a new mathematical basis is required to handle optimization problems in non-commutative spaces.

Furthermore, in reference [132] a novel theoretical method is represented for associating uncertainty with three-dimensional pose estimation. This method relies on transformation matrices for representing 3D pose variables. A simple and practical technique for uncertainty association with $4 \times 4$ transformation matrices are utilized to represent pose variables. This representation is challenging because in fact rotation variables are not *vectors* but rather member of a *non-commutative group*.

Currently, in pose based dynamic object's detection and tracking methods, estimation of state variables like *sensor's pose* ($\mathbf{X}_s$) and states of the dynamic tracks are considered to be of great importance. Through different kind of state estimation methods or Bayesian inference methodologies, the process of detection and estimation of future state variables are being formulated using discrete-time motion models.

These methodologies often use the discretized time steps to estimate trajectory of dynamic objects at those exact times. Although these methodologies are sufficient in many applications, they are not showing high-quality results especially when we are working with range-finder sensors. According to [133], a simple application of discrete-time approach would include a state at every measurement time, making estimation prohibitively expensive. Furthermore, discrete-time approaches may fail to find a unique solution while implemented on *scanning-while-moving* sensors. As poses are a part of *non-commutative group* we can not apply Bayesian filtering methods directly on them thus we must consider another methods for associating noise with pose variables instead of additive Gaussian noise.

**Project Goal:** ***Establish a MODaT procedure based on a real-time, smooth, continuous-time trajectory estimation algorithm for multi-object detection and tracking in 3D environments. A Physically-motivated Gaussian prior will be assigned to the each pose variable. As we tend to find the results in SE(3), linear time varying SDEs can not be utilized. Instead, we must assign our GP prior with regards to uncertainty injection methods introduced in [87] and associate uncertainty with transformation matrices via exponential map.***

In fact, instead of seeing necessary state variables as simple state variables and try to use Bayesian or common filtering methods, we tend to see these important aspects of our dynamical models as 2D or 3D poses, which are part of *Euclidean groups SE(3) or SE(2)*. Then we try to establish a mathematical framework to confront the problem of multi-object detection and tracking in this scheme. An important goal in this research would be batch continuous-time multi-object trajectory estimation to estimate all traffic participant's trajectory along with the ego-vehicle's motion states in real time and continuously.

I would rather to see these *sensor poses* as an SE(2) or SE(3) random variable:

$$\mathbf{X}_s := \exp(\xi^{\wedge})\bar{\mathbf{X}}_s \tag{6.1}$$

Where $\bar{\mathbf{X}}_s \in SE(3)$ is a large noise-free value and $\xi \in \mathbb{R}^6$ is a small noisy perturbation. This equation is actually called an *injection of noise* onto the three-dimensional Euclidean group [87]. After considering these poses as random Euclidean group variables, we must try to find a mean to estimate these parameters for our dynamic object's tracking purpose. Furthermore, we can see the problem of tracking as a *trajectory estimation* problem, and in this way we can also see the trajectory as a Gaussian process with respect to time variable and by placing a Gaussian prior on our problem. We may tend to find unique solution for our contінous tracking problem. A same methodology is proposed for **STEAM** problem by [133], but they assume that the motion between different measurement times are small and used their method in manipulator applications. This assumption must be re-evaluated in order to be capable of handling motions in driver-less car scenario, or another assumption might be used instead.

# Chapter 7

# Farsi Summary

# خلاصه فارسی

## خودگردانی وسایل نقلیه به چه معناست؟

بعد از برگزاری مسابقات شهری و گرند دارپا، همچنین تلاش بی سابقه شرکت گوگل برای ساخت و توسعه اتوموبیل خودران، انتظارات برای معرفی هرچه سریعتر اتوموبیل های کاملاً خودکاری که بتوانند به طور کامل در محیطهای شهری و پیچیده تردد کنند، افزایش یافته است. مأموریت های شامل رانندگی خودگردان شامل زیر مأموریتهای بسیار پیچیده ای می باشند که هر کدام از این زیر مأموریتها توسط زیر سیستم های مخصوص بخود انجام می شوند تا در نهایت و در ترکیب با هم، خودرو را قادر سازند تا محیط را به طوری هوشمندانه شناسایی کند و سپس به طور خودران تصمیم های کنترلی را جاری بسازد. با توجه به همین پیشرفتهای اخیر، مفهوم اتوموبیل های خودران و خودروهای بدون راننده، به مفهومی کاملاً ملموس برای مردم و بحث برانگیز برای خبرگان تبدیل گشته است. رسانه ها همین الان هم، نوید معرفی زود هنگام این محصولات را به بازار میدهند، از همین رو شرکتهای بزرگ خودروسازی، در حال رقابت برای توسعه فناوری های مرتبط با خودرانی و خودگردانی خودروها می باشند. رقابت در این حوزه به قدری شدید و جدی می باشد که حتی غولهای نرم افزاری در حال ورود به عرصه رقابت برای توسعه فناوریهای مرتبط میباشند.

از طرفی، مفهوم خودرانی خودروها ممکن است با توجه به کاربرد مورد نظر و شرایط محیطی متفاوت، به مفاهیم متعددی اتلاق شود. در حقیقت، با توجه به موقعیت های مختلف، درجات مختلفی از خودرانی ممکن است تعریف شود. به عنوان مثال، کیفیت و نحوه خودرانی خودرویی که در یک بزرگراه در حال حرکت میباشد با خودرو متحرک در محیط شهری بسیار متفاوت خواهد بود. معمولاً طبق استانداردهای مختلف، سه بازیگر مهم مسؤل فرآیند رانندگی در نظر گرفته میشوند: راننده انسان، سیستم خودگردان خودرو و دیگر اجزا و سیستم های موجود در خودرو. یک خودرو، به عنوان مثال میتواند توسط سیستم خودگردانی تجهیز شده باشد که قادر به ارایه ی ویژگی های متعدد در راستای خودگردانی خودرو و در لایه های مختلف باشد. ویژگی های درگیر شده در لایه، در هر مأموریت، نوع خودگردانی خودرو و لایه عملکردی آن را مشخص می کند.

جامعه مهندسی خودرو The Society of Automotive Engineers (SAE) دسته بندی بسیار سودمندی برای صنایع خودرو سازی ارایه کرده است که به دنبال تعاریف مشخصی برای دسته بندی طراحی های خود از لحاظ خودگردانی می باشند. این تعاریف، طبقه بندی مشخصی را برای دسته بندی نحوه خودگردانی وظایف پویای رانندگی برای سیستم های خودگردانی خودروها ارایه میدهند. عملیات خودروهای خودگردان ممکن است در لایه بدون خودگردانی رانندگی ( یعنی لایه ۰ )، تا لایه خودگردانی کامل خودرو ( لایه ۵) قرار بگیرند. بر اساس استاندارد جامعه مهندسی خودرو، سطوح مختلف خودگردانی رانندگی بر اساس نقش مشخصی که هرکدام از سه بازیگر اساسی وظایف پویای رانندگی بازی میکنند، تخصیص داده میشود. سیستم های فعال ایمنی خودرو از جمله انواع مشخصی از سیستم های کمک راننده از دایره بررسی های مربوط به سیستم های خودگردان خارج شده اند چرا که در واقع آنها هیچگاه مسؤلیت رانندگی خودرو را به طور کامل عهده نخواهند گرفت و تنها مسؤلیت دارند تا در هنگام مواجهه با خطر در فرآیند رانندگی مداخله کنند. این مداخله سیستم های کمک راننده، هیچ گاه مسؤلیت رانندگی را از روی دوش راننده انسانی برای انجام بخش و یا همه فرآیند رانندگی بر نخواهد داشت و از همین رو این سیستم ها را نمیتوان به عنوان سیستم های

خودگردانی رانندگی در نظر گرفت. سیستم های پیشرفته کمک رانندگی Advanced Driving Assistant Systems ( ADAS ) همانند سیستم های تطبیقی کنترل گشت زنی Adaptive Cruise Control Systems و یا سیستم های کمک پارک،بر اساس استاندارد ارایه شده در سطح سوم قرار خواهند گرفت. در این سطح مداخله انسانی هنوز در حین انجام رانندگی الزامی میباشد. این سیستم ها به این منظور طراحی شده اند تا بهترین عملکرد خود را در شرایط کم ترافیک و یا محیط های ساخت یافته همانند بزرگراه را به نمایش بگذارند.
سیستم های کمک پیشرفته محدود دیگری وجود دارند که برای انجام فعالیت هایی در محیط هایی با دینامیک بالا طراحی شده اند. از جمله می توان به سیستم های جلوگیری از برخورد با عابرین پیاده Pedestrian Collision Avoidance Systems(PCAS) و یا سیستم های کمک کننده در تقاطع ها ( Intersection Assistant Systems ( IAS اشاره کرد. درجه محدود خودگردانی سیستم های خودگردانی خودرو موجود، نشان دهنده این واقعیت می باشد که خودگردانی خودروها، بخصوص در محیط های پیچیده همانند محیط های شهری یک زمینه تحقیقاتی جاری میباشد. در این پایان نامه تعریف ما از یک خودرو خودران زمینی و یا ( Autonomous Land Vehicle ( ALV با توجه به تعریف سطح پنجم استاندارد جامعه مهندسی خودرو میباشد. در این سطح، خودرو نه تنها مسؤل مشاهده و شناخت محیط میباشد، بلکه مسؤل تصمیم گیری، تصمیم سازی و اجرای این تصمیمات نیز میباشد. در آخر همانند همه محققان فعال در این زمینه، هدف گیری ما یک هدف مشترک مشخص خواهد بود: حذف کردن کامل نقش انسان در اجرای یک وظیفه رانندگی.

**طراحی سیستم وسیله نقلیه خودگردان:**

رانندگی خودگردان کاری پیچیده است. بنابراین یک طراحی سیستمی پیچیده برای مقابله با همه ابهامات موجود در یک عملیات رانندگی خودگردان باید صورت گیرد. برای دست یافتن به درک بهتری از پیچیدگیهای طراحی سیستم خودروهای خودگردان، نیازمندیهایی عملیاتی آنها عامل مهمی محسوب می شوند. نیازمندیهای عملیاتی یک خودرو خودگردان شامل موارد زیر می شوند: فناوریهای ارتباطاتی مرتبط، مکانیابی مطلق و فراگیر، درک محیط و وضعیت خود خودرو، اجرایی کردن عملیات و اضافه کردن نقش انسانی چه به عنوان راننده، مسافر و یا به عنوان بازیگرهای خارجی موجود در ترافیک عمومی شهری و جاده ای مثلاً به عنوان عابر پیاده.
فرض می کنیم که هر خودرو خودران زمینی قادر است تا بر روی زمین زیر پای خود به طور آزادانه حرکت کند چرا که محدودیتی برای حرکت بر روی ریل، میله های اتوبوس و یا خط هی برقی ندارند. همچنین فرض میکنیم همه خودروهای خودران زمینی در حال حرکت در ترافیک عمومی جاده ای میباشند. طراحی سیستمی یک خودرو خودران باید به گونه ای باشد که عملیاتی کردن آن توسط انسان در ساده ترین سطح شهودی قرار بگیرد. بنابراین خودرو باید تنها توسط یک دستور عملیات ورودی مورد خطاب قرار بگیرد که عموماً به صورت یک وظیفه حمل و نقلی می باشد. ورودی عملیات باید قابل تطبیق با نیازهای اُپراتور ها باشدُ مخصوصاً اگر خودرو در حال جا به جا کردن انسان ها باشد، بع عنوان مثال خودرو باید برای ایستادن های اضطراری و یا ایستادن در رستوران بعدی آمادگی لازم را داشته باشد.
از طرفی دیگر از آنجایی که خودرو در حال اجرای عملیات در ترافیک عمومی می باشد، اجرایی کردن ایمن عملیات از اهمیت بسزایی برخوردار خواهد بود. از این رو در مورد طراحی سیستمی درک

عمومی، آگاهی، تصمیمات رانندگی و رفتار خودرو باید احتیاط و توجه بیشتری کرد. محیط های شهری در مقایسه با محیط های دیگر ممکن برای اجرای عملیات رانندگی، بسیار پویا می باشند. بنابراین خودرو های خودران باید قادر باشند تا در محیط های شهری اجسام و اجزای متحرک و ایستای موجود در محیط پیرامونی خود را به طور مقاوم شناسایی و دسته بندی کنند. حالت های مختلف رانندگی با حضور خودروهای خودگردان شده و دستی باید در نظر گرفته شود. از طرفی قوانین محلی ترافیکی باید به عنوان اولین شاخصه حداقلی، که کمیت و کیفیت محیط پیرامونی را در هر محل مشخص میکنند، در نظر گرفته بشوند. قوانین فراگیر و یا محلی در ارتباطی مؤثر با موقعیت مکانی و حرکتی وسیله نقلیه، رفتار خودرو را در موقعیت ها و محیط های بخصوص مشخص میکنند. بنابراین به عنوان یک نتیجه گیری ساده، خودرو خودران باید بتواند آگاهی خوبی از توانایی ها و مهارت های خود در هر لحظه داشته باشد و به نسبت موقعیتی که در آن قرار گرفته است، عمل کند.
تخمین زدن این مهارت های و توانایی ها از جمله نظارت بر عملکرد سخت افزار و نرم افزار، میتواند یکی دیگر از الزامات دلخواه در مورد خودرو های خودگردان باشد که به عنوان عیب یابی برخط شناخته میشود. از طرف دیگر خودرو باید در مقابل سوء استفاده و خرابکاری نیز مقاوم باشد. به طور خلاصه، خودروهای خوگردان باید بتوانند موارد زیر را مهار کنند:

- **عملیات**

  خودرو باید قادر باشد دستورات مورد نیاز خود را از یک عامل انسانی دریافت کند.
- **عملیات ورودی**

  خودرو خودگردان زمینی مورد نظر باید بتواند دستورات ورودی از عامل انسانی را تفسیر کند و سپس عملیات درخواست شده را اجرایی کند.
- **ارتباطات و داده های نقشه**

  وسیله نقلیه باید نسبت به موقعیت فراگیر و محلی خود واقف باشد تا بتواند در محیط به ناوبری بپردازد. از طرفی دیگر، مکانیابی فراگیر برای ارتباطات بین دو وسیله نقلیه و ارتباطات بین وسیله نقلیه و زیرساخت های ترافیکی و حساس لازم می باشد.
- **آگاهی محیطی ( ادراک )**

  وسیله نقلیه باید بتواند محیط پیرامونی خود را درک کند، بخصوص درک درست از اجسام پویا و ایستای محیط پیرامونی برای آگاهی از مقاصد آینده آنها برای گرفتن تصمیمات احتمالی برای مقابله، ضروری به نظر میرسد. از طرفی دیگر، آگاهی محیطی خودرو خودگردان را مطمین میسازد که آغاز کننده هیچ خطری برای محیط پیرامون خود نیست.
- **آگاهی از خود:**

  خودرو خودگردان باید حالت کنونی خود را در هر لحظه بداند. این حالت میتواند شامل توانایی های عملیاتی در لحظه، اجزا تشکیل دهنده و نشانه های حرکتی باشد.

اگر نشانه های رفتاری رانندگی انسانی را پایه مدلسازی عملیات رانندگی در نظر بگیریم، با توجه به مطالعات رفتاری انجام شده در این زمینه مدلی سه حالته برای توصیف رفتار انسان در حالت رانندگی استفاده میشود. بر اساس این مدل، عملیات های رانندگی را میتوان تشکیل شده از سه وظیفه متفاوت در نظر گرفت: ناوبری، هدایت و پایدار سازی.

**نحوه عملکرد سیستم:**
در مرحله اول، مأموریت در بالاترین سطح از معماری سیستم، که در آن برنامه ریزی مسیر و استراتژی ناوبری در حال انجام است، فرآوری خواهد شد. داده ورودی برای این مرحله باید شامل همه اطلاعات مورد نیاز در مورد شبکه های جاده ای مرتبط باشد. در سطح بعدی، اطلاعات پیرامونی خودرو خودگردان، همچنین اطلاعات صحنه محلی باید برای کنترل ایمن خودرو به حساب آورده شود. پس از این، در سطح پایدار سازی، مانور های انتخاب شده از ماژول هدایت جهت استفاده بعدی، فرآوری خواهند شد.

دو رهیافت عمده در زمینه معماری سیستم رانندگی خودروهای خودگردان منتشر شده است. بسیاری از روشهای متداول مرتبط با تیمهای شرکت کننده در چالش های دارپا، روش هایی بر اساس مکانیابی می باشند چرا که در چالش دارپا، نقشه ای دقیق از محیط در اختیار تیم ها قرار داشته است. در میان این روشها تنها یک تیم روشی ادراکی برای این مسأله ارایه داده بود. روش ادراکی برای به کارگیری در سناریوهای رانندگی در محیطهای شهری بسیار قابل اعتماد تر خواهند بود چرا که در اجرای عملیات های رانندگی در محیط های شهری، نقشه ای دقیق از محیط کم-ساخت یافته و شلوغ شهری در دسترس نیست. بنابراین ارایه کردن ساختاری کارآمد برای برای پردازش و تفسیر داده های پیرامونی، از اهمیت بسزایی در طراحی معماری خودروهای خودگردان برخوردار خواهد بود. برخی از روش ها همچون (۶۵)،(۹)،(۵۵) و (۷۷) فقط از یک واحد ادراکی در معماری سیستم خود استفاده کرده اند. طراحی معماری سیستم آنها فاقد یک معماری طبقاتی ساخت یافته برای درک محیط پیرامونی میباشند و به استفاده از زیر- ماژول ها برای فرآوری تمام داده های محیطی بسنده کرده اند.

در صحنه های رانندگی شهری، خودرو خودگردان باید قوانین بسیاری را با توجه به قانونگذاری های ترافیکی در نظر بگیرد. خودروهای خودگردان باید قادر باشند تا نتایج ادراکی خود را با خطوط جاده ای و نشانه های ترافیکی تطبیق بدهند تا بتوانند با وسایل نقلیه دیگر همکاری داشته باشند. بنابراین مدلسازی محیطی به سادگی ردیابی خودروهای دیگر حاضر در صحنه ترافیکی نیست بلکه یک چالش با پیچیدگی بالاست. ماژول بندی و ساختار دهی طبقاتی ممکن است برای خوش رفتار کردن این مسأله پیچیده کارآمد باشد.  بسیاری از معماری های سیستمی، با الگو برداری از مثالهای موجود در چالش شهری دارپا، عملیات رانندگی خودگردان را به زیر-عملیات های متعدد تقسیم میکنند اما در این معماری ها کمبود یک ماژول بندی و ساخت یافتگی در کل سیستم برای انجام عملیات ادراکی دیده میشود. این کمبود ساخت یافتگی همچنین در نحوه مدیریت کردن داده های متعدد ورودی به سیستم و فرآوری کردن آنها، داده های نقشه، سیستم ادارکی روی خودرو و بقیه قسمت های معماری دیده میشود(۵۶).

از دیگر جنبه های حیاتی ادراک محیطی، به روز بودن داده های محیطی میباشد که به نوبه خود وجود یک سیستم ادراکی زمان حقیقی برای تمام صحنه رانندگی را در طراحی عملکردی سیستم ضروری میکند. با توجه به وجود وقفه بین جمع آوری داده، ساخت نقشه و استفاده از داده های تقشه تولیدی، به روز بودن داده های محیطی را همواره نمیتوان تضمین کرد. بنابراین وجود یک سیستم ادراکی زمان حقیقی، چه بر روی خود خودرو و چه به عنوان بخشی از زیرساخت هوشمند برای احراز اطمینان از عملکرد سریع و زمان حقیقی خودرو خودگردان ضروری میباشد.

یک طراحی سه سطحی به عنوان پایه و اساس طراحی سیستم ادراکی خودرو خودگردان در این پایان نامه انتخاب شده است که در اصل توسط (۷۰) ارایه شده است و در شکل ۱.۲ نمایش داده شده است. این سه سطح افقی موجود در این طراحی، به طور عمده با یکدیگر در رزولوشن، افق دید و دقت در زمان و مکان متفاوت میباشند. سطوح مختلف این معماری سیستم از قرار زیر میباشند:

- **سطح استراتژیک:** در این سطح، برنامه ریزی صورت میگیرد و بزرگنمایی در ابعاد ماکرو وجود دارد.
- **سطح تاکتیکی:** در این سطح تصمیم گیری صورت میگیرد و بزرگنمایی در ابعاد مزو وجود دارد.
- **سطح عملیاتی:** در این سطح پایدارسازی راکتیو صورت می پذیرد و بزرگنمایی در ابعاد میکرو وجود دارد.

همانطور که مشاهده میشود، این معماری همچنین به بخش های عمودی متفاوتی تقسیم میشود:

- مکانیابی مطلق فراگیر
- داده های خارجی
- ادراک
  - ادراک محیطی
  - خود-ادراکی
- اجرای مأموریت

همچنین مشاهده میشود که کل طراحی سیستمی به دو بخش مجزا از لحاظ نحوه تعامل ماژول ها با محیط تقسیم میشود. مکانیابی مطلق فراگیر و داده های خارجی در بخشی قرار میگیرند که با نحوه تعامل با محیط مرتبط میباشند. ادراک و اجرای مأموریت نیز مربوط به بخشی می شوند که نحوه تعامل محیط با خودرو را شامل میشود. در ادامه به بررسی اجمالی هرکدام از زیر سیستم های عمودی در ارتباط با سطوح افقی معماری خواهیم پرداخت.

مکانیابی مطلق فراگیر: مکانیابی فراگیر موقعیت قرارگیری خودرو را در ارتباط با محیط مشخص میکند. این مفهوم در بسیاری از طراحی سیستم ها با توجه به سناریو های خاص چندان ضروری به نظر نمیرسد، به عنوان مثال در سناریو ما که اساس آن بر دید مرجع خودرو میباشد. در واقع در سناریو مورد نظر ما، ناظر مرجع بر روی خود خودرو ثابت شده است. در بسیاری از کاربرهای وسیع دیگر، مکانیابی مطلق فراگیر از اهمین بسزایی برخوردار است. به عنوان مثال در محیط هایی که هیچ ویژگی محلی در دسترس نیست، همانند بیابان ها و یا محیط های کم-ساخت یافته، مکانیابی مطلق فراگیر برای پایدار سازی خودرو به کار گرفته میشود. حتی در سناریوهای شهری، ترکیب داده های خارجی مرتبط با شرکت کنندگان متعدد حاضر در ترافیک با اهداف مختلف از جمله به اشتراک گذاری داده های تقشه مشترک، برقرار ارتباطات ( V2V ) و ( V2I ) و ... مکانیابی مطلق فراگیر حایز اهمیت می باشد. با استفاده از مکانیابی مطلق فراگیر میتوان یک پلتفرم مرکزی برای به روز رسانی نقشه قابل دسترسی برای همه استفاده کنندگان از معابر طراحی کرد.

دقت راه حل ارایه شده برای مکانیابی، سطح عملیاتی آن راه حل را مشخص میکند. دریافت کننده های ( GNSS ) دارای خطای موقعیت یابی تا حدود ۲۰ متر میباشند. از آنجایی که این دقت تنها به عنوان یک موقعیت با بزرگنمایی ماکرو به حساب می آید، این فناوری ها فقط در سطح استراتژیک

قابل استفاده میباشند. فناوریهای دیگر همچون DGPS ممکن است برای بهبود بخشیدن به دقت موقعیت یابی با استفاده از ترکیب کردن تخمین حرکت به تخمین موقعیت، به کار برده بشوند. این چنین روشهایی استفاده از روشهای متداول مکانیابی را در سطوح تاکتیکی و عملیاتی ممکن میسازند در حالی که بسیار به الزامات ایمنی وابسته میباشند.

**داده های خارجی**

تمام داده های محیطی که در داخل خودرو تولید میشوند و یا از محیط بیرونی، انبارهای داده موجود در زیرساخت های حمل و نقل، ارتباطات رادیویی، مدل فراگیری از محیط که در چارچوب مختصاتی جهانی تعریف شده است، داده های مرتبط با محیط پیرامونی ایستا همانند داده های نقشه، وضعیت چراغ راهنمایی، نشانه های ترافیکی، ساختمان ها، اجسام ایستای موجود در جاده همچنین داده های در مورد محیط پیرامونی دارای پتانسیل حرکتی ( اجسام قابل حرکت ) همانند اجسام خطرناک، انسدادهای جاده ای، گره های ترافیکی، لیست اجسام، درخواست های واصله از اجسام ترافیکی دیگر و یا اطلاعات حرکتی آنها همگی بخشی از داده های خارجی میباشند.

داده های خارجی سطوح مختلفی از دقت را در بازه های زمانی متفاوت و با توجه به سطح خلاصه سازی اطلاعات در سیستم، ارایه میدهند. به عنوان مثال، پیام های مرتبط با گره های ترافیکی و یا انسداد های مقطعی جاده ای که امروزه با استفاده از اینترنت و یا شبکه های رادیویی به خودروها ارسال میشوند را میتوان بخشی از داده های خارجی در سطوح داده ای استراتژیک و یا تاکتیکی در نظر گرفت. همچنین فهرستی از اجسام خطرناک در نزدیکی وسیله نقلیه ممکن است به عنوان یک داده خارجی در سطح عملیاتی شناخته بشود چراکه وسیله نقلیه باید از این اجسام آگاهی داشته باشد . برای مقابله کردن با خطرهای احتمالی از سوی آنها آماده باشد.

انواع مختلفی از داده های ورودی را در این معماری میتوانیم به عنوان داده های خارجی محسوب کنیم: یک موقعیت مطلق فراگیر، اطلاعات پیرامونی محلی که از ماژول ادراک محیطی بدست آمده اند، مقاصد رانندگان که از ماژول V2X وارد شده اند و داده های نقشه که از تأمین کنندگان نقشه وارد سیستم میشوند.

**ادراک:**

ماژول ادراکی، ماژول مرکزی یک خودرو خودگردان میباشد که هم شامل درک محیطی خودرو و هم درک خودرو از خودش میباشد( شکل ۱.۳). تمام اطلاعات جمع آوری شده در یک عملیات رانندگی خودگردان در مورد وسایل نقلیه پیرامونی و خود خودرو در واحد ادراکی جمع آوری میشوند. همانطور که ممکن است محیط پیرامونی در واحد داده های خارجی به گونه های متفاوتی به نمایش در آورده شود، نحوه نمایش این داده ها تا حد بسیار زیادی به سطح خلاصه سازی ادراک محیطی بستگی خواهد داشت. همانطور که در شکل ۱.۲ نشان داده شده است، حسگرهای خودرو، حسگرهای محیطی و داده های در دسترس خارجی از جمله منابع اصلی ورودی به ماژول ادراکی میباشند. همچنین خودروهای خودگردان مدرن ممکن است تحلیل های ادراکی خود را به صورت خروجی در اختیار وسایل نقلیه دیگر برای مقاصد همکاری قرار بدهند. سیستم های ردیابی تلفن همراه و مسیریاب ها از جمله خدماتی هستند که از به اشتراک گذاری درک بدست آمده از طریق یک پلتفرم نقشه میباشند که موقعیت ها در آن ها نسبت به نقشه مرکزی سنجیده میشود. از آنجایی که ماژول

ادراکی، مهمترین قسمت یک خودرو خودگردان را تشکیل میدهد، حداقل در مورد کاربردهای مد نظر ما در این پایان نامه، هرکدام از سطوح خلاصه سازی را بصورت جداگانه مورد بررسی قرار میدهیم.

**سطح عملیاتی:**

سطح عملیاتی به عنوان پایین ترین سطح خلاصه سازی در این معماری تعریف شده است. در این لایه، تمرکز ادراک محیطی بر روی استخراج دقیق و شبه-پیوسته ویژگی ها از داده ورودی سنسورها میباشد. ویژگیهای هندسی، مکان، سرعت و کیفیات بصری همچون رنگ اجسام حاضر در محیط پیرامون ممکن است در این سطح بدست بیاید. در عین حال واحد خود-ادراکی، داده های بدست آمده را برای مشخص کردن حالت های داخلی خودرو فرآوری میکند. روشهای متفاوت موجود برای بدست آورن درکی مفید از محیط بهمراه ترکیبات متفاوتی از سنسورهای ممکن، طراحی این بخش از سیستم را بیشتر با نیازمندیهای کاربردی، بودجه، شرایط محیط و سلیقه مهندسان مرتبط میکند.

**سطح تاکتیکی:**

سطح تاکتیکی را میتوان به عنوان فرآیند اصلی برای مدلسازی صحنه در نظر گرفت. تمرکز اصلی در سطح تاکتیکی و در ماژول ادراکی، بر روی ساخت زمینه ای کمک کننده بر اساس ویژگیهای محیطی اندازه گیری شده به صورت مستقل میباشد. در وهله اول در این سطح و در لایه ادراکی، محیط ایستای پیرامونی خودرو بازسازی خواهد شد، و سپس با اضافه کردن اجسام و عناصر متحرک به این محیط بازسازی شده اضافه میشود تا توصیفی تکمیلی از صحنه ارایه شود. اطلاعات مفهومی و شناختی در این سطح از خلاصه سازی، نسبت به اطلاعات هندسی و یا فضایی از اهمیت بیشتری برخوردار هستند.

**سطح استراتژیک:**

در این سطح، محیط پیرامونی در خلاصه ترین حالت ممکن در نظر گرفته شده است. همچنین در این سطح، مشاهداتی در ابعاد ماکرو مورد استفاده قرار میگیرند. ویژگیهای مورد استفاده در این سطح بیشتر، شبکه های جاده ای، جریان های ماکروسکوپی ترافیک، تقاطع ها و ویژگیهای توپولوژیکی کلان میباشند. نحوه ارتباط و اتصال جاده ها و ساختار توپولوژیکی محیط در این سطح برای برنامه ریزی مسیر در ابعاد ماکرو مورد استفاده قرار میگیرد. البته در این سطح، هنوز هم اطلاعات هندسی و شناختی سطوح قبلی برای برنامه ریزی بهینه مسیر الزامی میباشند.

**اجرای مأموریت:**

عملیات مورد نظر مسافرین و یا دیگر اُپراتور های خودرو، ورودی فرآیند **اجرای مأموریت** را تشکیل میدهند. در این قسمت هم سه سطح متفاوت استراتژیک، تاکتیکی و عملیاتی حضور خواهند داشت. عملیات های تعریف شده در ماژول اجرای مأموریت باید با موفقیت اجرا بشوند و نتیجه این عملیات ها باید با دیگر شرکت کنندگان در درجریان ترافیکی پیرامون، از کانال های متفاوت به اشتراک گذاشته بشود. در مورد یک خودرو خودگردان، دریافت کنندگان این پیام ها، مسافران خودرو میباشند و در مورد خودروهای مجهز به سیستم های کمک راننده دریافت کننده این پیامها، خود راننده خودرو میباشد. ارتباطات مورد نظر ممکن است به صورت آکوستیکی همانند بوق زدن در سطح تاکتیکی، و یا بصورت بصری همانند چراغ ترمز در سطح عملیاتی و یا هر نوع دیگری از ارتباطات بین دو وسیله نقلیه اتفاق بیفتد.

در سطح استراتژیک، برنامه ریزی تکمیل میشود. اطلاعات در ابعاد ماکرو همانند شبکه های جاده ای و جریان های ترافیکی از جمله ویژگیهایی هستند که برای این سطح آماده میشوند. از طرفی دیگر، یک مسافر و یا راننده میتواند به خودرو خودران در سطح دستور بدهد تا همان طور رفتار کند که اُپراتور ترجیح میدهد. عملیات های ناوبری در بالاترین قسمت سطح استراتژیک صورت میگیرند. شبکه جاده ای و جریان کنونی ترافیک، تنها اطلاعات ورودی مورد نیاز برای انجام هر گونه عملیات ناوبری میباشند. بر اساس این داده ها و مقصد مشخص شده و همچنین محدودیت های بهینه سازی که هز سوی مسافران خودرو خودگردان اعلام میشود، یک مسیر بهینه سازی شده اعلام میگردد که این عملیات معمولاً به ازای هر بار اعلام ورودی جدید صورت میگیرد. مسیر برنامه ریزی شده باید بتواند خود را به صورت برخط با توجه به تغییرات به وجود آمده در جریان ترافیکی و یا شناسایی ها و آشکارسازیهای جدید در جاده، تغییر بدهد و با شرایط جدید تطبیق یابد.

در سطح تاکتیکی، فرآیند تصمیم گیری صورت میگیرد. اطلاعات با ابعاد متوسط مثل صحنه های محلی خلاصه شده شامل اجسام متحرک ایستا و پویا اطلاعات و داده هایی هستند که برای استفاده این سطح آماده میشوند. پس از اجرای برنامه ریزی مسیر در سطح استراتژیک، با به حرکت در آمدن خودرو، محیط پیرامونی آن نسبت به خودرو و شیوه حرکت آن تغییر میکند. خروجی لایه استراتژیک به سیستم هدایت خودرو، نقطه ناوبری بعدی است که در سیستم های کمک رانندگی کنونی میتوان آن را در هشدارهای صوتی سیستم برای اعمال تغییر جهت در جهت های خاص مشاهده کرد. بنابراین سطح تاکتیکی و یا سیستم هدایت خودرو، مأموریت را به صورت غیر مستقیم دریافت میکند. ماژول ادراکی، موقعیت خودرو را در یک صحنه برآورد و آماده ارایه میکند و با استفاده از همین موقعیت آماده شده، صحنه با توجه به مأموریت اعلام شده بررسی خواهد شد. در این سطح، سیستم در صورت آشکار شدن یک موقعیت خطرناک به بازیگران دیگر صحنه ترافیکی هشدار لازم را اعلام میکند و خود برای مقابله با آن شرایط تصمیم میگیرد و عمل میکند. واحد تصمیم گیری، مانور های رانندگی مناسب برای شرایط کنونی را بررسی کرده . از میان آنها با در نظر گرفتن شرایط ترافیکی انتخاب میکند. این در حالی است که در سیستم های کمک رانندگی هشدار صرفاً به راننده ارسال میشود و راننده برای مقابله با این شرایط خطرناک، خود تصمیم میگیرد و عمل میکند.

در لایه عملیاتی، فرآیند اجرای تصمیمات اجرا میشود. اطلاعات با ابعاد میکرو در مورد مقادیر هندسی دقیق، برای فعالسازی سیستم فعال مقابله با تصادف و پایدار سازی خودرو تزریق میشود. ویژگیهای استفاده شده در این سطح، از واحد ادراکی وارد میشوند. ادراک محیطی ویژگیها را از روشهای بر پایه مدلسازی استخراج میکند و آنها را به این سطح میخوراند. واحد محاسبه مسیر، یک مسیر نامی را بر حسب مکان و زمان خودرو محاسبه میکند و با استفاده از این محاسبات، یک کنترل حلقه بسته بر اساس داده های محیطی کنونی پیاده میکند.

**اهمیت آشکارسازی و ردیابی اجسام متعدد متحرک:**

عملیات آشکارسازی و ردیابی اجسام متعدد متحرک، با استفاده از معماری ارایه شده در این پایان نامه در ماژول ادراکی و در سطح عملیاتی قرار میگیرد. سطح عملیاتی که پایین ترین سطح ماژول ادراکی در خلاصه سازی میباشد، به طور مستقیم با داده های سنسورهای محیطی روبرو میشود و همینطور در ارتباط مستقیم با پدیده ها و داده های جهان واقعی قرار دارد. از طرفی دیگر، این سطح مسؤل تغذیه اطلاعاتی سطوح بالایی دیگر را در ماژول ادراکی بر عهده دارد. این اطلاعات میبایست

شامل حالتها و اطلاعات محیطی مورد استفاده سطوح بالایی باشد. بنابراین ماژول ادراکی در سطح عملیاتی، واحدی حیاتی برای هر خودرو خودران میباشد چرا که این ماژول تنها مرجع اطلاعاتی برای ماژولهایی ادراکی سطوح بالاتر و در نتیجه کل سیستم میباشد.
الگوریتم ها و روشهای دقیق و مقاوم باید برای این مسائل در این سطح به کار برده شوند. بنابراین آشکارسازی و ردیابی اجسام متعدد متحرک باید در زمره یکی از حیاتی ترین عملیات هایی قرار بگیرد که یک خودرو خودران باید انجام بدهد. مفاهیم پایه ای و چالش های مرتبط با فرآیند ادراکی وسایل نقلیه و اجرای عملیات آشکارسازی و ردیابی اجسام متحرک، هسته اصلی این پایان نامه را تشکیل میدهد.

**مفاهیم و چالش های مرتبط با ادراک خودروهای خودران:**

در قسمتهای قبلی، نقش حیاتی ماژول ادراکی در یک خودرو خودران به طور اجمالی بررسی شده است. در این بخش، چالش های اصلی برای به کارگیری مؤثر ماژول ادراکی یک خودرو خودران مورد بررسی قرار میگیرد و روشهای متداول و بروز ادراکی خودروهای خودران شهری معرفی میشود. چالش های مهم بسیاری، در برای دستیابی به یک ماژول ادراکی قابل اطمینان برای خودروهای شهری باید پشت سر گذاشته بشوند.
در وهله اول، در سناریوهای رانندگی شهری، شرکت کنندگان در صحنه ترافیکی باید دسته بندی شوند و مسیر رانندگی آینده آنها باید به منظور در نظر گرفتن مسائل ایمنی و جلوگیری از برخورد احتمالی، محاسبه و تخمین زده بشود. این شرکت کنندگان صحنه ترافیکی، ممکن است دارای مدل های حرکتی متفاوتی باشند و یا حتی در کوتاه ترین بازه زمانی ممکن، مدل حرکتی خود را تغییر بدهند. بنابراین، از آنجایی که از خودرو خودران انتظار میرود بتواند به طور کامل مقاصد حرکتی حال حاضر و آینده اجسام پیرامونی را تخمین بزند، این تفاوت در مدل حرکتی اجسام پیرامونی و امکان تغییر این مدل حرکتی، لزوم مقابله با عدم قطعیت موجود را دوچندان میکند.
از طرفی دیگر، در صحنه های ترافیکی شهری، نسبت به بزرگراه ها شاهد حضور اجسام ترافیکی حساس بیشتری خواهیم بود. بنابراین روشهای مدونی باید برای نحوه رفتار خودرو خودران با استفاده کنندگان آسیب پذیر جاده بوجود بیاید. بعنوان مثال در مرجع (۲۰) به این موضوع اشاره شده است. این مرجع اشاره میکند که اگرچه فرآیند های ادراکی همچون تشخیص اجسام متحرک، بدون ارائه دادن مدلی مشخص برای زمین نیز قابل انجام میباشند، اما در محیط شهری و با توجه به حساسیت و آسیب پذیری احتمالی شرکت کنندگان در آن، تخمین و مدلسازی دقیق زمین به فرآیندی مهم در این شرایط تبدیل میشود.
علاوه بر اینها، شاید مهمترین چالش رانندگی خودگردان در محیط های شهری مربوط به تغییرات چشمگیر ساختار محیط پیرامونی خودرو خودگردان میباشد که منجر به دشوار شدن جداسازی اجسام غیرمرتبط با یکدیگر میشود (۱۶). شرکت خودروسازی تسلا، در حال هدایت مطالعات تحقیقاتی بسیاری در زمینه رانندگی خودگردان میباشد. معرفی فناوری رانندگی شبه خودگردان شرکت تسلا، به نام اتوپایلوت تسلا، شیوه رانندگی آینده انسان را با استفاده از فناوریهای هوش مصنوعی و سخت افزاری به طور کلی دگرگون ساخت. اتوپایلوت تسلا، امکان استفاده از داده های محیطی را برای دستیابی به خودگردانی نسبی در صحنه های رانندگی در بزرگراه ها را به اثبات رساند(۲۱). البته فناوری استفاده شده در اتوپایلوت تسلا، به طور ابتدایی قابل استفاده در شرایط

شهری نمیباشد (۲) چرا که در شرایط شهری، وسیله نقلیه باید بتواند بر بسیاری از پیچیدگیهای ادراکی ناشی از ساخت-نایافتگی و تغییر پذیری شرایط شهری، فائق آید تا بتواند به صورت ایمن عمل کند.
دو روش کلی ممکن برای ورود به مسأله ادراکی وجود دارد. در روش اول، با استفاده از یک دانش اولیه مسأله ادراکی بررسی خواهد شد، در حالی که در روش دوم مسأله ادراکی باید بر اساس داده های زمان-واقعی که از سنسور های نصب شده بر روی خودرو وارد میشوند، مورد بررسی قرار گیرد. روش مبتنی بر ادراک، برای کاربردهای مرتبط با سناریوهای رانندگی در محیط شهری، به دلیل طبیعت متغیر محیط شهری مناسب تر به نظر میرسد چرا که دانش اولیه ای معمولاً به این شیوه در دسترس نیست که بتواند خود را با نرخ تغییرات محیط، به طور زمان واقعی هماهنگ کند. نحوه انتخاب و قرارگیری سنسورها در خودروهای خودگردان به صورت کلی میتواند باشد: سنسورهای بصری و فاصله یاب های لیزری ممکن است به عنوان سنسور مرکزی این صورت های کلی انتخاب بشوند.
فاصله یاب های لیزری سه بعدی فشرده، قادر هستند تا داده های با وابستگی بالا و قرار گرفته در نقاط بسیار دور را از محیط پیرامونی جمع آوری کنند، در حالی که حسگرهایی همچون رادار و دوربین ها قادر به انجام این کار نیستند. بنابراین به طور کلی فاصله یابهای لیزری انتخاب مطمئن تری برای کاربرهایی همچون رانندگی خودگردان میباشند. با اینکه یک راننده انسانی میتواند به سادگی و به سرعت، ادراکی که از محیط پیرامونی دریافت میکند را دسته بندی معنایی کند، مسیر های و سرعت های احتمالی آنها را در آینده تخمین بزند و در عین حال بر روی فرآیند رانندگی نیز تمرکز داشته باشد و حتی این فرآیند را تقریباً بدون خطا به پایان برساند، انجام این مأموریت برای یک هوش مصنوعی با فناوریهای کنونی تقریباً به طور کامل غیر ممکن میباشد.
هوش مصنوعی ممکن است بتواند بدون در اختیار داشتن دانش اولیه، محل دقیق و ویژگیهای هندسی یک جسم ترافیکی را در اختیار ما قرار بدهد اما به هیچ عنوان قادر نیست تا اطلاعات مفهومی دقیقی را از آن جسم و یا ویژگیهایی حرکتی آن را ارائه کند. از این رو، دسته بندی اجسام ترافیکی، یا دسته بندی مفهومی آنها یکی از عملیات های ضروری و اولیه ایست که یک خودرو خودگردان باید بتواند از پس آن بر بیاید. سرنخ های پویای حرکتی و هندسی ممکن است برای اجرا کردن دسته بندی مفهومی مورد استفاده قرار بگیرد.
پیچیدگی ذاتی ادراک محیطی در مأموریت های رانندگی در محیطهای شهری، اهمیت معرفی مفهوم عدم قطعیت را در فرآیند مدلسازی مشخص میسازد. منبع اصلی عدم قطعیت در فرآیند ادراک محیطی، کمبود دانش برای نتیجه گیری های زمینه مفهومی و اندازه گیری های ناقص میباشند. غدم قطعیت های مرتبط با حسگرها، عدم قطعیت هایی هستند که در اندازه گیری متغیرهای فیزیکی همانند موقعیت، اندازه و سرعت اتفاق می افتد. همچنین، علاوه بر این عدم قطعیت ممکن است در نتیجه آشکارسازیهای نادرست بوجود بیاید. اندازه گیریهای مرتبط با یک جسم به درستی آشکارسازی شده ممکن است دارای خطا باشد و وسیله نقلیه نتواند در مورد وجود این جسم آشکارسازی شده اطمینان حاصل کند. روشهای مبتنی بر فیلتر بیز و رباتیک احتمالاتی رای بررسی عدم قطعیت های این زمینه مورد استفاده قرار میگیرد.

اولین قدم در راستای ادراک محیطی مطمین و قابل اعتماد، آشکارسازی دقیق و قابل اطمینان میباشد که قادر باشد موقعیت های دارای پتانسیل وجود اجسام ترافیکی را مشخص کند. بعد از صورت پذیرفتن فرآیند آشکارسازی، دسته بندی مفهومی قابل اطمینان باید صورت بگیرد تا تصمیم گیری در سطح تاکتیکی را میسر سازد. از طرفی دیگر، ردیابی قابل اطمینان اجسام متحرک از اهمیت بسزایی برخوردار است چرا که خودرو خودگردان باید قادر باشد مقاصد آینده شرکت کنندگان ترافیکی پیرامونی اش را پیش بینی کند. دسته بندی مفهومی در این مرحله نقش بسیار مهمی را بر عهده دارد، به این علت که سرنخ های مفهومی قابل اتکا و دقیق، در مورد یک جسم خاص خودرو خودگردان را قادر میسازد تا مدل حرکتی مختص آن جسم خاص را تخمین بزند.

**تشخیص جسم متحرک:**

تشخیص اجسام متحرک مهمترین بخش یک واحد ادراکی محیطی مرتبط با یک خودرو خودگردان است که میخواهد در یک محیط شهری به گشت و گذار ایمن بپردازد. در محیطهای شهری، تقریباً همواره خودرو خودگردان توسط تعداد زیادی از اشیاء احاطه شده است. بنابراین مسأله تشخیص جسم متحرک در محیطهای شهری به مسأله تشخصی اجسام متعدد متحرک تبدیل میشود. مؤثر بودن و دقت روشهای موجود در این زمینه تا حد بسیار زیادی به نحوه مدارا کردن با عدم قطعیت های موجود در مسأله و همچنین در سطح عملیاتی به حسگرهای به کار رفته شده برای درک محیط، بر میگردد. بیشتر روشهای تشخیص اجسام متحرک در ادبیات مربوطه، بر اساس معماری ردیابی بر اساس آشکارسازی عمل میکنند. اجسام بالقوه برای متحرک شدن، بر اساس داده های فراهم شده توسط حسگرهای به کار رفته آشکارسازی میشوند و موقعیت و سرعت اجسام متحرک بعد از این مرحله ردیابی میشود ( ۲۲، ۲۳، ۲۴، ۲۵، ۲۶، ۲۷، ۲۸، ۲۹).

بدست آوردن همواره آگاهی از حالت های کینماتیک اجسام ترافیکی پیرامونی، به طور پیوسته برای مدل سازی محیط درک شده و همینطور برای اجرای دستورات کنترلی و پایش ایمنی ضروری میباشد. این دانش و آگاهی باید بصورت زمان-واقعی و قابل تنظیم کردن باشد تا بتوان آن را در بازه های زمانی منطقی به کار برد. به عنوان مثال، حرکت کردن عابرین پیاده سریع و حتی تا حدودی غیر قابل پیش بینی میباشد. اگر یک عابر پیاده در حال عبور از مقابل یک خودرو باشد، دستور کنترلی مناسب برای جلوگیری از برخورد باید بصورت برخط طراحی و تنظیم شود تا خودرو بتواند در لحظه مناسب واکنش نشان بدهد و این واکنش باید در بازه زمانی زمان-واقعی قرار بگیرد تا اجرا شدن مؤثر آن تضمین بشود.

تشخیص جسم متحرک، ماژول های متعددی را در بر میگیرد. فرآیند تشخیص جسم متحرک با عملیات جمع آوری داده توسط سنسور های به کار رفته در خودرو آغاز میشود. در سناریوهای دیگر ممکن است این فرآیند جمع آوری اطلاعات از منابع دیگری نیز صورت بپذیرد. داده های خام محیطی وارد این ماژول میشوند. این داده های خام شامل هرگونه اطلاعاتی از جمله داده های مرتبط با شرکت کنندگان متفاوت حاضر در صحنه ترافیکی از جمله ایستا و یا متحرک، نشانه های ترافیکی، ساختمان ها و ... میباشند. دراین فاز از وارد مردن داده، الگوریتم های مرتبط با آشکارسازی گرفت میتوانند در صورت اجرا شدن روشهای بخش بندی قبلی، اجرا شوند. فرآیند های آشکارسازی و ردیابی بر روی داده های پالایش شده و فیلتر شده قابلیت اجرا خواهند داشت.

در مرحله آشکارسازی، داده پالایش میشود تا شامل داده های غیر ضروری نباشد. به عنوان مثال، داده های مرتبط با زمین حذف میشوند تا حجم محاسباتی عملیات کاهش پیدا کند. از طرف دیگر،اطلاعات معنایی در باره داده های در این مرحله تولید میشوند و نتیجه نهایی این مرحله، شامل اطلاعات معنایی و حرکتی شرکت کنندگان متفاوت جاده میباشد که به صورت خروجی ارسال خواهد شد. در واحد ردیابی، تخمین های متفاوت و روشهای ترکیبی حسگرها، برای ردیابی استفاده کنندگان از جاده استفاده میشوند. واحد ردیاب، همه اطلاعات مربوط به استفاده کنندگان از جاده را در یک انبار داده نگهداری میکند تا مطمین باشد اتفاق غیر قابل پیش بینی صورت نپذیرد.
تلاش های اولیه برای دستیابی به روشی پایدار برای آشکارسازی و ردیابی اهداف متحرک متعدد، بر روی ردیابی کردن اهداف نقطه ای جدا از هم و مستقل تمرکز کرده بودند. به زودی مشخص شد که ابعاد بسیاری از این مسأله در این روشها دیده نشده است و چالش های اساسی برای اجرای این فعالیت وجود دارد. یکی از اساسی ترین چالش های موجود، آمیزش درست اندازه گیری های دارای نویز و رد مربوط به ماشین ها میباشد. همچنین، چالش دیگری در مواجه با کاربردهایی همچون رانندگی خودگردان این است که اجسام متحرک دراین کاربردها معمولاً به صورت عمیقی با پس زمینه دارای تعامل و اثرات متقابل میباشند. بنابراین معمولاً اندازه گیریهای مرتبط با این اجسام توسط درهم آمیختگیهای قابل توجهی از پس زمینه آسیب میبینند.
یک چالش بزرگ دیگر در باره مسأله آشکارسازی و ردیابی اجسام متحرک متعدد، از این حقیقت نشأت میگیرد که همه مشاهدات در یک سناریو نسبت به یک حسگر متحرک انجام میشوند. این پدیده باعث بوجود آمدن پیچیدگی اضافه در مورد فرآیند ادراکی میباشد. به عنوان مثال اجسام ثابت در صحنه ممکن است متحرک به نظر برسند که این اشتباه ممکن است به علت گرفتگی و یا نویز های موجود باشد. بنابراین، مسأله آشکارسازی و ردیابی همه شرکت کنندگان در صحنه ترافیکی جاده، برای پیاده سازی و کاربرد در مواجه با مسائل دنیای واقعی، آنچنان ساده به نظر نمیرسند و باید توجه خاصی به آنها کرد.

**مفاهیم پایه ای آشکارسازی و ردیابی اجسام متحرک متعدد:**

مسأله درک یک جسم خاص، از سه فعالیت اصلی تشکیل میشود: **بخش بندی** و یا **دسته بندی**، **آشکارسازی** و **ردیابی** (۳۱). در ادامه، مفاهیم پایه ای مرتبط با این سه فرآیند مختلف را با ارایه جزییات تشریح و روشهای متفاوت ارایه شده در ادبیات روز را بررسی میکنیم. اگر فرض بر این باشد که داده قرار است پیش از استفاده مورد پالایش قرار بگیرند، فرآیند بخش بندی و دسته بندی قبل از شروع فرآیند اصلی آشکارسازی و ردیابی اجسام متحرک متعدد، باید صورت بگیرند. بنابراین، شناخت اجسام متعدد در دو مرحله اصلی صورت میگیرد: **آشکارسازی** و **ردیابی**. به همین دلیل این فرآیند آشکارسازی و ردیابی اجسام متعدد نامیده میشود. در شکل ۲.۱ نمایی کلی از این فرآیند نشان داده شده است.

در این پایان نامه، در مرحله آشکارسازی، داده های خام سه بعدیبخش بندی میشود تا اجسام متحرک از پس زمینه ثابت قابل تشخیص باشد. اجسام متحرک، تمام اجسامی در نظر گرفته میشوند که قابلیت به حرکت درآمدن را داشته باشند، اگرچه در حال حاضر و یا در کل مدت عملیات ثابت باشند. این تعریف به این دلیل میباشد که خودرو خودگردان باید همواره نسبت به خطرات به حرکت در آمدن ناگهانی این اجسام هشیار باشد. دسته بندی معنایی که پیش از این نیز از آن صحبت شده

است، به برچسب زدن بر روی این اجسام کمک شایانی میکند، به عنوان مثال یک خودرو ثابت با یک درخت ثابت بسیار متفاوت میباشند. همینطور، دسته بندی کردن داده ها، به تفکیک کردن اجسام متحرک و ثابت در هر لحظه کمک میکند. پس از بخش بندی کردن، موقعیت و جهت گیری اجسام با استفاده از مدل حرکتی مخصوصشان تخمین زده میشود.
بخش بندی داده در واقع مسیر را برای آشکارسازی اجسام باز میکند. برای انجام یک آشکارسازی اجسام کامل، ویژگیهای با معنی باید از داده خام استخراج بشود که این ویژگیها با توجه به زمینه مورد نظر و هدف نهایی خودرو تعیین میشوند. به بیان دیگر، اجسام آشکارسازی شده باید به طور حتمم از نظر معنایی دسته بندی بشوند چرا که ممکن است اجسام آشکار شده متحرک نباشند و یا حتی ممکن است این اجسام ثابت و غیر متحرک، ثابت باقی نمانند. بنابراین رفتار دینامیکی اجسام آشکار شده از اهمیت بسزایی برخورداد میباشند. به منظور نگهداری این رفتارهای دینامیکی، اجسام آشکار شده، به وروودی ماژول ردیابی خورانده خواهند شد که امکان ردیابی تحولات و رفتار دینامیکی اجسام آشکار شده موجود در محیط پیرامونی را فراهم میکند. پس از این مرحله، یک یک مدل حرکتی به هر یک از این اجسام ردیابی شده اختصاص داده خواهد شد. بر اساس همین مدل حرکتی اختصاص داده شده، ویژگی های حرکتی این اجسام در بازه های زمانی متفاوت محاسبه خواهند شد.
فرآیند آشکارسازی، پیش زمینه ای حیاتی برای اجرایی شدن فرآیند ردیابی به وجود می آورد. از این رو هر اشتباهی در فرآیند آشکارسازی، اثرات نا مطلوبی را در نتایج مربوط به ردیابی میگذارد. مسائل مرتبط با گرفتگی های موجود در داده خام، عدم قطعیت های موجود در فرآیند اندازه گیری، عابرین پیاده ای که ممکن است اجسام را بپوشانند، حدسهای نادرست در مورد جهت گیری اجسام آشکار شده و مسائل اینچنینی دیگر ممکن باعث بوجود آمدن مشاهدات ناکاملی بشود که به نوبه خود باعث پیدایش اطلاعات هندسی ناکامل و ناقص خواهد شد. در سناریوهای شهری، گرفتگی های حسگرها قابل پیشگیری نیستند و اغلب به دفعات اتفاق می افتند، بنابراین روشهایی باید برای مقابله با این اثران در نظر گرفته بشود.
همچنین ردیاب بخش مهمی از هر ماژول شناسایی اجسام متحرک متعدد میباشد، که مقادیر بیشتری از عدم اطمینان در فرآیند ردیابی نسبت به فرآیند آشکارسازی دخیل میباشد. الگوریتم ردیابی به کار گرفته شده در این روش سعی میکند تا به طور مؤثری همه تحولات احتمالی ممکن برای اجسام آشکار شده را در نظر بگیرد که در عین حال از وجود عدم قطعیت در اندازه گیری و مدلسازی رنج میبرد. فیلتر های بیزین بهینه، همچون فیلترهای کالمن و یا روشهای فیلتر احتمالاتی دیگر همچون فیلتر ذرات برای مواجه با این عدم قطعیت ها مورد استفاده قرار میگیرند.
همینطور که خودرو به مسیر خود به سمت اجسام متفاوت در محیط ادامه میدهد، الگوریتم آشکارساز در هر فریم ورودی داده به آشکارسازی اجسام جدید ادامه خواهد داد. این اجسام تازه یافت شده، چه متعلق به اجسام یافت شده قبلی باشند که در حال طی کردن تحول دینامیکی خود در محیط میباشند، چه اجسام یافت شده جدیدی باشند و یا حتی صرفاً یک خطای ناشی از گرفت باشند، الگوریتم آشکارسازی ما قادر نخواهند بود فرق بین این ها را مشخص کنند. در یک سناریو ردیابی اجسام با اهداف متعدد، اجسام متعددی با مسیرهای احتمالی متفاوت و متعددی یافت خواهند شد. فرآیند آمیزش داده برای مواجه با همین چالش طراحی شده است. الگوریتم آمیزش داده ها، هر جسم

تازه یافت شده را به یک ردپای موجود نسبت میدهد که دارای یک فرآیند فیلتر کردن حالت های از پیش اختصاص داده شده میباشد. همچنین الگوریتم میتواند بر اساس نشانه های حرکتی این اجسام تازه یافت شده یک فرآیند فیلتر کردن جدید به آنها نسبت داده میشود. این فرآیند آمیزش داده ها سعی میکند که یک هویت منحصر به فرد از ردپاهای موجود را در همه زمانها نگه دارند که در عین حال سعی میکند هیچ جسم متحرکی در این میان از قلم نیافتد.

بعد از طی شدن همه این مراحل، ممکن است تعدادی ردپای مشکوک در میان ردپاهای موجود پیدا شوند. ردپاهایی ممکن است پیدا شود که واقعی نیستند و یا تخمین مرتبط با آنها به دلایل متفاوت و خطاهای مختلف، به اندازه کافی واقعی نباشد. همچنین ممکن است ردپاهایی وجود داشته باشد که به دلیل خطاهای موجود در مرحله دسته بندی، مدل حرکتی درستی برای آنها در نظر گرفته نشده باشد.

در این پایان نامه سعی شده است تا در حد توان و حوصله زمانی به هرکدام از چالش های اشاره شده، در کنار طراحی معماری برای خودروهای خودگردان و همچنین ارائه الگوریتم ها و روشهای مؤثر پرداخته شود.

# Appendix A

# Gaussian Process Regression

Gaussian Processes can be viewed in two different fashion: *function-space view* and *weight-space view*.

**Weight-Space View** Consider a simple linear regression model where the output will be a linear combination of the inputs. While this method of handling the problem benefits from both the simplicity and interpret-ability at the same time, it has a main drawback that only allows a limited flexibility to real world conditions. This means that if the assumption of linear relationship between output and input fails to be true only with a very small divergence, the model gives us very poor predictions. In the Bayesian scheme, we can treat linear models with projecting inputs into a high dimensional feature space and then applying the linear model there which is known as "Kernel Trick". Assume that we have n observations, so our training set will be of the form $\mathcal{D} = \{(x_i, y_i)|i = 1, 2, ..., n\}$ with x being input vector of dimension $\mathcal{D}$ and y being scalar output or target (dependent variable). Also note that $x_i$ vectors are column vectors, for better presentation all the input column vectors are aggregated in $\mathcal{X}$ vector and all of outputs or targets are collected in the vector $\mathcal{Y}$, so we have the observations like $\mathcal{D} = (\mathcal{X}, \mathcal{Y})$.

In the Gaussian Process context we are interested in making inferences about the relationship between inputs and outputs, actually we want to obtain $P(\mathcal{Y}|\mathcal{X})$ or the conditional distribution of the targets given the inputs.

**The Standard Linear Regression Model** The main topic of this section is Bayesian analysis of the standard linear regression model with added Gaussian process noise. The standard linear model for regression task is of the form below:

$$f(x) = x^T \omega \qquad y = f(x) + \epsilon \tag{A.1}$$

Where x is the input vector and $\omega$ is a vector of weights or parameters of the linear model, also $f$ is the function value and y is the observed target. We have made the assumption that the observed values, are differed from function values by independent and identically distributed Gaussian additive noise with zero mean and variance $\sigma_n^2$:

$$\epsilon \sim \mathcal{N}(0, \sigma_n^2) \tag{A.2}$$

We actually tend to find the probability density of the observations given the parameters. We can further factor this probability density over training set because we assume that the cases in the training set are independent, this gives us equation below:

$$\begin{aligned} P(y|\mathcal{X}, \omega) &= \prod_{i=1}^{n} P(y_i|x_i, \omega) = \prod_{i=1}^{n} \frac{1}{\sqrt{2\pi}\sigma_n} exp\Big( - \frac{(y_i - x_i^T\omega)^2}{2\sigma_n^2}\Big) \\ &= \frac{1}{\big(2n\sigma_n^2\big)^{n/2}} exp\big( - \frac{1}{2\sigma_n^2}|\mathcal{Y} - \mathcal{X}^T\omega|^2\big) = \mathcal{N}(\mathcal{X}^T\omega, \sigma_n^2 I) \end{aligned} \tag{A.3}$$

The $|.|$ operator is The Euclidean norm (length) operator. In the bayesian formalism we need to put a prior over the parameters, this means that we should express our belief about these parameters from the beginning and by assigning this prior knowledge and this must be done before we look at the observations. As is previously mentioned in the introduction, now we must introduce a prior on weights, we set this prior distribution to be a zero mean Gaussian distribution with the covariance matrix $\Sigma_p$:

$$\omega \sim \mathcal{N}(0, \Sigma_p) \tag{A.4}$$

Now after assigning these priors, we need to perform the inference procedure on Bayesian linear model. In this step we make use of infamous Bayesian rule. In the linear fashion (like the case we have here) the posterior distribution $P(\omega|\mathcal{X}, y)$ is also the mode of distribution and therefore this method is often called Maximum A Posteriori (MAP) estimate of weights vector $\omega$:

$$\begin{aligned} Posterior &= \frac{Likelihood \times Prior}{Marginal\ Likelihood} \\ P(\omega|y, \mathcal{X}) &= \frac{P(y|\mathcal{X}, \omega)P(\omega)}{P(y|\mathcal{X})} \end{aligned} \tag{A.5}$$

In fact marginal likelihood term is a normalization constant which is independent of the weights. It is assumed to be normalizing constant because $P(y|\mathcal{X}) = \int P(y|\mathcal{X}, \omega)P(\omega)d\omega$ holds and obviously the denominator is obtaind with integration of nominator on whole omega space. After completing the squares and also writing weight-dependent terms the equation will be of the form:

$$\begin{aligned} P(\omega|y, \mathcal{X}) &\propto \exp\Big( - \frac{1}{2\delta_n^2}(y - \mathcal{X}^T\omega)^T(y - \mathcal{X}^T\omega)\Big) \exp(-\frac{1}{2}\omega^T\Sigma_p^{-1}\omega) \\ &\propto \exp\Big( - \frac{1}{2}(\omega - \bar{\omega})^T\Big(\frac{1}{\delta_n^2}\mathcal{X}\mathcal{X}^T + \Sigma_p^{-1}\Big)(\omega - \bar{\omega})\Big) \end{aligned} \tag{A.6}$$

With $\bar{\omega}$ being as $\bar{\omega} = \delta_n^{-2} A^{-1} \mathcal{X} y$ and A ebing as $A = \delta_n^{-2} \mathcal{X}\mathcal{X}^T + \Sigma_p^{-1}$. It is obvious that the posterior distribution in this fashion is Gaussian with the mean value $\bar{\omega}$ and covariance matrix $A^{-1}$.

$$P(\omega|y, \mathcal{X}) \sim \mathcal{N}(\bar{\omega}, A^{-1}) \tag{A.7}$$

According to [Ramussen] in order to find a prediction for a test case, we can average over all possible parameter space ($\omega$ space), multiplied by their posterior probability, thus the predicitive distribution for a test location like $x_*$ with the function relationship like $f_* = f(x_*)$ is in fact average of all possible linear models output with respect to the Gaussian posterior:

$$\begin{aligned}
P(f_*|x_*, \mathcal{X}, y) &= \int P(f_*|x_*, \omega) P(\omega|\mathcal{X}, y) d\omega \\
&= \int x_*^T \omega P(\omega|\mathcal{X}, y) d\omega \\
&= \int x_*^T \omega \Big( \mathcal{N}(\bar{\omega}, A^{-1}) \Big) d\omega \\
&= \int x_*^T \omega \Big( \mathcal{N}(\delta_n^{-2} A^{-1} \mathcal{X} y, A^{-1}) \Big) d\omega \\
&= \mathcal{N}(\frac{1}{\delta_n^2} x_*^T A^{-1} \mathcal{X} y, x_*^T A^{-1} x_*)
\end{aligned} \tag{A.8}$$

The result shows that the predictive distribution is also Gaussian with mean value given by test input multiplied by posterior mean of weights and the covariance matrix of the test output is a quadratic form of posterior covariance matrix and the test input vectors.

**Function-Space View:** If we directly consider the function space in a Gaussian Regression problem, we can view the concept of the distribution over functions as a stochastic process called Gaussian Process. Actually Gaussian Processes are one tractable way to assign prior distributions over functions in regression tasks. Using Gaussian Process Priors benefits from the fact that all the predictions can be made analytically at least in some levels, for fixed hyper-parameters and also we can define a global noise level for our regression task. A stochastic process is any collection of random variables that are indexed with a reference set of variables, like $\{f(x)|x \in X\}$ which in this thesis our random variables are the value of the function $f(x)$ in given location $x$ and we take $f_k \equiv f(x_k)$ corresponding notation for $(x_k, y_k)$. Also our index set $\mathcal{X}$ is in fact a set consisting of any possible input variable which often is assumed to be $\mathcal{R}^{d \times n}$ with $d$ being number of inputs and $n$ being number of data points.

Stochastic processes are fully defined by introducing the probability distribution for any given finite subset of variables $f(x_1), ..., f(x_n)$ so that we can access all of possible subsets of the process with the assigned distribution. A Gaussian Process is a stochastic process which any finite number of its random variables have a joint Gaussian distribution, Therefore, Gaussian Processes are completely defined by their mean function $m(x) = E\{f(x)\}$ and covariance function $k(x, x') = E\{f(x) - m(x)\}\{f(x') - m(x')\}$. We use the $\mathcal{GP}$ notation to define Gaussian Processes in this thesis.

$$f(x) \sim \mathcal{GP}\Big(m(x), k(x, x')\Big) \tag{A.9}$$

The most important specification of $\mathcal{GP}$s are the fact that any examination of larger set of variables does not change the distribution of smaller subsets of that variables. In other words, if we consider a $\mathcal{GP}$ as $(y_1, y_2) \sim \mathcal{N}(\mu, \Sigma)$, Then this Gaussian Process also specifies $y_1 \sim \mathcal{N}(\mu_1, \Sigma_1)$ where $\Sigma_1$ is a sub-matrix of $\Sigma$. This is called *consistency requirement* or *marginalization property* in the literature.

**Covariance Function:** In *Gaussian process* methodology, covariance matrix is of great importance as it specifies the covariance between pairs of random variables and this implies the distribution of functions by itself. In $\mathcal{GP}$ literature, covariance matrix between the outputs is written as a function of inputs. One of the most popular covariance kernels is the so called *Squared Exponential Covariance Kernel* which is defined as follows:

$$k(x_p, x_q) = \exp(-\frac{1}{2}|x_p - x_q|^2) \tag{A.10}$$

Based on this kernel, the noise-free covariance function is constructed:

$$K(f(x_p), f(x_q)) = cov\{f(x_p), f(x_q)\} = \sigma_f^2 k(x_p, x_q) \tag{A.11}$$

This covariance function is almost equal to unity for the inputs that are very close together, while its value decreases with the increase in euclidean distance of input points from each other. If we wish to obtain the resulting distribution of functions evaluated by $\mathcal{GP}$ method we can draw samples from the distribution of functions at any number of points and the corresponding covariance can be obtained in Squared Exponential fashion with the use of latter equation and the functions distribution is of the form given below:

$$f_* \sim \mathcal{N}\Big(0, K(\mathcal{X}_*, \mathcal{X}_*)\Big) \tag{A.12}$$

With $\mathcal{X}_* \in \mathbb{R}^{d \times n}$. Also we will get in to concept of *Characteristic Length-Scales* in proceeding sections of this chapter. Characteristic Length-Scale $\mathcal{L}$ is the distance that you are able to displace in input space before a very specific change in the function value can be sensed. This concept often gives us a rough perception about how smoothly our data is distributed in its space. This concept often shows itself in Squared Exponential covariance functions as the denominator of exponential argument or $\frac{|x_p - x_q|}{\mathcal{L}}$ which we can see the relationship with the covariance value and change in length-scale value by just looking at the equation.
Often covariance functions that we use in our $\mathcal{GP}$ regression task have some free parameters inside themselves that we are not always so sure about their value, these parameters are called hyper-parameters and different methods of declaring them has got a quiet diversity in $\mathcal{GP}$ literature. For example consider one-dimensional squared exponential covariance kernel for noisy targets $y$ that has the following form :

$$k_y(x_p, x_q) = \sigma_f^2 \exp(-\frac{(x_p - x_q)^2}{2\mathcal{L}}) + \sigma_n^2 \delta_p q \tag{A.13}$$

In this one-dimensional expression for covariance function, the parameters $\sigma_f^2$ is the signal variance, $\mathcal{L}$ is the length-scale and $\sigma_n^2$ is the noise variance. These parameters are the parameters that are not always predefined in real world situations, also we even may have not any physical interpretation for them in our application, therefore these parameters are called *hyper parameters* in $\mathcal{GP}$ regression tasks.
Also we will encounter the concept of *stationarity* for covariance functions in proceeding materials. A stationary covariance function is a covariance function that is invariant to translation in input space, this means that this covariance function is only related to the euclidean norm between two input points or $||x - x'||$ and not their values itself. Also isotropic covariance functions are covariance functions that are only functions of euclidean distance of two input points or $|x - x'|$. For an example for a covariance function that is both stationary and isotropic we can name the squared exponential covariance function.

**Prediction With Noise-Free Observation:** The predicted distribution of a regression task $\{(x_i, f_i)\}$ and $i = 1, ..., n$ with noise-free observation for the test output $f_*$ and training output $f$ can be botained with equation below given below. the distribution of $f$ and $f_*$ will be as :

Using this Gaussian joint distribution the predictive distribution will be as :

$$P(f_* | \mathcal{X}_*, \mathcal{X}, f) \sim \tag{A.14}$$

$$\mathcal{N}\Big(K(\mathcal{X}_*, \mathcal{X})K(\mathcal{X}, \mathcal{X})^{-1} f, K(\mathcal{X}_*, \mathcal{X}_*) - K(\mathcal{X}_*, \mathcal{X})K(\mathcal{X}, \mathcal{X})^{-1} K(\mathcal{X}, \mathcal{X}_*)\Big)$$

**Prediction With Noisy Observation:** By having noisy observation we mean that we do not have access to the real function values and the only thing that we can get our hand on, is the noisy observations or noisy measurements of the function which we specify by $y = f(x) + \epsilon$, This case is actually the case that we mostly have to work with and we assume that $\epsilon$ is an additive independent and identically distributed Gaussian noise with variance $\sigma_n^2$. We must also introduce the covariance kernel function for noisy measurements which finally leads to induction of prior on noisy measurements.

$$cov(y_p, y_q) = k(x_p, x_q) + \sigma_n^2 \delta_p q \tag{A.15}$$

$$Cov(\mathcal{Y}) = K(\mathcal{X}, \mathcal{X}) + \sigma_n^2 I \tag{A.16}$$

This definition of covariance for noisy measurements will result the joint distribution of observed values and the function values at the test point as follows:

$$\begin{bmatrix} y \\ f_* \end{bmatrix} \sim \left(0, \begin{bmatrix} K(\mathcal{X}, \mathcal{X}) + \sigma_n^2 I & K(\mathcal{X}, \mathcal{X}_*) \\ K(\mathcal{X}_*, \mathcal{X}) & K(\mathcal{X}_*, \mathcal{X}_*) \end{bmatrix}\right) \tag{A.17}$$

With these informations we can obtain the predictive distribution equation for Gaussian Process Regression which is a key concept in the proceeding sections:

$$\begin{aligned} P(f_*|\mathcal{X}, y, \mathcal{X}_*) &\sim \mathcal{N}(\bar{f}_*, cov(f_*)) \\ \bar{f}_* &= E\{P(f_*|\mathcal{X}, y, \mathcal{X}_*\} \\ &= K(\mathcal{X}_*, \mathcal{X})\Big(K(\mathcal{X}, \mathcal{X}) + \sigma_n^2 I\Big)^{-1} y \\ cov(f_*) &= K(\mathcal{X}_*, \mathcal{X}_*) - K(\mathcal{X}_*, \mathcal{X})\Big(K(\mathcal{X}, \mathcal{X}) + \sigma_n^2 I\Big)^{-1} K(\mathcal{X}, \mathcal{X}_*) \end{aligned} \tag{A.18}$$

The most important noteworthy thing that these equations can bring up to us, is that the variance that is presented here does not depend on the observed targets, but only on the inputs, which is actually a property of Gaussian distributions. As we are often going to find ourselves involved with expressions including $K(\mathcal{X}, \mathcal{X}) \in \mathbb{R}^{n\times n}$, $K(\mathcal{X}_*, \mathcal{X}) \in \mathbb{R}^{d\times n}$, $K(\mathcal{X}, \mathcal{X}_*) \in \mathbb{R}^{n\times d}$ and $K(\mathcal{X}_*, \mathcal{X}_*) \in \mathbb{R}^{d\times d}$ we are going to introduce a new notation for simplicity. We set $K = K(\mathcal{X}, \mathcal{X})$ and $K_* = K(\mathcal{X}, \mathcal{X}_*)$ to shorten the notation.

**Prediction Distribution With Only One Test Point** In the case that there is only one test point $x_*$ we will write $k_* = k(x_*)$ to denote the vector of covariance between test point and all of the n training points. then we can change the latter equations to the form below:

$$\begin{aligned} \bar{f}_* &= k_*^T \Big( K + \sigma_n^2 I \Big)^{-1} y \\ \mathcal{V}(f_*) &= k(x_*, x_*) - k_*^T \Big( K + \sigma_n^2 I \Big)^{-1} k_* \end{aligned} \quad \text{(A.19)}$$

In this case we can see that the mean prediction is in fact a linear combination of observations so this is called a Linear Predictor in literature. This relation also can be seen as a linear combination of n kernel functions, each one centered on the training point:

$$\bar{f}(x_*) = \sum_{i=1}^{n} (K + \sigma_n^2 I)_i k(x_i, x_*) \quad \text{(A.20)}$$

The amazing fact about this result is that the mean prediction can be written as a finite sequence despite that $\mathcal{GP}$ is often represented in terms of possibly infinite number of kernel functions (*representer theorem*). This theorem can be described by noting that although $\mathcal{GP}$ defines a joint Gaussian distribution over all of measurement space, which is considered to be one-to-one in the index set $\mathcal{X}$, the only thing that we should care about when we are making predictions about $x_*$ is all of the previous $n$ training points and the test point.

**Marginal Likelihood** Marginal likelihood or evidence $P(y|\mathcal{X})$ is the integral of likelihood times the prior:

$$P(y|\mathcal{X}) = \int P(y|f, \mathcal{X}) P(f|\mathcal{X}) df \quad \text{(A.21)}$$

The term marginal goes to the marginalization over function $f$ values. Obviously, under Gaussian process model the prior is Gaussian $f|\mathcal{X} \sim \mathcal{N}(0, K)$. This leads us to the equation for logarithm of marginal likelihood:

$$\log P(f|\mathcal{X}) = -\frac{1}{2} f^T K^{-1} f - \frac{1}{2} \log |K| - \frac{n}{2} \log(2\pi) \quad \text{(A.22)}$$

In general the product of two Gaussian gives another Gaussian distribution which is un-normalized:

$$\mathcal{N}(x|a, A)\mathcal{N}(x|b, B) = Z^{-1}\mathcal{N}(x|c, C) \quad \text{(A.23)}$$

Where in this equation the resulting Gaussian has an inverse variance matrix (precision) equal to the sum of the precisions $C = (A^{-1} + B^{-1})^{-1}$ and a mean equal to the convex sum of the means weighted by the precisions $c = C(A^{-1}a + B^{-1}b)$. Also the normalizing constant looks itself like a Gaussian distribution:

$$Z^{-1} = (2\pi)^{-\frac{D}{2}}|A+B|^{-\frac{1}{2}} \exp\Big( -\frac{1}{2}(a-b)^T(A+B)^{-1}(a-b)\Big) \quad \text{(A.24)}$$

Also the likelihood itself is a factorized Gaussian distribution $y|f \sim \mathcal{N}(f, \sigma_n^2 I)$. combining these equations we can find the log marginal likelihood:

$$\log P(y|\mathcal{X}) = -\frac{1}{2}y^T(K+\sigma_n^2 I)^{-1}y - \frac{1}{2}\log|K+\sigma_n^2| - \frac{n}{2}\log(2\pi) \quad \text{(A.25)}$$

This result is also equivalent to the result by viewing the measurement az a zero mean Gaussian process with the covariance $K + \sigma_n^2 I$ or $y \sim \mathcal{N}(0, K + \sigma_n^2 I)$.

**Bayesian Model Selection** In Bayesian methodology, the hierarchical specification of models are divided into 3 different levels, the lowest level of this hierarchical model belongs to the parameters $\omega$, the middle level belongs to the hyper-parameters $\theta$ which control the quality of distribution of parameters in the first level and at last and at the top level there are discrete sets of possible model structures $\mathcal{H}_i$ that is originally the center of our concentration in different problems.

**First Level Inference:** At this level for the inference procedure we use *Bayes rule* to obtain the posterior on the parameters:

$$P(\omega|y, \mathcal{X}, \theta, \mathcal{H}_i) = \frac{P(y|\mathcal{X}, \omega, \mathcal{H}_i)P(\omega|\theta, \mathcal{H}_i)}{P(y|\mathcal{X}, \theta, \mathcal{H}_i)} \tag{A.26}$$

If we take a glance at this equation we will find out that the term $P(y|\mathcal{X}, \omega, \mathcal{H}_i)$ is actually the likelihood term, and the term $P(\omega|\theta, \mathcal{H}_i)$ is the prior over parameter. The prior over the parameter actually is giving us some knowledge about the parameters prior to observing the data, If we have little information about the parameters before observing the data this prior is chosen to be broad to reflect effects of little information of ours. Note that the inference procedure combines the information from prior and the data through the likelihood term. The term in the denominator can be obtained using marginal likelihood definition:

$$P(y|\mathcal{X}, \theta, \mathcal{H}_i) = \int P(y|\mathcal{X}, \omega, \mathcal{H}_i)P(\omega|\theta, \mathcal{H}_i)d\omega \tag{A.27}$$

**Second Level Inference:** In this level we want to express the posterior over hyper-parameters as the marginal likelihood from first level inference is the likelihood of this level inference:

$$P(y|\theta, \mathcal{X}, \mathcal{H}_i) = \frac{P(y|\mathcal{X}, \theta, \mathcal{H}_i)P(\theta|\mathcal{H}_i)}{P(y|\mathcal{X}, \mathcal{H}_i)} \tag{A.28}$$

The normalization constant can be obtained using the following equation:

$$P(y|\mathcal{X}, \mathcal{H}_i) = \int P(y|\mathcal{X}, \theta, \mathcal{H}_i)P(\theta|\mathcal{H}_i)d\theta \tag{A.29}$$

**Third Level Inference:** At last and at the top level we tend to find the posterior of the model:

$$P(\mathcal{H}_i|y, \mathcal{X}) = \frac{P(y|\mathcal{X}, \mathcal{H}_i)P(\mathcal{H}_i)}{P(y|\mathcal{X})} \tag{A.30}$$

$$P(y|\mathcal{X}) = \sum_i P(y|\mathcal{X}, \mathcal{H}_i)P(\mathcal{H}_i) \tag{A.31}$$

**Model Complexity:** As one may mention, often implementing the Bayesian model selection method requires computation of several integrals therefore as in alternative method one may try to maximize the likelihood given in the first level inference with respect to the hyper-parameters $\theta$ to obtain actual values for hyper-parameters instead of using the second level inference to find the whole distribution of hyper-parameters. This actually opens a new division of methods that are fundamentally different from Bayesian scheme of inference in the part that they use optimization of likelihood over hyper-parameters instead of involving integrals over whole hyper-parameters space. Marginal Likelihood plays a vital role in model selection because it inherently does a trade-off between model-fit criterion and model complexity.

Consider a model selection task with fixed n data points and input set X. If we consider a simple model for our task, this model only remains reliable for a limited range of possible target values because the model is actually a probability distribution over y and data sets that our model accounts for must have large value of marginal likelihood. If we choose a model for our task that is too complex, although we make our model to be capable of being accountable for larger range of data sets, we make the marginal likelihood unable to have large value ( Occam's razor effect). This would cause that marginal likelihood factor the model that fits the data set the best, Occam's razor effect simply says that plurality should not be assumed without necessity which encourages us to be more simple-minded in our explanations.

The result is that in order to have better marginal likelihood, the model should be of intermediate complexity over too simple or too complex alternatives. the amazing thing about Bayesian methods is that this trade-off between complexity and data-fit doesn't need us to externally set parameters and it is happening automatically when we are using Bayesian scheme methods. We should not confuse the automatic Occam's razor principle with the use of priors in the Bayesian method because even if we choose a completely flat priors over our parameters, the marginal likelihood still tends to select a less complex model itself.

**Model Selection For $\mathcal{GP}$ Regression:** While Bayesian principles represent a very persuasive and consistent framework towards inference, in the most machine learning real world problems computations demand integration over all parameter space and consequently much of them are analytically intractable and often good approximations are not easily derived. The very rare exception of this fact can be the Gaussian process regression with Gaussian noise involved which present kind of integrals over parameters which are analytically tractable and also flexible models.

Gaussian process regression method is a non-parametric model in the way that model parameters are not known prior to implementation of inference procedure and it is not always obvious that what the parameters of the model are. In order to take a step toward formalization of the model selection for $\mathcal{GP}$ regression task, we first derive the marginal likelihood or evidence $P(y|\mathcal{X})$ equation. As we said before marginal likelihood is the integral of likelihood time the prior. Under the Gaussian process assumption the prior on the function is Gaussian $f|\mathcal{X} \sim \mathcal{N}(0, K)$ and in other notation:

$$\log P(f|\mathcal{X}) = -\frac{1}{2} f^T K^{-1} f - \frac{1}{2} \log |K| - -\frac{n}{2} \log 2\pi \qquad \text{(A.32)}$$

As we mentioned before, The likelihood itself is a factorized Gaussian distribution $y \sim \mathcal{N}(0, K_f + \sigma_n^2 I)$. combining these equations we can find the log marginal likelihood:

$$\log P(y|\mathcal{X}, \theta) = -\frac{1}{2} y^T (K + \sigma_n^2 I)^{-1} y - \frac{1}{2} \log |K + \sigma_n^2| - \frac{n}{2} \log(2\pi) \qquad \text{(A.33)}$$

$$\log P(y|\mathcal{X}, \theta) = -\frac{1}{2} y^T K_y^{-1} y - \frac{1}{2} \log |K_y| - \frac{n}{2} \log(2\pi) \qquad \text{(A.34)}$$

Note that in this setup, $K_y = K + \sigma_n^2 I$ is the covariance matrix for the noisy targets $y$ and $K_f$ is the covariance matrix for the noise-free latent $f$. The result is that we have written the log marginal likelihood function based on covariance matrices which are a function of hyper-parameters itself. In this representation three items can play significant role that can help us to interpret the behavior of log marginal likelihood function and in result the behavior of our $\mathcal{GP}$ regression task.

First term is the only term involving the noisy observed targets and it is called the *data-fit* or $-\frac{1}{2} y^T K_y y$, this term decreases monotonically with the length-scale because the model becomes less and less flexible with growth of length-scale. The next term is $-\frac{1}{2} \log |K_y|$ which is actually the *complexity penalty* term which only depends on the covariance functions and the input, the negative complexity penalty increases with the length-scale because as the length-scale increases, causes less complex model by this growth, and finally the term $-\frac{n}{2} \log(2\pi)$ is a constant that is helpful for normalization and thus it is called *normalization constant*.

In Gaussian Process regression fashion we can set the hyper-parameters by optimizing the log marginal likelihood function, therefore we should seek the partial derivatives of the log marginal likelihood function with respect to the *hyper-parameters* $\theta$ to implement optimization methods.

$$\begin{aligned}\frac{\partial}{\partial\theta_j}\log P(y|\mathcal{X},\theta) &= \frac{1}{2}y^T K^{-1}\frac{\partial K}{\partial\theta_j}K^{-1}y - \frac{1}{2}tr\left(K^{-1}\frac{\partial K}{\partial\theta_j}\right) \\ &= \frac{1}{2}tr\left((K^{-1}y(K^{-1}y)^T - K^{-1})\frac{\partial K}{\partial\theta_j}\right) \qquad \text{(A.35)}\end{aligned}$$

**Computational Complexity** Computational complexity of Gaussian process regression method is mostly concentrated on computing marginal likelihood, and computing marginal likelihood's computation cost is dominated by the matrix inversion operation for $K^{-1}$ term which is in order of $\mathcal{O}(n^3)$. Also computing derivatives for hyper-parameter evaluation is of order $\mathcal{O}(n^2)$ which shows that the computational overhead of computing derivatives is small and this makes the use of gradient-based optimization methods applicable. log marginal likelihood is presented for the case that there is a data set available and the hyper-parameters of log marginal likelihood can be assumed to be just the result of this data set, but in our case we are often confronted with the situation that we are given several number of data sets and it is assumed that all of these data sets share the same hyper-parameters, this is called *multi-task* marginal likelihood. In this case we apply the optimization on the sum of all log marginal likelihoods of respected problems with respect to these shared hyper-parameters.

# Appendix B

# Non-stationary Spatial Covariance Function

In Gaussian process setting, specifications of a covariance structure is of great importance and is a vital part of algorithm design when one wants to deal a with real-world problem. In the ground segmentation problem using 3D LiDAR data, we are confronting with locally varying geometric anisotropies in our data, demanding a special treatment for covariance deployment purposes. Although, ground segmentation could be a certain example of using *spatial statistics* methodologies, even at least by the name, but there are very rare methods regarding or citing to this literature while there are many useful ideas in this area for contribution to the field. We did not do such contribution but used the existing concepts to introduce some of statistical concepts that we have used in our problem setting. Kernel parameters can vary smoothly in the space and with growing demand for the explicit expression for a non-stationary spatial covariance function that allows local geometric non-smoothness, many models have been proposed. Anisotropy is the property of being directionally dependent which implies different properties in different directions and it is opposed to isotropy, although it may be helpful for more precise results to consider anisotropy in our algorithm setup, this work is postponed for later research and we do not choose a covariance function with different specification in different directions had we done we could have more conclusive results.

Reference [112] have discussed the use of non-stationary covariance models for Gaussian process regression. they discussed that as the regression models can have any number of regressors, there is a need for a non-stationary covariance function in arbitrary dimensions and for ease of computation and also interpretation of physical realities it is going to be more beneficial if the proposed non-stationary covariance matrix is expressed in explicit form, which does not need to be approximated or estimated for each data point and we can calculate the co-variance function for every point in our dataset if we have just decided to do so. [112] have described a method for generating explicit expressions for valid spatial covariance functions with locally varying geometric anisotropies, Although this approach does not allow one to vary other aspects of the covariance structure and to take into the account the specifications like differentiability of process or long distance dependence in the structure.

Spatial Statistics mainly concerns with real-world phenomenas whose spatial location is of great importance when it comes to make inference about that phenomena or setting statistical features for that phenomena. Probably the most infamous problem of this kind may be the Buffon's needle problem. Suppose a needle of length $x$ is thrown randomly onto a ground marked by parallel lines with a distance $d > x$ apart from each other. The question is that what is the problem that the needle crosses one of the parallel lines ? The hardship for finding a exact solution for this problem lies in the fact that one has to define what he/she means by "randomly" and in what sense he/she tries to solve the problem. Actually the mathematical solution for this problems could be find in *theory of point processes* which is one of main contributions of spatial statistics literature.

**Spatial Stochastic Processes** The spatial variable can vary over a continuous domain namely $\mathcal{D} \subset \mathcal{R}^d$, and the spatial dimension could be $d = 2, 3$. The spatial stochastic process $\mathcal{Y}(s) : s \in \mathcal{D} \subset \mathcal{R}^d$ has got an element of chance in itself as is defined as $\mathcal{Y}(s) = \mathcal{Y}(s, \omega)$ with $\omega \in \Omega$ is the element of chance in the spatial stochastic processes and $s$ is the spatial location, actually $\mathcal{Y}(s, w)$ is a collection of random variables with a well-defined joint distribution. At any single spatial location $s \in \mathcal{D}$, $\mathcal{Y}(s)$ is a random variable that can be written as $\mathcal{Y}(s, w)$ and $\omega$ belongs to some abstract sample space $\Omega$. The spatial stochastic random vector $(\mathcal{Y}(s_1), ..., \mathcal{Y}(s_n))^T$ is regarding to a finite set of spatial locations $\{s_1, ..., s_n\} \subset \mathcal{D}$ and the spatial dependencies can be represented by multivariate distributions of the variable of interests. If we consider the elementary event $\omega \in \Omega$ to be fixed, then $\mathcal{Y}(s, w)$ are realizations of the spatial stochastic process, always *observed data* are considered to be a *realization* of the certain spatial stochastic process.

In different applications the sample process remains unknown and abstract but as one may mention, it is of great importance to ensure we may have a valid and trustworthy mathematical specification of the spatial stochastic process. Often the distribution of the process is given by the corresponding collection of finite joint distribution of random variables in the variable space of the spatial stochastic process for every $n$ and every collection of sets $s_1, ..., s_n$ in the domain $\mathcal{D}$.

$$F(y_1, s_1; y_2, s_2; ....; y_n, s_n) = P(\mathcal{Y}(s_1) \leq y_1, ..., \mathcal{Y}(s_n) \leq y_n) \tag{B.1}$$

**Kolmogorov Existence Theorem** The *kolmogorov existence theorem* states that the stochastic process model, only would assumed to be valid if family of the finite-dimension joint distributions are consistent under marginalization and re-ordering. ***Gaussian processes*** which we will be encountered with in further parts of this thesis are an important special case of *kolmogorov theorem* where the finite-dimensional distributions are *multivariate normal* and thus can be characterized by their mean and covariance matrices. Also the consistency condition of the *kolmogorov theory* reduces to non-negativeness of the covariance matrix in Gaussian processes.

**Stationary Spatial Stochastic Processes** A spatial stochastic process is called *strictly stationary* if the finite dimensional joint distributions are invariant under spatial shifts, this definition implies for all vectors $\mathbf{h} \in \mathbb{R}$ that:

$$F(y_1, s_1 + \mathbf{h}; y_2, s_2 + \mathbf{h}; ....; y_n, s_n + \mathbf{h}) = F(y_1, s_1; y_2, s_2; ....; y_n, s_n) \quad \text{(B.2)}$$

For the Gaussian processes where their finite-dimensional distributions are only determined with their second-order moments, we can interprete the stationarity to be of like:

$$\mathbb{E}(\mathcal{Y}(s)) = \mathbb{E}(\mathcal{Y}(s+h)) = \mu \quad \text{(B.3)}$$

and

$$Cov(\mathcal{Y}(s), \mathcal{Y}(s+h)) = Cov(\mathcal{Y}(0), \mathcal{Y}(h)) = C(h) \quad \text{(B.4)}$$

A process, be it Gaussian or not, which satisfies these two conditions is called *weakly stationary* or *second-order stationary* process, it is obvious that a Gaussian process which is *weakly stationary* is also *strictly stationary*.

**The Nugget Effect** Often a spatial stochastic process is decomposed as $\mathcal{Y}(s) = \mu(s)+\eta(s)+\epsilon(s)$ where $\mu(s) = \mathbb{E}(\mathcal{Y}(s))$ is the mean function which is assumed to be deterministic and smooth and the process $\eta(s)$ is a assumed to have a zero mean and continuous realizations, and at alst $\epsilon(s)$ is a field of spatially uncorrelated zero-mean errors that is assumed to be independent of process $\eta$. The covariance function for error process is defined as follows:

$$Cov(\epsilon(s), \epsilon(s+h)) = \begin{cases} \sigma^2 \geq 0 & : h = 0 \\ 0 & : h \neq 0 \end{cases} \quad \text{(B.5)}$$

The *nugget effect* describes the situation in which the observational errors in potentially repeated measurements at any single site, or even to microscale variability, are occurring at such small scales that it cannot be distinguished from the effects of measurement errors. In our problem setup it is assumed that our prboblem's covariance function is continuous and is Gaussian, thus second-order spatial stochastic process.

**Bochner's Theorem** If we have a second-order stationary spatial stochastic process, for a given finite set of spatial locations $s_1, ..., s_n \in \mathbb{R}^d$ will have the covariance matrix of the finite dimensional joint distribution:

$$\begin{bmatrix} C(0) & C(s_1 - s_2) & \dots & C(s_1 - s_n) \\ C(s_2 - s_1) & C(0) & \dots & C(s_2 - s_n) \\ \vdots & \vdots & \ddots & \vdots \\ C(s_n - s_1) & C(s_n - s_2) & \dots & C(0) \end{bmatrix} \tag{B.6}$$

This matrix should be valid in the sense that it should be nonnegative for the spatial stochastic process to be valid. Actually the covariance function must be *positive definite*, also one can state that given any valid covariance function there exist a spatial Gaussian process with that covariance function.

**Isotropic Covariance Functions** Isotropic covariance functions are the covariance functions regarding to the second-order stationary spatial stochastic processes that only depends on the *spatial separation vector* **h** and only through its Euclidean length or $||\mathbf{h}||$. iso-covariance curves for an isotropic process are circles or spheres around the location that we are calculating covariance around them. *Geometrically anisotropic* processes are spatial stochastic processes which the iso-covariance curves are ellipsoids rather than circles or spheres. In the real-world applications, the *isotropy* assumption is frequently violated but the isotropic processes still construct fundamental part of the theory because they form the basic building blocks of more complex, anisotropic and non-stationary spatial stochastic process models. We can assume without loss of generality that for a standardized process the covariance function is equal to unity at the zero point. then for an isotropic covariance function we can write:

$$C(\mathbf{h}) = \phi(||\mathbf{h}||); \mathbf{h} \in \mathbb{R}^d \tag{B.7}$$

Where $\phi : [0, \infty) \to \mathbb{R}$ is a continuous function with $\phi(0) = 1$.

**Smoothness Properties** A spatial stochastic process is called ***mean square continuous*** if following relations holds:

$$\mathbb{E}(\mathcal{Y}(s) - \mathcal{Y}(s + h))^2 \to 0 \; as \; ||h|| \to 0 \tag{B.8}$$

Also for a second-order stationary process we would have:

$$\mathbb{E}(\mathcal{Y}(s) - \mathcal{Y}(s+h))^2 = 2(C(0) - C(h)) \tag{B.9}$$

We can conclude that mean square continuity is equivalent to the covariance function being continuous at the origin and thus everywhere. The important fact is that a process that is mean square continuous needs not to have continuous sample path and conversely.
For the isotropic processes, the properties of function $\phi$ translates to the properties of the associated Gaussian spatial process on $\mathbb{R}^d$, specifically the behavior of $\phi(t)$ at the origin determines the smoothness of the sample path, which is of great important particularly in our field of research which relies on the prediction problems.

**Building Blocks For Non-Stationary Processes** Often in order to manipulate data that are sampled regarding to spatial specifications, we would need to model the spatial dependencies and structures of environmental processes. Also we saw that we almost decompose spatial stochastic processes into a *residual* and a *mean* part, actually we can write the process as $\mathcal{Y}(s) = \mu(s) + \eta(s)$. Also we often assume that our process is a second-order stationary one so that the covariance between any two locations only depends on the spatial distance vector connecting them. The important fact relies in the fact that we can assume all the spatially stochastic processes to be approximately stationary over relatively small or so called *locally* spatial regions which often have anisotropic covariance structure.

**Smoothing And Kernel Based Methods** The simplest approach for dealing with non-stationary spatial covariance function structures can be considered to be the perspective of locally stationary models, which are empirically smoothed over the space or from a perspective that, we can assume to have empirical covariances estimated among a finite number of desired sites and calculating the interpolation between them.

# Appendix C

# Gradient Evaluation

## Gradient Evaluation For Input-dependent Smoothness Method

### Grardient of $L(\theta)$:

In order to implement gradient-based optimization algorithms we shall obtain gradients of our log marginal likelihood objective function $L(\theta)$ analytically which turns out to be of the form below:

$$\begin{aligned}
\frac{\partial L(\theta)}{\partial *} &= \frac{\partial}{\partial *}\Bigg[-\frac{1}{2}\sum_{m=1}^{M}\Big((z^m)^T A_m^{-1} z^m\Big) - \frac{1}{2}\log\Big(\prod_{m=1}^{M}|A_m|\Big) \\
&\quad -\frac{1}{2}\log\Big(\prod_{m=1}^{M}|B_m|\Big) - \frac{\log(2\pi)}{2}\Big(\sum_{m=1}^{M} n_m\Big)\Bigg] \\
&= \frac{\partial}{\partial *}\Bigg[-\frac{1}{2}\sum_{m=1}^{M}\Big((z^m)^T A_m^{-1} z^m\Big) - \frac{1}{2}\log\Big(\prod_{m=1}^{M}|A_m|\Big) \\
&\quad -\frac{1}{2}\log\Big(\prod_{m=1}^{M}|B_m|\Big)\Bigg] \\
&= \frac{1}{2}\sum_{m=1}^{M}\Big((z^m)^T A_m^{-1}\frac{\partial A_m}{\partial *}A_m^{-1} z^m\Big) - \frac{1}{2}\sum_{m=1}^{M}\mathbf{tr}\Big(A_m^{-1}\frac{\partial A_m}{\partial *}\Big) \\
&\quad -\frac{1}{2}\sum_{m=1}^{M}\mathbf{tr}\Big(B_m^{-1}\frac{\partial B_m}{\partial *}\Big)
\end{aligned} \tag{C.1}$$

It is then obvious that if we calculate the $\frac{\partial A_m}{\partial *}$ and $\frac{\partial B_b}{\partial *}$ for all divided segments, all the remaining calculations seems to be straight forward.

## Gradient of matrices $A_m$ and $B_m$ And $L(\theta)$ w.r.t $\sigma_n$

Having explicit equations for covariance kernels and also $A_m$ and $B_m$ matrices, we can get :

$$\frac{\partial A_m}{\partial \sigma_n} = \frac{\partial(K_m(r,r) + \sigma_n^2 I)}{\partial \sigma_n} = 2\sigma_n I \tag{C.2}$$

$$\frac{\partial B_m}{\partial \sigma_n} = \frac{\partial(K_m(\bar{r},\bar{r}) + \bar{\sigma}_n^2 I)}{\partial \sigma_n} = 0 \tag{C.3}$$

Thus we can calculate $\frac{\partial L(\theta)}{\partial \sigma_n}$ as follows:

$$\begin{aligned}
\frac{\partial L(\theta)}{\partial \sigma_n} &= \frac{1}{2}\sum_{m=1}^{M}\left((z^m)^T A_m^{-1}\frac{\partial A_m}{\partial \sigma_n}A_m^{-1}z^m\right) - \frac{1}{2}\sum_{m=1}^{M}\textbf{tr}\left(A_m^{-1}\frac{\partial A_m}{\partial \sigma_n}\right) \\
&\quad -\frac{1}{2}\sum_{m=1}^{M}\textbf{tr}\left(B_m^{-1}\frac{\partial B_m}{\partial \sigma_n}\right) \\
&= \frac{1}{2}\sum_{m=1}^{M}\left((z^m)^T A_m^{-1}2\sigma_n I A_m^{-1}z^m\right) - \frac{1}{2}\sum_{m=1}^{M}\textbf{tr}\left(A_m^{-1}2\sigma_n I\right) \\
&\quad -\frac{1}{2}\sum_{m=1}^{M}\textbf{tr}\left(B_m^{-1}\times 0\right) \\
&= \frac{1}{2}\sum_{m=1}^{M}\left((z^m)^T A_m^{-1}2\sigma_n I A_m^{-1}z^m\right) - \frac{1}{2}\sum_{m=1}^{M}\textbf{tr}\left(A_m^{-1}2\sigma_n I\right)
\end{aligned} \tag{C.4}$$

## Gradient of matrices $A_m$ and $B_m$ And $L(\theta)$ w.r.t $\sigma_f$

Again we write the explicit equations that we have as follows:

$$\frac{\partial A_m}{\partial \sigma_f} = \frac{\partial(K_m(r,r) + \sigma_n^2 I)}{\partial \sigma_f} = 2\sigma_f K_m(r,r) \tag{C.5}$$

$$\frac{\partial B_m}{\partial \sigma_f} = \frac{\partial(K_m(\bar{r},\bar{r}) + \bar{\sigma}_n^2 I)}{\partial \sigma_f} = 0 \tag{C.6}$$

Thus we may calculate $\frac{\partial L(\theta)}{\partial \sigma_f}$ as follows:

$$\begin{aligned}
\frac{\partial L(\theta)}{\partial \sigma_f} &= \frac{1}{2}\sum_{m=1}^{M}\Bigg((z^m)^T A_m^{-1}\frac{\partial A_m}{\partial \sigma_f}A_m^{-1}z^m\Bigg) - \frac{1}{2}\sum_{m=1}^{M}\mathbf{tr}\Bigg(A_m^{-1}\frac{\partial A_m}{\partial \sigma_f}\Bigg) \\
&\quad -\frac{1}{2}\sum_{m=1}^{M}\mathbf{tr}\Bigg(B_m^{-1}\frac{\partial B_m}{\partial \sigma_f}\Bigg) \\
&= \frac{1}{2}\sum_{m=1}^{M}\Bigg((z^m)^T A_m^{-1}2\sigma_f K_m(r,r)A_m^{-1}z^m\Bigg) - \frac{1}{2}\sum_{m=1}^{M}\mathbf{tr}\Bigg(A_m^{-1}2\sigma_f K_m(r,r)\Bigg) \\
&\quad -\frac{1}{2}\sum_{m=1}^{M}\mathbf{tr}\Bigg(B_m^{-1}\times 0\Bigg) \\
&= \frac{1}{2}\sum_{m=1}^{M}\Bigg((z^m)^T A_m^{-1}2\sigma_f K_m(r,r)A_m^{-1}z^m\Bigg) - \frac{1}{2}\sum_{m=1}^{M}\mathbf{tr}\Bigg(A_m^{-1}2\sigma_f K_m(r,r)\Bigg)
\end{aligned} \tag{C.7}$$

# Gradient of matrices $A_m$ and $B_m$ And $L(\theta)$ w.r.t $\bar{\mathcal{L}}$

$$\frac{\partial A_m}{\partial \bar{\mathcal{L}}} = \frac{\partial (K_m(r,r)+\sigma_n^2 I)}{\partial \bar{\mathcal{L}}} = \frac{\partial K_m(r,r)}{\partial \bar{\mathcal{L}}} \tag{C.8}$$

$$\frac{\partial B_m}{\partial \bar{\mathcal{L}}} = \frac{\partial (K_m(\bar{r},\bar{r})+\bar{\sigma}_n^2 I)}{\partial \bar{\mathcal{L}}} = 0 \tag{C.9}$$

Thus we may calculate $\frac{\partial L(\theta)}{\partial \bar{\mathcal{L}}}$ as follows:

$$\begin{aligned}
\frac{\partial L(\theta)}{\partial \bar{\mathcal{L}}} &= \frac{1}{2}\sum_{m=1}^{M}\Bigg((z^m)^T A_m^{-1}\frac{\partial A_m}{\partial \bar{\mathcal{L}}}A_m^{-1}z^m\Bigg) - \frac{1}{2}\sum_{m=1}^{M}\mathbf{tr}\Bigg(A_m^{-1}\frac{\partial A_m}{\partial \bar{\mathcal{L}}}\Bigg) \\
&\quad -\frac{1}{2}\sum_{m=1}^{M}\mathbf{tr}\Bigg(B_m^{-1}\frac{\partial B_m}{\partial \bar{\mathcal{L}}}\Bigg) \\
&= \frac{1}{2}\sum_{m=1}^{M}\Bigg((z^m)^T A_m^{-1}\frac{\partial K_m(r,r)}{\partial \bar{\mathcal{L}}}A_m^{-1}z^m\Bigg) - \frac{1}{2}\sum_{m=1}^{M}\mathbf{tr}\Bigg(A_m^{-1}\frac{\partial K_m(r,r)}{\partial \bar{\mathcal{L}}}\Bigg) \\
&\quad -\frac{1}{2}\sum_{m=1}^{M}\mathbf{tr}\Bigg(B_m^{-1}\times 0\Bigg) \\
&= \frac{1}{2}\sum_{m=1}^{M}\Bigg((z^m)^T A_m^{-1}\frac{\partial K_m(r,r)}{\partial \bar{\mathcal{L}}}A_m^{-1}z^m\Bigg) - \frac{1}{2}\sum_{m=1}^{M}\mathbf{tr}\Bigg(A_m^{-1}\frac{\partial K_m(r,r)}{\partial \bar{\mathcal{L}}}\Bigg)
\end{aligned} \tag{C.10}$$

Now we must calculate the term $\frac{\partial K_m(r,r)}{\partial \bar{\mathcal{L}}}$ in order to be able to calculate the gradient:

$$
\begin{aligned}
&\frac{\partial K_m(r,r)}{\partial \bar{\mathcal{L}}} = \sigma_f^2(\frac{1}{2})^{-\frac{1}{2}}\Bigg[\\
&\Bigg(\frac{\partial\Big[\mathcal{L}^T\mathcal{L}I_n^T\Big]^{\frac{1}{4}}}{\partial \bar{\mathcal{L}}}\Bigg)\Big[I_n^T\mathcal{L}^T\mathcal{L}\Big]^{\frac{1}{4}}\Big[\mathcal{L}^T\mathcal{L}I_n^T + I_n^T\mathcal{L}^T\mathcal{L}\Big]^{\frac{-1}{2}}\exp\Bigg[\frac{-s(R)}{[\mathcal{L}^T\mathcal{L}I_n^T + I_n^T\mathcal{L}^T\mathcal{L}]}\Bigg]\\
&+\Big[\mathcal{L}^T\mathcal{L}I_n^T\Big]^{\frac{1}{4}}\Bigg(\frac{\partial\Big[I_n^T\mathcal{L}^T\mathcal{L}\Big]^{\frac{1}{4}}}{\partial \bar{\mathcal{L}}}\Bigg)\Big[\mathcal{L}^T\mathcal{L}I_n^T + I_n^T\mathcal{L}^T\mathcal{L}\Big]^{\frac{-1}{2}}\exp\Bigg[\frac{-s(R)}{[\mathcal{L}^T\mathcal{L}I_n^T + I_n^T\mathcal{L}^T\mathcal{L}]}\Bigg]\\
&+\Big[\mathcal{L}^T\mathcal{L}I_n^T\Big]^{\frac{1}{4}}\Big[I_n^T\mathcal{L}^T\mathcal{L}\Big]^{\frac{1}{4}}\Bigg(\frac{\partial\Big[\mathcal{L}^T\mathcal{L}I_n^T + I_n^T\mathcal{L}^T\mathcal{L}\Big]^{\frac{-1}{2}}}{\partial \bar{\mathcal{L}}}\Bigg)\exp\Bigg[\frac{-s(R)}{[\mathcal{L}^T\mathcal{L}I_n^T + I_n^T\mathcal{L}^T\mathcal{L}]}\Bigg]\\
&+\Big[\mathcal{L}^T\mathcal{L}I_n^T\Big]^{\frac{1}{4}}\Big[I_n^T\mathcal{L}^T\mathcal{L}\Big]^{\frac{1}{4}}\Big[\mathcal{L}^T\mathcal{L}I_n^T + I_n^T\mathcal{L}^T\mathcal{L}\Big]^{\frac{-1}{2}}\Bigg(\frac{\partial\exp\Big[\frac{-s(R)}{[\mathcal{L}^T\mathcal{L}I_n^T + I_n^T\mathcal{L}^T\mathcal{L}]}\Big]}{\partial \bar{\mathcal{L}}}\Bigg)\Bigg]
\end{aligned}
\tag{C.11}
$$

First we write the relationship between $\mathcal{L}_i$ and $\bar{\mathcal{L}}_i$ as $\mathcal{L}_i = \tau(\bar{\mathcal{L}}) = \exp[\eta_i\bar{\mathcal{L}}]$ with $\eta_i = (\bar{K}(r_i,\bar{r}))^T[\bar{K}(\bar{r},\bar{r}) + \bar{\sigma}_n^2 I]^{-1}$ which is calculated for each data point $i$, thus we can calculate the first partial derivarive. Note that operations in above equations are element-wise, therefore obtained expressions for derivatives are element-wise too, but we should make sure that we correctly consider matrix multiplication rules:

$$
\begin{aligned}
\frac{\partial\Big[\mathcal{L}_i^T\mathcal{L}_i\Big]^{\frac{1}{4}}}{\partial\bar{\mathcal{L}}_i} &= \frac{1}{4}\times\Big(\mathcal{L}_i^T\mathcal{L}_i\Big)^{-\frac{3}{4}}\times\frac{\partial\Big[\mathcal{L}_i^T\mathcal{L}_i\Big]}{\partial\bar{\mathcal{L}}}\\
&= \frac{1}{4}\times\Big(\mathcal{L}^T\mathcal{L}1_n^2\Big)^{-\frac{3}{4}}\times 2\mathcal{L}1_n^2\times\frac{\partial\big(\exp[\eta\bar{\mathcal{L}}]\big)}{\partial\bar{\mathcal{L}}}\\
&= \frac{1}{4}\times\Big(\mathcal{L}^T\mathcal{L}1_n^2\Big)^{-\frac{3}{4}}\times 2\mathcal{L}\times\exp[\eta_i\bar{\mathcal{L}}]\times\eta_i^T
\end{aligned}
\tag{C.12}
$$

Also the third partial derivative will be of the form:

$$
\begin{aligned}
\frac{\partial\left[\mathcal{L}_i^T\mathcal{L}_i + \mathcal{L}_j^T\mathcal{L}_j\right]^{\frac{-1}{2}}}{\partial\bar{\mathcal{L}}} &= -\frac{1}{2}\times(\mathcal{L}_i^T\mathcal{L}_i + \mathcal{L}_j^T\mathcal{L}_j)^{-\frac{3}{2}} \\
&\quad \times\frac{\partial}{\partial\bar{\mathcal{L}}}(\mathcal{L}_i^T\mathcal{L}_i + \mathcal{L}_j^T\mathcal{L}_j) \\
&= -\frac{1}{2}\times(\mathcal{L}_i^T\mathcal{L}_i + \mathcal{L}_j^T\mathcal{L}_j)^{-\frac{3}{2}} \\
&\quad \times\left(2\mathcal{L}_i\exp[\eta_i\bar{\mathcal{L}}]\eta_i^T + 2\exp[\eta_j\bar{\mathcal{L}}]\eta_j^T\right)
\end{aligned} \tag{C.13}
$$

The last partial derivative:

$$
\begin{aligned}
\frac{\partial\exp\left[\frac{-s(R)}{[\mathcal{L}_i^T\mathcal{L}_i 1_n^T + 1_n^T\mathcal{L}_j^T\mathcal{L}_j]}\right]}{\partial\bar{\mathcal{L}}} &= \exp\left[\frac{-s(R)}{[\mathcal{L}_i^T\mathcal{L}_i 1_n^T + 1_n^T\mathcal{L}_j^T\mathcal{L}_j]}\right] \\
&\quad \times\frac{\partial}{\partial\bar{\mathcal{L}}}\left(\frac{-s(R)}{[\mathcal{L}_i^T\mathcal{L}_i 1_n^T + 1_n^T\mathcal{L}_j^T\mathcal{L}_j]}\right) \\
&= \exp\left[\frac{-s(R)}{[\mathcal{L}_i^T\mathcal{L}_i 1_n^T + 1_n^T\mathcal{L}_j^T\mathcal{L}_j]}\right] \\
&\quad \times\left(-S(R) - (\mathcal{L}_i^T\mathcal{L}_i 1_n^T + 1_n^T\mathcal{L}_j^T\mathcal{L}_j)^{-2}\right) \\
&\quad \times\frac{\partial}{\partial\bar{\mathcal{L}}}\left(\mathcal{L}_i^T\mathcal{L}_i 1_n^T + 1_n^T\mathcal{L}_j^T\mathcal{L}_j\right) \\
&= \exp\left[\frac{-s(R)}{[\mathcal{L}_i^T\mathcal{L}_i 1_n^T + 1_n^T\mathcal{L}_j^T\mathcal{L}_j]}\right] \\
&\quad \times\left(-S(R)\times-(\mathcal{L}_i^T\mathcal{L}_i 1_n^T + 1_n^T\mathcal{L}_j^T\mathcal{L}_j)^{-2}\right) \\
&\quad \times\left(2\mathcal{L}_i\exp[\eta_i\bar{\mathcal{L}}]\eta_i^T + 2\exp[\eta_j\bar{\mathcal{L}}]\eta_j^T\right)
\end{aligned} \tag{C.14}
$$

After finding corresponding gradients for hyper-parameters regarding the $\mathcal{GP}_z$, we have to find the gradients regarding to latent process $\mathcal{GP}_l$ with respect to the hyper-parameters $\{\bar{\sigma}_f, \bar{\sigma}_l, \bar{\sigma_n}\}$.

## Gradient of matrices $A_m$ and $B_m$ And $L(\theta)$ w.r.t $\bar{\sigma}_f$

For the $\bar{\sigma}_f$, the gradient of $A$ with respect to it wil be:

$$\frac{\partial A}{\partial \bar{\sigma}_f} = \frac{\partial (K(r,r) + \sigma_n^2 I)}{\partial \bar{\sigma}_f} = 0 \tag{C.15}$$

For the $\bar{\sigma}_f$, the gradient of $B$ with respect to it wil be:

$$\frac{\partial B}{\partial \bar{\sigma}_f} = \frac{\partial (K(\bar{r},\bar{r}) + \bar{\sigma}_n^2 I)}{\partial \bar{\sigma}_f} = 2\bar{\sigma}_f \exp\bigg( - \frac{1}{2} s(\bar{\sigma}_l^{-2} \bar{R}^2)\bigg) \tag{C.16}$$

Thus we may calculate $\frac{\partial L(\theta)}{\partial \bar{\sigma}_f}$ as follows:

$$\begin{aligned}
\frac{\partial L(\theta)}{\partial \bar{\sigma}_f} &= \frac{1}{2}\sum_{m=1}^{M}\bigg((z^m)^T A_m^{-1}\frac{\partial A_m}{\partial \bar{\sigma}_f} A_m^{-1} z^m\bigg) \\
&\quad -\frac{1}{2}\sum_{m=1}^{M}\mathbf{tr}\bigg(A_m^{-1}\frac{\partial A_m}{\partial \bar{\sigma}_f}\bigg) - \frac{1}{2}\sum_{m=1}^{M}\mathbf{tr}\bigg(B_m^{-1}\frac{\partial B_m}{\partial \bar{\sigma}_f}\bigg) \\
&= \frac{1}{2}\sum_{m=1}^{M}\bigg((z^m)^T A_m^{-1} \times 0 \times A_m^{-1} z^m\bigg) - \frac{1}{2}\sum_{m=1}^{M}\mathbf{tr}\bigg(A_m^{-1} \times 0\bigg) \\
&\quad -\frac{1}{2}\sum_{m=1}^{M}\mathbf{tr}\bigg(B_m^{-1} 2\bar{\sigma}_f \exp(-\frac{1}{2} s(\bar{\sigma}_l^{-2}\bar{R}^2))\bigg) \\
&= -\frac{1}{2}\sum_{m=1}^{M}\mathbf{tr}\bigg(B_m^{-1} 2\bar{\sigma}_f \exp(-\frac{1}{2} s(\bar{\sigma}_l^{-2}\bar{R}^2))\bigg)
\end{aligned} \tag{C.17}$$

## Gradient of matrices $A_m$ and $B_m$ And $L(\theta)$ w.r.t $\bar{\sigma}_l$

For the $\bar{\sigma}_l$, the gradient of $A_m$ with respect to it wil be:

$$\frac{\partial A_m}{\partial \bar{\sigma}_l} = \frac{\partial (K_m(r,r) + \sigma_n^2 I)}{\partial \bar{\sigma}_l} = 0 \tag{C.18}$$

For the $\bar{\sigma}_l$, the gradient of $B_m$ with respect to it wil be:

$$\begin{aligned}
\frac{\partial B_m}{\partial \bar{\sigma}_l} &= \frac{\partial (K_m(\bar{r},\bar{r}) + \bar{\sigma}_n^2 I)}{\partial \bar{\sigma}_l} \\
&= \frac{\partial \Big[ \bar{\sigma}_f^2 \exp\Big( -\frac{1}{2} s(\bar{\sigma}_l^{-2}(\bar{R}^m)^2) \Big) + \bar{\sigma}_n^2 I \Big]}{\partial \bar{\sigma}_l} \\
&= \bar{\sigma}_f^2 \frac{\partial \Big[ \exp\Big( -\frac{1}{2} s(\bar{\sigma}_l^{-2}(\bar{R}^m)^2) \Big) \Big]}{\partial \bar{\sigma}_l} \\
&= \bar{\sigma}_f^2 \frac{\partial \Big[ \exp\Big( -\frac{1}{2} \frac{(\bar{r}_i^m - \bar{r_j}^m)^2}{\bar{\sigma}_l^2} \Big) \Big]}{\partial \bar{\sigma}_l} \\
&= \frac{(\bar{r}_i^m - \bar{r_j}^m)^2 \bar{\sigma}_f^2 \Big[ \exp\Big( -\frac{1}{2} \frac{(\bar{r}_i^m - \bar{r_j}^m)^2}{\bar{\sigma}_l^2} \Big) \Big]}{\bar{\sigma}_l^3} \qquad \text{(C.19)}
\end{aligned}$$

Thus we may calculate $\frac{\partial L(\theta)}{\partial \bar{\sigma}_l}$ as follows:

$$\begin{aligned}
\frac{\partial L(\theta)}{\partial \bar{\sigma}_f} &= \frac{1}{2} \sum_{m=1}^{M} \Big( (z^m)^T A_m^{-1} \frac{\partial A_m}{\partial \bar{\sigma}_f} A_m^{-1} z^m \Big) - \frac{1}{2} \sum_{m=1}^{M} \mathbf{tr} \Big( A_m^{-1} \frac{\partial A_m}{\partial \bar{\sigma}_f} \Big) \\
&\quad - \frac{1}{2} \sum_{m=1}^{M} \mathbf{tr} \Big( B_m^{-1} \frac{\partial B_m}{\partial \bar{\sigma}_f} \Big) \\
&= \frac{1}{2} \sum_{m=1}^{M} \Big( (z^m)^T A_m^{-1} \times 0 \times A_m^{-1} z^m \Big) - \frac{1}{2} \sum_{m=1}^{M} \mathbf{tr} \Big( A_m^{-1} \times 0 \Big) \\
&\quad - \frac{1}{2} \sum_{m=1}^{M} \mathbf{tr} \Big( B_m^{-1} \frac{\partial B_m}{\partial \bar{\sigma}_f} \Big) \\
&= -\frac{1}{2} \sum_{m=1}^{M} \mathbf{tr} \Bigg( B_m^{-1} \frac{(\bar{r}_i^m - \bar{r_j}^m)^2 \bar{\sigma}_f^2 \exp\Big( -\frac{1}{2} \frac{(\bar{r}_i^m - \bar{r_j}^m)^2}{\bar{\sigma}_l^2} \Big)}{\bar{\sigma}_l^3} \Bigg) \\
&= -\frac{1}{2} \sum_{m=1}^{M} \mathbf{tr} \Bigg( B_m^{-1} \frac{(\bar{R}^m)^2 \bar{\sigma}_f^2 \exp\Big( -\frac{1}{2} \frac{(\bar{R}^m)^2}{\bar{\sigma}_l^2} \Big)}{\bar{\sigma}_l^3} \Bigg) \qquad \text{(C.20)}
\end{aligned}$$

### Gradient of matrices $A_m$ and $B_m$ And $L(\theta)$ w.r.t $\bar{\sigma}_n$

For the $\bar{\sigma}_n$, the gradient of $A_m$ with respect to it wil be:

$$\frac{\partial A_m}{\partial \bar{\sigma}_n} = \frac{\partial (K_m(r,r) + \sigma_n^2 I)}{\partial \bar{\sigma_n}} = 0 \tag{C.21}$$

For the $\bar{\sigma}_n$, the gradient of $B_m$ with respect to it wil be:

$$\frac{\partial B_m}{\partial \bar{\sigma}_n} = \frac{\partial (K_m(\bar{r},\bar{r}) + \bar{\sigma}_n^2 I)}{\partial \bar{\sigma_n}} = 2\bar{\sigma}_n I \tag{C.22}$$

Thus we may calculate $\frac{\partial L(\theta)}{\partial \bar{\sigma}_l}$ as follows:

$$\begin{aligned}
\frac{\partial L(\theta)}{\partial \bar{\sigma}_f} &= \frac{1}{2}\sum_{m=1}^{M}\left((z^m)^T A_m^{-1}\frac{\partial A_m}{\partial \bar{\sigma}_f}A_m^{-1}z^m\right) - \frac{1}{2}\sum_{m=1}^{M}\textbf{tr}\left(A_m^{-1}\frac{\partial A_m}{\partial \bar{\sigma}_f}\right) \\
&\quad -\frac{1}{2}\sum_{m=1}^{M}\textbf{tr}\left(B_m^{-1}\frac{\partial B_m}{\partial \bar{\sigma}_f}\right) \\
&= \frac{1}{2}\sum_{m=1}^{M}\left((z^m)^T A_m^{-1}\times 0\times A_m^{-1}z^m\right) - \frac{1}{2}\sum_{m=1}^{M}\textbf{tr}\left(A_m^{-1}\times 0\right) \\
&\quad -\frac{1}{2}\sum_{m=1}^{M}\textbf{tr}\left(B_m^{-1}2\bar{\sigma}_n I\right) \\
&= -\frac{1}{2}\sum_{m=1}^{M}\textbf{tr}\left(B_m^{-1}2\bar{\sigma}_n I\right)
\end{aligned} \tag{C.23}$$

Having these explicit expressions for gradient of log-marginal likelihood with respect to the independent hyper-parameters, enables us to run our gradient-based optimizations algorithm to find the best MAP value for our hyper-parameters, and this in turn enables us to predict the ground model for each frame and in each segment.

## Gradient Evaluation For Input-dependent Noise Method

### Grardient of $L_{\mathcal{I}}(\theta)$ w.r.t $*$ :

In order to implement gradient-based optimization algorithms we shall obtain gradients of our new log marginal likelihood objective function $L_{\mathcal{I}}(\theta)$ analytically which turns out to be of the form below:

$$\frac{\partial L(\theta)}{\partial *} = \frac{\partial}{\partial *}\Bigg[ -\frac{1}{2}\sum_{m=1}^{M}\left((z^m)^T A_{\mathcal{I}_m}^{-1}z^m\right) - \frac{1}{2}\log\left(\prod_{m=1}^{M}|A_{\mathcal{I}_m}|\right)$$

$$-\frac{1}{2}\log\Bigg(\prod_{m=1}^{M}|B_{\mathcal{I}_m}|\Bigg)-\frac{1}{2}\log\Bigg(\prod_{m=1}^{M}|C_{\mathcal{I}_m}|\Bigg)-\frac{3\log(2\pi)}{2}\Bigg(\sum_{m=1}^{M}n_m\Bigg)\Bigg]$$

$$=\frac{\partial}{\partial *}\Bigg[-\frac{1}{2}\sum_{m=1}^{M}\Bigg((z^m)^T A_{\mathcal{I}_m}^{-1}z^m\Bigg)-\frac{1}{2}\log\Bigg(\prod_{m=1}^{M}|A_{\mathcal{I}_m}|\Bigg)$$

$$-\frac{1}{2}\log\Bigg(\prod_{m=1}^{M}|B_{\mathcal{I}_m}|\Bigg)-\frac{1}{2}\log\Bigg(\prod_{m=1}^{M}|C_{\mathcal{I}_m}|\Bigg)\Bigg]$$

$$=\frac{1}{2}\sum_{m=1}^{M}\Bigg((z^m)^T A_{\mathcal{I}_m}^{-1}\frac{\partial A_{\mathcal{I}_m}}{\partial *}A_m^{-1}z^m\Bigg)-\frac{1}{2}\sum_{m=1}^{M}\textbf{tr}\Bigg(A_{\mathcal{I}_m}^{-1}\frac{\partial A_{\mathcal{I}_m}}{\partial *}\Bigg)$$

$$-\frac{1}{2}\sum_{m=1}^{M}\textbf{tr}\Bigg(B_{\mathcal{I}_m}^{-1}\frac{\partial B_{\mathcal{I}_m}}{\partial *}\Bigg)-\frac{1}{2}\sum_{m=1}^{M}\textbf{tr}\Bigg(C_{\mathcal{I}_m}^{-1}\frac{\partial C_{\mathcal{I}_m}}{\partial *}\Bigg) \tag{C.24}$$

It is then obvious that if we calculate the $\frac{\partial A_{\mathcal{I}_m}}{\partial *}$ and $\frac{\partial B_{\mathcal{I}_m}}{\partial *}$ for all divided segments, all the remaining calculations seems to be straight forward. For the hyper-parameters of $\mathcal{GP}_z$, $\mathcal{GP}_l$ and $\mathcal{GP}_n$ the gradients will be calculated with respect to all the hyper parameters represented by $\theta_{id}=\{\sigma_f,\bar{\mathcal{L}},\bar{\mathcal{I}},\bar{\sigma}_f,\bar{\sigma}_l,\bar{\sigma}_n,\bar{\bar{\sigma}}_f,\bar{\bar{\sigma}}_l,\bar{\bar{\sigma}}_n\}$.

## Gradient of matrices $A_{\mathcal{I}_m}$, $B_{\mathcal{I}_m}$, $C_{\mathcal{I}_m}$ And $L_{\mathcal{I}}(\theta)$ w.r.t $\sigma_f$

:

Again we write the explicit equations that we have as follows:

$$\frac{\partial A_{\mathcal{I}_m}}{\partial \sigma_f}=\frac{\partial(K^m(r,r)+(\mathcal{I}^m)^2 I)}{\partial \sigma_f}=2\sigma_f K^m(r,r) \tag{C.25}$$

$$\frac{\partial B_{\mathcal{I}_m}}{\partial \sigma_f}=\frac{\partial(\bar{K}_m(\bar{r},\bar{r})+\bar{\sigma}_n^2 I)}{\partial \sigma_f}=0 \tag{C.26}$$

$$\frac{\partial C_{\mathcal{I}_m}}{\partial \sigma_f}=\frac{\partial(\bar{\bar{K}}_{\mathcal{I}}^m(\bar{r}_{\mathcal{I}},\bar{r}_{\mathcal{I}})+\bar{\bar{\sigma}}_n^2 I)}{\partial \sigma_f}=0 \tag{C.27}$$

Thus we may calculate $\frac{\partial L_{\mathcal{I}}(\theta)}{\partial \sigma_f}$ as follows:

$$\frac{\partial L_{\mathcal{I}}(\theta)}{\partial \sigma_f}=\frac{1}{2}\sum_{m=1}^{M}\Bigg((z^m)^T A_{\mathcal{I}_m}^{-1}\frac{\partial A_{\mathcal{I}_m}}{\partial \sigma_f}A_{\mathcal{I}_m}^{-1}z^m\Bigg)-\frac{1}{2}\sum_{m=1}^{M}\textbf{tr}\Bigg(A_{\mathcal{I}_m}^{-1}\frac{\partial A_{\mathcal{I}_m}}{\partial \sigma_f}\Bigg)$$

$$-\frac{1}{2}\sum_{m=1}^{M}\textbf{tr}\Big(B_{\mathcal{I}_m}^{-1}\frac{\partial B_{\mathcal{I}_m}}{\partial \sigma_f}\Big)-\frac{1}{2}\sum_{m=1}^{M}\textbf{tr}\Big(C_{\mathcal{I}_m}^{-1}\frac{\partial C_{\mathcal{I}_m}}{\partial \sigma_f}\Big)$$

$$=\frac{1}{2}\sum_{m=1}^{M}\Big((z^m)^T A_{\mathcal{I}_m}^{-1} 2\sigma_f K^m(r,r) A_{\mathcal{I}_m}^{-1} z^m\Big)-\frac{1}{2}\sum_{m=1}^{M}\textbf{tr}\Big(A_{\mathcal{I}_m}^{-1} 2\sigma_f K^m(r,r)\Big)$$

$$-\frac{1}{2}\sum_{m=1}^{M}\textbf{tr}\Big(B_{\mathcal{I}_m}^{-1}\times 0\Big)-\frac{1}{2}\sum_{m=1}^{M}\textbf{tr}\Big(C_{\mathcal{I}_m}^{-1}\times 0\Big)$$

$$\frac{\partial L_{\mathcal{I}}(\theta)}{\partial \sigma_f}=\frac{1}{2}\sum_{m=1}^{M}\Big((z^m)^T A_{\mathcal{I}_m}^{-1} 2\sigma_f K^m(r,r) A_{\mathcal{I}_m}^{-1} z^m\Big)-\frac{1}{2}\sum_{m=1}^{M}\textbf{tr}\Big(A_{\mathcal{I}_m}^{-1} 2\sigma_f K^m(r,r)\Big) \tag{C.28}$$

## Gradient of matrices $A_{\mathcal{I}_m}$, $B_{\mathcal{I}_m}$, $C_{\mathcal{I}_m}$ And $L_{\mathcal{I}}(\theta)$ w.r.t $\bar{\mathcal{L}}$

:

$$\frac{\partial A_{\mathcal{I}_m}}{\partial \bar{\mathcal{L}}}=\frac{\partial(K^m(r,r)+(\mathcal{I}^m)^2 I)}{\partial \bar{\mathcal{L}}}=\frac{\partial K^m(r,r)}{\partial \bar{\mathcal{L}}} \tag{C.29}$$

$$\frac{\partial B_{\mathcal{I}_m}}{\partial \bar{\mathcal{L}}}=\frac{\partial(\bar{K}^m(\bar{r},\bar{r})+\bar{\sigma}_n^2 I)}{\partial \bar{\mathcal{L}}}=0 \tag{C.30}$$

$$\frac{\partial C_{\mathcal{I}_m}}{\partial \sigma_f}=\frac{\partial(\bar{K}_{\mathcal{I}}^m(\bar{r}_{\mathcal{I}},\bar{r}_{\mathcal{I}})+\bar{\bar{\sigma}}_n^2 I)}{\partial \sigma_f}=0 \tag{C.31}$$

Thus we may calculate $\frac{\partial L_{\mathcal{I}}(\theta)}{\partial \sigma_f}$ as follows:

$$\frac{\partial L_{\mathcal{I}}(\theta)}{\partial \sigma_f}=\frac{1}{2}\sum_{m=1}^{M}\Big((z^m)^T A_{\mathcal{I}_m}^{-1}\frac{\partial A_{\mathcal{I}_m}}{\partial \sigma_f} A_{\mathcal{I}_m}^{-1} z^m\Big)-\frac{1}{2}\sum_{m=1}^{M}\textbf{tr}\Big(A_{\mathcal{I}_m}^{-1}\frac{\partial A_{\mathcal{I}_m}}{\partial \sigma_f}\Big)$$

$$-\frac{1}{2}\sum_{m=1}^{M}\textbf{tr}\Big(B_{\mathcal{I}_m}^{-1}\frac{\partial B_{\mathcal{I}_m}}{\partial \sigma_f}\Big)-\frac{1}{2}\sum_{m=1}^{M}\textbf{tr}\Big(C_{\mathcal{I}_m}^{-1}\frac{\partial C_{\mathcal{I}_m}}{\partial \sigma_f}\Big)$$

$$=\frac{1}{2}\sum_{m=1}^{M}\Big((z^m)^T A_{\mathcal{I}_m}^{-1}\frac{\partial K^m(r,r)}{\partial \bar{\mathcal{L}}} A_{\mathcal{I}_m}^{-1} z^m\Big)-\frac{1}{2}\sum_{m=1}^{M}\textbf{tr}\Big(A_{\mathcal{I}_m}^{-1}\frac{\partial K^m(r,r)}{\partial \bar{\mathcal{L}}}\Big)$$

$$-\frac{1}{2}\sum_{m=1}^{M}\textbf{tr}\Big(B_{\mathcal{I}_m}^{-1}\times 0\Big)-\frac{1}{2}\sum_{m=1}^{M}\textbf{tr}\Big(C_{\mathcal{I}_m}^{-1}\times 0\Big)$$

$$\frac{\partial L_{\mathcal{I}}(\theta)}{\partial \sigma_f} = \frac{1}{2}\sum_{m=1}^{M}\Bigg((z^m)^T A_{\mathcal{I}_m}^{-1}\frac{\partial K^m(r,r)}{\partial \bar{\mathcal{L}}} A_{\mathcal{I}_m}^{-1} z^m\Bigg) - \frac{1}{2}\sum_{m=1}^{M}\mathbf{tr}\Bigg(A_{\mathcal{I}_m}^{-1}\frac{\partial K^m(r,r)}{\partial \bar{\mathcal{L}}}\Bigg) \tag{C.32}$$

The term $\frac{\partial K^m(r,r)}{\partial \bar{\mathcal{L}}}$ is calculated exactly in previous section, but we do not hesitate to bring the results here again for the sake of consistency:

$$\frac{\partial K_m(r,r)}{\partial \bar{\mathcal{L}}} = \sigma_f^2(\frac{1}{2})^{-\frac{1}{2}}\Bigg(...$$

$$+\Bigg(\frac{\partial\Big[\mathcal{L}^T\mathcal{L}I_n^T\Big]^{\frac{1}{4}}}{\partial \bar{\mathcal{L}}}\Bigg)\Big[I_n^T\mathcal{L}^T\mathcal{L}\Big]^{\frac{1}{4}}\Big[\mathcal{L}^T\mathcal{L}I_n^T + I_n^T\mathcal{L}^T\mathcal{L}\Big]^{\frac{-1}{2}}\exp\Bigg[\frac{-s(R)}{[\mathcal{L}^T\mathcal{L}I_n^T + I_n^T\mathcal{L}^T\mathcal{L}]}\Bigg]$$

$$+\Big[\mathcal{L}^T\mathcal{L}I_n^T\Big]^{\frac{1}{4}}\Bigg(\frac{\partial\Big[I_n^T\mathcal{L}^T\mathcal{L}\Big]^{\frac{1}{4}}}{\partial \bar{\mathcal{L}}}\Bigg)\Big[\mathcal{L}^T\mathcal{L}I_n^T + I_n^T\mathcal{L}^T\mathcal{L}\Big]^{\frac{-1}{2}}\exp\Bigg[\frac{-s(R)}{[\mathcal{L}^T\mathcal{L}I_n^T + I_n^T\mathcal{L}^T\mathcal{L}]}\Bigg]$$

$$+\Big[\mathcal{L}^T\mathcal{L}I_n^T\Big]^{\frac{1}{4}}\Big[I_n^T\mathcal{L}^T\mathcal{L}\Big]^{\frac{1}{4}}\Bigg(\frac{\partial\Big[\mathcal{L}^T\mathcal{L}I_n^T + I_n^T\mathcal{L}^T\mathcal{L}\Big]^{\frac{-1}{2}}}{\partial \bar{\mathcal{L}}}\Bigg)\exp\Bigg[\frac{-s(R)}{[\mathcal{L}^T\mathcal{L}I_n^T + I_n^T\mathcal{L}^T\mathcal{L}]}\Bigg]$$

$$+\Big[\mathcal{L}^T\mathcal{L}I_n^T\Big]^{\frac{1}{4}}\Big[I_n^T\mathcal{L}^T\mathcal{L}\Big]^{\frac{1}{4}}\Big[\mathcal{L}^T\mathcal{L}I_n^T + I_n^T\mathcal{L}^T\mathcal{L}\Big]^{\frac{-1}{2}}\Bigg(\frac{\partial\exp\Big[\frac{-s(R)}{[\mathcal{L}^T\mathcal{L}I_n^T + I_n^T\mathcal{L}^T\mathcal{L}]}\Big]}{\partial \bar{\mathcal{L}}}\Bigg)$$

$$...\Bigg) \tag{C.33}$$

First we write the relationship between $\mathcal{L}$ and $\bar{\mathcal{L}}$ as $\mathcal{L} = \tau(\bar{\mathcal{L}}) = \exp[\eta\bar{\mathcal{L}}]$ with $\eta = (\bar{K}(r,\bar{r}))^T[\bar{K}(\bar{r},\bar{r}) + \bar{\sigma}_n^2 I]^{-1}$ and then we can calculate the first partial derivative as follows:

$$\frac{\partial\Big[\mathcal{L}^T\mathcal{L}I_n^T\Big]^{\frac{1}{4}}}{\partial \bar{\mathcal{L}}} = \frac{1}{4}\times 2\mathcal{L}I_n^2\Big(\mathcal{L}^T\mathcal{L}I_n^2\Big)^{-\frac{3}{4}}\times\eta\exp[\eta\bar{\mathcal{L}}] \tag{C.34}$$

Also the second partial derivative would as follows:

$$\frac{\partial\Big[I_n^T\mathcal{L}^T\mathcal{L}\Big]^{\frac{1}{4}}}{\partial \bar{\mathcal{L}}} = \frac{1}{4}\times 2I_n^2\mathcal{L}\Big(I_n^2\mathcal{L}^T\mathcal{L}\Big)^{-\frac{3}{4}}\times\eta\exp[\eta\bar{\mathcal{L}}] \tag{C.35}$$

The third partial derivative:

$$\frac{\partial\Big[\mathcal{L}^T\mathcal{L}I_n^T + I_n^T\mathcal{L}^T\mathcal{L}\Big]^{\frac{-1}{2}}}{\partial\bar{\mathcal{L}}} = -\frac{1}{2}\times(2\mathcal{L}I_n^T+I_n^T 2\mathcal{L})(\mathcal{L}^T\mathcal{L}I_n^T+I_n^T\mathcal{L}^T\mathcal{L})^{-\frac{3}{2}}\times\eta\exp[\eta\bar{\mathcal{L}}] \tag{C.36}$$

The last partial derivative:

## Gradient of matrices $A_{\mathcal{I}_m}$, $B_{\mathcal{I}_m}$, $C_{\mathcal{I}_m}$ And $L_{\mathcal{I}}(\theta)$ w.r.t $\bar{\mathcal{I}}$

:

$$\frac{\partial A_{\mathcal{I}_m}}{\partial\bar{\mathcal{I}}} = \frac{\partial(K^m(r,r) + (\mathcal{I}^m)^2 I)}{\partial\bar{\mathcal{I}}} = 2\bar{\mathcal{I}} \tag{C.37}$$

$$\frac{\partial B_{\mathcal{I}_m}}{\partial\bar{\mathcal{I}}} = \frac{\partial(\bar{K}^m(\bar{r},\bar{r}) + \bar{\sigma}_n^2 I)}{\partial\bar{\mathcal{I}}} = 0 \tag{C.38}$$

$$\frac{\partial C_{\mathcal{I}_m}}{\partial\bar{\mathcal{I}}} = \frac{\partial(\bar{K}_{\mathcal{I}}^m(\bar{r}_{\mathcal{I}},\bar{r}_{\mathcal{I}}) + \bar{\bar{\sigma}}_n^2 I)}{\partial\bar{\mathcal{I}}} = 0 \tag{C.39}$$

Thus we may calculate $\frac{\partial L_{\mathcal{I}}(\theta)}{\partial\bar{\mathcal{I}}}$ as follows:

$$\begin{aligned}
\frac{\partial L_{\mathcal{I}}(\theta)}{\partial\bar{\mathcal{I}}} &= \frac{1}{2}\sum_{m=1}^{M}\Bigg((z^m)^T A_{\mathcal{I}_m}^{-1}\frac{\partial A_{\mathcal{I}_m}}{\partial\bar{\mathcal{I}}}A_{\mathcal{I}_m}^{-1}z^m\Bigg) - \frac{1}{2}\sum_{m=1}^{M}\textbf{tr}\Bigg(A_{\mathcal{I}_m}^{-1}\frac{\partial A_{\mathcal{I}_m}}{\partial\bar{\mathcal{I}}}\Bigg) \\
&-\frac{1}{2}\sum_{m=1}^{M}\textbf{tr}\Bigg(B_{\mathcal{I}_m}^{-1}\frac{\partial B_{\mathcal{I}_m}}{\partial\bar{\mathcal{I}}}\Bigg) - \frac{1}{2}\sum_{m=1}^{M}\textbf{tr}\Bigg(C_{\mathcal{I}_m}^{-1}\frac{\partial C_{\mathcal{I}_m}}{\partial\bar{\mathcal{I}}}\Bigg) \\
&= \frac{1}{2}\sum_{m=1}^{M}\Bigg((z^m)^T A_{\mathcal{I}_m}^{-1}2\bar{\mathcal{I}}A_{\mathcal{I}_m}^{-1}z^m\Bigg) - \frac{1}{2}\sum_{m=1}^{M}\textbf{tr}\Bigg(A_{\mathcal{I}_m}^{-1}2\bar{\mathcal{I}}\Bigg) \\
&-\frac{1}{2}\sum_{m=1}^{M}\textbf{tr}\Bigg(B_{\mathcal{I}_m}^{-1}\times 0\Bigg) - \frac{1}{2}\sum_{m=1}^{M}\textbf{tr}\Bigg(C_{\mathcal{I}_m}^{-1}\times 0\Bigg)
\end{aligned}$$

$$\frac{\partial L_{\mathcal{I}}(\theta)}{\partial\bar{\mathcal{I}}} = \frac{1}{2}\sum_{m=1}^{M}\Bigg((z^m)^T A_{\mathcal{I}_m}^{-1}2\bar{\mathcal{I}}A_{\mathcal{I}_m}^{-1}z^m\Bigg) - \frac{1}{2}\sum_{m=1}^{M}\textbf{tr}\Bigg(A_{\mathcal{I}_m}^{-1}2\bar{\mathcal{I}}\Bigg) \tag{C.40}$$

## Gradient of matrices $A_{\mathcal{I}_m}$, $B_{\mathcal{I}_m}$, $C_{\mathcal{I}_m}$ And $L_{\mathcal{I}}(\theta)$ w.r.t $\bar{\sigma}_f$

:

Again we write the explicit equations that we have as follows:

$$\frac{\partial A_{\mathcal{I}_m}}{\partial \bar{\sigma}_f} = \frac{\partial (K^m(r,r) + (\mathcal{I}^m)^2 I)}{\partial \bar{\sigma}_f} = 0 \tag{C.41}$$

$$\frac{\partial B_{\mathcal{I}_m}}{\partial \bar{\sigma}_f} = \frac{\partial (\bar{K}_m(\bar{r},\bar{r}) + \bar{\sigma}_n^2 I)}{\partial \bar{\sigma}_f} = 2\bar{\sigma}_f \bar{K}^m(\bar{r},\bar{r}) \tag{C.42}$$

$$\frac{\partial C_{\mathcal{I}_m}}{\partial \bar{\sigma}_f} = \frac{\partial (\bar{K}_{\mathcal{I}}^m(\bar{r}_{\mathcal{I}},\bar{r}_{\mathcal{I}}) + \bar{\bar{\sigma}}_n^2 I)}{\partial \bar{\sigma}_f} = 0 \tag{C.43}$$

Thus we may calculate $\frac{\partial L_{\mathcal{I}}(\theta)}{\partial \bar{\sigma}_f}$ as follows:

$$\frac{\partial L_{\mathcal{I}}(\theta)}{\partial \bar{\sigma}_f} = \frac{1}{2}\sum_{m=1}^{M}\left((z^m)^T A_{\mathcal{I}_m}^{-1} \frac{\partial A_{\mathcal{I}_m}}{\partial \bar{\sigma}_f} A_{\mathcal{I}_m}^{-1} z^m\right) - \frac{1}{2}\sum_{m=1}^{M}\mathbf{tr}\left(A_{\mathcal{I}_m}^{-1}\frac{\partial A_{\mathcal{I}_m}}{\partial \bar{\sigma}_f}\right)$$

$$-\frac{1}{2}\sum_{m=1}^{M}\mathbf{tr}\left(B_{\mathcal{I}_m}^{-1}\frac{\partial B_{\mathcal{I}_m}}{\partial \bar{\sigma}_f}\right) - \frac{1}{2}\sum_{m=1}^{M}\mathbf{tr}\left(C_{\mathcal{I}_m}^{-1}\frac{\partial C_{\mathcal{I}_m}}{\partial \bar{\sigma}_f}\right)$$

$$\frac{\partial L_{\mathcal{I}}(\theta)}{\partial \bar{\sigma}_f} = \frac{1}{2}\sum_{m=1}^{M}\left((z^m)^T A_{\mathcal{I}_m}^{-1} \times 0 \times A_{\mathcal{I}_m}^{-1} z^m\right) - \frac{1}{2}\sum_{m=1}^{M}\mathbf{tr}\left(A_{\mathcal{I}_m}^{-1} \times 0\right)$$

$$-\frac{1}{2}\sum_{m=1}^{M}\mathbf{tr}\left(B_{\mathcal{I}_m}^{-1} 2\bar{\sigma}_f \bar{K}^m(\bar{r},\bar{r})\right) - \frac{1}{2}\sum_{m=1}^{M}\mathbf{tr}\left(C_{\mathcal{I}_m}^{-1} \times 0\right)$$

$$\frac{\partial L_{\mathcal{I}}(\theta)}{\partial \bar{\sigma}_f} = -\frac{1}{2}\sum_{m=1}^{M}\mathbf{tr}\left(B_{\mathcal{I}_m}^{-1} 2\bar{\sigma}_f \bar{K}^m(\bar{r},\bar{r})\right) \tag{C.44}$$

## Gradient of matrices $A_{\mathcal{I}_m}$, $B_{\mathcal{I}_m}$, $C_{\mathcal{I}_m}$ And $L_{\mathcal{I}}(\theta)$ w.r.t $\bar{\sigma}_l$

:

Again we write the explicit equations that we have as follows:

$$\frac{\partial A_{\mathcal{I}_m}}{\partial \bar{\sigma}_l} = \frac{\partial (K^m(r,r) + (\mathcal{I}^m)^2 I)}{\partial \bar{\sigma}_l} = 0 \tag{C.45}$$

$$\frac{\partial B_{\mathcal{I}_m}}{\partial \bar{\sigma}_l} = \frac{\partial (\bar{K}^m(\bar{r},\bar{r}) + \bar{\sigma}_n^2 I)}{\partial \bar{\sigma}_l} = \frac{\partial \left[\bar{\sigma}_f^2 \exp\left(-\frac{1}{2}s(\bar{\sigma}_l^{-2}(\bar{R}^m)^2)\right) + \bar{\sigma}_n^2 I\right]}{\partial \bar{\sigma}_l}$$

$$= \bar{\sigma}_f^2 \frac{\partial \Big[ \exp\Big( -\frac{1}{2} s(\bar{\sigma}_l^{-2}(\bar{R}^m)^2) \Big) \Big]}{\partial \bar{\sigma}_l} = \bar{\sigma}_f^2 \frac{\partial \Big[ \exp\Big( -\frac{1}{2} \frac{(\bar{r}_i^m - \bar{r_j}^m)^2}{\bar{\sigma}_l^2} \Big) \Big]}{\partial \bar{\sigma}_l}$$

$$\frac{\partial B_{\mathcal{I}_m}}{\partial \bar{\sigma}_l} = \frac{(\bar{r}_i^m - \bar{r_j}^m)^2 \bar{\sigma}_f^2 \exp\Big( -\frac{1}{2} \frac{(\bar{r}_i^m - \bar{r_j}^m)^2}{\bar{\sigma}_l^2} \Big)}{\bar{\sigma}_l^3} \tag{C.46}$$

$$\frac{\partial C_{\mathcal{I}_m}}{\partial \bar{\sigma}_f} = \frac{\partial (\bar{K}_{\mathcal{I}}^m(\bar{r}_{\mathcal{I}}, \bar{r}_{\mathcal{I}}) + \bar{\bar{\sigma}}_n^2 I)}{\partial \bar{\sigma}_f} = 0 \tag{C.47}$$

Thus we may calculate $\frac{\partial L_{\mathcal{I}}(\theta)}{\partial \bar{\sigma}_l}$ as follows:

$$\frac{\partial L_{\mathcal{I}}(\theta)}{\partial \bar{\sigma}_l} = \frac{1}{2} \sum_{m=1}^{M} \Big( (z^m)^T A_{\mathcal{I}_m}^{-1} \frac{\partial A_{\mathcal{I}_m}}{\partial \bar{\sigma}_l} A_{\mathcal{I}_m}^{-1} z^m \Big) - \frac{1}{2} \sum_{m=1}^{M} \mathbf{tr} \Big( A_{\mathcal{I}_m}^{-1} \frac{\partial A_{\mathcal{I}_m}}{\partial \bar{\sigma}_l} \Big)$$

$$- \frac{1}{2} \sum_{m=1}^{M} \mathbf{tr} \Big( B_{\mathcal{I}_m}^{-1} \frac{\partial B_{\mathcal{I}_m}}{\partial \bar{\sigma}_l} \Big) - \frac{1}{2} \sum_{m=1}^{M} \mathbf{tr} \Big( C_{\mathcal{I}_m}^{-1} \frac{\partial C_{\mathcal{I}_m}}{\partial \bar{\sigma}_l} \Big)$$

$$= \frac{1}{2} \sum_{m=1}^{M} \Big( (z^m)^T A_{\mathcal{I}_m}^{-1} \times 0 \times A_{\mathcal{I}_m}^{-1} z^m \Big) - \frac{1}{2} \sum_{m=1}^{M} \mathbf{tr} \Big( A_{\mathcal{I}_m}^{-1} \times 0 \Big)$$

$$- \frac{1}{2} \sum_{m=1}^{M} \mathbf{tr} \Big( B_{\mathcal{I}_m}^{-1} \frac{\partial B_{\mathcal{I}_m}}{\partial \bar{\sigma}_l} \Big) - \frac{1}{2} \sum_{m=1}^{M} \mathbf{tr} \Big( C_{\mathcal{I}_m}^{-1} \times 0 \Big)$$

$$\frac{\partial L_{\mathcal{I}}(\theta)}{\partial \bar{\sigma}_l} = -\frac{1}{2} \sum_{m=1}^{M} \mathbf{tr} \Big( B_{\mathcal{I}_m}^{-1} \frac{(\bar{r}_i^m - \bar{r_j}^m)^2 \bar{\sigma}_f^2 \exp\Big( -\frac{1}{2} \frac{(\bar{r}_i^m - \bar{r_j}^m)^2}{\bar{\sigma}_l^2} \Big)}{\bar{\sigma}_l^3} \Big) \tag{C.48}$$

## Gradient of matrices $A_{\mathcal{I}_m}$, $B_{\mathcal{I}_m}$, $C_{\mathcal{I}_m}$ And $L_{\mathcal{I}}(\theta)$ w.r.t $\bar{\sigma_n}$

:

Again we write the explicit equations that we have as follows:

$$\frac{\partial A_{\mathcal{I}_m}}{\partial \bar{\sigma}_n} = \frac{\partial (K^m(r, r) + (\mathcal{I}^m)^2 I)}{\partial \bar{\sigma}_n} = 0 \tag{C.49}$$

$$\frac{\partial B_{\mathcal{I}_m}}{\partial \bar{\sigma}_n} = \frac{\partial (\bar{K}_m(\bar{r}, \bar{r}) + \bar{\sigma}_n^2 I)}{\partial \bar{\sigma}_n} = 2\bar{\sigma}_n \tag{C.50}$$

$$\frac{\partial C_{\mathcal{I}_m}}{\partial \bar{\sigma}_n} = \frac{\partial (\bar{K}_{\mathcal{I}}^m(\bar{r}_{\mathcal{I}}, \bar{r}_{\mathcal{I}}) + \bar{\bar{\sigma}}_n^2 I)}{\partial \bar{\sigma}_n} = 0 \tag{C.51}$$

Thus we may calculate $\frac{\partial L_{\mathcal{I}}(\theta)}{\partial \bar{\sigma}_n}$ as follows:

$$\frac{\partial L_{\mathcal{I}}(\theta)}{\partial \bar{\sigma}_n} = \frac{1}{2}\sum_{m=1}^{M}\left((z^m)^T A_{\mathcal{I}_m}^{-1}\frac{\partial A_{\mathcal{I}_m}}{\partial \bar{\sigma}_n}A_{\mathcal{I}_m}^{-1}z^m\right) - \frac{1}{2}\sum_{m=1}^{M}\mathbf{tr}\left(A_{\mathcal{I}_m}^{-1}\frac{\partial A_{\mathcal{I}_m}}{\partial \bar{\sigma}_n}\right)$$

$$-\frac{1}{2}\sum_{m=1}^{M}\mathbf{tr}\left(B_{\mathcal{I}_m}^{-1}\frac{\partial B_{\mathcal{I}_m}}{\partial \bar{\sigma}_n}\right) - \frac{1}{2}\sum_{m=1}^{M}\mathbf{tr}\left(C_{\mathcal{I}_m}^{-1}\frac{\partial C_{\mathcal{I}_m}}{\partial \bar{\sigma}_n}\right)$$

$$= \frac{1}{2}\sum_{m=1}^{M}\left((z^m)^T A_{\mathcal{I}_m}^{-1} \times 0 \times A_{\mathcal{I}_m}^{-1}z^m\right) - \frac{1}{2}\sum_{m=1}^{M}\mathbf{tr}\left(A_{\mathcal{I}_m}^{-1} \times 0\right)$$

$$-\frac{1}{2}\sum_{m=1}^{M}\mathbf{tr}\left(B_{\mathcal{I}_m}^{-1}2\bar{\sigma}_n\right) - \frac{1}{2}\sum_{m=1}^{M}\mathbf{tr}\left(C_{\mathcal{I}_m}^{-1} \times 0\right)$$

$$\frac{\partial L_{\mathcal{I}}(\theta)}{\partial \bar{\sigma}_n} = -\frac{1}{2}\sum_{m=1}^{M}\mathbf{tr}\left(B_{\mathcal{I}_m}^{-1}2\bar{\sigma}_n\right) \tag{C.52}$$

## Gradient of matrices $A_{\mathcal{I}_m}$, $B_{\mathcal{I}_m}$, $C_{\mathcal{I}_m}$ And $L_{\mathcal{I}}(\theta)$ w.r.t $\bar{\bar{\sigma}}_f$

:

Again we write the explicit equations that we have as follows:

$$\frac{\partial A_{\mathcal{I}_m}}{\partial \bar{\bar{\sigma}}_f} = \frac{\partial (K^m(r,r) + (\mathcal{I}^m)^2 I)}{\partial \bar{\bar{\sigma}}_f} = 0 \tag{C.53}$$

$$\frac{\partial B_{\mathcal{I}_m}}{\partial \bar{\bar{\sigma}}_f} = \frac{\partial (\bar{K}_m(\bar{r},\bar{r}) + \bar{\sigma}_n^2 I)}{\partial \bar{\bar{\sigma}}_f} = 0 \tag{C.54}$$

$$\frac{\partial C_{\mathcal{I}_m}}{\partial \bar{\bar{\sigma}}_f} = \frac{\partial (\bar{K}_{\mathcal{I}}^m(\bar{r}_{\mathcal{I}},\bar{r}_{\mathcal{I}}) + \bar{\bar{\sigma}}_n^2 I)}{\partial \bar{\bar{\sigma}}_f} = 2\bar{\bar{\sigma}}_f \bar{K}_{\mathcal{I}}^m(\bar{r}_{\mathcal{I}},\bar{r}_{\mathcal{I}}) \tag{C.55}$$

Thus we may calculate $\frac{\partial L_{\mathcal{I}}(\theta)}{\partial \bar{\bar{\sigma}}_f}$ as follows:

$$\frac{\partial L_{\mathcal{I}}(\theta)}{\partial \bar{\bar{\sigma}}_f} = \frac{1}{2}\sum_{m=1}^{M}\left((z^m)^T A_{\mathcal{I}_m}^{-1}\frac{\partial A_{\mathcal{I}_m}}{\partial \bar{\bar{\sigma}}_f}A_{\mathcal{I}_m}^{-1}z^m\right) - \frac{1}{2}\sum_{m=1}^{M}\mathbf{tr}\left(A_{\mathcal{I}_m}^{-1}\frac{\partial A_{\mathcal{I}_m}}{\partial \bar{\bar{\sigma}}_f}\right)$$

$$-\frac{1}{2}\sum_{m=1}^{M}\mathbf{tr}\left(B_{\mathcal{I}_m}^{-1}\frac{\partial B_{\mathcal{I}_m}}{\partial \bar{\bar{\sigma}}_f}\right) - \frac{1}{2}\sum_{m=1}^{M}\mathbf{tr}\left(C_{\mathcal{I}_m}^{-1}\frac{\partial C_{\mathcal{I}_m}}{\partial \bar{\bar{\sigma}}_f}\right)$$

$$\frac{\partial L_{\mathcal{I}}(\theta)}{\partial \bar{\bar{\sigma}}_f} = \frac{1}{2}\sum_{m=1}^{M}\left((z^m)^T A_{\mathcal{I}_m}^{-1} \times 0 \times A_{\mathcal{I}_m}^{-1}z^m\right) - \frac{1}{2}\sum_{m=1}^{M}\mathbf{tr}\left(A_{\mathcal{I}_m}^{-1} \times 0\right)$$

$$-\frac{1}{2}\sum_{m=1}^{M}\mathbf{tr}\bigg(B_{\mathcal{I}_m}^{-1}\times 0\bigg)-\frac{1}{2}\sum_{m=1}^{M}\mathbf{tr}\bigg(C_{\mathcal{I}_m}^{-1}\times 2\bar{\bar{\sigma}}_f\bar{K}_{\mathcal{I}}^{m}(\bar{r}_{\mathcal{I}},\bar{r}_{\mathcal{I}})\bigg)$$

$$\frac{\partial L_{\mathcal{I}}(\theta)}{\partial\bar{\bar{\sigma}}_f}=-\frac{1}{2}\sum_{m=1}^{M}\mathbf{tr}\bigg(C_{\mathcal{I}_m}^{-1}\times 2\bar{\bar{\sigma}}_f\bar{K}_{\mathcal{I}}^{m}(\bar{r}_{\mathcal{I}},\bar{r}_{\mathcal{I}})\bigg) \tag{C.56}$$

## Gradient of matrices $A_{\mathcal{I}_m}$, $B_{\mathcal{I}_m}$, $C_{\mathcal{I}_m}$ And $L_{\mathcal{I}}(\theta)$ w.r.t $\bar{\bar{\sigma}}_l$

:

Again we write the explicit equations that we have as follows:

$$\frac{\partial A_{\mathcal{I}_m}}{\partial\bar{\bar{\sigma}}_l}=\frac{\partial(K^m(r,r)+(\mathcal{I}^m)^2I)}{\partial\bar{\bar{\sigma}}_l}=0 \tag{C.57}$$

$$\frac{\partial B_{\mathcal{I}_m}}{\partial\bar{\bar{\sigma}}_l}=\frac{\partial(\bar{K}^m(\bar{r},\bar{r})+\bar{\sigma}_n^2I)}{\partial\bar{\bar{\sigma}}_l}=0 \tag{C.58}$$

$$\frac{\partial C_{\mathcal{I}_m}}{\partial\bar{\bar{\sigma}}_l}=\frac{\partial(\bar{K}_{\mathcal{I}}^m(\bar{r}_{\mathcal{I}},\bar{r}_{\mathcal{I}})+\bar{\bar{\sigma}}_n^2I)}{\partial\bar{\bar{\sigma}}_l}=\frac{\partial\bigg[\bar{\bar{\sigma}}_f^2\exp\bigg(-\frac{1}{2}s(\bar{\bar{\sigma}}_l^{-2}(\bar{R}_{\mathcal{I}}^m)^2)\bigg)+\bar{\bar{\sigma}}_n^2I\bigg]}{\partial\bar{\bar{\sigma}}_l}$$

$$=\bar{\bar{\sigma}}_f^2\frac{\partial\bigg[\exp\bigg(-\frac{1}{2}s(\bar{\bar{\sigma}}_l^{-2}(\bar{R}_{\mathcal{I}}^m)^2)\bigg)\bigg]}{\partial\bar{\bar{\sigma}}_l}=\bar{\bar{\sigma}}_f^2\frac{\partial\bigg[\exp\bigg(-\frac{1}{2}\frac{(\bar{r}_{\mathcal{I}_i}^m-\bar{r}_{\mathcal{I}_j}^m)^2}{\bar{\bar{\sigma}}_l^2}\bigg)\bigg]}{\partial\bar{\bar{\sigma}}_l}$$

$$\frac{\partial C_{\mathcal{I}_m}}{\partial\bar{\bar{\sigma}}_l}=\frac{(\bar{r}_{\mathcal{I}_i}^m-\bar{r}_{\mathcal{I}_j}^m)^2\times\bar{\bar{\sigma}}_f^2\times\exp\bigg(-\frac{1}{2}\frac{(\bar{r}_{\mathcal{I}_i}^m-\bar{r}_{\mathcal{I}_j}^m)^2}{\bar{\bar{\sigma}}_l^2}\bigg)}{\bar{\bar{\sigma}}_l^3} \tag{C.59}$$

Thus we may calculate $\frac{\partial L_{\mathcal{I}}(\theta)}{\partial\bar{\bar{\sigma}}_l}$ as follows:

$$\frac{\partial L_{\mathcal{I}}(\theta)}{\partial\bar{\bar{\sigma}}_l}=\frac{1}{2}\sum_{m=1}^{M}\bigg((z^m)^TA_{\mathcal{I}_m}^{-1}\frac{\partial A_{\mathcal{I}_m}}{\partial\bar{\bar{\sigma}}_l}A_{\mathcal{I}_m}^{-1}z^m\bigg)-\frac{1}{2}\sum_{m=1}^{M}\mathbf{tr}\bigg(A_{\mathcal{I}_m}^{-1}\frac{\partial A_{\mathcal{I}_m}}{\partial\bar{\bar{\sigma}}_l}\bigg)$$

$$-\frac{1}{2}\sum_{m=1}^{M}\mathbf{tr}\bigg(B_{\mathcal{I}_m}^{-1}\frac{\partial B_{\mathcal{I}_m}}{\partial\bar{\bar{\sigma}}_l}\bigg)-\frac{1}{2}\sum_{m=1}^{M}\mathbf{tr}\bigg(C_{\mathcal{I}_m}^{-1}\frac{\partial C_{\mathcal{I}_m}}{\partial\bar{\bar{\sigma}}_l}\bigg)$$

$$=\frac{1}{2}\sum_{m=1}^{M}\bigg((z^m)^TA_{\mathcal{I}_m}^{-1}\times 0\times A_{\mathcal{I}_m}^{-1}z^m\bigg)-\frac{1}{2}\sum_{m=1}^{M}\mathbf{tr}\bigg(A_{\mathcal{I}_m}^{-1}\times 0\bigg)$$

$$-\frac{1}{2}\sum_{m=1}^{M}\mathbf{tr}\bigg(B_{\mathcal{I}_m}^{-1}\times 0\bigg)-\frac{1}{2}\sum_{m=1}^{M}\mathbf{tr}\bigg(C_{\mathcal{I}_m}^{-1}\times\frac{\partial C_{\mathcal{I}_m}}{\partial\bar{\bar{\sigma}}_l}\bigg)$$

$$\frac{\partial L_{\mathcal{I}}(\theta)}{\partial \bar{\bar{\sigma}}_l} = -\frac{1}{2}\sum_{m=1}^{M} \mathbf{tr}\Bigg( C_{\mathcal{I}_m}^{-1} \times \frac{(\bar{r}_{\mathcal{I}_i}^m - \bar{r}_{\mathcal{I}_j}^m)^2 \times \bar{\bar{\sigma}}_f^2 \times \exp\Big( -\frac{1}{2}\frac{(\bar{r}_{\mathcal{I}_i}^m - \bar{r}_{\mathcal{I}_j}^m)^2}{\bar{\bar{\sigma}}_l^2}\Big)}{\bar{\bar{\sigma}}_l^3} \Bigg) \tag{C.60}$$

## Gradient of matrices $A_{\mathcal{I}_m}$, $B_{\mathcal{I}_m}$, $C_{\mathcal{I}_m}$ And $L_{\mathcal{I}}(\theta)$ w.r.t $\bar{\bar{\sigma}}_n$

:

Again we write the explicit equations that we have as follows:

$$\frac{\partial A_{\mathcal{I}_m}}{\partial \bar{\bar{\sigma}}_n} = \frac{\partial (K^m(r,r) + (\mathcal{I}^m)^2 I)}{\partial \bar{\bar{\sigma}}_n} = 0 \tag{C.61}$$

$$\frac{\partial B_{\mathcal{I}_m}}{\partial \bar{\bar{\sigma}}_n} = \frac{\partial (\bar{K}_m(\bar{r},\bar{r}) + \bar{\sigma}_n^2 I)}{\partial \bar{\bar{\sigma}}_n} = 0 \tag{C.62}$$

$$\frac{\partial C_{\mathcal{I}_m}}{\partial \bar{\bar{\sigma}}_n} = \frac{\partial (\bar{K}_{\mathcal{I}}^m(\bar{r}_{\mathcal{I}},\bar{r}_{\mathcal{I}}) + \bar{\bar{\sigma}}_n^2 I)}{\partial \bar{\bar{\sigma}}_n} = 2\bar{\bar{\sigma}}_n \tag{C.63}$$

Thus we may calculate $\frac{\partial L_{\mathcal{I}}(\theta)}{\partial \bar{\bar{\sigma}}_n}$ as follows:

$$\frac{\partial L_{\mathcal{I}}(\theta)}{\partial \bar{\bar{\sigma}}_n} = \frac{1}{2}\sum_{m=1}^{M}\Bigg((z^m)^T A_{\mathcal{I}_m}^{-1}\frac{\partial A_{\mathcal{I}_m}}{\partial \bar{\bar{\sigma}}_n} A_{\mathcal{I}_m}^{-1} z^m\Bigg) - \frac{1}{2}\sum_{m=1}^{M}\mathbf{tr}\Bigg(A_{\mathcal{I}_m}^{-1}\frac{\partial A_{\mathcal{I}_m}}{\partial \bar{\bar{\sigma}}_n}\Bigg)$$

$$-\frac{1}{2}\sum_{m=1}^{M}\mathbf{tr}\Bigg(B_{\mathcal{I}_m}^{-1}\frac{\partial B_{\mathcal{I}_m}}{\partial \bar{\bar{\sigma}}_n}\Bigg) - \frac{1}{2}\sum_{m=1}^{M}\mathbf{tr}\Bigg(C_{\mathcal{I}_m}^{-1}\frac{\partial C_{\mathcal{I}_m}}{\partial \bar{\bar{\sigma}}_n}\Bigg)$$

$$= \frac{1}{2}\sum_{m=1}^{M}\Bigg((z^m)^T A_{\mathcal{I}_m}^{-1} \times 0 \times A_{\mathcal{I}_m}^{-1} z^m\Bigg) - \frac{1}{2}\sum_{m=1}^{M}\mathbf{tr}\Bigg(A_{\mathcal{I}_m}^{-1} \times 0\Bigg)$$

$$-\frac{1}{2}\sum_{m=1}^{M}\mathbf{tr}\Bigg(B_{\mathcal{I}_m}^{-1} \times 0\Bigg) - \frac{1}{2}\sum_{m=1}^{M}\mathbf{tr}\Bigg(C_{\mathcal{I}_m}^{-1} \times 2\bar{\bar{\sigma}}_n\Bigg)$$

$$\Rightarrow \frac{\partial L_{\mathcal{I}}(\theta)}{\partial \bar{\bar{\sigma}}_n} = -\frac{1}{2}\sum_{m=1}^{M}\mathbf{tr}\Bigg(C_{\mathcal{I}_m}^{-1} 2\bar{\bar{\sigma}}_n\Bigg) \tag{C.64}$$

Having the objective function along with the gradients of this with respect to all the hyper-parameters enables us to use gradient-based optimization algorithms to solve for the hyper-parameters.

# Appendix D

# Review of Different Line Extraction Algorithms

**Split and Merge method** Probably the most popular algorithm for line extraction can be $Split\ and\ Merge$ or with another version of this algorithm called $Iterative - End - Point - Fit$ which was firstly introduced in computer vision literature.

---

**Algorithm 10** Split and Merge

---

Initial step: the set $s_1$ consists of N data points. put $s_1$ in a list $\mathcal{L}$
Fit a line to the next set $s_i$ in $\mathcal{L}$
Detect point P with the maximum distance $d_P$ to the line
If $d_P$ is less than a predefined threshold continue **goto** 2
If not, split $s_i$ at the point $P$ into two different sets $s_{i1}$ and $s_{i2}$
Replace $s_i$ in $\mathcal{L}$ by $s_{i1}$ and $s_{i2}$ then **continue** (**goto** 2)
Continue until all of the sets (segments) in $\mathcal{L}$ are checked, then merge the collinear segments.

---

**Hough-Transform Algorithm** Hough-Transform (HT) algorithm turns out to be the most popular algorithm in the field of intensity image line extraction. This method suffers from some major drawbacks, first of all basic version of this algorithm does not take uncertainty into account in its estimation procedure, and the second, it is not always so easy to set the grid size for this method efficiently.

**Expectation-Maximization Algorithm** This algorithm is in fact a probabilistic method which is often used in missing variable problems, this method is also used in computer vision and robotics line extraction problems. The most challenge in

**Algorithm 11** Hough-Transform

**Input**: A set of N points.
Initialize the model space or the accumulator array
construct values for the array
Choose the element with the maximum votes $V_{max}$
If $V_{max}$ is less than a predefined threshold then terminate
If not, determine the inliers
Fit a line through the inliers
Store the fitted line
Remove the in-liers from the set, **goto** 2

dealing line extraction scenarios using $EM$ algorithm is either it is always not so simple to reach a right initial value or it can be trapped in a local minimum often. $EM$'s algorithm have the following form:

**Algorithm 12** Expectation Maximization

1: **INPUT**: A set containing N points.
2: **repeat**
3: Randomly generate a set of line parameters
4: Initialize weights for remaining points.
5: **repeat**
6: Compute the weights of the points from the line model
7: Recompute the line model parameters
8: **until** $Maximum\ of\ N\ Steps\ reached\ or\ convergence$
9: **until** $Maximum\ of\ N\ Trials$ reached or a line is found
10: **if** found a line **then**
11: store the line
12: remove the inliers **goto** 2
13: **end if**

**Line-Regression Algorithm** This algorithm originated from map-based localization tasks and the key idea in this method is to transform the line extraction problem into a a search problem in the model space or in fact the line parameters domain and then applies the a clustering approach to construct adjacent line segments. The clustering approach which is often used in this method is Agglomerative Hierarchical Clustering(AHC). The main drawback of this algorithm is that it would be quiet complex for implementation. The sliding window size $N_f$, is a very sensitive parameter as it is highly dependent on environment and also highly influential on algorithm performance.

**Algorithm 13** Line Regression

1: **set** sliding window size $N_f$
2: Fit a line to every consecutive window points in $N_f$
3: Compute a line fidelity array, each element of which is the sum of Mahalanobis distances between every 3 adjacent windows.
4: construct line segments by scanning the fidelity array for consecutive elements having values less than a threshold, using an AHC algorithm.
5: Merge overlapped line segments and recompute line parameters for each segment.

**Incremental Algorithm** The incremental algorithm or as it is called in some sources Line-Tacking algorithm highly benefits from simplicity and at the same time, precise performance in many situations. If it is necessary one can speed up this algorithm with adding more points in adding step of this algorithm which speeds up the algorithm by calculating total-least-square parameters for a bunch of points at the same time.

**Algorithm 14** Incremental Algorithm

**INPUT** A set of N two-dimensional points
Construct a line for the first two points
Add the next point to the current line model
Recompute the line parameters
**if** recomputed parameters satisfies line condition **then**
  **goto** 2
**end if**
**if not** put back the last point, recompute the line parameters, return the line
**Continue** with the next two points, **goto** 2

**RANSAC** Random Sample Consensus or RANSAC is a method for robust model fitting in presence of data outliers, it is originally a generic segmentation method and is not limited to line segments and it can be used for segmenting any shapes just in case that we have the model.

**Incremental Algorithm over RANSAC** In fact, the incremental algorithm is favorable on other introduced methods due to the fact that it benefits at the same time from its speed and precision which the former plays a crucial role on our kind of implementation as the result of our ground segmentation algorithm is going to be input of next processing procedures, and those are enough time-consuming

**Algorithm 15** Random Sample Consensus or RANSAC Algorithm

```
INPUT: A set of N points
repeat
Choose a sample of two points at random, uniformly
Fit a line through this two points
Compute the distance of other point to the line
Construct the inlier set
if there are enough inliers then
    Recompute the line parameters again
    Store the line
    Remove the inliers from the set
end if
until Too few points left or Maximum of N iterations reached
```

and complex for us to decide to choose a faster method over precision, while incremental method holds the precision fairly.

| Algorithm | Complexity | Speed | Correctness | | Precision | |
|---|---|---|---|---|---|---|
| | | | TruePos [%] | FalsePos [%] | $\sigma_{\Delta_r}$ $[cm]$ | $\sigma_{\Delta_\alpha}$ $[deg]$ |
| Split-Merge | $N \times \log N$ | 1470 | 86.0 | 8.9 | 1.95 | 0.74 |
| Incremental | $S \times N^2$ | 344 | 77.8 | 5.9 | 2.04 | 0.72 |
| Line Regression | $N \times N_f$ | 364 | 76.4 | 10.1 | 1.99 | 0.80 |
| RANSAC | $S \times N \times N.Trials$ | 29 | 77.8 | 5.9 | 2.04 | 0.72 |

It seems fairly to use simpler algorithms like incremental algorithm or Split-Merge algorithm in this thesis, because famous RANSAC algorithm doesn't show speed and precision at the same time.

# Appendix E

# Ground Segmentation with Input-Dependent Noise

So far, the problem of ground segmentation with the Gaussian process is discussed with two methods with the assumption of constant noise variance on the output. Situation in which the output of a Gaussian process task could be reliably assumed to have constant noise variance is very rare. $\mathcal{GP}$ regression methodology provides very efficient means to assign non-parametric distributions over regression functions. In this section those kind of regression problems are discussed where the variance of the main process noise is still dominating outputs but dependent on the inputs value. The LiDAR sensor's output shows such quality. In much of the works regarding regression problems both in the neural networks and statistical literature, the global noise level is assumed to be independent of the input vector. Reference [134] has discussed the matter of input-dependent noise only for parametric models and reference [135] has discussed this matter for the non-parametric Gaussian regression problems with Monte Carlo based inference method which is not intractable in many case as of ours. Goldberg [135] did not suggest any methods to use the original idea with other schemes. Input-dependent noise for the Gaussian process regression tasks is discussed here based on MAP inference methods. We assume that our probability distribution in the statistics space, is interpretable using the probabilistic graphical model presented in the Figure E.1.

## Case of Three Gaussian Process

The input-dependent noise is assumed to be a smooth function of input. Thus a Gaussian latent is put the noise variance. Therefore, basically the noise variance is modeled with a third Gaussian process in addition to the $\mathcal{GP}_z$ (the Gaussian process governing noise-free outputs) and $\mathcal{GP})_l$, (the Gaussian process governing length-scales). The outputs are assumed to have been generated from the true z-values by adding independent Gaussian noise whose variance is $r - dependent$. Thus the noise variance is assumed to be $\boldsymbol{n} = (n(r_1), n(r_2), n(r_3), ..., n(r_n))$. Hyper-parameters related to this Gaussian process would be $\{\bar{\bar{\sigma}}_f, \bar{\bar{\sigma}}_l, \bar{\bar{\sigma}}_n\}$.

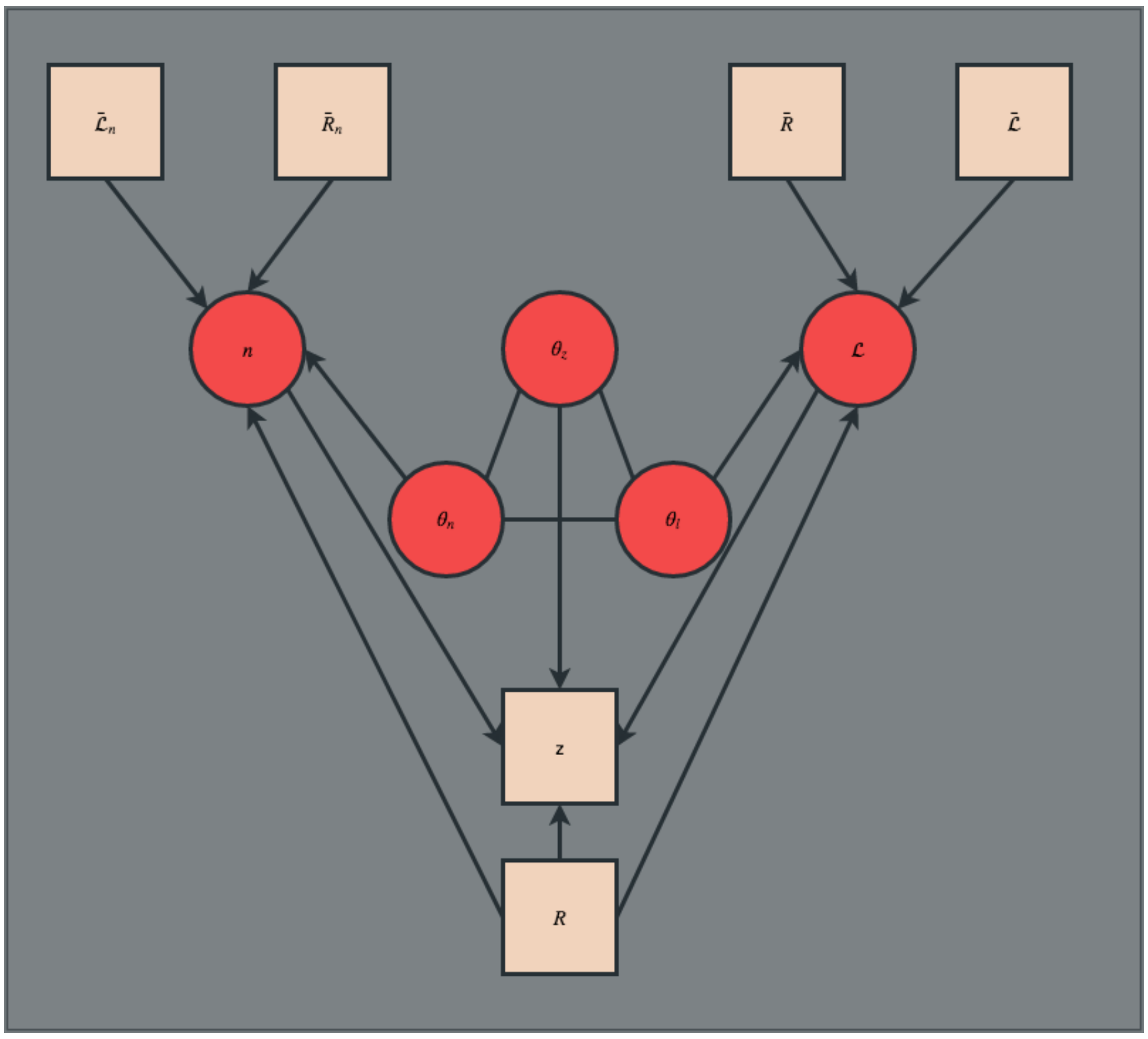

**Figure E.1:** The probabilistic graphical model for the Gaussian process with input-dependent noise.

The important fact regarding this solution is that in the new setup, hyper-parameters regarding the new $\mathcal{GP}_z$ is the set $\{\sigma_f\}$ and the hyper-parameters regarding to the new latent length-scale Gaussian process will be the set $\{\bar{\sigma}_f, \bar{\sigma}_l, \bar{\sigma}_n\}$. Therefore, the new problem setup tends to updates seven different parameters $\theta_{id} = \{\sigma_f, \bar{\mathcal{L}}, \bar{\mathcal{I}}, \bar{\sigma}_f, \bar{\sigma}_l, \bar{\sigma}_n, \bar{\bar{\sigma}}_f, \bar{\bar{\sigma}}_l, \bar{\bar{\sigma}}_n\}$. Assuming all these parameters, the covariance function we have used in previous section changes, so we have to do the math again. The predictive distributions in this situation is discussed in proceeding materials.

## Gaussian Process $\mathcal{GP}_n$ for Input-Dependent Noise

First, the Gaussian process governing the input-dependent noise process must be updated:

$$\mathcal{GP}_n :\Rightarrow P(\bar{n}^*|\bar{R}_n^*, \mathcal{R}_m) \sim \mathcal{N}(n^*, \sigma_{n*}^2) \tag{E.1}$$

$$n^* = \mu_{\mathcal{GP}_n} = \log n = (\bar{K}_n(r, \bar{r}_n))^T [\bar{K}_n(\bar{r}_n, \bar{r}_n) + \bar{\bar{\sigma}}_n^2 I]^{-1} \bar{n}$$

$$\Rightarrow n = \exp\left[(\bar{K}_n(r, \bar{r}_n))^T [\bar{K}_n(\bar{r}_n, \bar{r}_n) + \bar{\bar{\sigma}}_n^2 I]^{-1} \bar{n}\right] \tag{E.2}$$

With covariance kernel being as follows:

$$\bar{K}_n(\bar{r}_n, \bar{r}_n) = \bar{\bar{\sigma}}_f^2 \exp\left( -\frac{1}{2} \frac{s(\bar{R}_n)}{\bar{\bar{\sigma}}_l^2} \right)$$

$$\Rightarrow \bar{k}_n(\bar{r}_{n_i}, \bar{r}_{n_j}) = \bar{\bar{\sigma}}_f^2 \exp\left( -\frac{1}{2} \frac{(\bar{r}_{n_i} - \bar{r}_{n_j})^2}{\bar{\bar{\sigma}}_l^2} \right) \tag{E.3}$$

Also:

$$\bar{K}_n(r, \bar{r}_n) = \bar{\bar{\sigma}}_f^2 \exp\left( -\frac{1}{2} \frac{s(R, \bar{R}_n)}{\bar{\bar{\sigma}}_l^2} \right)$$

$$\bar{k}_n(r_{n_i}, \bar{r}_{n_j}) = \bar{\bar{\sigma}}_f^2 \exp\left( -\frac{1}{2} \frac{(r_i - \bar{r}_{n_j})^2}{\bar{\bar{\sigma}}_l^2} \right) \tag{E.4}$$

## Gaussian Process $\mathcal{GP}_l$ for the Latent Length-Scale

After updating $\mathcal{GP}_n$, the latent length-scale Gaussian process $\mathcal{GP}_l$ must be update which is quite the same as the previous section:

$$\mathcal{GP}_l :\Rightarrow P(\mathcal{L}^*|\bar{R}^*, \mathcal{R}_m) \sim \mathcal{N}(\mathcal{L}^*, \sigma_*^2)$$

$$\mathcal{L}^* = \mu_{\mathcal{GP}_l} = \log \mathcal{L} = (\bar{K}(r, \bar{r}))^T [\bar{K}(\bar{r}, \bar{r}) + \bar{\sigma}_n^2 I]^{-1} \bar{\mathcal{L}}$$

$$\Rightarrow \mathcal{L} = \exp\left[(\bar{K}(r, \bar{r}))^T [\bar{K}(\bar{r}, \bar{r}) + \bar{\sigma}_n^2 I]^{-1} \bar{\mathcal{L}}\right] \tag{E.5}$$

With covariance kernel being as follows:

$$\bar{K}(\bar{r}, \bar{r}) = \bar{\sigma}_f^2 \exp\left( -\frac{1}{2} \frac{s(\bar{R})}{\bar{\sigma}_l^2} \right)$$

$$\Rightarrow \bar{k}(\bar{r}_i, \bar{r}_j) = \bar{\sigma}_f^2 \exp\left( -\frac{1}{2} \frac{(\bar{r}_i - \bar{r}_j)^2}{\bar{\sigma}_l^2} \right) \tag{E.6}$$

Also:

$$\bar{K}(r, \bar{r}) = \bar{\sigma}_f^2 \exp\left( -\frac{1}{2} \frac{s(R, \bar{R})}{\bar{\sigma}_l^2} \right)$$

$$\bar{k}(r_i, \bar{r}_j) = \bar{\sigma}_f^2 \exp\left( -\frac{1}{2} \frac{(r_i - \bar{r}_j)^2}{\bar{\sigma}_l^2} \right) \tag{E.7}$$

## Gaussian Process $\mathcal{GP}_z$ for the Main Observation Process

Now for the main Gaussian process $\mathcal{GP}_z$ the predicting distribution would be of the form below:

$$\mu_{\mathcal{GP}_z} = \bar{z} = K(r^*, r)^T \Big[K(r,r) + \mathbf{n}^2 I\Big]^{-1} z \tag{E.8}$$

$$K(r,r) =$$

$$\sigma_f^2 \cdot \Big[\mathcal{L}^T \mathcal{L} I_n^T\Big]^{\frac{1}{4}} \Big[I_n^T \mathcal{L}^T \mathcal{L}\Big]^{\frac{1}{4}} (\frac{1}{2})^{-\frac{1}{2}} \Big[\mathcal{L}^T \mathcal{L} I_n^T + I_n^T \mathcal{L}^T \mathcal{L}\Big]^{\frac{-1}{2}} \exp\left(\frac{-s(R)}{[\mathcal{L}^T \mathcal{L} I_n^T + I_n^T \mathcal{L}^T \mathcal{L}]}\right) \tag{E.9}$$

## Learning Hyper-Parameters

In this situation, according to the graphical model presented in Figure E.1, the distribution factorizes as follows:

$$P(\mathcal{L}|z, R, \theta_{id}) =$$

$$P(z|R, \exp(\mathcal{L}), \exp(n), \theta_l) \times P(\mathcal{L}|R, \bar{\mathcal{L}}, \bar{R}, \theta_l) \times P(n|R, \bar{n}, \bar{R}_n, \theta_n) \times cte \tag{E.10}$$

We tend to maximize the log-likelihood of this distribution instead of taking integral on all over the statistical space, the log-likelihood would be of the form below:

$$\log P(\mathcal{L}|z, R, \theta_{id}) =$$

$$\log P(z|R, \exp(\mathcal{L}), \exp(n), \theta_l) + \log P(\mathcal{L}|R, \bar{\mathcal{L}}, \bar{R}, \theta_l) + \log P(n|R, \bar{n}, \bar{R}_n, \theta_n) + cte \tag{E.11}$$

The first term only enjoys the presence of data fit, complexity penalty and the consatntterm, so it can be written as follows:

$$\log P(z|R, \exp(\mathcal{L}), \exp(n), \theta_l) =$$

$$-\frac{1}{2} z^T (K(r,r) + n^2 I)^{-1} z - \frac{1}{2} \log |K(r,r) + n^2 I| - \frac{n}{2} \log(2\pi) \tag{E.12}$$

The second term only enjoys the complexity penalty and the constant term due to the fact that it doesn't use any information from output measurement so there is no need to include data fit.

$$\log P(\mathcal{L}|R, \bar{\mathcal{L}}, \bar{R}, \theta_l) = -\frac{1}{2} \log |\bar{K}(\bar{r}, \bar{r}) + \bar{\sigma}_n^2| - \frac{n}{2} \log(2\pi) \tag{E.13}$$

The third term is also written like the second term:

$$\log P(\mathcal{I}|R, \bar{n}, \bar{R}_n, \theta_n) = -\frac{1}{2} \log |\bar{K}_{\mathcal{I}}(\bar{r}_{\mathcal{I}}, \bar{r}_{\mathcal{I}}) + \bar{\bar{\sigma}}_n^2| - \frac{n}{2} \log(2\pi) \tag{E.14}$$

Adding up these terms will give us the objective function for each segement:

$$\log P(\mathcal{L}^m|z^m, R^m, \theta_{id}) =$$

$$\log P(z^m|R^m, \exp(\mathcal{L}^m), \exp(\mathcal{I}^m), \theta_l) + \log P(\mathcal{L}^m|R^m, \bar{\mathcal{L}}^m, \bar{R}^m, \theta_l)$$
$$+ \log P(\mathcal{I}^m|R^m, \bar{\mathcal{I}}^m, \bar{R}^m_n, \theta_{\mathcal{I}}) + \textit{cte}$$
$$= -\frac{1}{2}(z^m)^T(K^m(r,r) + (\mathcal{I}^m)^2 I)^{-1} z^m - \frac{1}{2}\log|K^m(r,r) + (\mathcal{I}^m)^2 I| - \frac{n}{2}\log(2\pi)$$
$$-\frac{1}{2}\log|\bar{K}^m(\bar{r},\bar{r}) + \bar{\sigma}_n^2| - \frac{n}{2}\log(2\pi) - \frac{1}{2}\log|\bar{K}_{\mathcal{I}}(\bar{r}_{\mathcal{I}},\bar{r}_{\mathcal{I}}) + \bar{\bar{\sigma}}_n^2| - \frac{n}{2}\log(2\pi)$$
$$= -\frac{1}{2}(z^m)^T(K^m(r,r) + (\mathcal{I}^m)^2 I)^{-1} z^m - \frac{1}{2}\log|K^m(r,r) + (\mathcal{I}^m)^2 I|$$
$$-\frac{1}{2}\log|\bar{K}^m(\bar{r},\bar{r}) + \bar{\sigma}_n^2| - \frac{1}{2}\log|\bar{K}_{\mathcal{I}}(\bar{r}_{\mathcal{I}},\bar{r}_{\mathcal{I}}) + \bar{\bar{\sigma}}_n^2| - \frac{3n}{2}\log(2\pi) \quad \text{(E.15)}$$
$$= -\frac{1}{2}\Bigg((z^m)^T(K^m(r,r) + (\mathcal{I}^m)^2 I)^{-1} z^m + \log|K^m(r,r) + (\mathcal{I}^m)^2 I| + ...$$
$$+ \log|\bar{K}^m(\bar{r},\bar{r}) + \bar{\sigma}_n^2| + \log|\bar{K}^m_{\mathcal{I}}(\bar{r}_{\mathcal{I}},\bar{r}_{\mathcal{I}}) + \bar{\bar{\sigma}}_n^2|\Bigg) - \frac{3n}{2}\log(2\pi) \quad \text{(E.16)}$$
$$= \alpha_{id}\Bigg((z^m)^T A^{-1}_{\mathcal{I}_m} z^m + \log|A_{\mathcal{I}_m}| + \log|B_{\mathcal{I}_m}| + \log|C_{\mathcal{I}_m}|\Bigg) + \beta_{id} \quad \text{(E.17)}$$

Where$A_{\mathcal{I}_m} := (K^m(r,r) + (\mathcal{I}^m)^2 I)$, also $B_{\mathcal{I}_m} := \bar{K}^m(\bar{r},\bar{r}) + \bar{\sigma}_n^2$, and $C_{\mathcal{I}_m} := \bar{K}^m_{\mathcal{I}}(\bar{r}_{\mathcal{I}},\bar{r}_{\mathcal{I}}) + \bar{\bar{\sigma}}_n^2$ is the corresponding covariance functions of Gaussian processes which are detailed before.

## Objective Function

We are again confronting a ***multi-task*** problem, which our problem is distributed over $M$ different and divided problems across $M$ segments of data which all share the same random field, or parameters, so we should sum up all the objective functions across the space to get our desirable objective function:

$$L_{id}(\theta) = \sum_{m=1}^{M}\Bigg(\log P(\mathcal{L}^m|z^m, R^m, \theta)\Bigg)$$

$$= \sum_{m=1}^{M}\left[\alpha_{id}\Bigg((z^m)^T A^{-1}_{\mathcal{I}_m} z^m + \log|A_{\mathcal{I}_m}| + \log|B_{\mathcal{I}_m}| + \log|C_{\mathcal{I}_m}|\Bigg) + \beta_{id}\right]$$

$$= \alpha_{id} \sum_{m=1}^{M} \left( (z^m)^T A_{\mathcal{I}_m}^{-1} z^m + \log |A_{\mathcal{I}_m}| + \log |B_{\mathcal{I}_m}| + \log |C_{\mathcal{I}_m}| \right) + \sum_{m=1}^{M} \beta_{id}$$

$$= \alpha_{id} \sum_{m=1}^{M} \left( (z^m)^T A_{\mathcal{I}_m}^{-1} z^m \right) + \alpha_{id} \sum_{m=1}^{M} \left( \log |A_{\mathcal{I}_m}| \right) + \alpha_{id} \sum_{m=1}^{M} \left( \log |B_{\mathcal{I}_m}| \right)$$

$$+\alpha_{id} \sum_{m=1}^{M} \left( \log |C_{\mathcal{I}_m}| \right) + \sum_{m=1}^{M} \beta_{id}$$

$$= \alpha_{id} \sum_{m=1}^{M} \left( (z^m)^T A_{\mathcal{I}_m}^{-1} z^m \right) + \alpha_{id} \log \left( \prod_{m=1}^{M} |A_{\mathcal{I}_m}| \right) + \alpha_{id} \log \left( \prod_{m=1}^{M} |B_{\mathcal{I}_m}| \right)$$

$$+\alpha_{id} \log \left( \prod_{m=1}^{M} |C_{\mathcal{I}_m}| \right) + \sum_{m=1}^{M} \beta_{id}$$

$$= -\frac{1}{2} \sum_{m=1}^{M} \left( (z^m)^T A_{\mathcal{I}_m}^{-1} z^m \right) - \frac{1}{2} \log \left( \prod_{m=1}^{M} |A_{\mathcal{I}_m}| \right) - \frac{1}{2} \log \left( \prod_{m=1}^{M} |B_{\mathcal{I}_m}| \right)$$

$$-\frac{1}{2} \log \left( \prod_{m=1}^{M} |C_{\mathcal{I}_m}| \right) + \sum_{m=1}^{M} \left( -\frac{3n}{2} \log(2\pi) \right)$$

$$= -\frac{1}{2} \sum_{m=1}^{M} \left( (z^m)^T A_{\mathcal{I}_m}^{-1} z^m \right) - \frac{1}{2} \log \left( \prod_{m=1}^{M} |A_{\mathcal{I}_m}| \right) - \frac{1}{2} \log \left( \prod_{m=1}^{M} |B_{\mathcal{I}_m}| \right)$$

$$-\frac{1}{2} \log \left( \prod_{m=1}^{M} |C_{\mathcal{I}_m}| \right) - \frac{3 \log(2\pi)}{2} \left( \sum_{m=1}^{M} n_m \right) \tag{E.18}$$

# Bibliography

[1] M Himmelsbach, F von Hundelshausen, and H Wuensche. Fast segmentation of 3D point clouds for ground vehicles. *IEEE Int. Intell. Veh. Symp.*, pages 560–565, 2010.

[2] Achim Kampker, Mohsen Sefati, Arya Abdul Rachman, Kai Kreisköther, and Pascual Campoy. Towards Multi-Object Detection and Tracking in Urban Scenario under Uncertainties. (Vehits):978–989, 2018.

[3] R Behringer, S Sundareswaran, B Gregory, R Elsley, B Addison, W Guthmiller, R Daily, and D Bevly. The DARPA grand challenge - development of an autonomous vehicle. *Intell. Veh. Symp. 2004 IEEE*, pages 226–231, 2004.

[4] SAE Society of Automotive Engineers. Taxonomy and Definitions for Terms Related to Driving Automation Systems for On-Road Motor Vehicles J3016_201806, 2018.

[5] B.W. Smith. Summary of SAE Levels of Driving Automation, 2015.

[6] NVIDIA. Advanced Driving Assitance Systems.

[7] L. C. Davis. Effect of adaptive cruise control systems on traffic flow. *Phys. Rev. E - Stat. Nonlinear, Soft Matter Phys.*, 69(6 2):1–8, 2004.

[8] A. Vahidi and A. Eskandarian. Research advances in intelligent collision avoidance and adaptive cruise control. *IEEE Trans. Intell. Transp. Syst.*, 4(3):143–153, 2003.

[9] Lingyun Xiao and Feng Gao. A comprehensive review of the development of adaptive cruise control systems. *Veh. Syst. Dyn.*, 48(10):1167–1192, 2010.

[10] T Gandhi and M M Trivedi. Pedestrian collision avoidance systems: a survey of computer vision based recent studies. *IEEE Intell. Transp. Syst. Conf.*, pages 976–981, 2006.

[11] Bettina Abendroth and Ralph Bruder. *Handbook of Driver Assistance Systems*. Number Wickens 1992. 2014.

[12] Richard Matthaei and Markus Maurer. Autonomous driving - A top-down-approach. *At-Automatisierungstechnik*, 63(3):155–167, 2015.

[13] Jens Rasmussen. Skills, Rules, and Knowledge; Signals, Signs, and Symbols, and Other Distinctions in Human Performance Models. *IEEE Trans. Syst. Man Cybern.*, SMC-13(3):257–266, 1983.

[14] Edmund Donges. A Conceptual Framework for Active Safety in Road Traffic. *Veh. Syst. Dyn.*, 32(2-3):113–128, 1999.

[15] J. Montgomery, S. I. Roumeliotis, A. Johnson, and L. Matthies. The jet propulsion laboratory autonomous helicopter testbed: A platform for planetary exploration technology research and development. *J. F. Robot.*, 23(2004):245–267, 2006.

[16] John Leonard, Jonathan How, Seth Teller, Mitch Berger, Stefan Campbell, Gaston Fiore, Luke Fletcher, Emilio Frazzoli, Albert Huang, Sertac Karaman, Olivier Koch, Yoshiaki Kuwata, David Moore, Edwin Olson, Steve Peters, Justin Teo, Robert Truax, Matthew Walter, David Barrett, Alexander Epstein, Keoni Maheloni, Katy Moyer, Troy Jones, Ryan Buckley, Matthew Antone, Robert Galejs, Siddhartha Krishnamurthy, and Jonathan Williams. A perception-driven autonomous urban vehicle. *J. F. Robot.*, 25(10):727–774, oct 2008.

[17] Andrew Bacha, Cheryl Bauman, Ruel Faruque, Michael Fleming, Chris Terwelp, Charles Reinholtz, Dennis Hong, Al Wicks, Thomas Alberi, David Anderson, Stephen Cacciola, Patrick Currier, Aaron Dalton, Jesse Farmer, Jesse Hurdus, Shawn Kimmel, Peter King, Andrew Taylor, David Van Covern, and Mike Webster. Odin: Team VictorTango's entry in the DARPA Urban Challenge. *J. F. Robot.*, 25(8):467–492, aug 2008.

[18] Sören Kammel, Julius Ziegler, Benjamin Pitzer, Moritz Werling, Tobias Gindele, Daniel Jagzent, Joachim Schröder, Michael Thuy, Matthias Goebl, Felix von Hundelshausen, Oliver Pink, Christian Frese, and Christoph Stiller. Team AnnieWAY's autonomous system for the 2007 DARPA Urban Challenge. *J. F. Robot.*, 25(9):615–639, sep 2008.

[19] Michael Montemerlo, Jan Becker, Suhrid Bhat, Hendrik Dahlkamp, Dmitri Dolgov, Scott Ettinger, Dirk Haehnel, Tim Hilden, Gabe Hoffmann, Burkhard Huhnke, Doug Johnston, Stefan Klumpp, Dirk Langer, Anthony

Levandowski, Jesse Levinson, Julien Marcil, David Orenstein, Johannes Paefgen, Isaac Penny, Anna Petrovskaya, Mike Pflueger, Ganymed Stanek, David Stavens, Antone Vogt, and Sebastian Thrun. Junior: The stanford entry in the urban challenge. *Springer Tracts Adv. Robot.*, 56(October 2005):91–123, 2009.

[20] Jens Rieken, Richard Matthaei, and Markus Maurer. Benefits of using explicit ground-plane information for grid-based urban environment modeling. *Fusion*, pages 2049–2056, 2015.

[21] Shantanu Ingle and Madhuri Phute. Tesla Autopilot : Semi Autonomous Driving, an Uptick for Future Autonomy. *Int. Res. J. Eng. Technol.*, pages 2395–56, 2016.

[22] Anna Petrovskaya and Sebastian Thrun. Model Based Vehicle Detection and Tracking for Autonomous Urban Driving. 139:123–139, 2009.

[23] Michael Darms, Chris Baker, Paul Rybksi, and Chris Urmson. Vehicle Detection and Tracking for the Urban Challenge. *Work. Fahrerassistenzsysteme*, 2008.

[24] Josip Cesic, Ivan Markovic, Srecko Juric-Kavelj, and Ivan Petrovic. Detection and tracking of dynamic objects using 3D laser range sensor on a mobile platform. *Informatics Control. Autom. Robot. (ICINCO), 2014 11th Int. Conf.*, 2:110–119, 2014.

[25] Cristiano Premebida, Gonçalo Monteiro, Urbano Nunes, and Paulo Peixoto. A Lidar and vision-based approach for pedestrian and vehicle detection and tracking. *IEEE Conf. Intell. Transp. Syst. Proceedings, ITSC*, pages 1044–1049, 2007.

[26] M. Himmelsbach and H. J. Wuensche. Tracking and classification of arbitrary objects with bottom-up/top-down detection. *IEEE Intell. Veh. Symp. Proc.*, pages 577–582, 2012.

[27] S Gidel, P Checchin, C Blanc, T Chateau, and L Trassoudaine. Pedestrian Detection and Tracking in an Urban Environment Using a Multilayer Laser Scanner. *Intell. Transp. Syst. IEEE Trans.*, 11(3):579–588, 2010.

[28] Nicolai Wojke and Marcel Haselich. Moving Vehicle Detection and Tracking in Unstructured Environments. *2012 IEEE Int. Conf. Robot. Autom.*, pages 3082–3087, 2012.

[29] Tongtong Chen, Bin Dai, Daxue Liu, Hao Fu, Jinze Song, and Chongyang Wei. Likelihood-Field-Model-Based Vehicle Pose Estimation with Velodyne. *IEEE Conf. Intell. Transp. Syst. Proceedings, ITSC*, 2015-Octob:296–302, 2015.

[30] Yaakov Bar-Shalom, X. Rong Li, and Thiagalingam Kirubarajan. *Estimation with Applicaiton to Tracking and Navigation*. 2001.

[31] Trung-Dung Vu, Olivier Aycard, and Fabio Tango. Object Perception for Intelligent Vehicle Applications: A Multi-Sensor Fusion Approach. *Iv '14*, (Iv):774–780, 2014.

[32] Anton Milan, Laura Leal-Taixe, Ian Reid, Stefan Roth, and Konrad Schindler. MOT16: A Benchmark for Multi-Object Tracking. pages 1–12, 2016.

[33] Wen Xiao, Bruno Vallet, Konrad Schindler, and Nicolas Paparoditis. Simultaneous Detection and Tracking of Pedestrian From Panoramic Laser Scanning Data. *ISPRS Ann. Photogramm. Remote Sens. Spat. Inf. Sci.*, III-3(July):295–302, 2016.

[34] Longyin Wen, Dawei Du, Zhaowei Cai, Zhen Lei, Ming-Ching Chang, Honggang Qi, Jongwoo Lim, Ming-Hsuan Yang, and Siwei Lyu. UA-DETRAC: A New Benchmark and Protocol for Multi-Object Detection and Tracking. 2015.

[35] Ju Hong Yoon, Chang-Ryeol Lee, Ming-Hsuan Yang, and Kuk-Jin Yoon. Online Multi-object Tracking via Structural Constraint Event Aggregation. *2016 IEEE Conf. Comput. Vis. Pattern Recognit.*, pages 1392–1400, 2016.

[36] Yu Xiang, Alexandre Alahi, and Silvio Savarese. Learning to track: Online multi-object tracking by decision making. *Proc. IEEE Int. Conf. Comput. Vis.*, 2015 Inter:4705–4713, 2015.

[37] Jose C. Rubio, Joan Serrat, Antonio M. López, and Daniel Ponsa. Multiple-target tracking for intelligent headlights control. *IEEE Trans. Intell. Transp. Syst.*, 13(2):594–605, 2012.

[38] Sayanan Sivaraman and Mohan Manubhai Trivedi. Looking at vehicles on the road: A survey of vision-based vehicle detection, tracking, and behavior analysis. *IEEE Trans. Intell. Transp. Syst.*, 14(4):1773–1795, 2013.

[39] Mykhaylo Andriluka, Stefan Roth, and Bernt Schiele. People-tracking-by-detection and people-detection-by-tracking. *26th IEEE Conf. Comput. Vis. Pattern Recognition, CVPR*, 2008.

[40] Wenhan Luo, Junliang Xing, Anton Milan, Xiaoqin Zhang, Wei Liu, Xiaowei Zhao, and Tae-Kyun Kim. Multiple Object Tracking: A Literature Review. pages 1–18, 2014.

[41] Michael Montemerlo, Jan Becker, Suhrid Bhat, Hendrik Dahlkamp, Dmitri Dolgov, Scott Ettinger, Dirk Haehnel, Tim Hilden, Gabe Hoffmann, Burkhard Huhnke, Doug Johnston, Stefan Klumpp, Dirk Langer, Anthony Levandowski, Jesse Levinson, Julien Marcil, David Orenstein, Johannes Paefgen, Isaac Penny, Anna Petrovskaya, Mike Pflueger, Ganymed Stanek, David Stavens, Antone Vogt, and Sebastian Thrun. Junior: The stanford entry in the urban challenge. *Springer Tracts Adv. Robot.*, 56(October 2005):91–123, 2009.

[42] K. Anders, M. Hämmerle, G. Miernik, T. Drews, A. Escalona, C. Townsend, and B. Höfle. 3D GEOLOGICAL OUTCROP CHARACTERIZATION: AUTOMATIC DETECTION OF 3D PLANES (AZIMUTH AND DIP) USING LiDAR POINT CLOUDS. *ISPRS Ann. Photogramm. Remote Sens. Spat. Inf. Sci.*, III-5(July):105–112, 2016.

[43] Keni Bernardin and Rainer Stiefelhagen. Evaluating multiple object tracking performance: The CLEAR MOT metrics. *Eurasip J. Image Video Process.*, 2008, 2008.

[44] Heng Wang, Bin Wang, Bingbing Liu, Xiaoli Meng, and Guanghong Yang. Pedestrian recognition and tracking using 3D LiDAR for autonomous vehicle. *Rob. Auton. Syst.*, 88:71–78, 2017.

[45] Liang Zhang, Qingquan Li, Ming Li, Qingzhou Mao, and Andreas Nüchter. Multiple vehicle-like target tracking based on the velodyne lidar? *IFAC Proc. Vol.*, 8(PART 1):126–131, 2013.

[46] Bertrand Douillard, J. Underwood, N. Kuntz, V. Vlaskine, a. Quadros, P. Morton, and a. Frenkel. On the segmentation of 3D lidar point clouds. *Icra*, pages 2798–2805, 2011.

[47] Jens Rieken, Richard Matthaei, and Markus Maurer. Toward Perception-Driven Urban Environment Modeling for Automated Road Vehicles. *IEEE Conf. Intell. Transp. Syst. Proceedings, ITSC*, 2015-Octob:731–738, 2015.

[48] Tobias Nothdurft, Peter Hecker, Sebastian Ohl, Falko Saust, Markus Maurer, Andreas Reschka, and Jürgen Rudiger Böhmer. Stadtpilot: First fully autonomous test drives in urban traffic. *IEEE Conf. Intell. Transp. Syst. Proceedings, ITSC*, pages 919–924, 2011.

[49] Shiyang Song, Zhiyu Xiang, and Jilin Liu. Object tracking with 3D LIDAR via multi-task sparse learning. *2015 IEEE Int. Conf. Mechatronics Autom. ICMA 2015*, (61071219):2603–2608, 2015.

[50] Michael Kusenbach, Michael Himmelsbach, and Hans Joachim Wuensche. A new geometric 3D LiDAR feature for model creation and classification of moving objects. *IEEE Intell. Veh. Symp. Proc.*, 2016-Augus(Iv):272–278, 2016.

[51] R. MacLachlan and C. Mertz. Tracking of Moving Objects from a Moving Vehicle Using a Scanning Laser Rangefinder. *2006 IEEE Intell. Transp. Syst. Conf.*, pages 301–306, 2006.

[52] Sayanan Sivaraman and Mohan M. Trivedi. A review of recent developments in vision-based vehicle detection. *2013 IEEE Intell. Veh. Symp.*, (Iv):310–315, 2013.

[53] Zhongzhen Luo, Saeid Habibi, and Martin Mohrenschildt. LiDAR Based Real Time Multiple Vehicle Detection and Tracking. 10(6):1083–1090, 2016.

[54] Kristian Kovačić, Edouard Ivanjko, and Hrvoje Gold. Computer Vision Systems in Road Vehicles: A Review. 6(2):6–14, 2013.

[55] Songlin Piao, Tanittha Sutjaritvorakul, and Karsten Berns. Compact Data Association in Multiple Object Tracking: Pedestrian Tracking on Mobile Vehicle as Case Study. *IFAC-PapersOnLine*, 49(15):175–180, 2016.

[56] Siyang Han, Xiao Wang, Linhai Xu, Hongbin Sun, and Nanning Zheng. Frontal object perception for Intelligent Vehicles based on radar and camera fusion. *Chinese Control Conf. CCC*, 2016-Augus:4003–4008, 2016.

[57] Bin Tian, Ye Li, Bo Li, and Ding Wen. Rear-view vehicle detection and tracking by combining multiple parts for complex Urban surveillance. *IEEE Trans. Intell. Transp. Syst.*, 15(2):597–606, 2014.

[58] R. H. Rasshofer and K. Gresser. Automotive radar and lidar systems for next generation driver assistance functions. *Adv. Radio Sci.*, 3:205–209, 2005.

[59] Velodyne. High definition lidar HDL-64E S2 - datasheet, 2010.

[60] Radu Bogdan Rusu and S. Cousins. 3D is here: point cloud library. *IEEE Int. Conf. Robot. Autom.*, pages 1 – 4, 2011.

[61] Andreas Geiger, Philip Lenz, Cristoph Stiller, and Raquel Urtasun. Vision meets robotics: The KITTI dataset. *Int. J. Rob. Res.*, 32(11):1231–1237, 2013.

[62] Christophe Coué, Cédric Pradalier, Christian Laugier, Thierry Fraichard, and Pierre Bessière. Bayesian occupancy filtering for multitarget tracking: An automotive application. *Int. J. Rob. Res.*, 25(1):19–30, 2006.

[63] Georg Tanzmeister, Julian Thomas, Dirk Wollherr, and Martin Buss. Grid-based mapping and Tracking in dynamic environments using a uniform evidential environment representation. *Proc. - IEEE Int. Conf. Robot. Autom.*, pages 6090–6095, 2014.

[64] Radu Danescu, Florin Oniga, and Sergiu Nedevschi. Modeling and tracking the driving environment with a particle-based occupancy grid. *IEEE Trans. Intell. Transp. Syst.*, 12(4):1331–1342, 2011.

[65] Matthias Schreier. Bayesian environment representation, prediction, and criticality assessment for driver assistance systems. *At-Automatisierungstechnik*, 65(2):151–152, 2017.

[66] Tongtong Chen, Bin Dai, Ruili Wang, and Daxue Liu. Gaussian-Process-Based Real-Time Ground Segmentation for Autonomous Land Vehicles. *J. Intell. Robot. Syst. Theory Appl.*, 76(3-4):563–582, 2014.

[67] Klaas Klasing, Dirk Wollherr, and Martin Buss. A clustering method for efficient segmentation of 3D laser data. *Icra*, pages 4043–4048, 2008.

[68] Soonmin Hwang, Namil Kim, Yukyung Choi, Seokju Lee, and In So Kweon. Fast multiple objects detection and tracking fusing color camera and 3D LIDAR for intelligent vehicles. *2016 13th Int. Conf. Ubiquitous Robot. Ambient Intell. URAI 2016*, pages 234–239, 2016.

[69] Bjorn Barrois, Stela Hristova, Christian Wohler, Franz Kummert, and Christoph Hermes. 3D Pose Estimation of Vehicles Using a Stereo Camera. *Intell. Veh. Symp.*, pages 267–272, 2009.

[70] Matti Lehtomaki, Anttoni Jaakkola, Juha Hyyppa, Jouko Lampinen, Harri Kaartinen, Antero Kukko, Eetu Puttonen, and Hannu Hyyppa. Object Classification and Recognition From Mobile Laser Scanning Point Clouds in a Road Environment. *IEEE Trans. Geosci. Remote Sens.*, 54(2):2–3, 2015.

[71] Yutong Ye, Liming Fu, and Bijun Li. Object detection and tracking using multi-layer laser for autonomous urban driving. *IEEE Conf. Intell. Transp. Syst. Proceedings, ITSC*, pages 259–264, 2016.

[72] W. Zhang, Y. Guo, M. Lu, and J. Zhang. Ground target detection in LiDAR point clouds using AdaBoost. *ICCAIS 2015 - 4th Int. Conf. Control. Autom. Inf. Sci.*, pages 22–26, 2015.

[73] Heng Wang, Bin Wang, Bingbing Liu, Xiaoli Meng, and Guanghong Yang. Pedestrian recognition and tracking using 3D LiDAR for autonomous vehicle. *Rob. Auton. Syst.*, 88:71–78, 2017.

[74] Takeo Miyasaka, Yoshihiro Ohama, and Yoshiki Ninomiya. Ego-motion estimation and moving object tracking using multi-layer LIDAR. *2009 IEEE Intell. Veh. Symp.*, pages 151–156, 2009.

[75] Dominic Zeng Wang, Paul Newman, and Ingmar Posner. Laser-Based Detection and Tracking of Dynamic Objects. (October), 2014.

[76] R Biswas, B Limketkai, S Sanner, and S Thrun. Towards Object Mapping in Dynamic Environments With Mobile Robots. *Proc. Conf. Intell. Robot. Syst.*, 2002.

[77] Denis F. Wolf and Gaurav S. Sukhatme. Mobile robot simultaneous localization and mapping in dynamic environments. *Auton. Robots*, 19(1):53–65, 2005.

[78] David Held, Jesse Levinson, Sebastian Thrun, and Silvio Savarese. Robust real-time tracking combining 3D shape, color, and motion. *Int. J. Rob. Res.*, 35(1-3):1–28, 2015.

[79] R Kaestner, J Maye, Y Pilat, and Roland Siegwart. Generative object detection and tracking in 3D range data. *Robot. Autom. (ICRA), 2012 IEEE Int. Conf.*, pages 3075–3081, 2012.

[80] Olivier Aycard. Laser-based detection and tracking moving objects using data-driven Markov chain Monte Carlo. *2009 IEEE Int. Conf. Robot. Autom.*, pages 3800–3806, 2009.

[81] B. Fortin, J.C. Noyer, and R. Lherbier. A particle filtering approach for joint vehicular detection and tracking in lidar data. *2012 IEEE Int. Instrum. Meas. Technol. Conf. Proc.*, (3):391–396, 2012.

[82] A Gorji and M B Menhaj. Multiple Target Tracking for Mobile Robots Using the JPDAF Algorithm. *19th IEEE Int. Conf. Tools with Artif. Intell. 2007. ICTAI 2007*, 1:137–145, 2007.

[83] Songhwai Oh, S. Russell, and S. Sastry. Markov chain Monte Carlo data association for general multiple-target tracking problems. *2004 43rd IEEE Conf. Decis. Control (IEEE Cat. No.04CH37601)*, pages 735–742 Vol.1, 2004.

[84] Qian Yu, Gerard Medioni, and Isaac Cohen. Multiple Target Tracking Using Spatio-Temporal Markov Chain Monte Carlo Data Association. *IEEE Conf. Comp. Vis. \& Pattern Recognit.*, 2007.

[85] N Kaempchen and K Dietmayer. IMM Vehicle Tracking for Traffic Jam Situations on Highways. *Proc. 7th Intl. Conf. Multisens. Inf. Fusion*, 1(2):868–875, 2004.

[86] Subhash Challa, Mark R. Morelande, Darko Mušicki, and Robin J. Evans. *Fundamentals of Object Tracking - Knovel*. 2011.

[87] Timothy D Barfoot. State Estimation for Robotics. 2018.

[88] S.J. Julier and J. K. Uhlmann. Unscented filtering and nonlinear estimation. *Proc. IEEE*, 92(3):401–422, 2004.

[89] Marco Allodi, Alberto Broggi, Domenico Giaquinto, Marco Patander, and Antonio Prioletti. Machine learning in tracking associations with stereo vision and lidar observations for an autonomous vehicle. *IEEE Intell. Veh. Symp. Proc.*, 2016-Augus(Iv):648–653, 2016.

[90] Feihu Zhang, Daniel Clarke, and Alois Knoll. Vehicle detection based on LiDAR and camera fusion. *2014 17th IEEE Int. Conf. Intell. Transp. Syst. ITSC 2014*, pages 1620–1625, 2014.

[91] Matthias Schreier, Volker Willert, and Jürgen Adamy. Compact Representation of Dynamic Driving Environments for ADAS by Parametric Free Space and Dynamic Object Maps. pages 1–18, 2015.

[92] Yong-shik Kim and Keum-Shik Hong. An IMM algorithm for tracking maneuvering vehicles in an adaptive cruise control environment. *Int. J. Control Autom. Syst.*, 2(3):310–318, 2004.

[93] M. de Feo, A. Graziano, R. Miglioli, and A. Farina. IMMJPDA versus MHT and Kalman filter with NN correlation: performance comparison. *IEE Proc. - Radar, Sonar Navig.*, 144(2):49, 1997.

[94] Guan Zhai, Huadong Meng, and Xiqin Wang. A Constant Speed Changing Rate and Constant Turn Rate Model for Maneuvering Target Tracking. pages 5239–5253, 2014.

[95] S S Blackman. Multiple hypothesis tracking for multiple target tracking. *IEEE Aerosp. Electron. Syst. Mag.*, 19(1):5–18, 2004.

[96] Yaakov Bar-Shalom, Fred Daum, and Jim Huang. The probabilistic data association filter. *IEEE Control Syst. Mag.*, 29(6):82–100, 2009.

[97] Xuezhi Wang, Subhash Challa, and Rob Evans. Gating techniques for maneuvering target tracking in clutter. *IEEE Trans. Aerosp. Electron. Syst.*, 38(3):1087–1097, 2002.

[98] Matthias Schreier and Volker Willert. Grid Mapping in Dynamic Road Environments : Classification of Dynamic Cell Hypothesis via Tracking. pages 3995–4002, 2014.

[99] C.K.I. Williams C.E. Rasmussen. *Gaussian Processes for Machine Learning*. 2006.

[100] Dimitris Zermas, Izzat Izzat, and Nikolaos Papanikolopoulos. Fast Segmentation of 3D Point Clouds : A Paradigm on LiDAR Data for Autonomous Vehicle Applications. *IEEE Int. Conf. Robot. Autom.*, (May):5067–5073, 2017.

[101] Myung-Ok Shin, Gyu-Min Oh, Seong-Woo Kim, and Seung-Woo Seo. Real-Time and Accurate Segmentation of 3-D Point Clouds Based on Gaussian Process Regression. *IEEE Trans. Intell. Transp. Syst.*, pages 1–15, 2017.

[102] Dmitriy Korchev, Shinko Cheng, Yuri Owechko, and Kk Kim. On Real-Time LIDAR Data Segmentation and Classification. *Worldcomp-Proceedings.Com*, (February), 2016.

[103] Alireza Asvadi, Cristiano Premebida, Paulo Peixoto, and Urbano Nunes. 3D Lidar-based static and moving obstacle detection in driving environments: An approach based on voxels and multi-region ground planes. *Rob. Auton. Syst.*, 2016.

[104] Frank Moosmann, Oliver Pink, and Christoph Stiller. Segmentation of 3D lidar data in non-flat urban environments using a local convexity criterion. *IEEE Intell. Veh. Symp. Proc.*, pages 215–220, 2009.

[105] Christopher J Paciorek and Mark J Schervish. Nonstationary Covariance Functions for Gaussian Process Regression. pages 1–27, 2003.

[106] ML Stein. Nonstationary spatial covariance functions. *Unpubl. Tech. report. Available*, 60637:0–7, 2005.

[107] Christian Plagemann, Kristian Kersting, and Wolfram Burgard. Nonstationary Gaussian process regression using point estimates of local smoothness. *Lect. Notes Comput. Sci. (including Subser. Lect. Notes Artif. Intell. Lect. Notes Bioinformatics)*, 5212 LNAI(PART 2):204–219, 2008.

[108] Yulai Zhang and Guiming Luo. Recursive prediction algorithm for non-stationary Gaussian Process. *J. Syst. Softw.*, 127:295–301, 2017.

[109] Geir-Arne Fuglstad, Finn Lindgren, Daniel Simpson, and Håvard Rue. Exploring a new class of non-stationary spatial Gaussian random fields with varying local anisotropy. *arXiv*, 25(1):1–28, 2013.

[110] Geir Arne Fuglstad, Daniel Simpson, Finn Lindgren, and Håvard Rue. Does non-stationary spatial data always require non-stationary random fields? *Spat. Stat.*, 14:505–531, 2015.

[111] V. Nguyen, A. Martinelli, N. Tomatis, and R. Siegwart. A comparison of line extraction algorithms using 2D laser rangefinder for indoor mobile robotics. pages 1929–34, 2005.

[112] Christopher Joseph Paciorek. Nonstationary Gaussian processes for regression and spatial modelling. *Thesis*, 6(May):1–34, 2003.

[113] Daniel Oñoro Rubio, Artem Lenskiy, and Jee Hwan Ryu. Connected components for a fast and robust 2D lidar data segmentation. *Proc. - Asia Model. Symp. 2013 7th Asia Int. Conf. Math. Model. Comput. Simulation, AMS 2013*, pages 160–165, 2013.

[114] Siddharth Gupta, Diana Palsetia, Md Mostofa Ali Patwary, Ankit Agrawal, and Alok Choudhary. A new parallel algorithm for two-pass connected component labeling. *Proc. Int. Parallel Distrib. Process. Symp. IPDPS*, pages 1355–1362, 2014.

[115] Kesheng Wu, Ekow Otoo, and Kenji Suzuki. Optimizing two-pass connected-component labeling algorithms. *Pattern Anal. Appl.*, 12(2):117–135, 2009.

[116] Gill Barequet and Sariel Har-Peled. Efficiently Approximating the Minimum-Volume Bounding Box of a Point Set in Three Dimensions. *J. Algorithms*, 38(1):91–109, 2001.

[117] Oscar Wellenstam. LiDAR Clustering and Shape Extraction for Automotive Applications. 2017.

[118] Stefan Wender, Michael Schoenherr, Nico Kaempchen, and Klaus Dietmayer. Classification of Laserscanner measurements at intersection scenarios with automatic parameter optimization. *IEEE Intell. Veh. Symp. Proc.*, 2005:94–99, 2005.

[119] Felipe Jiménez and José Eugenio Naranjo. Improving the obstacle detection and identification algorithms of a laserscanner-based collision avoidance system. *Transp. Res. Part C Emerg. Technol.*, 19(4):658–672, 2011.

[120] Xiao Zhang, Wenda Xu, Chiyu Dong, and John M. Dolan. Efficient L-shape fitting for vehicle detection using laser scanners. *2017 IEEE Intell. Veh. Symp.*, (Iv):54–59, 2017.

[121] R MacLachlan and C Mertz. Tracking of Moving Objects from a Moving Vehicle Using a Scanning Laser Rangefinder. *IEEE Intell. Transp. Syst. Conf.*, pages 301–306, 2006.

[122] Phan Thanh An. A modification of Graham's algorithm for determining the convex hull of a finite planar set. *Ann. Math. Informaticae*, 34:3–8, 2007.

[123] R. L. Graham. An efficient algorith for determining the convex hull of a finite planar set, 1972.

[124] Gang Mei, John C. Tipper, and Nengxiong Xu. An algorithm for finding convex hulls of planar point sets. *Proc. 2nd Int. Conf. Comput. Sci. Netw. Technol. ICCSNT 2012*, pages 888–891, 2012.

[125] Yong-Shik Kim Yong-Shik Kim and Keum-Shik Hong Keum-Shik Hong. A tracking algorithm for autonomous navigation of AGVs in a container terminal. *30th Annu. Conf. IEEE Ind. Electron. Soc. 2004. IECON 2004*, 1, 2004.

[126] Beomjun Kim, Dongwook Kim, Sungyoul Park, Yonghwan Jung, and Kyongsu Yi. Automated Complex Urban Driving based on Enhanced Environment Representation with GPS/map, Radar, Lidar and Vision. *IFAC-PapersOnLine*, 49(11):190–195, 2016.

[127] Abdul Hadi Abd Rahman, Hairi Zamzuri, Saiful Amri Mazlan, and Mohd Azizi Abdul Rahman. Model-Based Detection and Tracking of Single Moving Object Using Laser Range Finder. *2014 5th Int. Conf. Intell. Syst. Model. Simul.*, pages 556–561, 2014.

[128] H.A.P. Blom and E.A. Bloem. Interacting multiple model joint probabilistic data association avoiding track coalescence. *Proc. 41st IEEE Conf. Decis. Control. 2002.*, 3(December):3408–3415, 2002.

[129] D Mušicki and R Evans. Joint Integrated Probabilistic Data Association - \textrm{JIPDA}. *Proc. Fusion 2002*, (I):1120–1125, 2002.

[130] Robin Schubert, Eric Richter, and Gerd Wanielik. Comparison and Evaluation of Advanced Motion Models for Vehicle Tracking. (1):730–735.

[131] Giorgio Grisetti. Notes on Least-Squares and SLAM. Technical report, 2014.

[132] Timothy D. Barfoot and Paul T. Furgale. Associating uncertainty with three-dimensional poses for use in estimation problems. *IEEE Trans. Robot.*, 30(3):679–693, 2014.

[133] Sean Anderson and Timothy D. Barfoot. Full STEAM ahead: Exactly sparse Gaussian process regression for batch continuous-time trajectory estimation on SE(3). *IEEE Int. Conf. Intell. Robot. Syst.*, 2015-Decem(3):157–164, 2015.

[134] Christopher M Bishop and Cazhaow S Qazaz. Regression with input-dependent noise: A Bayesian treatment. *Adv. Neural Inf. Process. Syst. 9*, 9(1991):347–353, 1997.

[135] PW Goldberg, CKI Williams, and CM Bishop. Regression with input-dependent noise: A gaussian process treatment. *Adv. Neural Inf. Process. Syst.*, 10:493–499, 1997.